\documentclass[a4paper,12pt, oneside, onecolumn, final]{book}
\usepackage{amsmath}
\usepackage{amsfonts}
\usepackage{clusteringCommands}
\usepackage{multirow}
\usepackage{color}
\usepackage[usenames,dvipsnames,svgnames,table]{xcolor}
\usepackage{graphicx, epsfig, tipa}
\usepackage{graphicx}
\usepackage{pdfpages}
\usepackage{epsfig}
\usepackage{mathrsfs}
\usepackage{amssymb}
\usepackage{trfsigns}
\usepackage{setspace}
\usepackage{placeins}
\usepackage{algorithmic}
\usepackage{algorithm}
\usepackage{lscape}
\numberwithin{algorithm}{chapter}
\usepackage{rotating}
\usepackage{pifont}
\usepackage{multicol}
\usepackage{exscale,relsize}
\usepackage{float}
\usepackage{rotating}
\usepackage{booktabs}
\usepackage{arydshln}
\usepackage{caption}
\usepackage{todonotes}
\usepackage[lofdepth,lotdepth]{subfig}
\usepackage{afterpage}
\usepackage{url}
\usepackage{rotating}
\usepackage[toc,page]{appendix}
\usepackage{tipa}
\usepackage{textcomp}
\usepackage{bm}
\usepackage{nicefrac}
\usepackage{pdfpages}

\newcommand{\phat}[1]{\hat{#1}}
\begin{document}

\pagenumbering{roman}
%!TEX root = main.tex 
\begin{titlepage} 
\singlespacing 
 
\begin{center} 
\vspace*{2.0cm} 
 
\begin{Huge} 
\begin{spacing}{1} 
\textbf{Decoding visemes: improving machine lip-reading (PhD thesis)} 
\end{spacing} 
\end{Huge} 
 
\vfill 
{\LARGE Helen L. Bear} 
  
\vspace*{1cm} 
\begin{Large} 
\begin{spacing}{1} 
University of East Anglia\\ 
School of Computing Sciences 
\end{spacing} 
\end{Large} 
 
\vfill 
 
\includegraphics[width=5cm,keepaspectratio=true]{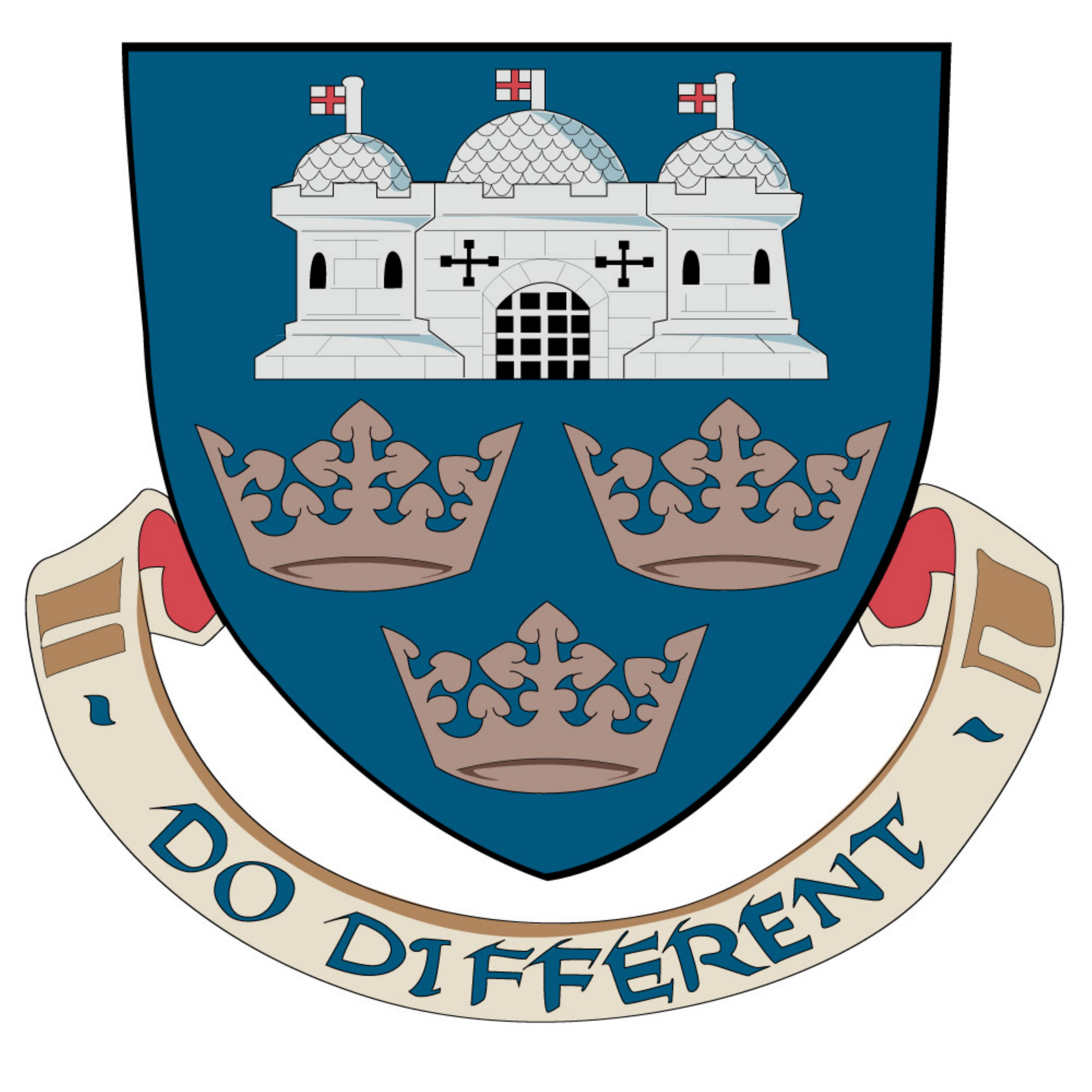} 
 
\large{July 2016} 
 
\end{center} 
 
\vspace{-\parsep} 
 
\vfill 
\noindent 
\footnotesize 
\copyright This copy of the thesis has been supplied on condition that anyone who consults it is understood to recognise that its copyright rests with the author and that no quotation from the thesis, nor any information derived therefrom, may be published without the author's prior written consent. 
\end{titlepage} 
 
\chapter*{Abstract} 
\begin{spacing}{1.5} 

This thesis is about improving machine lip-reading, that is, the classification of speech from only visual cues of a speaker. Machine lip-reading is a niche research problem in both areas of speech processing and computer vision. %Current machine lip-reading performance is good but it is not robust and classification rates need to be higher. Our research question is: can we augment or improve current lip-reading classifiers to improve machine lip-reading? 

Current challenges for machine lip-reading fall into two groups: the content of the video, such as the rate at which a person is speaking or; the parameters of the video recording for example, the video resolution. We begin our work with a literature review to understand the restrictions current technology limits machine lip-reading recognition and conduct an experiment into resolution affects. We show that high definition video is not needed to successfully lip-read with a computer. 

The term ``viseme'' is used in machine lip-reading to represent a visual cue or gesture which corresponds to a subgroup of phonemes where the phonemes are indistinguishable in the visual speech signal. Whilst a viseme is yet to be formally defined, we use the common working definition `a viseme is a group of phonemes with identical appearance on the lips'. A phoneme is the smallest acoustic unit a human can utter. Because there are more phonemes per viseme, mapping between the units creates a many-to-one relationship. Many mappings have been presented, and we conduct an experiment to determine which mapping produces the most accurate classification. Our results show Lee's \cite{lee2002audio} is best. Lee's classification also outperforms machine lip-reading systems which use the popular Fisher \cite{fisher1968confusions} phoneme-to-viseme map.  

Further to this, we propose three methods of deriving speaker-dependent phoneme-to-viseme maps and compare our new approaches to Lee's. Our results show the sensitivity of phoneme clustering and we use our new knowledge for our first suggested augmentation to the conventional lip-reading system. 

Speaker independence in machine lip-reading classification is another unsolved obstacle. It has been observed, in the visual domain, that classifiers need training on the test subject to achieve the best classification. Thus machine lip-reading is highly dependent upon the speaker. Speaker independence is the opposite of this, or in other words, is the classification of a speaker not present in the classifier's training data. We investigate the dependence of phoneme-to-viseme maps between speakers. Our results show there is not a high variability of visual cues, but there is high variability in trajectory between visual cues of an individual speaker with the same ground truth. This implies a dependency upon the number of visemes within each set for each individual.

Finally, we investigate how many visemes is the optimum number within a set. We show the phoneme-to-viseme maps in literature rarely have enough visemes and the optimal number, which varies by speaker, ranges from 11 to 35. The last difficulty we address is decoding from visemes back to phonemes and into words. Traditionally this is completed using a language model. The language model unit is either: the same as the classifier, e.g. visemes or phonemes; or the language model unit is words. In a novel approach we use these optimum range viseme sets within hierarchical training of phoneme labelled classifiers. This new method of classifier training demonstrates significant increase in classification with a word language network.
 
\end{spacing} 
\tableofcontents 
%\listoffigures 
%\listoftables 
%\input{abbreviations} 
 %%---------------------------------- 
\chapter*{Acknowledgements} 
%\addcontentsline{toc}{chapter}{Acknowledgements} 
 
\begin{spacing}{1.5} 
 
 This is my opportunity to say thank you to some extraordinary people without whom I really couldn't have completed my PhD. My heartfelt thanks go to each and every one of you for so much more than I am capable of conveying in words. 
 
Professor Richard Harvey (aka PhD supervisor extraordinare), you are my dream supervisor and friend. Your intelligence, patience, support and humour have been invaluable and I have loved working with you this past four years. To the rest of my supervisory team: Dr Barry-John Theobald, Dr Yuxuan Lan, Professor Stephen Cox, and Dr Anthony Bagnall, you are amazing. Thank you all for your patient education, support and guidance. I am also grateful for my examiners Professor Andy Day and Dr Naomi Harte for assessing my viva performance. 
 
To my lab colleagues Mr Thomas Le Cornu and Mr Danny Websdale - you guys are the best lab buddies I could ever have dreamed of. Thank you for making coming in to work every day so good. 

Finally, thank you to my family, Barbara (aka Mum), Jeremy (aka Dad), Philip, Michelle, and Amelia, who know barely anything about what I've been doing for the past four years and understand it even less, but have supported me throughout this crazy endeavour, \#proudtobeabear 

\vspace*{0.5in}

\end{spacing} 

\mainmatter
%----------------------------------
\begin{spacing}{1}

\pagenumbering{arabic}
%!TEX root = main.tex 
\chapter[Introduction]{Introduction} 
\label{chap:intro} 
 
Speech is bimodal. This means there are two modes of information: acoustic and visual. Humans use both signals to understand the speech of others \cite{mcgurk}. Given that acoustic recognition has been studied for over fifty years \cite{davis1952automatic}, it is not surprising that acoustic recognition is far more mature than visual-only recognition and there have been significant increases in performance in speech recognition systems, although they remain susceptible to noise \cite{galatas2011audio}. Imagine trying to recognise a pilot's speech over the background noise of the aeroplane engine in a cockpit. In this case, the audio signal is severely deteriorated by the noise of the environment. However, this noise does not affect the visual signal. Thus, a desire to recognise speech from the visual signal alone is born. The visual signal can be used in combination with the acoustic signal, this is audio-visual speech recognition (AVSR) \cite{potamianos2004audio}, or, there is the possibility of using the visual signal alone. This latter configuration is machine lip-reading which is the topic of this thesis.

Lip-reading is a challenging task. When researchers investigate AVSR, it is common for audio recognition to dominate any benefit from lip-reading, nevertheless, if we can make pure lip-reading successful there would be benefits for audio-visual recognition. Furthermore, there are a few scenarios where it is impractical or senseless to install a close microphone. An example might be an interactive booth in a busy station or airport where there is poor signal-to-noise ratio (SNR) or some distance between the person and the screen. In practice however, a major use of a good machine lip-reading system would be as part of an AVSR system. 
 
\section{Applications of machine lip-reading} 
There are a range of scenarios where a machine lip-reading system would be beneficial. We discuss a few examples here. 
 
During sports events there are often headlines about arguments between players, referees and even supporters. In the 2006 football World Cup Final between France and Italy, it was 19 minutes into extra time when Zinedine Zidane, on the opposite end of the pitch to the football, head-butted an Italian player without apparent justification. This action earned him a red card and consequently France went on to lose both the match and the world cup \cite{zidane}. It later transpired, as admitted by Materazzi (the recipient of the head-butt), Zidane was provoked by a targeted insult of a late family member. In this case, if a machine lip-reading system had been present to confirm the provocation, whilst Zidane would have still been red carded, so would have Materazzi. Thus playing ten men against ten, the outcome of the match, and the World Cup, could have been different. 
 
In history there are a great number of silent videos. Common examples are silent entertainment films and historical documentaries. In \cite{battlesomme} we see a professional human lip-reader assist researchers comprehend what soldier's conversations were before they went into battle and during battle preparations. Similarly, in \cite{hitler} we are shown how lip-readers used on the home movies of Hitler give historians an insight to an infamous figure of interest. 

There has been long debate about if, in silent entertainment films of the era 1895-1927, films were ever scripted as the audio could not be captured with the video channel. In \cite{silentmovie} we learn that, not only were these films in fact fully scripted, but in human lip-reading experiments, variation from the scripts were fully noticeable. Collectively, this human nature to be interested in history and learn from historical evidence is a further motivator for achieving robust automatic lip-reading systems. 
 
Theobald \textit{et al.} \cite{theobald:640205} examine lip-reading for law enforcement. They note that in law-enforcement there are many departments who would benefit from an automatic lip-reading system. They present a new technique for improved lip-reading whereby the extracted features are modified to increase the classification performance. The modification is amplifying the feature parameters (they use Active Appearance Models which we explain fully in Chapter~\ref{chap:featuretypes}), to exaggerate the lip gestures recorded on camera. The technique was tested using a phonetically balanced corpus of syntactically correct sentences. The data set had very little contextual information \cite{theobaldPHD} to remove effects of context network support.  Machine lip-reading would help in law enforcement as robust lip-reading of filmed conversations during criminal acts, e.g. on CCTV could be evidence for the prosecution of offenders. 
 
In the murder case of Arelene Fraser, Nat Fraser was caught and imprisoned. Evidence used by the prosecution included transcripts provided by professional lip-reader Jessica Rees \cite{natfraser}. Whilst the perpetrator thought he had committed the `perfect' murder, and took steps to avoid any conversations being overheard, he had not thought about those who could read lips. With the transcripts of Fraser's conversations, prosecutors turned the co-conspirators into witnesses and Nat Fraser was prosecuted. However later, the reliability of lip-reading transcripts as evidence was successfully challenged, because human lip-readers are unreliable. 

The reliability of human lip-readers is debatable. It has been said that this reliability varies not just between different pairings of speakers within a conversation, but also subject to the situation (context and environment) of the conversation (with the same speakers) \cite{lott1960influence}. This means that a good lip-reader on one day with a particular speaker could either misinterpret an alternative speaker or if lip-reading the same person in another place, fail to comprehend the speech uttered. Furthermore, human lip-readers are expensive, examples of Consuelo Gonsales \cite{consuelo} and Jessica Rees \cite{rees} operate on an as quoted basis. So we know that robust lip-readers are rare \cite{lott1960influence} and often we have no way of verifying the accuracy of the lip-reading performance as a ground truth is rarely available. It is only in controlled experiments that a ground truth exists \cite{Benoit1996, stork1996speechreading, comparHumMacLipRead}. 

In \cite{lott1960influence} an investigation into the effect of likeability between individuals in a lip-read conversation, such as the status of their relationship, showed that a good relationship increases the accuracy of the lip-reading interpretation. To apply this observation to a real world scenario of introducing a lip-reader to someone they do not know personally, such as on a video documentary, deteriorates the confidence that their lip-reading ability will be robust. This idea is supported in Nichie's lip-reading and practice handbook \cite{nrrchm1912lipreading} where in Chapter two it is suggested that the value of practicing lip-reading is rightly attached to the teacher's personality for success. 

In \cite{summerfield1992lipreading} Summerfield describes some reasons which can distinguish poor from good lip-readers. This list is deduced from the results of a series of experiments (\cite{heider1940experimental, dodd1989teaching, macleod1987quantifying, lyxell1989beyond, woodward1960phoneme}) which show that the achievement rates in lip-reading tests can range from 10\% to over 70\%. These achievement rates vary due to the parameter selections for each experiment which are chosen for the specific task being addressed. In particular, the accuracy metric (some present word error rate, $w.e.r$ whereas others present percent true positive matches, $\%$, others alternative metrics like the HTK correctness and accuracy scores (explained in full in section~\ref{sec:htk}, Equations~\ref{eq:correctness} \&~\ref{eq:accuracy} respectively) and the classification unit (there are a number of options here - matching on phonemes, visemes or words) have a significant affect on how one should compare such investigations.   

Some affects on human lip-reading performance are: 

\begin{itemize}
\item{intelligence and verbal reasoning - McGrath~\cite{mcgrath1985examination} showed that a fundamental level of intelligence and verbal reasoning are essential to be able to lip-read at all, but beyond a limit these skills could not raise human comprehension further.}
\item{Training - human lip-readers who have either self-studied or have been trained in some manner to practice the skill of lip-reading are shown to be no better than those who have received no training \cite{conrad1977lip, dodd1989teaching}. Also it has been shown that human lip-readers can actually get worse with training \cite{binnie1976visual}, and this effect is more present when humans lip read from videos rather than in the presence of the speaker \cite{lan2012insights}. }
\item{Low-level visual-neural processing - Summerfield \cite{summerfield1992lipreading} discusses the physiological matter of the processing speed of these neural processes in the human brain. The suggestion is that lip-reading is difficult to learn because it is dependent upon these low-level neural processes. This suggestion has however, not received reproducible results to support the proposition which comforts us that human lip-reading is possible, however challenging.}
\item{Closeness between the conversation participants - studies show that a relationship of some description between those talking, or personable knowledge of the speaker by the interpreter can improve human lip-reading \cite{lott1960influence, ronnberg1998conceptual, ewing1944lipreading}. }
\item{Knowledge of conversation context - without the constraint that is the `rules' of a language to limit what a probable utterance is, lip-reading becomes almost impossible, or akin to guessing \cite{ samuelsson1993implicit}. In \cite{samuelsson1991script} experiments showed that recognising isolated sentences was as low scoring as simply guessing from the context alone.}
\end{itemize} 

In summary, the main application of a machine lip-reading system would be any situation where the audio signal in a video is either absent or too noisy to comprehend, or where the alternative, human lip-readers, are too expensive or too unreliable. % In these types of scenarios having a machine able to lip-read would be invaluable.

\section{The research problem} 
 \label{sec:resq}
\begin{figure}[!ht] 
\centering 
\includegraphics[width=0.9\textwidth]{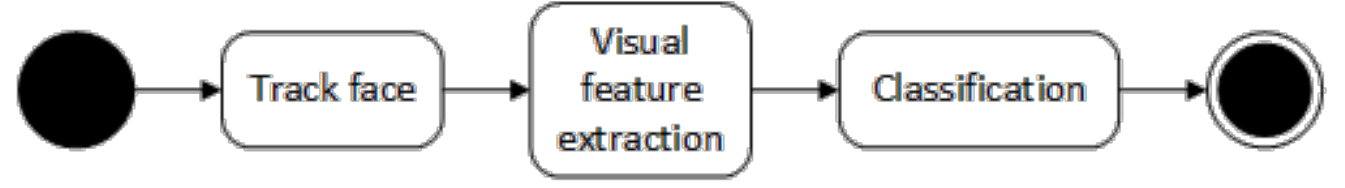} 
\caption{The three main functions in a traditional lip-reading system} 
\label{fig:lip_read} 
\end{figure}

A conventional lip-reading system consists of a sequence of tasks as shown in Figure~\ref{fig:lip_read}. Our work focuses on the classification task. Currently we have to make some assumptions by tracking a face in a video in order to extract some features before we can undertake machine lip-reading. 
 
The first task on the left hand side of Figure~\ref{fig:lip_read}, is face tracking. This means to locate a face in an image (one frame of a video) and track it throughout the whole video sequence. By the end of the tracking process, often completed by fitting a model to each frame, we have a data structure containing information about the face through time. Examples of work showing face finding and tracking are in \cite{schwerdt2000robust} and \cite{tomitaka1995human}. Example tracking methods are, with Active Appearance Models \cite{927467}, or with Linear Predictors \cite{ong2011robust}. We discuss these two methods in Chapter~\ref{chap:featuretypes}. The second task, in the centre of Figure~\ref{fig:lip_read}, is visual feature extraction. Using the fitted data parameters from task one, we can extract features which contain solely information pertaining to the speaker's lips. The third and final task on the right hand side of Figure~\ref{fig:lip_read} is classification. This is where we train some kind of classification model, using some visual features as training data, and use the classifiers to classify some unseen test data. Classification produces an output which can be compared with a ground truth to evaluate the accuracy of the classifiers. 

%Even with perfect tracking and features, if classifiers are too poor, then there is no point in trying to lip-read. However, if we can classify well, even with non-perfect tracking and features, we have hope for a robust machine lip-reading system in the future. In other words, at this time it is possible to conceive a system with effective tracking and reliable feature extraction but, still there is not enough information available for accurate classification. Therefore, the classification task is the nub of the lip-reading problem, and much of this thesis is focused on classification, rather than additional tasks such as tracking and feature extraction.

There is a lot of literature on methods of feature extraction methods \cite{ong2008robust, 4362878, hong2006pca,yang1996real, potamianos1998image, luettin1997speechreading} and tracking faces through images, \cite{927467, 1027648, mckenna1996tracking, lerdsudwichai2003algorithm, crowley1997multi} for lip-reading. However, to date, there is no one accepted method as the de facto method for extracting lip-reading features. In lieu of this, in \cite{zhou2014review}, Zhou \textit{et al.} ask two questions about feature extraction, specifically for lip reading: primarily, how to cope with the speaker identity dependency in visual data? But also, how to incorporate the temporal information of visual speech? The intent of this second question is for capturing co-articulation effects into features. Zhou \textit{et al.} categorise a comprehensive range of feature extraction techniques into four groups: Image-based e.g. \cite{gowdy2004dbn}, Motion-based e.g. \cite{mase1991automatic}, Geometric-feature-based e.g. \cite{nefian2002dynamic}, or, Model-based e.g. \cite{eveno2004accurate}.
 
This categorisation serves to show the breadth of current research into features. However, this attention on feature extraction does not address the only challenges in machine lip-reading. Improvements can still be made in the classification stage of lip-reading also. Therefore much of this thesis is focused on classification, rather than additional tasks such as tracking and feature extraction. That is not to say we are dismissive of the feature extraction and tracking requirements, rather that we wish focus our work to improve the classification methods. 
 
\begin{figure}[h!] 
\includegraphics[width=0.9\linewidth]{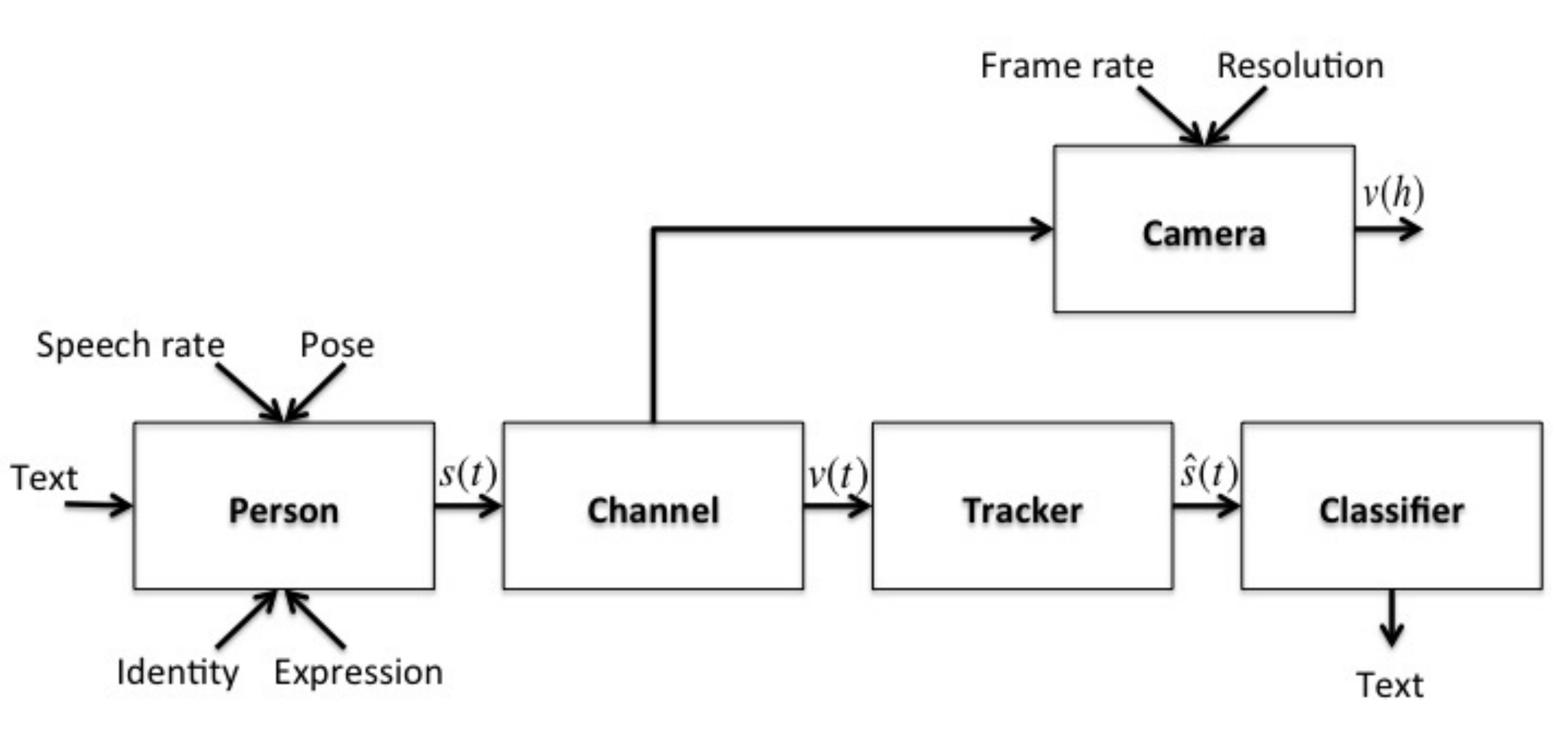} 
\caption{Sources of variability in computer lip-reading: affects on automatic lip-reading systems} 
\label{fig:variability} 
\end{figure} 
 
Figure~\ref{fig:variability} shows the situation in which we are trying to recreate the text in the mind of the speaker. Each speaker articulates differently, and so the identity of the individual speaker is a significant affect on the efficacy of lip-reading. The visual signal is also affected by the speaker's pose, motion and expression. Cameras typically have many parameters that might affect lip-reading. Of these, we mention frame rate and resolution as highly probable to be significant. 
 
 \begin{table}[!ht]
 \centering
 \caption{A list of affects on automatic lip-reading systems}
 \begin{tabular}{|l|l|l|r|}
 \hline
 Evaluation & Previously studied, in & Likely sensitivity \\
 \hline \hline
 Motion & Yes, \cite{ong2008robust, Matthews_Baker_2004} & Low  \\
 Pose & Yes, \cite{6298439}& Medium  \\
 Expression & Yes, \cite{Moore2011541} & Low \\
 Frame rate & Yes, \cite{Blokland199897, saitoh2010study}& Low  \\
 Resolution & No & Unknown  \\
 Colour & Yes, \cite{kaucic1998accurate} & Low  \\
 Classifier unit choice & Yes, \cite{cappelletta2012phoneme}& High  \\
 Feature type & Yes, \cite{matthews1998nonlinear, lan2009comparing}& High  \\
 Classifier technology & Yes, \cite{982900, young2006htk}& Medium  \\
 Multiple persons & Yes, \cite{871067} & Medium  \\
 Speaker identity & Yes, \cite{607030} & High  \\
 Rate of speech & Yes, \cite{6854158} & High \\
 \hline
 \end{tabular}
 \label{tab:mlraffects}
 \end{table}
 
In Table~\ref{tab:mlraffects} we have listed and assessed a number of environmental affects on machine lip-reading. There are a number of factors that can be difficult to control in machine lip-reading. These include, but are not limited to, lighting, identity, motion, emotion, and expression. Table~\ref{tab:mlraffects} is an attempt at a systematic study of the affects. Considering initially the problem of speaker-dependent lip-reading, then three factors are of immediate interest: resolution because it does not appear to have been studied systematically, and unit choice, and feature type because they are likely to be highly significant to performance. For the time-being, speaker identity and rate of speech can be ignored since they are constant for a given speaker.

The choice of feature has been studied quite well and there have been a number of `contests' between feature types (e.g. \cite{lan2009comparing,cappelletta2012phoneme}) which have led to the conclusion that state of the art Active Appearance Models (AAMs) are highly likely to give the best known performance. These are the features we use and the subject of the next chapter. However the choice of visual unit, the analogous quantity to a phoneme is more intriguing. 

A phoneme is the smallest sound which can be uttered \cite{international1999handbook}. A viseme is not so precisely defined \cite{chen1998audio,fisher1968confusions, Hazen1027972}. However, a working definition is that a viseme is a set of phonemes that have identical appearance on the lips. Therefore many phonemes fall into one viseme class: a many-to-one mapping. There are alternative definitions of visemes in which the viseme is, for example, seen as a repeatable, visual gesture. In \cite{7074217} two alternative definitions are explored: visemes based upon articulatory gestures or on similar visual appearance. The tentative conclusion is that visemes based upon the articulatory gestures definition perform better. This study only looks at recognition, in synthesis studies, visemes are considered as `temporal units that describe distinctive speech movements of the visual speech articulators' \cite{taylor2012dynamic}. As there are many definitions to choose from, we continue with the recognition working definition of `a viseme is a group of phonemes with identical appearance on the lips'. Thus, our study starts with two key problems: resolution which has not been systematically studied before in isolation from observing the effects of noise, and unit selection because it is likely to be highly significant. But, before we can study these items, it is necessary to discuss the third affect to which classification is highly sensitive: feature selection. 

Thus, our research question is; `can we augment or replace the current lip-reading classifiers to improve machine lip-reading?'
 %  Introduction, includes application & motivations, describes 3 stages of recognition (track, extract and recognise) - add in Aims & Objectives

%!TEX root = main.tex 
\chapter{Features and classification methods}\label{chap:featuretypes} 
In the previous chapter it was asserted that feature choice was likely to be highly significant. In this chapter therefore, we examine the full processing chain in more detail from tracking to classification, dwelling on the methods of special relevance to this thesis. 
 
\section{Linear predictors} 
 
Linear Predictors (LPs) are a person-specific and data-driven facial tracking method. Devised primarily for observing visual changes in the face during speech, these make it possible to cope with facial feature configurations not present in the training data by treating each feature independently. For speech, this means isolating the lips from the eyes, outline of the face, etc. 
 
The linear predictor itself is a part of the tracking mechanism. It is the central point around which support pixels are used to identify the change in position of the central point over time. The central point is visually seen as a landmark on the outline of a feature. A set of these landmarks represent the changing shape of something (in our case lips) morphing over time. In this method both the shape (comprised of landmarks) and the pixel information surrounding the linear predictor position are intrinsically linked. 
 
A single LP alone is not enough to provide robust and accurate tracking, so \cite{ong2008robust} explains how rigid flocks (a small group) of selected LPs are grouped around a central feature (not the linear predictor central point, but as an example, the feature mean position) restrict the motion of the LPs within a boundary and reduce their susceptibility to noise. These LPs have been successfully used to track objects in motion \cite{matas2006learning}. 
 
Further improvements to the LPs selection method are described in \cite{ong2008robust, ong2011robust}, both of which show improvement of over original LP tracking accuracy. 
 
An interactive LP tracking tool has been made at the University of Surrey. Its benefits are the real-time tracking and autonomous use, but a limitation of this tool is when a face is partially off-screen, the real-time tracking requires the user to guess in real time where the appropriate LP should be. This is not a simple task to perform with any accuracy or consistency and when tested with our Rosetta Raven dataset (see Chapter~\ref{chap:datasets_dics}) we found the AAM features still outperformed the LP features. 
 
\section{Active shape and appearance models} 
\label{sec:aams}
An Active Appearance model (AAM) \cite{927467} is a combined shape and appearance trained model used in tracking a face throughout a video sequence. The model is constructed from a small training subset (Table \ref{table:frames}) and is a type of Point Distribution Model (PDM) used to represent the shape of a face and how it varies during speech. The shape $s$ of an AAM is the coordinates of the $v$ vertices which make up a mesh,

\begin{equation}
s = (x_1,y_1, x_2,y_2, ..., x_v,y_v)^T
\label{eq:meshvertices}
\end{equation}

\begin{figure}[h] 
\centering 
\includegraphics[width=0.2\textwidth]{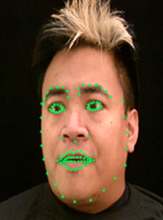} 
\caption{Example Active Appearance Model shape mesh.} 
\label{fig:egshape} 
\end{figure}

Training creates a mean model permitting deviations within a predetermined range of variance. Any of the training co-ordinate vectors used for model creation with 30\% or more of occluded landmarks are omitted from the mean shape formation. Normalised meshes are built from the manually trained data (landmarks) for translation, scale and rotation (i.e. movement between the image frames).We now have a vector of $2n$ values for $n$ landmarks upon the face. Principal Component Analysis (PCA) provides us with eigenvectors so an independent shape model becomes a set of meshes, 
 
\begin{equation} 
s = s_0 + \sum_{i=1}^np_is_i 
\label{equation:Shape} 
\end{equation} 
 
where $s_0$ is the mean shape, $p_i$ are coefficient shape parameters, and $s_i$ are the eigenvectors of the covariance matrix of the $n$ largest eigenvalues. We can assume $s_i$ is orthonormal because we can always perform a linear reparameterisation \cite{Matthews_Baker_2004}. The landmarks are chosen to model the sub-shapes within the face such as: the outline of the hairline and jaw, eyes, nose or lips. We have to hand label these training images. The meshes constructed with our hand labelling are normalised by Procrustes analysis \cite{procrustes} before we apply PCA. An example of a full face shape model is shown in Figure~\ref{fig:egshape}. In this Figure there are 104 landmarks, the majority (44) of which are modelling the inner and outer lip contours. 
 
An independent appearance AAM  uses appearance data over the base mesh, $S_0$. This allows linear variation in the shape whilst maintaining a compact model. $S_0$ also denotes the set of pixels that lie inside the base mesh. Thus $A(x)$ (or AMM appearance) is an image defined over the pixels $x \in S_0$.  This means pixels are mapped into the triangles of the shape model by Procrustes analysis \cite{procrustes} over the shape model vector (the aligned the set of points) to build the statistical model. Each training image is warped to match the mean shape to identify a shape-free area of the training image. This shape-free area is normalised with a linear transform before the texture model is built by eigen-analysis \cite{927467}. 
 
\begin{equation} 
A(x) = A_0(x) + \sum_{i=1}^m\lambda_iA_i(x) \quad \forall x \in s_0 
\label{equation:app} 
\end{equation} 
 
In Equation~\ref{equation:app} the coefficients $\lambda_i$ are the appearance parameters, $A_0$ is the base appearance, and $A_i(x)$ are the appearance image eigenvectors of the covariance matrix. Our appearance $A(x)$ is $A_0$ plus a combination of images $A_i(x)$. $A_0$ is the mean image, and $A_i$ are the $m$ eigenimages with the $m$ largest eigenvalues. 
 
It has been demonstrated that the combination of appearance and shape models significantly improves lip-reading performance \cite{982900,927467} and we use these in the work presented here unless explicitly stated otherwise. The combination of these model types requires a single parameter set to represent the relationship between shape and appearance. In independent shape and appearance AAMs \cite{Matthews_Baker_2004}, the shape parameters, $p$, and appearance parameters $\lambda$, are distinct. In a combined model, we use one set of parameters, $C = (c_1, c_2, c_3, ..., c_n)^T$. This is shown in Equation~\ref{eq:shapecombined} and Equation~\ref{eq:appcombined}. This usage of a common parameter set, $c_i$, intrinsically ties the models together by warping the image over the shape model to represent both the appearance and shape variation in a face. 
 
For shape 
\begin{equation} 
s = s_0 + \sum_{i=1}^nc_is_i 
\label{eq:shapecombined} 
\end{equation} 
and for appearance 
\begin{equation} 
A(x) = A_0(x) + \sum_{i=1}^n{c}_iA_i(x) 
\label{eq:appcombined} 
\end{equation} 
 
A combined AAM requires a third application of PCA on the weighted shape, $p$ and appearance, $\lambda$ parameters. The correlation between the shape and texture (appearance) model is learned and integrated into the combined model. 

To initialise the AAM we use the shape parameters $p=(p_1, p_2, ..., p_n)^T$ in Equation~\ref{fig:egshape} to generate the shape $s$, and the appearance parameters $\lambda = (\lambda_1, \lambda_2, ..., \lambda_m)^T$ to generate the appearance $A(x)$ in $s_0$.  This AAM instance is built by a piecewise affine warp of $A$ from the base mesh $s_0$ to the AAM shape $s$. 

Finally we fit the AAM using the Inverse Compositional algorithm \cite{inversecompAlg} to all frames in the video sequence \cite{Matthews_Baker_2004}. This algorithm uses the coordinate frame of the image $I$ and the coordinate frame of the AAM. To initiate the fit with the best starting position, the first image frame in a video sequence receives a manually labelled shape, $s$. Iterating through each frame of the video in turn, a backwards warp $W$ is used to warp each image $I$ onto the base mesh $s_0$  until the landmark positions converge into place to match corresponding pixels between frames. The more movement there is between frames, or the lower the frame rate, tracking is more difficult as these create greater variation between frame images. 
 
\section{Discrete cosine transforms} 

The Discrete Cosine Transform (DCT) \cite{ahmed1974discrete} is a technique for converting a signal into elementary frequency components, or in other words, it transforms an image from a spatial to frequency domain by separating an image into parts of unequal importance. There are many variants of DCT and in lip-reading and AVSR authors use 2D-DCT (Equation~\ref{eq:dct}) as it is applied too each two-dimensional frame image throughout a video. For example in \cite{improveVis, cappelletta2012phoneme} and \cite{milner2015analysing}. To create 2D-DCT features co-efficient vectors are extracted from the information from the region of interest in an image, for machine lip-reading, this is the lips. 

\begin{equation*}
q_{u,v} = W_u W_v\sum_{i=0}^{N-1} \sum_{j=0}^{M-1} {p}_{i,j} \cos \left( \frac{u\pi(2i+1)}{2N} \right) 
\cos \left( \frac{v\pi(2j+1)}{2M} \right) \\
\end{equation*}
\begin{equation} 
W_u = \left\{ 
  \begin{array}{l l}
    \sqrt{\nicefrac{1}{N}} &  \text{if $u=0$}\\
    \sqrt{\nicefrac{2}{N}} &  \text{otherwise}
  \end{array} \right. \\
W_v = \left\{ 
  \begin{array}{l l}
    \sqrt{\nicefrac{1}{M}} & \text{if $v=0$}\\
    \sqrt{\nicefrac{2}{M}} & \text{otherwise}
  \end{array} \right.
  \label{eq:dct}
\end{equation}

In Equation~\ref{eq:dct} we show that 2D-DCT is pixel-based, features are extracted from a region of interest matrix of size $M$ by $N$, where $P$ is the mouth centre. $p_{ij}$ is pixel intensity in row $i$ and column $j$. This creates $q_{uv}$.

\section{Comparison of available feature types} \label{sec:featuretypes} 
 
Lan \textit{et al.} present in \cite{improveVis} a comparison of different features first presented in \cite{927467}. Revisited in \cite{Matthews_Baker_2004}, AAM features are produced as either model-based (using shape information) or pixel-based (using appearance information). In \cite{improveVis} Lan \textit{et al.} observed that state of the art AAM features with appearance parameters outperform other feature types like sieve features, 2D DCT, and eigen-lip features, suggesting appearance is more informative than shape. Also pixel methods benefit from image normalisation to remove shape and affine variation from region of interest (in this example, the mouth and lips). The method in \cite{improveVis} classified words with the RMAV dataset but recommended in future creating classifiers with viseme labels for lipreading, and advises that most information is from the inner of the mouth.

A comparison of two current key methods for fitting and extraction of facial features for computer lip-reading is summarised in Table~\ref{tab:featComp}. 
 
\begin{table}[h] 
\caption{A summary of shape and appearance models and linear predictors.} 
\centering 
\begin{tabular}{| p {6.5cm} | p{6.5cm} |} 
\hline 
Linear Predictors (LP) & Shape Appearance Model (SAM) \\ 
\hline \hline 
Data driven. & Face knowledge required from training for modelling. \\ 
\hdashline
Unsupervised. & Supervised. \\ 
\hdashline
Feature independent. & Feature dependence improves tracking. \\ 
\hdashline
Use only intensity information ie. grey scale images. & The fitted model can be either solely shape model, an appearance model (pixel information) or a combined model of shape and appearance where each pixel is related to a triangular section of the shape model. \\ 
\hdashline
Prior training shape models or temporal models for dynamics are not required or used. & An active appearance model is built from training data to fit new images. \\ 
\hdashline
Can cope with feature configurations not present in training data. & Training needs to encapsulate all variance in the video to be tracked.\\ 
\hdashline
Multiple LPs are grouped into flocks for robustness. & Primary landmarks are used for the important positions in training data.  \\ 
\hline 
\end{tabular} 
\label{tab:featComp} 
\end{table}

For the work presented in this thesis, we chose to use AAMs. This is because whilst DCT features can outperform geometric features (as shown in \cite{heckmann2002dct}), a state of the art AAM can outperform DCT features. In \cite{neti2000audio} the results suggest that DCT features outperform AAMs because they complete most experiments with them after initial AAMs performed poorly, (65.9\% $w.e.r$ for AAMs compared to 61.80\% $w.e.r$ with DCT features). However, the authors also note that their AAMs were not good ones and the reasons for this could be attributed to either; modeling or tracking errors. This is because insufficient training data can have two effects. First, that the AAM is not generalised enough from the training data to classify the test data, and secondly, an undertrained AAM will not fit well when tracking a face. It should be noted that in comparing DCT and AAM features, Neti \textit{et al.} use different regions of interest for the feature types. For the DCT features, the ROI is the mouth, compared to the whole face for the AAMs \cite{neti2000audio}. 

In the work presented in Chapters~\ref{chap:five} to~\ref{chap:eight}, particularly for continuous speech experiments with newer datasets, we have confidence that our AAMs are state of the art, have tracked well between all frames (this is confirmed by producing a jpg image of each frame with the AAM landmarks plotted on and the fit is manually checked) and is achieved by using a higher number of landmarks, we use 104 \cite{bear2014resolution} rather than the 68 in \cite{neti2000audio}). 

\section{Hidden Markov models} 
\label{sec:hmm}

Hidden Markov Models (HMMs) have been used in speech classification for some time for acoustic, audio-visual and visual-only classification. Both channels of speech can be considered as a time series, i.e. they will produce data points in a causal manner. Other domains which have applied HMMs are sets of temporal data such as handwriting, DNA sequences and energy consumption. 

A HMM has two stochastic processes: the first process is based around state transition probabilities, and the second, is based upon state emission probabilities.

A Markov model (also known as a Markov chain), is made of a number of states connected to all other states. Each connection has transition probabilities for moving between the states it is connected to. In a $n^{th}$-order Markov chain, an inherent assumption is that state transitions are dependent upon the $n$ previous states. In a Markov chain the stochastic process output is the sequence of states. Practically, in speech classification, a first order model is normally used. In a first order HMM the state transitions are dependent only on the current state. The probabilities of all possible actions (transitions) at time, $t$, are dependent upon the state the HMM is in at time $t$, not the value of $t$.

The second stochastic process is concerned with emission probabilities. Each HMM state has an associated Probability Density Function (PDF). A PDF used on feature vectors determines the emission probabilities of any particular feature vector being output (emitted) by the state, when the HMM is in that state. Whereas in a Markov chain the output is the sequence of states, in an HMM the PDF means the output is a feature vector. Because the emission probabilities are a function of the state, the knowledge of the state is hidden from the observer \cite{holmes2001speech}. 

In a network of HMMs, each HMM is labelled by its representative unit. In visual speech, these units are referred to as visemes, in acoustic speech phoneme labels are used. In some simple speech classification tasks, or with limited datasets, words may be used as the HMM unit label. Additional HMMs can also be built to model the silence at the start and end of utterances and the shorter silence pauses between words. In the work presented in this thesis, all HMMs are monophones. 

\subsection{HTK: an HMM toolkit} 
\label{sec:htk}
 
HTK provides a set of tools which enable users to build speech processing tools, including recognisers and estimators. The main algorithm used in HMM estimation is the Baum-Welch algorithm \cite{baum1970maximization}, and the algorithm used in classification is the Viterbi algorithm \cite{viterbi1967error}. The HTK book \cite{123564} details the background of HTK in full, up to its current version for full information of its implementation and use. 
 
The use of HTK is commonplace in acoustic speech classification \cite{almajai2006analysis, potamianos2004audio, howellPhD, matthews1998nonlinear} and current lip-reading literature \cite{6298439, lan2009comparing, 871067, improveVis, comparHumMacLipRead}. So using HTK for machine lip-reading allows very easy replication of our results. HTK has achieved ubiquity due to its generally high performance, so we can be confident that our results will be close to the best achievable performance when we adopt similar strategies as described in previous works.
  
In HTK recognition. performance of the HMMs can be measured by both correctness, $C$, and accuracy, $A$, 
\begin{equation} 
C = \displaystyle \frac{N-D-S}{N},\quad  
\label{eq:correctness}
\end{equation}
\begin{equation}
A = \displaystyle \frac{N-D-S-I}{N} \quad
\label{eq:accuracy}
\end{equation}
%therefore; 
%\begin{equation}
%C = \displaystyle \frac{-I}{N}
%\label{eq:corr_short} 
%\end{equation}
where $S$ is the number of substitution errors, $D$ is the number of deletion errors, $I$ is the number of insertion errors and $N$ the total number of labels in the reference transcriptions~\cite{young2006htk34}. 

We can explain these types of errors with an example. Suppose we have a ground truth utterance, ``John wanted to visit the shop to buy groceries". Our classifiers can produce different outputs. Possible output 1: `` John wanted visit the to groceries" has three words missing. `to', `shop', and `buy'. In this instance, these are deletion errors. In another possible output: ``John wanted to visit visit the shop to buy groceries'', the word `visit' is included twice. This is an insertion error. Finally, if we achieved a classifier output of ``John wanted to shop the shop to buy groceries". The word `shop' has been identified where the word `visit' should be. This is a substitution error.

Common tools used for a classification task in HTK are: \texttt{HCompV}, \texttt{HERest}, \texttt{HHed}, \texttt{HVite} and \texttt{HResults}.

\texttt{HCompV} - used to flat start each HMM subject to a prototype file determining number of states and mixtures. It does this based upon the data within the whole dataset so all states are equal. It uses a prototype HMM definition, some training data and initialises each new HMM where every local HMM mean is the same as the global mean across the whole set. Only the covariances are updated.

\texttt{HERest} - is the Balm-Welsh re-estimation of each HMM using the training fold samples and a transcription using the HMM labels. HERest uses embedded training to simultaneously updated all HMMs within a systems using all training data available within a fold. This is particularly important for systems where the HMM labels are sub word models as HERest ignores boundary information in transcripts of training samples. 

\texttt{HHEd} - permits the tying together of states within an HMM model to allow fast transitions between states and shorter Markov chains. This is particularly useful for similar or short models such as silence (at the start and end of utterances) and short pauses between words. 

\texttt{HVite} - is commonly used for both forced alignment of HMMs using the ground truth transcription, and also for the crucial classification task. Using the trained HMMs, \texttt{HVite} attempts to recognise test samples and produces a classification output.

\texttt{HResults} - compares the classification output to the ground truth, HResults provides statistics about how accurate the HMM recognisers have been, primarily correctness (Equation~\ref{eq:correctness}) at both the unit and network level, and also includes model-level accuracy (Equation~\ref{eq:accuracy}). 
 
%\section{Deep neural networks} 
% 
%Examples of Deep Neural Network (DNN) solutions include; Kaldi \cite{povey2011kaldi} and Torch \cite{collobert2002torch}. Very recent work using DNNs in visual speech has shown that their performance can be higher than HMMs \cite{thangthaiimproving, le2015voicing} particularly in audio-visual classification. However, the disadvantage of using a DNN for classification is that we do not know what it is they learn, for example, which properties or attributes of training data which enable them to outperform, say HMM classifiers. As HMM's are less of a black box and enable us to glean more scientific observations from the training and testing phases of classification, we continue our work with HMM recognisers. Whilst we do not use DNNs they are mentioned here to confirm why they have not been selected in our work. 
 % Technical background: Tracking & Extraction methods (LPs, AAMs, SIFT, DCT), Recognition methods: HMMS, DNN's, TSC's, Make sure to compare all
%!TEX root = main.tex 
\chapter{Datasets}\label{chap:datasets_dics} 
This chapter summarises the datasets used in the work presented throughout this thesis. Note that while this thesis is about machine lip-reading (visual speech recognition), audio-visual datasets are commonplace since researchers often wish to compare visual-only performance to audio and audio-visual performance for the purposes of audio-visual integration such as in \cite{neti2000audio}. A summary of the most common AVSR databases is presented in Table~\ref{tab:datasetsummary}. The result values listed are those from the original presented papers referenced in column 1. The results vary based upon the specific experiments, content, classification units (e.g. words, visemes, or phonemes), and original intent of each dataset.  Other databases are available, such as those in \cite{1394602, 6920467, doi:10.1142/S1469026811003045} but these are non-English (Mandarin, Arabic and French respectively) and therefore not considered here. 
 
 \begin{table}[h]
 \centering
 	\caption{Common databases available for machine lip-reading research.}
	\resizebox{\columnwidth}{!}{% 
 	\begin{tabular}{| c | c | c | c |}
	\hline
	Name & Speakers & Content & Results \\
	\hline \hline
	AVLetters \cite{982900} & 10 & Alphabet letters &  $<27\%$\\
	AVLetters2 \cite{cox2008challenge} & 5 & High definition alphabet letters & $80\%><90\%$ \\
	AV-TIMIT \cite{Hazen1027972} & 223 & TIMIT sentences & $35\%$ p.e.r \\
	CUAVE \cite{patterson2002cuave} & 36 & Digits & $87\%$ Acc \\
	GRID \cite{cooke2006audio} & 36 & Command sentences & $<1.85\%$ w.e.r \\
	IBM LVCSR (ViaVoice) \cite{matthews2001comparison} & 290 & Continuous speech & $58\%$ w.e.r\\
	OuluVS \cite{5208233} & 20 & 10 everyday phrases & 70\% Acc \\
	RMAV (LILIR) \cite{improveVis} & 20 & Context dependent sentences &  $20\%><60\%$\\
	Rosetta Raven \cite{bear2014resolution} & 2 & E. A. Poe's The Raven & $20\%><60\%$\\
	TCD-TIMIT \cite{7050271} & 62  & 98 sentences & $>55\%$ Acc \\
	\hline
	\end{tabular} }%
	\label{tab:datasetsummary}
 \end{table}
 
For the work presented in this thesis, the Rosetta Raven database was selected for the resolution robustness experiment in Chapter~\ref{chap:five} because it is both continuous and structured speech. This means that there is a good quantity of data but also that the speech itself is constrained meaning that the task is simpler than that of say AV-TIMIT, this is better for a controlled experiment to measure the affects of a single parameter. Note that AusTalk, AV-TIMIT and IBM LVCSR are proprietary and thus not freely available. 

We have confidence that the larger (in regards to number of speakers) continuous speech datasets have a good phoneme coverage and so, subject to the viseme mapping selected, will also have good viseme coverage, however the smaller datasets, including those with limited vocabularies, the quantity of visemes (and the consequential volume of training samples per viseme class) will be at risk of inter-class skew. Therefore preliminary experiments in later chapters were undertaken first with AVLetters2 for proof of concept and confirmation that hypotheses were sound, before repeating experiments with RMAV. RMAV has sentences selected from the resource management data \cite{fisher1986darpa} which ensures a good phoneme coverage in its content.  RMAV was selected as extracted features were available which enabled focusing on the classification task rather than that of tracking and extracting features. 
 
\section{Pronunciation dictionaries}
\label{sec:dict}
To accommodate the breadth of possible pronunciations, a number of dictionaries are available for use in machine lip-reading. These dictionaries map words to phoneme sequences subject to the pronunciation habits of the speaker. Two are described here: firstly, CMU \cite{cmudict}, has been used in conjunction with the Rosetta Raven data, and secondly, BEEP \cite{beepdict}, is used in later chapters with AVLetters2 and RMAV. 
 
The Carnegie Mellon University North American Pronunciation Dictionary \cite{cmudict}, known as CMU, uses 39 phonemes and also encodes whether vowels carry levels of lexical stress \cite{hieronymus1992use} of either 0-None, 1-Primary or 2-Secondary. Lexical stress is the relative emphasis placed upon certain syllables within a word. Including lexical stress representations, this dictionary has 57 phonemes. Containing over 125,000 words, it is based on the ARPAbet symbol set (which relates to the standard IPA symbol set) developed for speech recognition uses. This dictionary is used for American speakers speaking English i.e. American English. %/r/ /iy0/ /d/ /r/ /eh1/ /d/
  
The Cambridge University British English Pronunciation dictionary, known as BEEP, \cite{beepdict} has 49 phonemes mapped to over 250,000 words allowing for duplicate pronunciations of the same word. For example, the word `read' phonetically can be, `/r/ /eh/ /d/' as in `I read my book last night' or, `/r/ /\textsci/ /d/' as in `I like to read'. This dictionary is used for British speakers of English.

\section{AVLetters2 - an isolated word dataset} 
\label{sec:avl2}
\begin{figure}[!ht] 
\centering 
\setlength{\tabcolsep}{1pt} 
\begin{tabular}{c c c c} 
\includegraphics[width=0.22\textwidth]{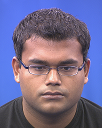}  & 
\includegraphics[width=0.22\textwidth]{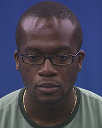} & 
\includegraphics[width=0.22\textwidth]{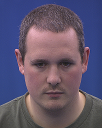}  & 
\includegraphics[width=0.22\textwidth]{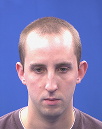}  \\ 
(a) Speaker 1 & (b) Speaker 2 & (c) Speaker 3 & (d) Speaker 4 \\ 
\end{tabular} 
\caption{Example faces from the AVLetters2 videos (four speakers).} 
\label{fig:AVL2egs} 
\end{figure}

AVLetters 2 (AVL2) \cite{cox2008challenge} is an HD version of the AVLetters dataset \cite{matthews1998nonlinear}. It is a single word dataset of four British English speakers (all male) each reciting the 26 letters of the alphabet seven times. We can not present the quantity of visemes in the data set at this stage as it is dependent upon the viseme set being used (see Section~\ref{chap:maps}). The speakers in this dataset can be seen in Figure~\ref{fig:AVL2egs}. AVL2 has 28 videos of between $1,169$ and $1,499$ frames between $47s$ and $58s$ in duration. As the dataset provides isolated words of single letters, it lends itself to controlled experiments without needing to address matters such as co-articulation. 

\begin{figure}[h] 
\centering 
\includegraphics[width=0.95\textwidth]{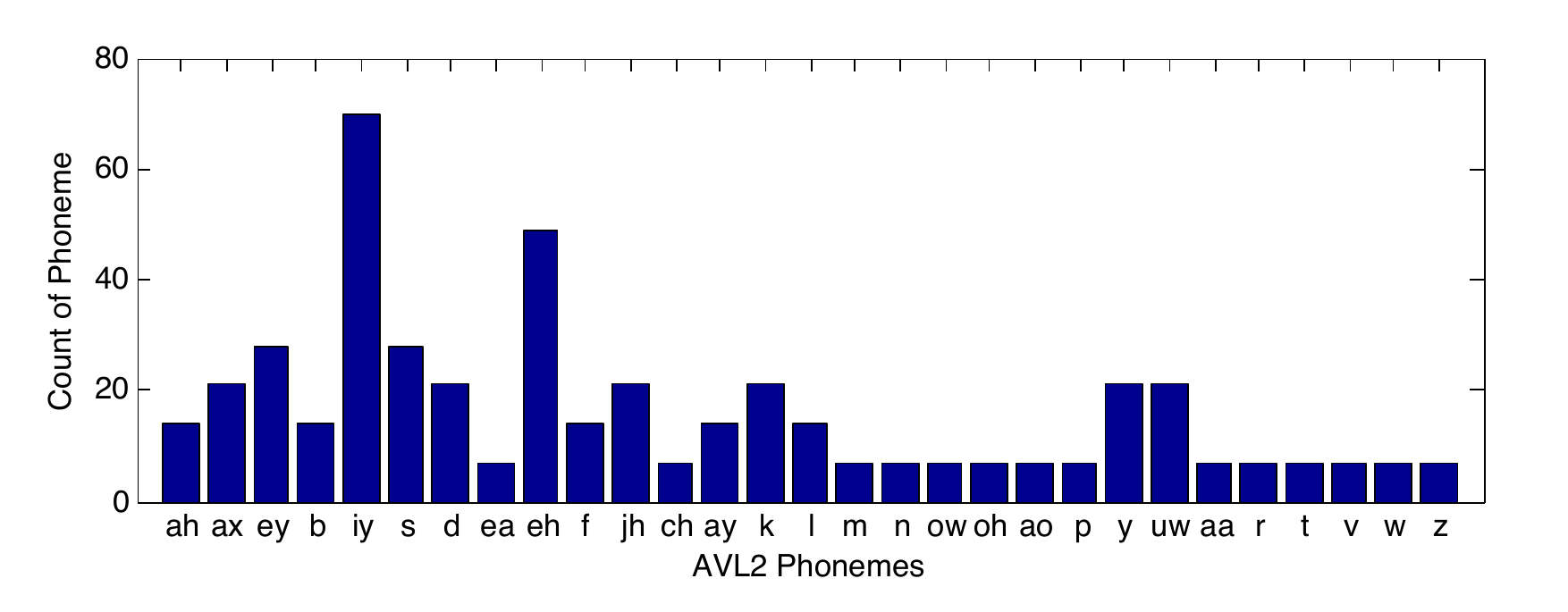} 
\caption{Occurrence frequency of phonemes in the AVLetters2 dataset.} 
\label{fig:avl2histogram} 
\end{figure}

There are 30 unique British English phonemes in AVL2, the occurrence frequency of these is shown in Figure~\ref{fig:avl2histogram}. Therefore, the data set is missing 19 phonemes found in spoken British English.

Table~\ref{tab:feature_parameters} describes the features extracted from the AVL2 videos. These features have been derived after tracking a full-face Active Appearance Model throughout the video before extracting features containing only the lip area. Therefore, they contain information representing only the speaker's lips and none of the rest of the face. Speakers 2, 3 and 4 are similar in number of parameters contained in the features. The combined features are the concatenation of the shape and appearance features \cite{Matthews_Baker_2004}. All features retain 95\% variance of facial shape and appearance information. 

 \begin{table}[!h] 
\centering
\caption{The number of parameters in shape, appearance and combined shape \& appearance AAM features for each speaker in the AVLetters2 dataset for each speaker. Features retain 95\% variance of facial information.}  
\begin{tabular}{| l | r | r | r |} 
\hline 
Speaker	& Shape & Appearance & Combined \\ 
\hline \hline 
S1	& 11 & 27	& 38 \\ 
S2 	& 9 & 19 & 28 \\ 
S3 	& 9 & 17 & 25 \\ 
S4	& 9 & 17 & 25 \\ 
\hline 
\end{tabular} 
\label{tab:feature_parameters} 
\end{table}

%\begin{table}[h] 
%\centering 
%\begin{tabular}{| l | r | r |} 
%\hline 
%Video& Frames & Duration\\ 
%\hline \hline 
%Speaker1\_v1 & 1,385 & 00:00:56 \\ 
%Speaker1\_v2 & 1,325 & 00:00:53 \\ 
%Speaker1\_v3 & 1,245 & 00:00:50 \\ 
%Speaker1\_v4 &  1,310 & 00:00:53 \\ 
%Speaker1\_v5 & 1,217 & 00:00:49 \\ 
%Speaker1\_v6 & 1,236 & 00:00:50 \\ 
%Speaker1\_v7 & 1,233 & 00:00:50 \\ 
%Speaker2\_v1 & 1,416 & 00:00:57 \\ 
%Speaker2\_v2 & 1,428 & 00:00:58 \\ 
%Speaker2\_v3 & 1,388 & 00:00:56 \\ 
%Speaker2\_v4 & 1,428 & 00:00:58 \\ 
%Speaker2\_v5 & 1,326 & 00:00:54 \\ 
%Speaker2\_v6 & 1,470 & 00:00:59 \\ 
%Speaker2\_v7 & 1,387 & 00:00:56 \\ 
%Speaker3\_v1 & 1,211 & 00:00:49 \\ 
%Speaker3\_v2 & 1,214 & 00:00:49 \\ 
%Speaker3\_v3 & 1,185 & 00:00:48 \\ 
%Speaker3\_v4 & 1,197 & 00:00:48 \\ 
%Speaker3\_v5 & 1,171 & 00:00:47 \\ 
%Speaker3\_v6 & 1,169 & 00:00:47 \\ 
%Speaker3\_v7 & 1,176 & 00:00:48 \\ 
%Speaker4\_v1 & 1,412 & 00:00:57 \\ 
%Speaker4\_v2 & 1,455 & 00:00:59 \\ 
%Speaker4\_v3 & 1,413 & 00:00:57 \\ 
%Speaker4\_v4 & 1,357 & 00:00:55 \\ 
%Speaker4\_v5 & 1,351 & 00:06:55 \\ 
%Speaker4\_v6 & 1,371 & 00:00:55 \\ 
%Speaker4\_v7 & 1,499 & 00:00:58 \\ 
%\hline 
%\end{tabular} 
%\caption{Summary of each video in the AVL2 dataset} 
%\label{table:frames} 
%\end{table} 
This dataset is used for comparing visemes, testing new speaker-dependent visemes (Chapter~\ref{chap:maps}) and for evaluating the robustness of speaker-dependent phoneme-to-viseme maps in Chapter~\ref{chap:seven}.

%\clearpage 
\section {Rosetta Raven - a stylised continuous speech dataset} 
\label{sec:rr}
\begin{figure}[!h] 
\centering 
\setlength{\tabcolsep}{1pt} 
\begin{tabular}{c c} 
\includegraphics[width=0.4\textwidth]{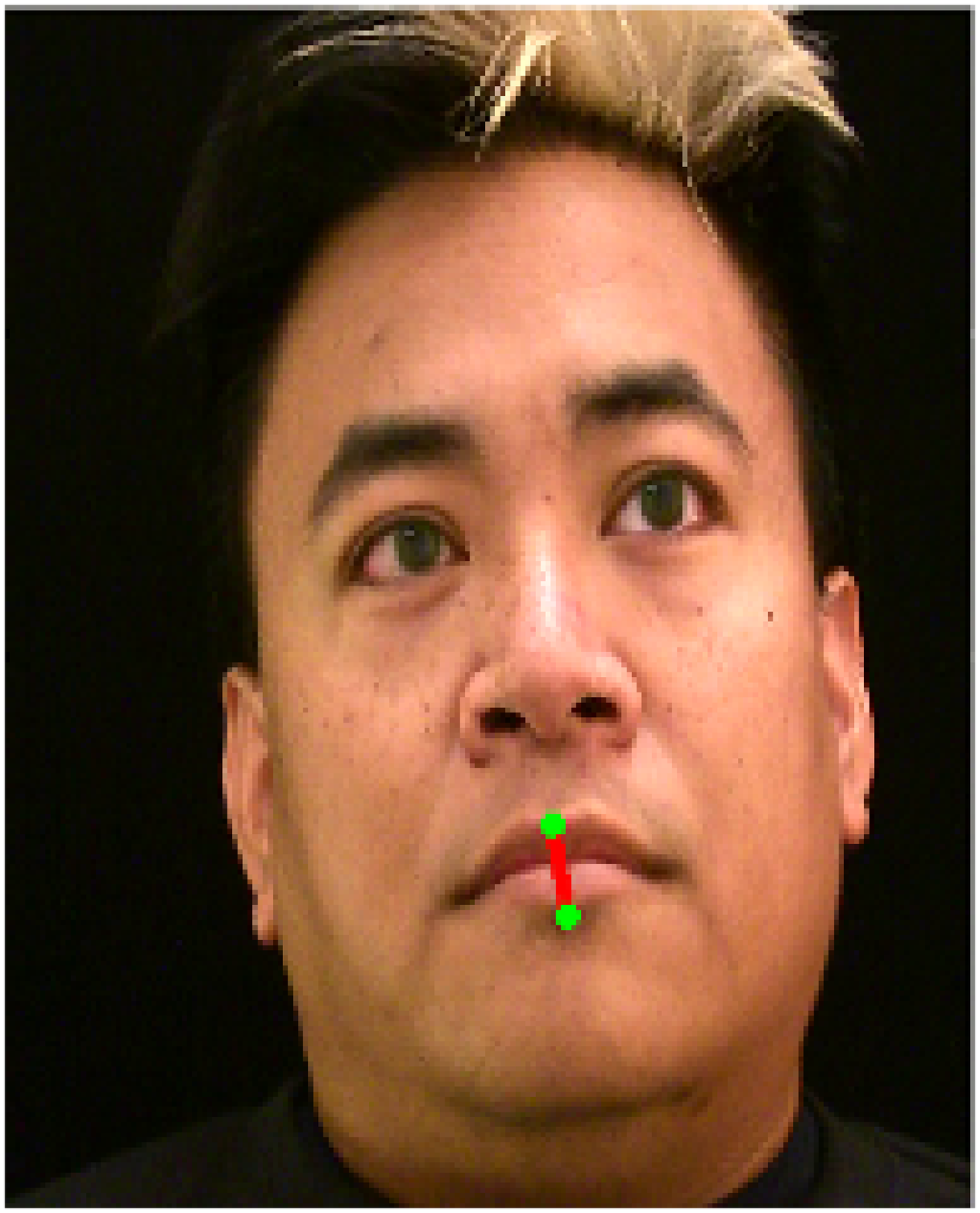} 
\label{fig:bud} & 
\includegraphics[width=0.4\textwidth]{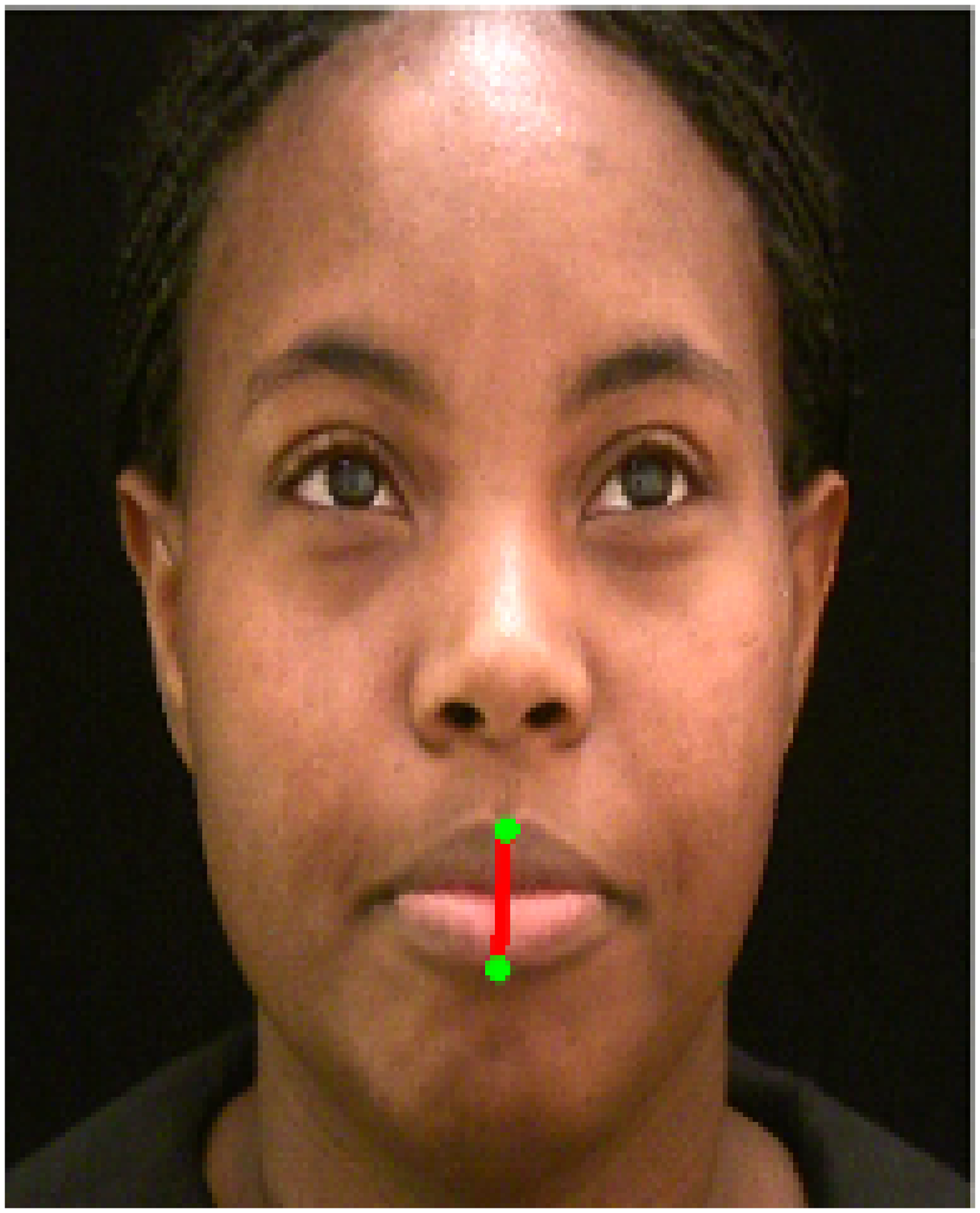} 
\label{fig:dumebi} \\ 
(a) Speaker 1 & (b) Speaker 2 \\ 
\end{tabular}
\caption{Example faces from the Rosetta Raven videos (two speakers).} 
\label{fig:exampleRR}
\end{figure}

This dataset was recorded at UCLA in January 2012 by Dr Eamon Keogh and was formulated as an attempt to provide a standardised audio-visual machine learning problem \cite{bear2014resolution}. It comprises four videos which consist of two North American untrained speakers (one male, one female, seen in Figure~\ref{fig:exampleRR}) each reciting E.A.Poe's `The Raven'. The poem was published in 1845 and the linguistic content of the Raven make this an interesting dataset as the narrative uses a stylised language including internal rhyme and alliteration. The poem is described as being generally trochaic octameter \cite{quinn1980critical}. 
 
Trochaic octameter is a rarely used meter in poetry. Within each line of a trochaic octametric poem, there are eight trochaic metrical feet. Each of these eight feet consist of two syllables, the first of the two is stressed, the latter unstressed giving rise to an `up and down' effect to a professional recitation. This pairing of a stressed and an unstressed syllable (or poetic foot) is trochaic \cite{beaver1968grammar}. However, this does not appear to have been followed by the speakers in this dataset.
  
\begin{table}[!h] 
\centering 
\caption{Summary of video content in the Rosetta Raven dataset.} 
\begin{tabular}{| l | r | r | r |} 
\hline 
Video & AAM train frames & AAM fit frames & Duration\\ 
\hline \hline 
Speaker1\_v1 & 11 & 31,858 & 00:08:52 \\ 
Speaker1\_v2 & 11 & 33,328 & 00:09:17 \\ 
Speaker2\_v1 & 10 & 21,648 & 00:06:01 \\ 
Speaker2\_v2 & 10 & 21,703 & 00:06:02 \\ 
\hline 
\end{tabular} 
\label{table:frames} 
\end{table} 

 \begin{figure}[!h] 
\centering 
\includegraphics[width=1\textwidth]{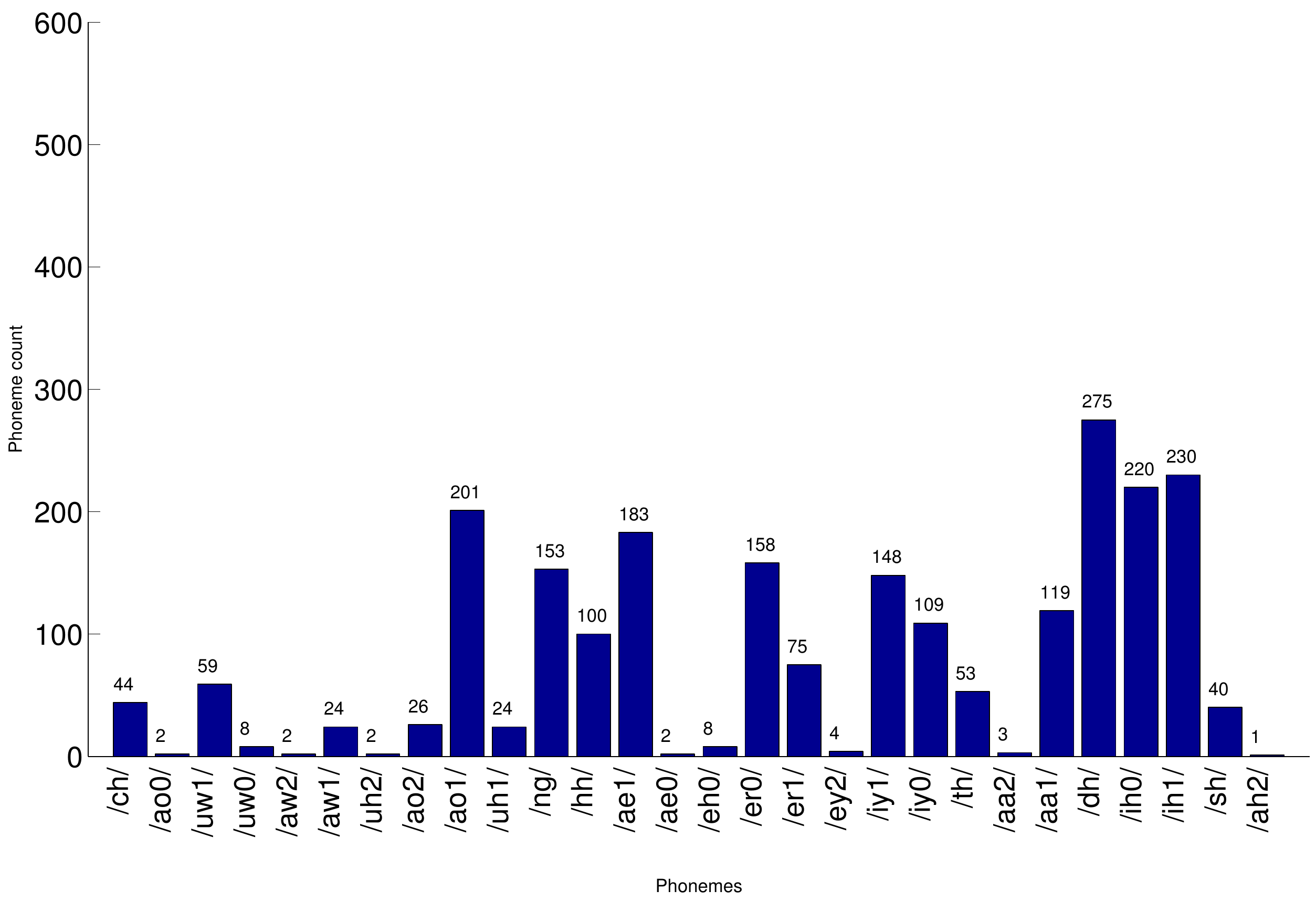} \\
\includegraphics[width=1\textwidth]{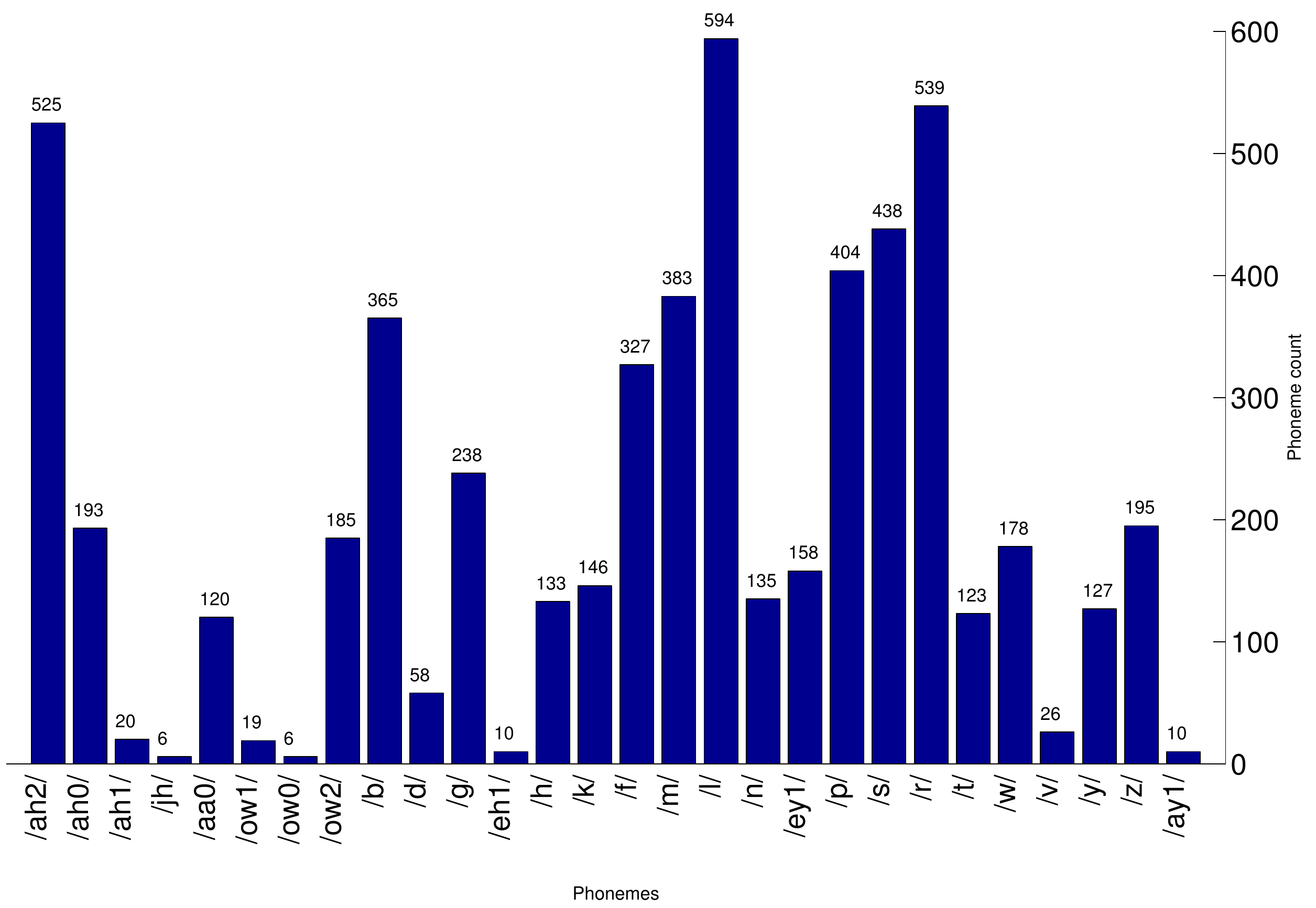} 
\caption{Occurrence frequency of phonemes in the Rosetta Raven dataset.} 
\label{fig:rrhistogram} 
\end{figure} 

In linguistic terms the the videos have 56 phonemes present with minor variation on their occurrences in each video (Figure~\ref{fig:rrhistogram}). It is noted some phonemes namely /\textopeno0/, /uw0/, /\textscripta\textupsilon2/, /\textupsilon2/, /ae0/, /eh0/, /ey2/, /\textscripta2/, /\textturnv2/, /\textscripta0/ and /\textschwa\textupsilon2/ have less than ten instances within the whole data set. These phonemes all have lexical stress shown by the numbers in their naming convention, this comes from the American English set of phonemes used in the CMU pronunciation dictionary. Again, we can not quantify the viseme counts in this dataset as it varies with the viseme set used in any particular experiment. 
   
For these data to be used in a machine lip-reading system, we need to extract features. The training images from each speaker video (Table~\ref{table:frames}) were used together to make a single AAM model for tracking the rest of the video. A full face AAM was used to track the face for a robust fitting, whereas a lip-only AAM was used to extract lip-only feature. These features retained 95\% of the speakers face shape and appearance variance throughout the video and are used in the resolution work described in Chapter~\ref{chap:five} and for assessing the contribution of individual visemes within a set in Chapter~\ref{chap:contributions}. 
 
\begin{table}[!h] 
\centering 
\caption{The number of parameters in shape, appearance, and combined shape and appearance AAM features for the Rosetta Raven dataset speakers. Features retain 95\% variance of facial information.} 
\begin{tabular}{| l | r | r | r |} 
\hline 
Speaker	& Shape & Appearance & Combined \\ 
\hline \hline 
S1	& 6 & 14 & 20 \\ 
S2 	& 7 & 14 & 21 \\ 
\hline 
\end{tabular} 
\label{tab:rr_parameters}
 \end{table}
%\clearpage

\section{RMAV - a context-independent continuous speech dataset} 
\label{sec:lilir}

Formerly known as LiLIR, the RMAV dataset consists of 20 British English speakers (we use 12, seven male and five female), 200 utterances per speaker of the Resource Management (RM) context independent sentences from \cite{fisher1986darpa} which totals around 1000 words each. It should be noted the sentences selected for the RMAV speakers are a significantly cut down version of the full RM dataset transcripts. They were selected by a phonetician to maintain as much coverage of all phonemes as possible. The original videos were recorded in high definition and in a full-frontal position. Individual speakers are tracked using Active Appearance Models \cite{Matthews_Baker_2004} and AAM features of concatenated shape and appearance information have been extracted. %without co-articulation deltas 
 
 \begin{figure}[h] 
\centering 
\includegraphics[width=1\textwidth]{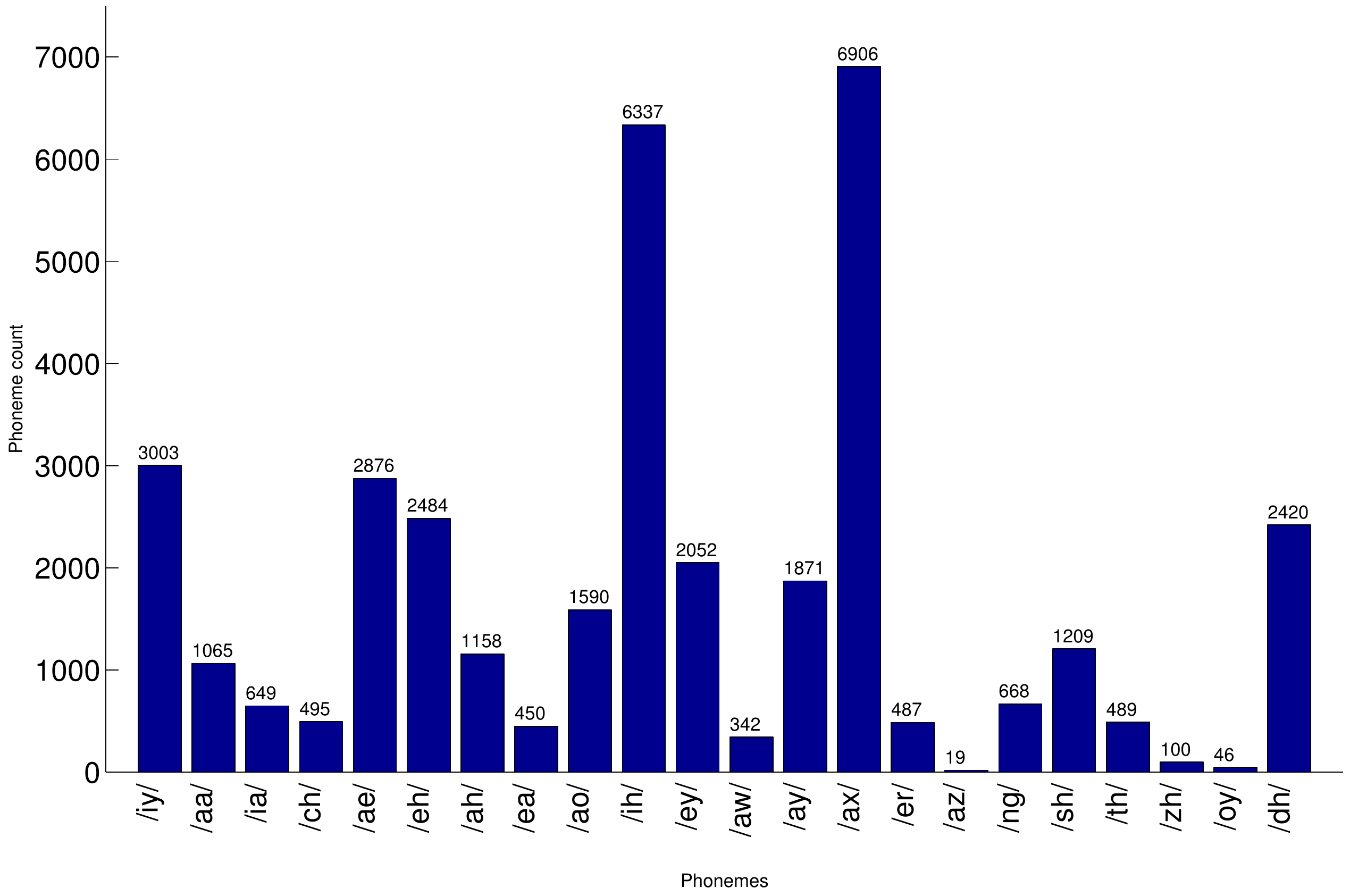} \\
\includegraphics[width=1\textwidth]{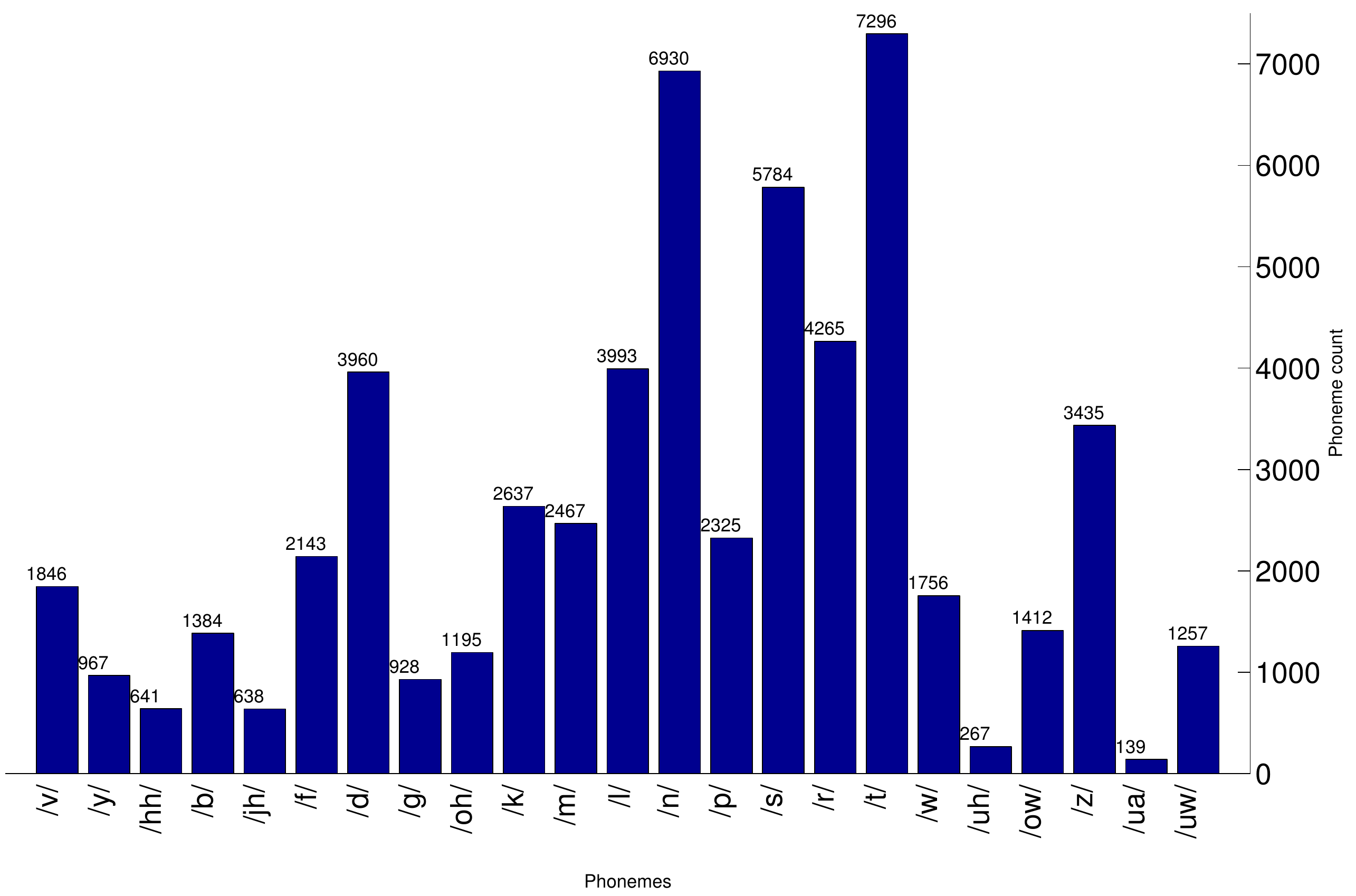}
\caption{Occurrence frequency of phonemes in the RMAV dataset.} 
\label{fig:histogram} 
\end{figure} 
 
 \begin{figure}[h] 
\centering 
\setlength{\tabcolsep}{1pt} 
\begin{tabular}{c c c} 
\includegraphics[width=0.3\textwidth]{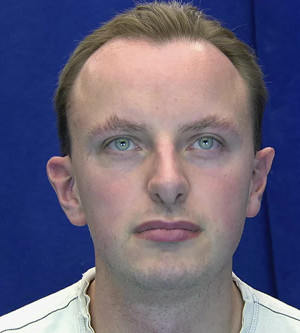} 
\label{fig:sp01} & 
\includegraphics[width=0.3\textwidth]{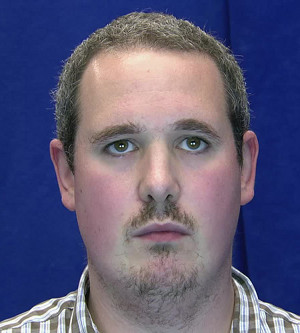} 
\label{fig:sp02} &
\includegraphics[width=0.3\textwidth]{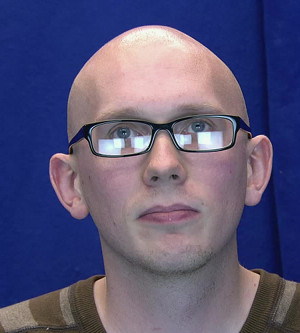} 
\label{fig:sp03} \\ 
(a) Speaker 1 & (b) Speaker 2 & (c) Speaker 3 \\ 
\includegraphics[width=0.3\textwidth]{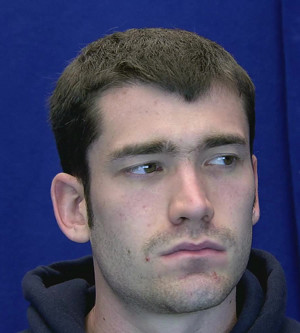} 
\label{fig:sp09} & 
\includegraphics[width=0.3\textwidth]{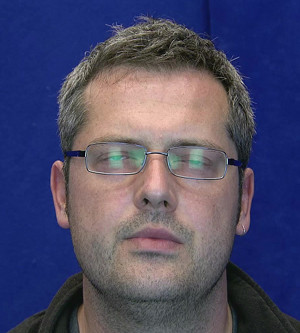} 
\label{fig:sp10} & 
\includegraphics[width=0.3\textwidth]{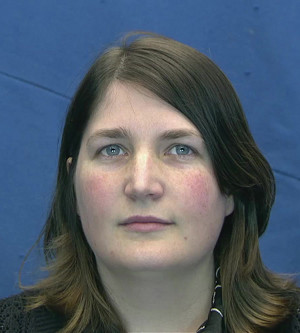} 
\label{fig:sp11} \\ 
(c) Speaker 4 & (d) Speaker 5 & Speaker 6\\ 
\includegraphics[width=0.3\textwidth]{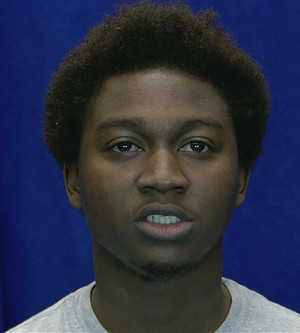} 
\label{fig:sp15} & 
\includegraphics[width=0.3\textwidth]{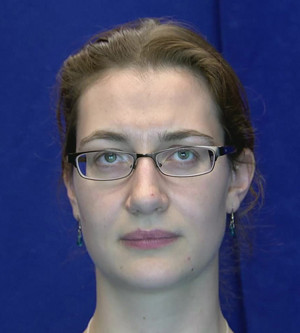} 
\label{fig:sp05} & 
\includegraphics[width=0.3\textwidth]{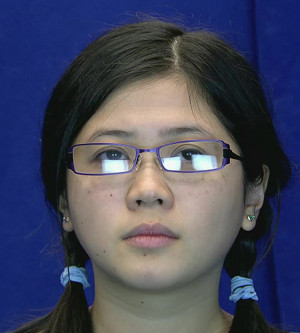} 
\label{fig:sp06} \\ 
(c) Speaker 7 & (d) Speaker 8 & Speaker 9\\
\includegraphics[width=0.3\textwidth]{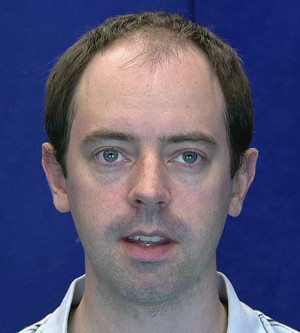} 
\label{fig:sp08} & 
\includegraphics[width=0.3\textwidth]{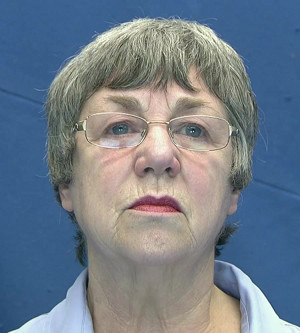} 
\label{fig:sp13} & 
\includegraphics[width=0.3\textwidth]{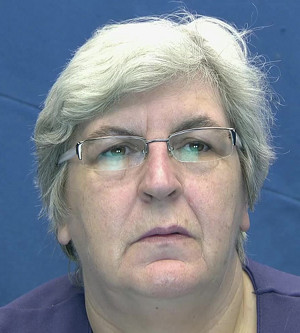} 
\label{fig:sp14} \\ 
(c) Speaker 10 & (d) Speaker 11 & Speaker 12\\  
\end{tabular} 
\label{fig:liilregs} 
\caption{Example faces from the RMAV videos (12 speakers).} 
\end{figure}

 \begin{table}[!h] 
\centering 
\caption{The number of parameters of shape, appearance, and combined shape and appearance AAM features for the RMAV dataset speakers. Features retain 95\% variance of facial information.} 
\begin{tabular}{| l | r | r | r |} 
\hline 
Speaker	& Shape & Appearance & Combined \\ 
\hline \hline 
S1	& 13 & 46	& 59 \\ 
S2 	& 13 & 47 & 60 \\ 
S3	& 13 & 43	& 56 \\ 
S4	& 13 & 47	& 60 \\ 
S5 	& 13 & 45 & 58 \\ 
S6	& 13 & 47	& 60 \\ 
S7 	& 13 & 37 & 50 \\ 
S8 	& 13 & 46 & 59 \\ 
S9	& 13 & 45	& 58 \\ 
S10 	& 13 & 45 & 58 \\ 
S11 	& 14 & 72 & 86 \\ 
S12	& 13 & 45	& 58 \\ 
\hline 
\end{tabular} 
\label{tab:parameterslilirfeatures}
 \end{table}
 
 % Datasets & Tools: Rosetta Raven, AVL2, Lilir, The HTK toolkit, 
%!TEX root = main.tex 

\chapter[Current difficulties in machine lip-reading]{Current difficulties in machine lip-reading} 
\label{sec:current_diff} 
In Chapter~\ref{chap:intro}, we identified a number of factors, or affects, in machine lip-reading which are often difficult to control such as lighting, pose, identity, motion, emotion, linguistic content and expression. We now address these challenges in turn. 
 
\section{Motion} 
The ability to recognise lip gestures throughout a video is addressed in the tracking part of the lip-reading task. There are two systems most commonly used for tracking faces in videos for machine lip-reading. These systems are Active Appearance Models (which can be shape, appearance or shape and appearance models) \cite{927467} and Linear Predictors \cite{ong2011robust}. Both of these systems are effective, even on low quality videos, for tracking the motion of a face during speech. Chapter~\ref{chap:featuretypes} has described these two systems in full. Both methods make some assumptions about motion within videos, LPs are locally affine whereas AAMs are globally affine. Therefore the only minor issue that remains is for non-affine transformations. 
 
\section{Pose} 
There is literature about the effects of pose on computer lip-reading. Some look at expression recognition for Human Computer Interactions (HCI) \cite{Moore2011541} and present an improvement in expression recognition by computers and humans when the pose is rotated to 45$^{\circ}$. Others by Kumar \textit{et al.} and Kaucic \textit{et al.} \cite{4218129, kaucic1998accurate}, look at visual speech classification and suggest that the profile view gives a better classification. However, they also processed the visual features over a longer time period than the duration marked by the endpoints of each speech utterance to consider co-articulation within their tests and so can not isolate which of the longer time window or the pose improved classification. 
 
When considering lip-reading, the study in \cite{11011995} examines the effects of human sentence perception across three viewing angles in relation to the camera position: full-frontal view (0$^{\circ}$), angled view (45$^{\circ}$), and side view (90$^{\circ}$). The performance of a female adult with post-lingual hearing loss was measured for accuracy at each angle. This study used a single-subject, with alternating treatment design where three treatment angles were randomly presented in every session. The accuracy for each session was compared to determine the most effective viewing angle of the speaker. The results indicated that the side-view angle was most effective, as the percentage gain of improvement was greatest in combination with the consistent upward trend of the data points across treatment sessions. The performance of frontal-view and angled-view angles were also successful but not significantly more so than full-frontal. The results of this preliminary effort indicate the value of treatment for visual sentence perception at all three angles, including the non-traditionally targeted side view for human lip-reading. 

Preliminary studies into non-frontal pose affects in lip-reading can be found in \cite{4064511} \&~\cite{kumar2007profile}. In both a small vocabulary is used in order to simplify the recognition task for measuring the effects of features extracted from non-frontal camera positions. In ~\cite{4064511} the classifiers were trained on frontal features and tested on non-frontal features and the results showed that the greater the off-frontal angle became, then the word error rate increased. However, the frontal view features provided inferior recognition to off-angle features in \cite{kumar2007profile}. The key distinction between these studies is the visual noise of image backgrounds in the original videos.

Most AVSR databases are recorded face frontal, an alternative idea of lip-reading non-frontal camera angles with frontal-trained classifiers using a mapping from the recorded angle to the estimated actual angle of the speaker to the camera is presented in \cite{pass2010investigation}. In this work, we see a new dataset recorded for the specifically for the mapping technique and the results support the observations in ~\cite{4064511} \&~\cite{kumar2007profile} but add the observation that with the larger off-camera angles, then a smaller feature vector of only the higher order features is preferable. 
 
 These studies into the affect of pose on machine lip-reading are taken further by Lucey \textit{et al.} \cite{lucey2009visual} with a proprietary dataset. Here the authors undertake three activities with a small vocabulary (connected digit strings) on 38 speakers; comparing the frontal and profile view lip-reading performance (akin to the experiments in \cite{4064511} \&~\cite{kumar2007profile}), but they also take the challenge further by experimenting with concatenating both the frontal and profile view features into \textit{multi-view} features, and attempting to lip-read using a single pose-invariant normalisation method. The results for task one support those seen in \cite{4064511} whereby the frontal features outperformed the profile features. This is considered due to both datasets being recorded in controlled conditions with minimal noise. 
 
The results for the \textit{multi-view} features in \cite{lucey2009visual}, marginally better than frontal, and significantly better than profile features. The $w.e.r$ reduces from 38.88\% for profile features, for 27.66\% for frontal features and the best \textit{multi-view} features achieved a 25.36\% $w.e.r$. This was achieved by simply concatenating the two sets of features. This observation is important that it is important to not simply pick a pose for lip-reading, but rather, there are useful visual cues from all angles. 

Finally, in the third test, Lucey \textit{et al.} develop a single pose-invariant model for lip-reading, regardless of the pose of the test data. They compare different pairings of features over the training/testing split. For example, using frontal features $F$, for training and testing with frontal features. Then using the same features $F$, to test profile features $P$ and vice versa. A third training model using a 50/50 split of $F$ and $P$ is included in the experiment setup. Also adopted is the projection of each set of features, $F$ and $P$ into the alternative feature space for new features $F'$ and $P'$ for alternative testing data for the three training options, $F$, $P$, and $[F_{50},P_{50}]$. These tests showed best recognition where the training and test features matched. Where these didn't match the $w.e.r$ dramatically increased, for example for an ($F$,$F$) train/test pairing the $w.e.r$ was 29.18\%. The train/test pair of ($F$,$P$) achieves a $w.e.r$ of 87.07\%. However, the authors also show that this can be mitigated by the projection of the test profile data back into the frontal feature space where the train/test split ($F$,$P'$) recovers the $w.e.r$ back down to 54.85\%. This transformation principle is also used in \cite{6298439} by Lan \textit{et al.} who presented an view-independent lip-reading system. This investigation uses a continuous speech corpus compared to the small vocabulary dataset in \cite{lucey2009visual}. This later study acknowledges a human lip-readers preference for a non-frontal view and suggests it could be attributed to lip protrusion. A different approach for the feature transform is presented, (a linear mapping between poses) but the development of a such system shows computer lip-reading can be independent of speaker pose. 
 
\section{Multiple people} 
 
The challenge of machine lip-reading a video with more than one person, meaning to track their faces, has a number of solutions. \cite{871067} demonstrates multiple person tracking (albeit not lip-reading) and has also implemented this into a simple HCI system. Also, in \cite{layne2014re} we see how a person can be re-identified between videos, either a second view of the same space at an alternate perspective or, as a person moves through a location. An example of a speaker identification method is detailed in \cite{607030}, and \cite{simstruct, visualvowelpercept} detail lip-reading of multiple people, \cite{simstruct} recognises consonants, and \cite{visualvowelpercept} visual vowels. Whilst none of these papers have directly tested concurrent speech, it would be interesting to know what effect, if any, speakers talking in unison would cause upon current lip-reading systems. \cite{871073} presents an audio-visual system for HCI which automatically detects a talking person (both spatially and temporally) using video and audio data from a single microphone. Until visual-only classifiers have improved, a robust visual-only system for machine lip-reading still needs to be developed and the classifiers are a essential part of the system. 
 
\section{Video conditions} 
 
Studies such as \cite{Blokland199897} on the effect of low video frame-rate on human speech intelligibility during video communications, suggest that lower frame rates encourage humans to over-articulate to compensate for the reduced visual information available, akin to a visual Lombard effect. (N.B. this is only when the speakers are aware of the low quality parameters e.g. during a video conference.) Therefore, it should be asked: does a computer need more information (higher frame rate/resolution) to lip-read a speaker in a recorded video sequence? The study in \cite{Blokland199897} observes in face-to-face human interactions, articulation is relaxed. So one could ask, in the instance where a computer needs extra visual information throughout the recording, (think of the example where a face-to-face conversation is being recorded incognito), how much does this lack of visual information impact on the classification performance? That is, how far does the lack of video recording quality affect classification? 
 
Another study into frame rate in computer lip-reading, \cite{saitoh2010study}, tells us the greatest classification is achieved when the same frame rate is used for both training and testing data. This is perhaps unsurprising as it is shown that when both training and test data sets are at low frame rates, classification drops when the frame rate of the training data is lower than the test data. They show longer words are easier to classify. It would be interesting to see if this is the same for visemes. \cite{saitoh2010study} also shows a dependency between frame rates and classification accuracy by speaker. When training and test data do not have the same, or very similar frame rates, it is recommended training data has a higher frame rate (for feature extraction) than the test (fit) data. It observes word classification rates vary in a non-linear fashion as the frame rate is reduced which is caused by the particular words being recognised. The duration of an utterance does not have an effect on the classification rate in this paper. 
 
\section{Speech methods and rates} 
People have different speaking styles, accents and rates of speech. Some people talk fast, some slow, some talk out of the side of their mouth, others naturally over-articulate and others have facial hair which occludes the visibility of lip movement during speech. The rate of speech alters both an utterance duration and articulator positions. Therefore, both the sounds produced, but particularly, visible appearance are altered. In \cite{6854158}, the authors present an experiment which measures the effect of speech rate and shows the effect is significantly higher on visual speech than in acoustic. 

Because of this variable, some people undertake elocution classes for a myriad of reasons. Examples include call centre employees undertaking `accent neutralisation' courses to make them more approachable for their target customers \cite{cowie2007accents}. This is supported in \cite{hall2005introduction} where they state ``Speakers of non-prestige dialects in some countries take elocution courses, or respond to newspaper adverts which promise to `eliminate' their `embarrassing' accents, and second language learners fret that they'll never sound like a native.".
 
\section{Resolution}
In this chapter we have reviewed the environmental affects of lip-reading classification. Whilst many can be controlled, and we have seen in the literature how some of the effects can be managed, we also note previously considered challenges such as, outdoor video, poor lighting, and agile motion can all be overcome \cite{bowden2013recent}. 

In regards to studies about the affects of resolution, there is limited literature found at the time of writing which examines this. Some experiments touch on this area of interest with investigations into recognition from noisy images. 

An investigation into the effects of compression artefacts, visual noise (simulated with white noise), localisation errors in training is presented in \cite{heckmann2003effects}, and in \cite{ACP:ACP371} the authors undertake two experiments, of which the first includes some attention to spatial resolution (the number of pixels). This inclusion of features from three different resolutions is interesting but the resolutions selected have differing aspect ratios and as such it is not a controlled method of resolution variation. Also, the effect of this spatial resolution is not measured or presented, rather it is included as a property of tests on frame rate and contrast. Neither of these papers consider the simple removal of information from a smaller image compared to a larger one.

Therefore testing of this is necessary (see Chapter~\ref{chap:five}). Given that, up to this point, with a known speaker and reduced linguistic context, classification rates can be high, it is a fair bet the most sensitivity is to be found on the parameters associated with the left hand side of Figure~\ref{fig:variability} (identity, expression etc). Nevertheless, there has been surprisingly little attention paid to a systematic review of the cameras parameters. Therefore, in our first practical experiment we ask `what is the lowest resolution at which a machine can lip-read?'. 
 
 % Environmental limitations on lip-reading, setup resolution work.
%!TEX root = main.tex 

\chapter[Resolution limits in lip-reading]{Resolution limits in lip-reading} 
\label{chap:five} 
 
We have discussed how machine lip-reading depends on factors which can be difficult to control, such as: lighting \cite{6106749}, identity \cite{cox2008challenge}, motion \cite{lan2009comparing} and pose \cite{kaucic1998accurate, 6298439, 4218129, 11011995}, rate of speech \cite{6854158}, and expression \cite{Moore2011541}. But some factors, such as video resolution, are controllable. So it is surprising there is not yet a specific, systematic and complete study of the effect of resolution on lip-reading in non-noisy conditions. There is a tendency, without evidence, to assume a high resolution video will produce better classification results and so a study to measure the effect of resolution on classification is needed and this is undertaken in this chapter. 

\section{Image pre-processing for feature modification} 
\label{sec:resolution} 
 
\begin{figure}[p] 
\centering 
\setlength{\tabcolsep}{1pt} 
\begin{tabular}{l c  c  c  c  c  } 
%picture row 
once & \includegraphics[width=0.17\textwidth]{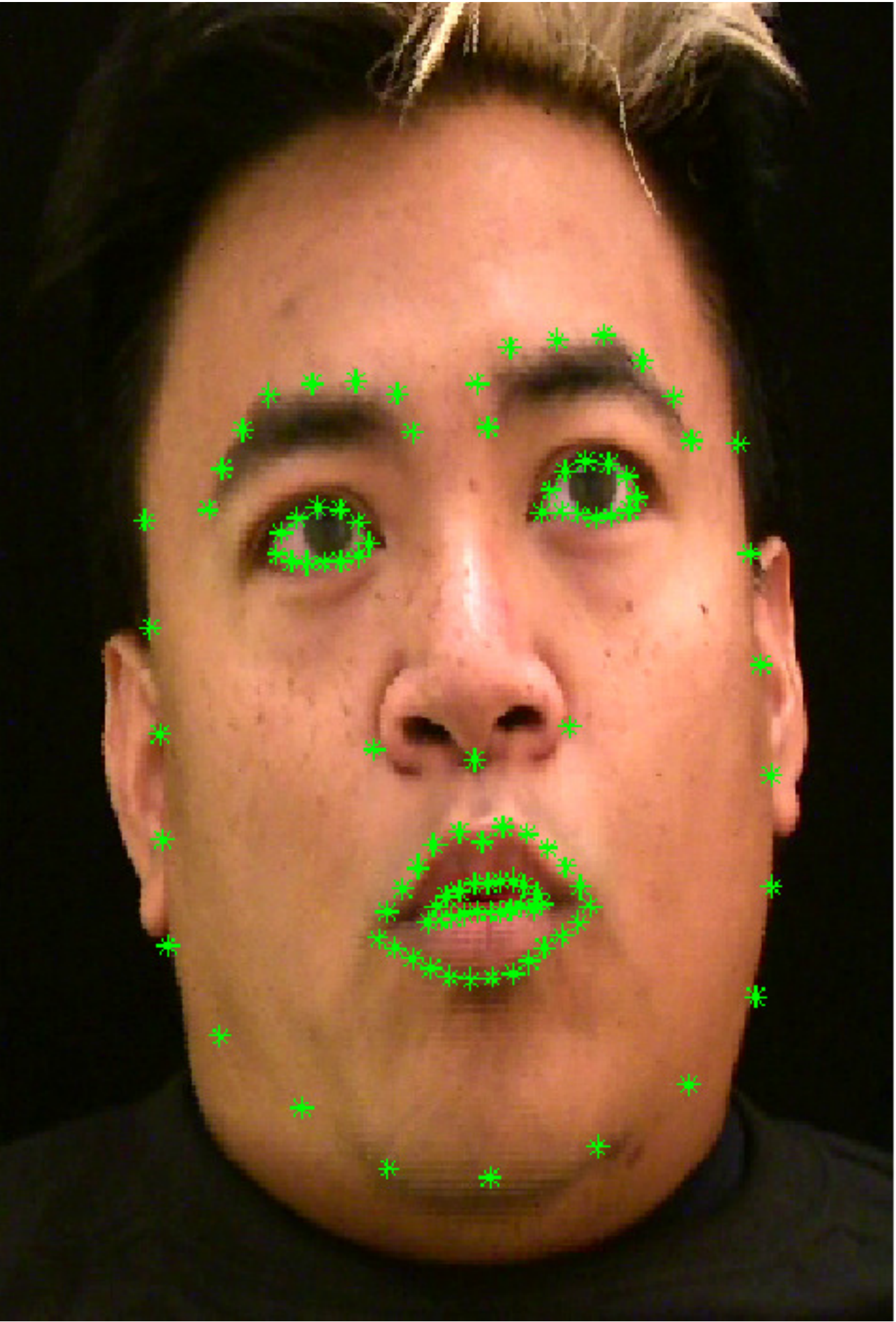} & \includegraphics[width=0.17\textwidth]{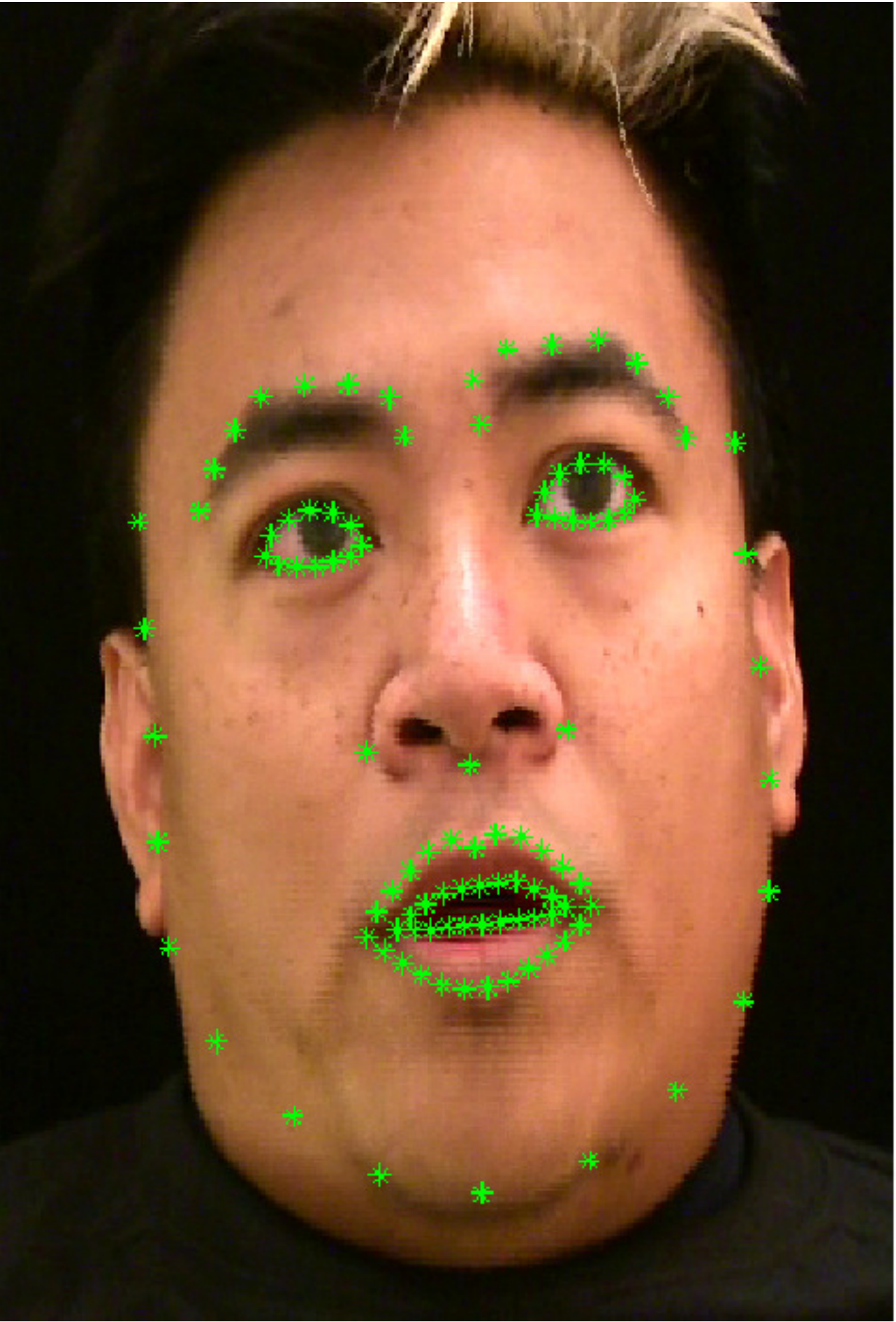} & \includegraphics[width=0.17\textwidth]{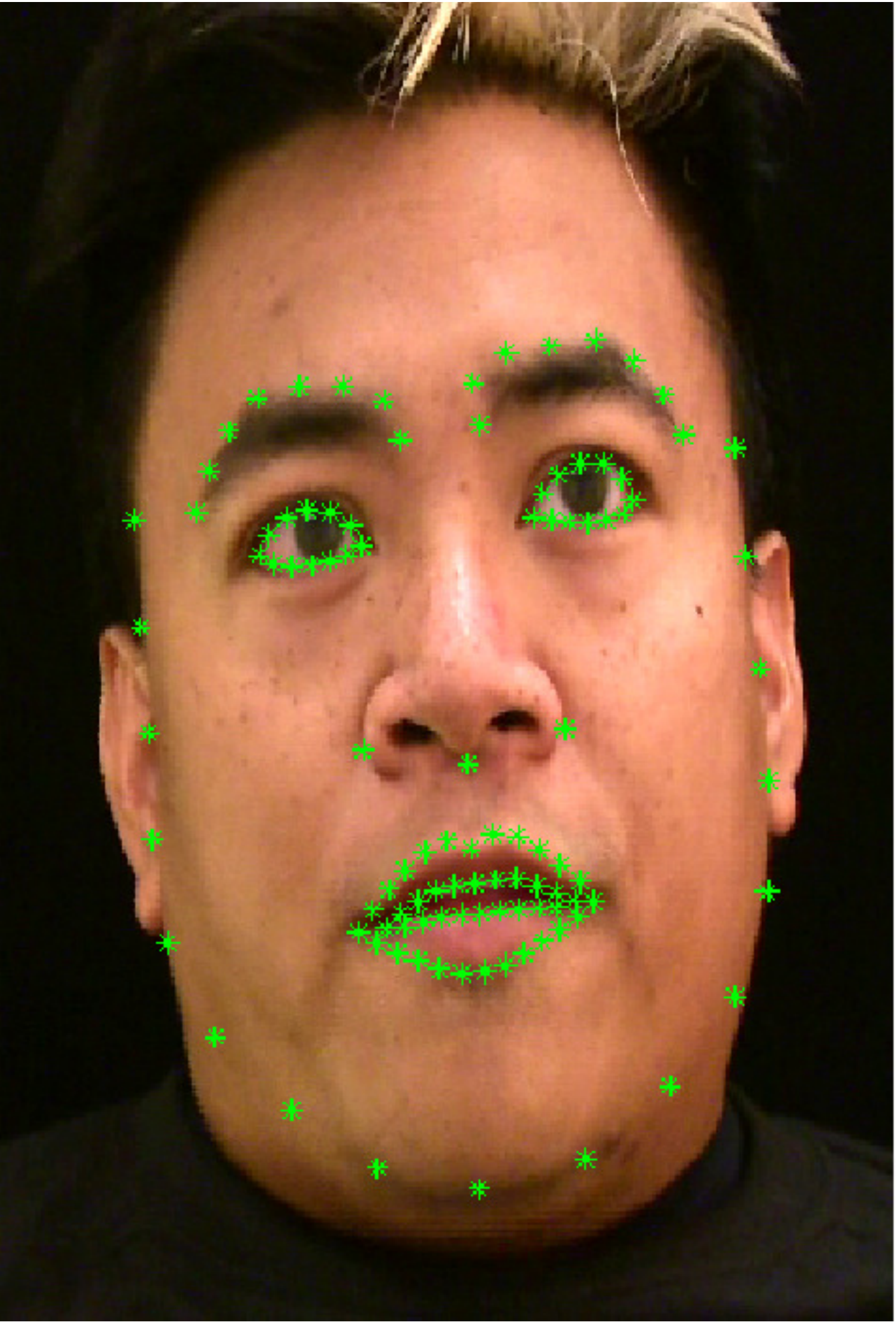} &  \includegraphics[width=0.17\textwidth]{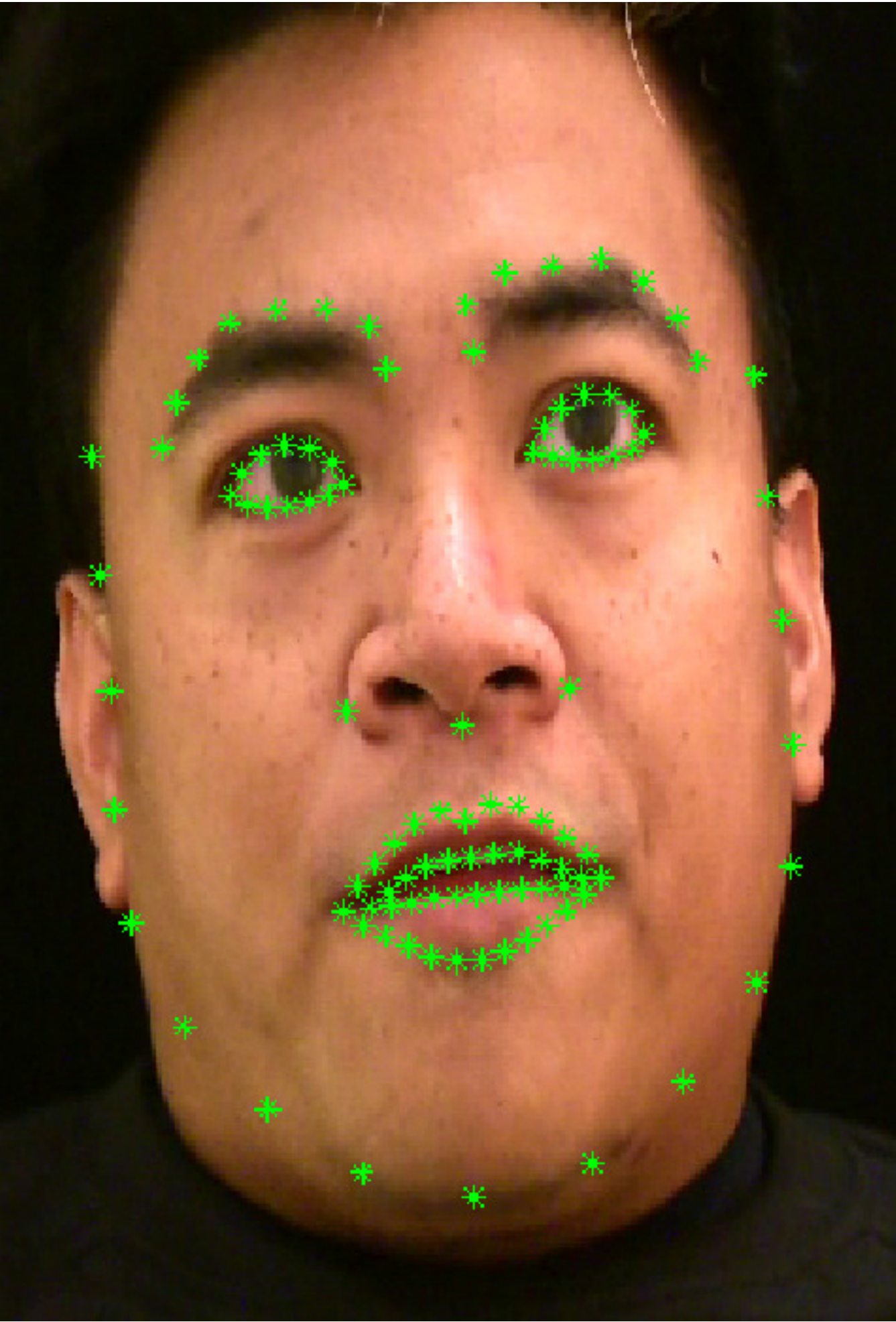}&  \\ 
%viseme row 
%& & & &  \\ 
% phoneme row 
& /w/ & /\textturnv1/ & /n/ & /s/  &  \\ 
%word row 
%\multicolumn{5} { c } {once} \\ 
%picture row 
upon & \includegraphics[width=0.17\textwidth]{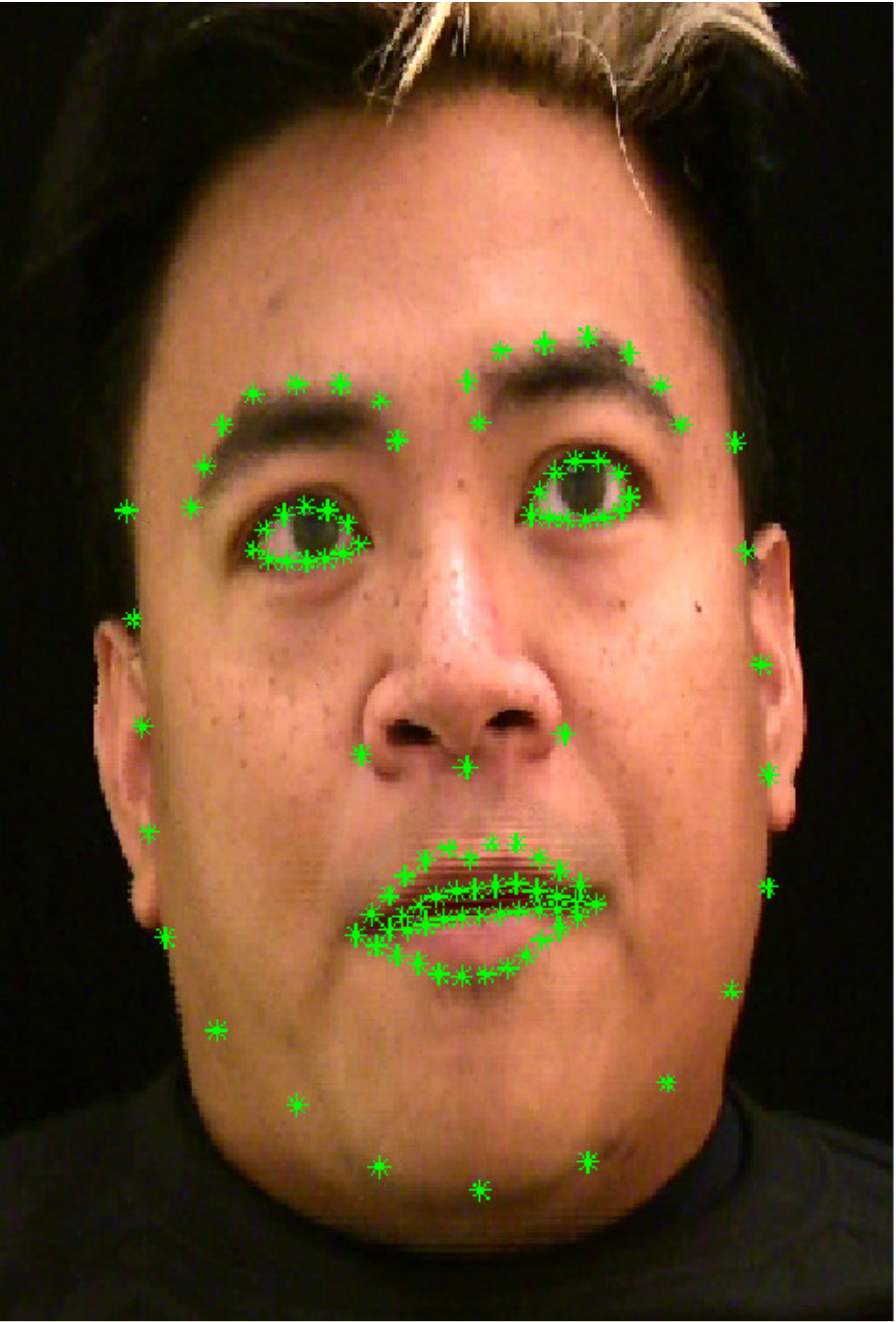} & \includegraphics[width=0.17\textwidth]{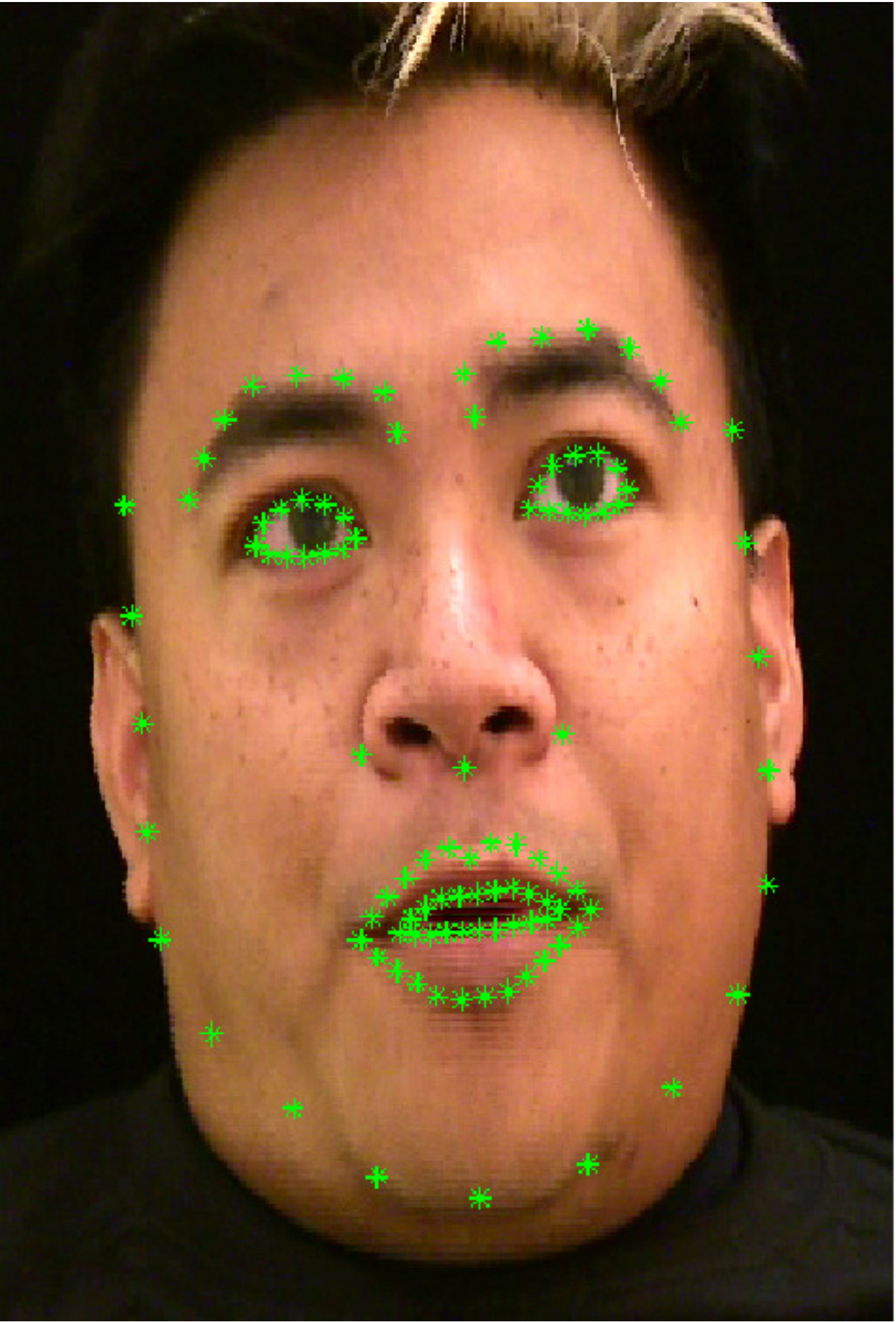} & \includegraphics[width=0.17\textwidth]{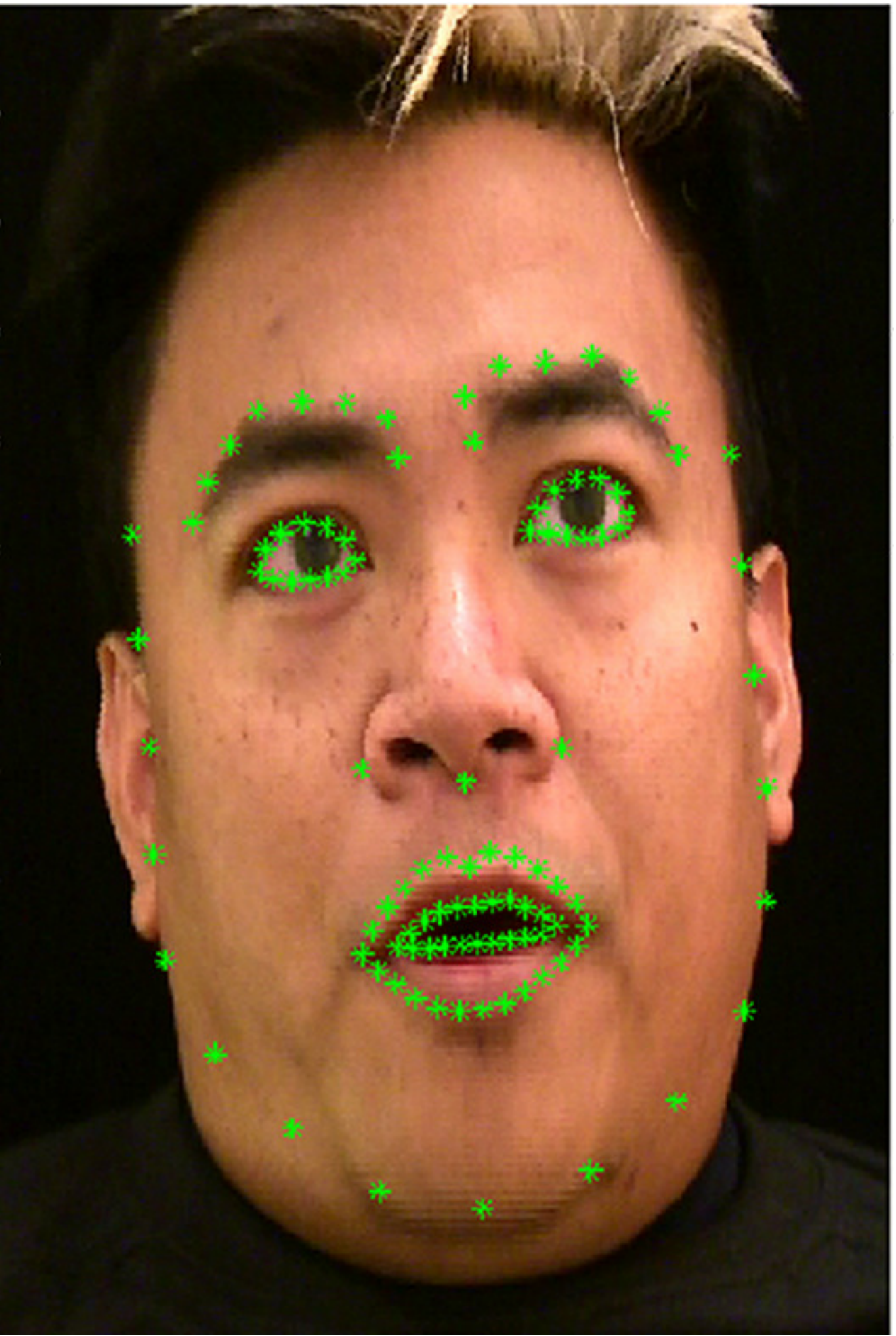} & \includegraphics[width=0.17\textwidth]{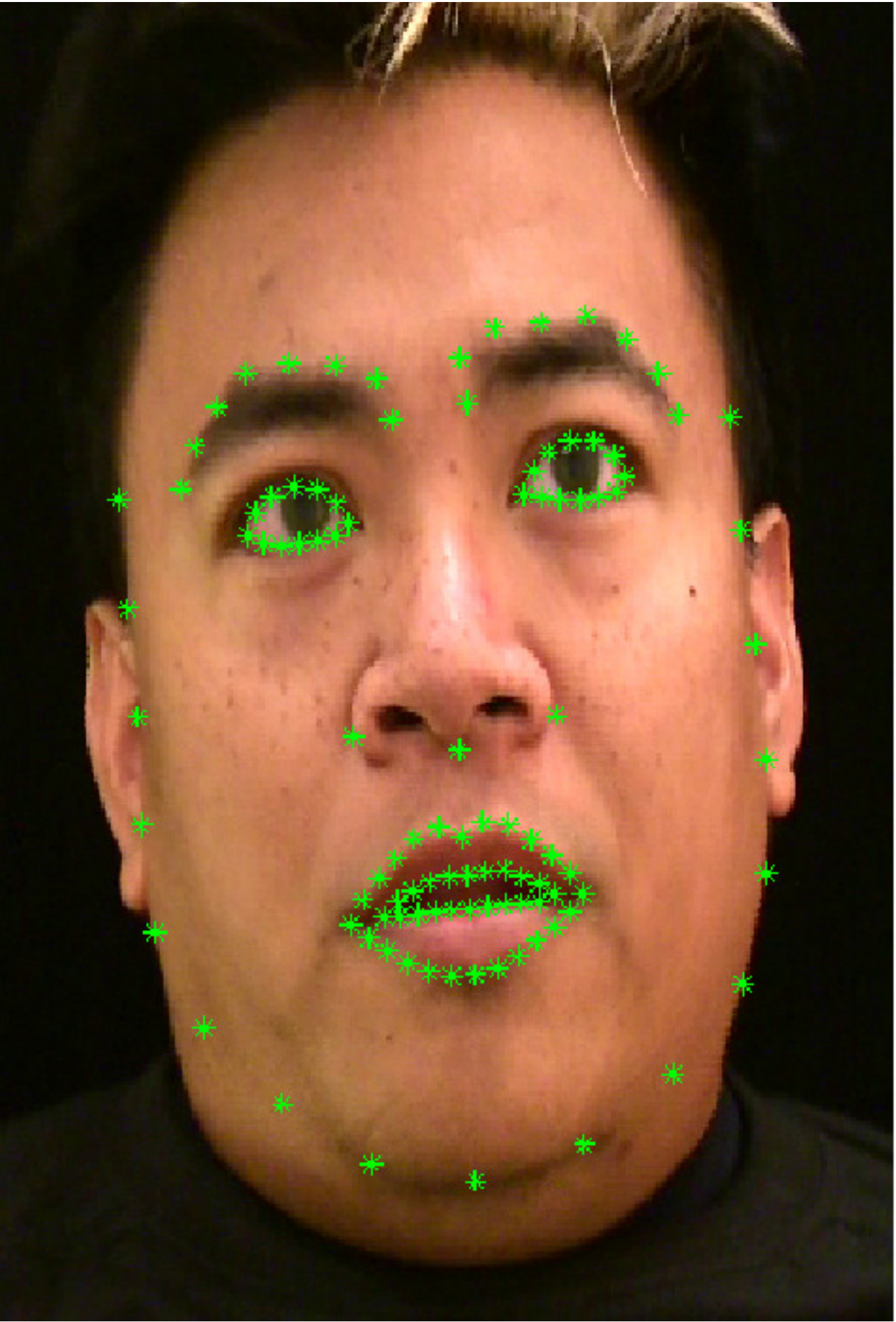} & \\ 
%viseme row 
%& & & &   \\ 
% phoneme row 
& /\textturnv0/ & /p/ & /\textopeno0/ & /n/  & \\ 
%word row 
%\multicolumn{5} { c }  {upon} \\ 
%picture row 
a mid-& \includegraphics[width=0.17\textwidth]{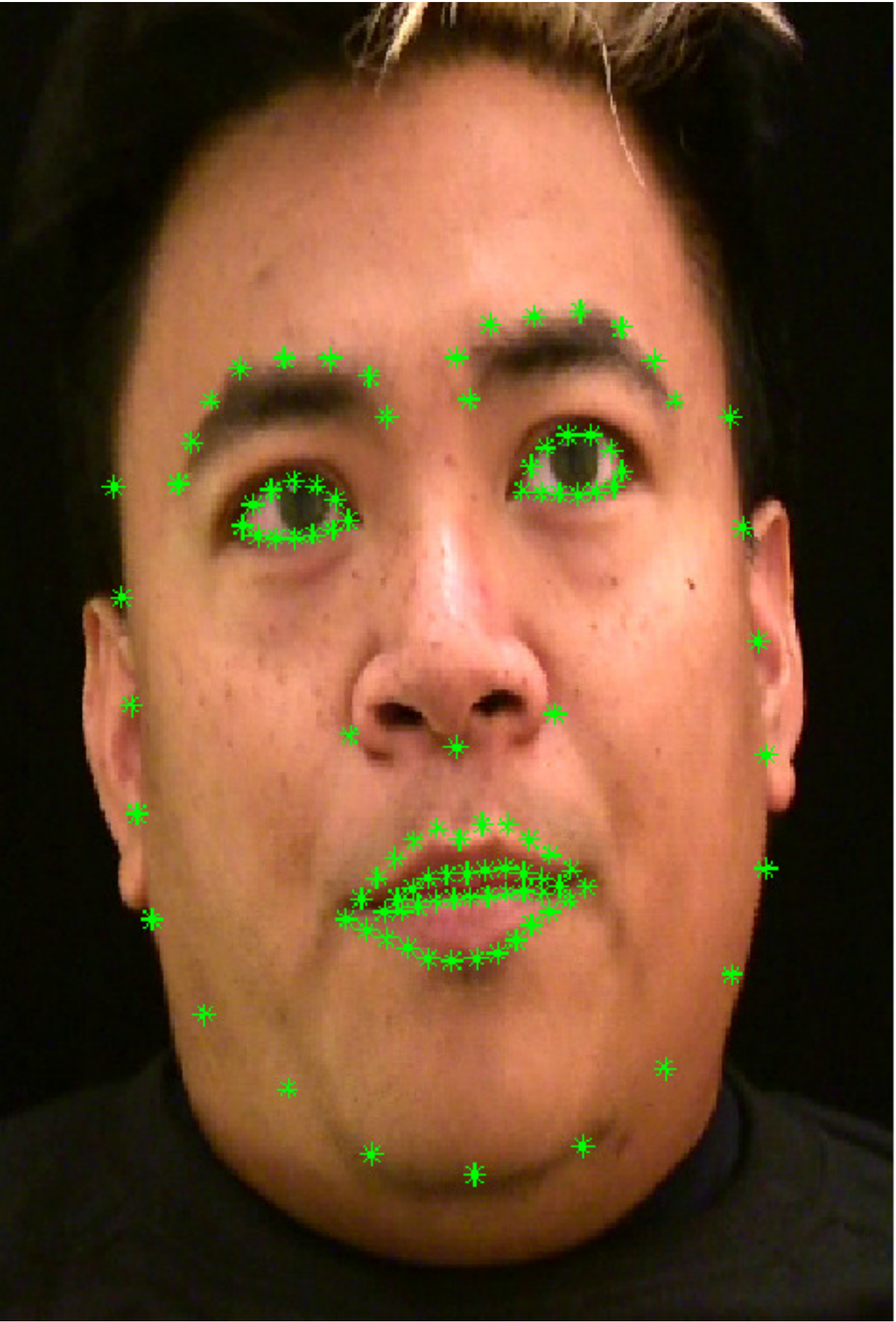} & &  \includegraphics[width=0.17\textwidth]{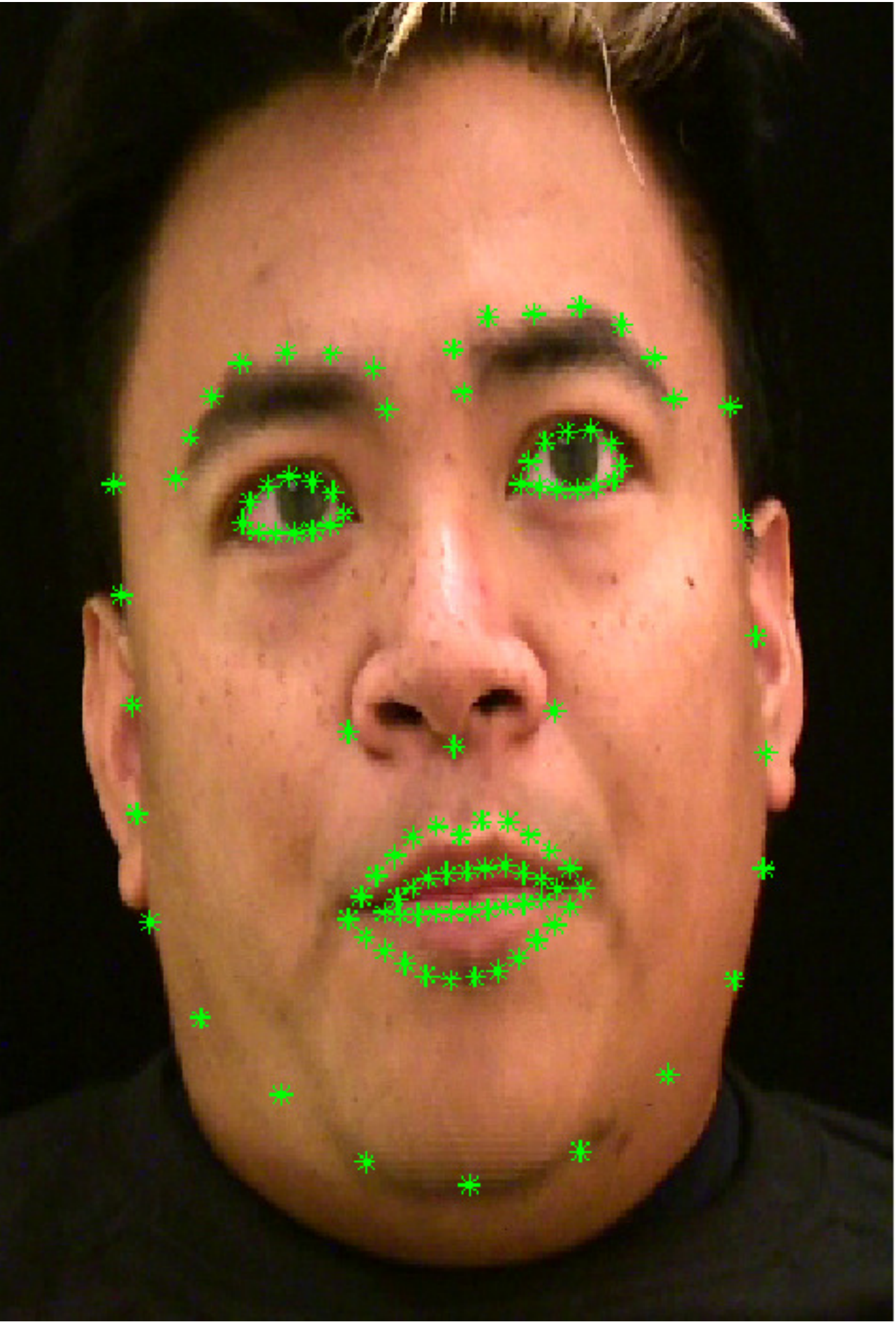} & \includegraphics[width=0.17\textwidth]{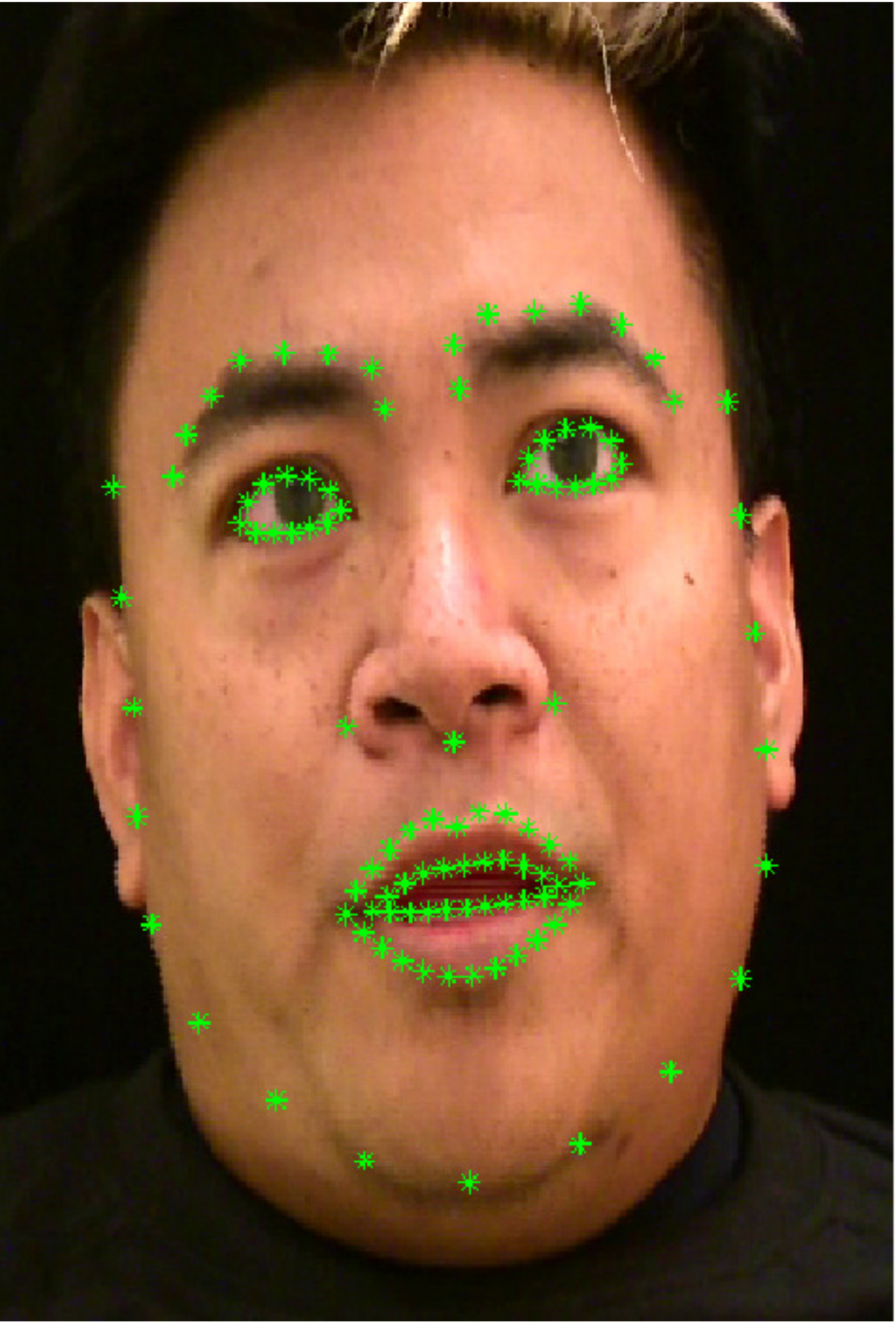}  & \includegraphics[width=0.17\textwidth]{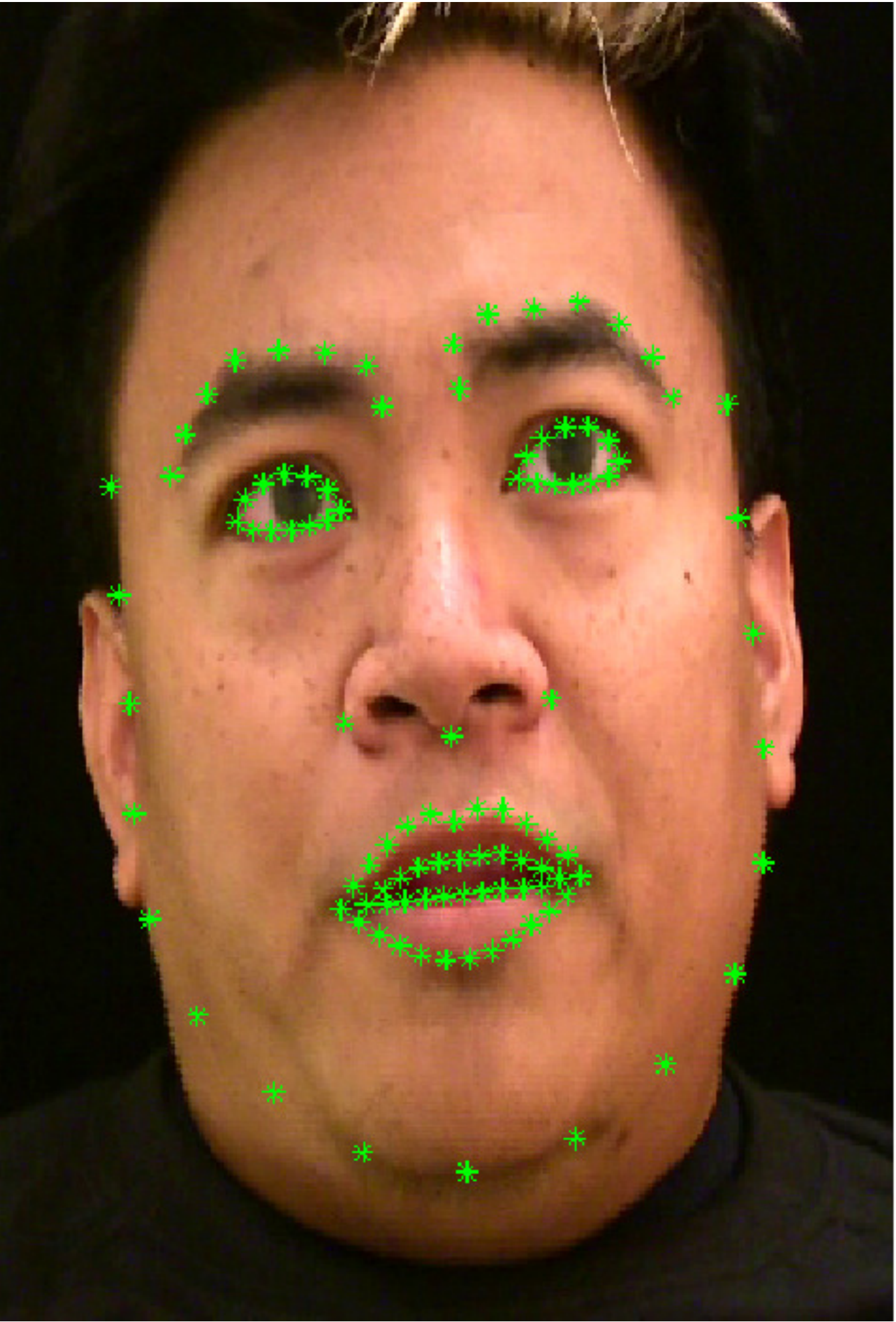}  \\ 
%viseme row 
%& & & & \\ 
% phoneme row 
& /\textturnv0/ &  & /m/ & /\textsci1/ & /d/ \\ 
%word row 
%a & & \multicolumn{3} { c }  {mid-} \\ 
%picture row 
night & \includegraphics[width=0.17\textwidth]{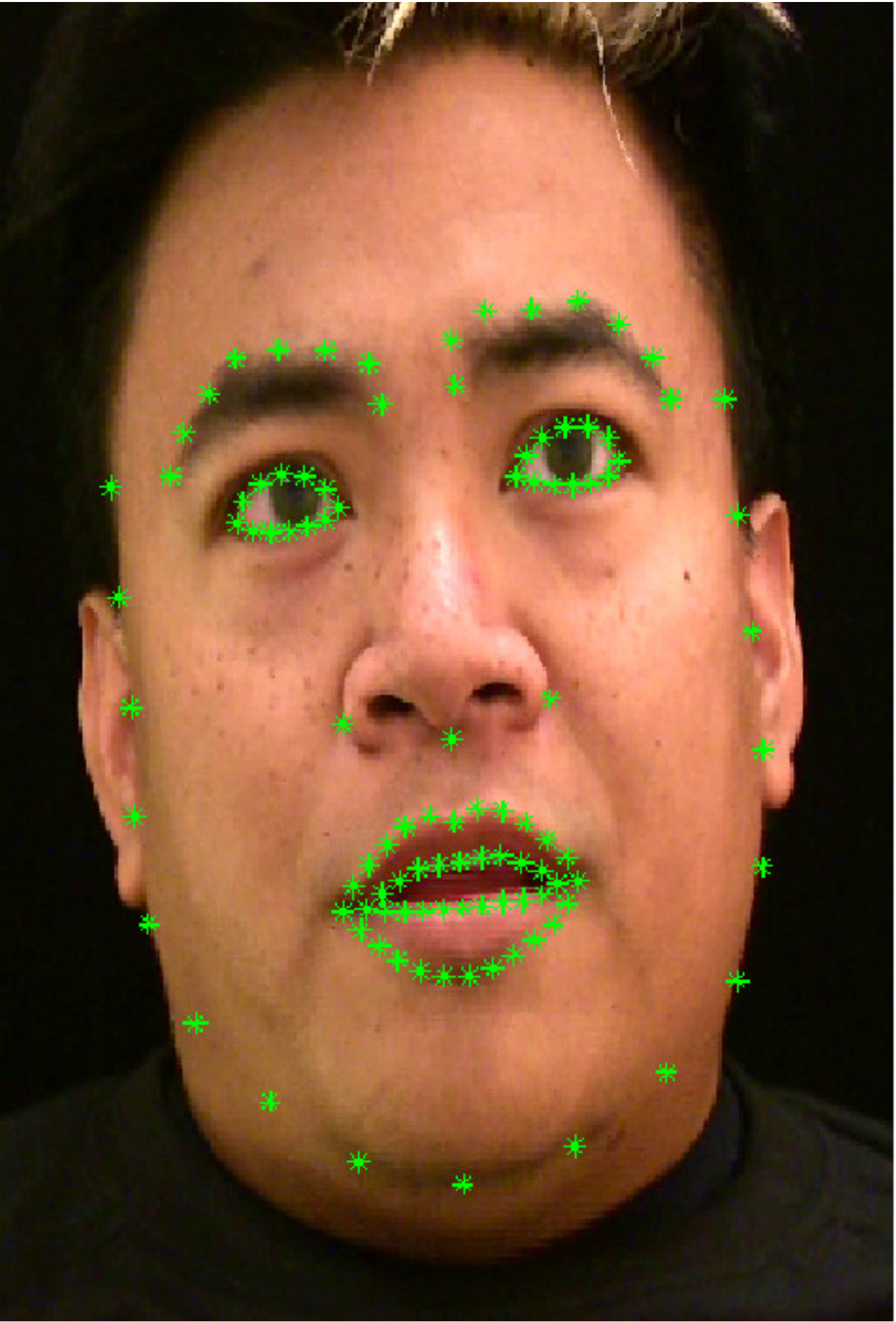} & \includegraphics[width=0.17\textwidth]{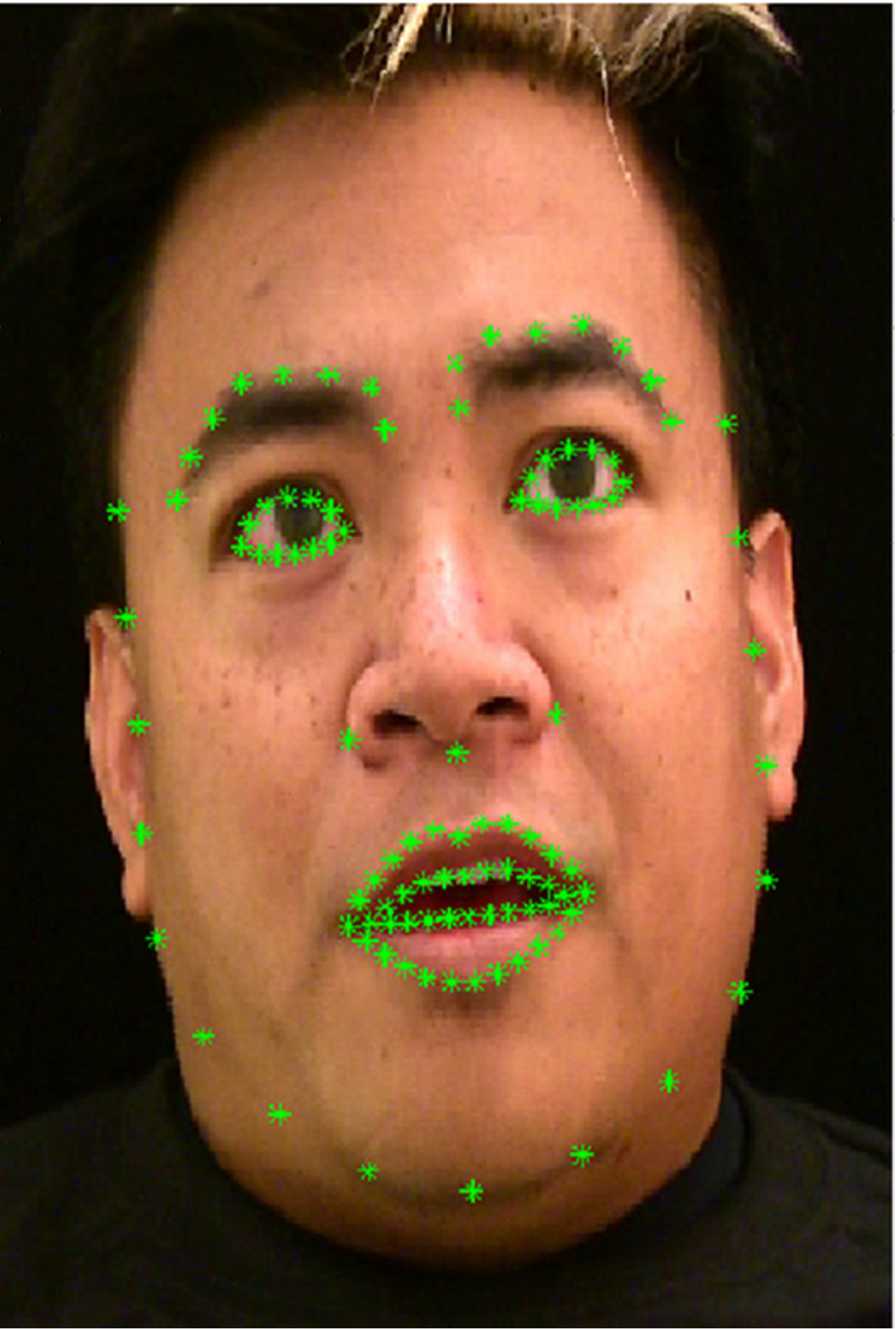} & \includegraphics[width=0.17\textwidth]{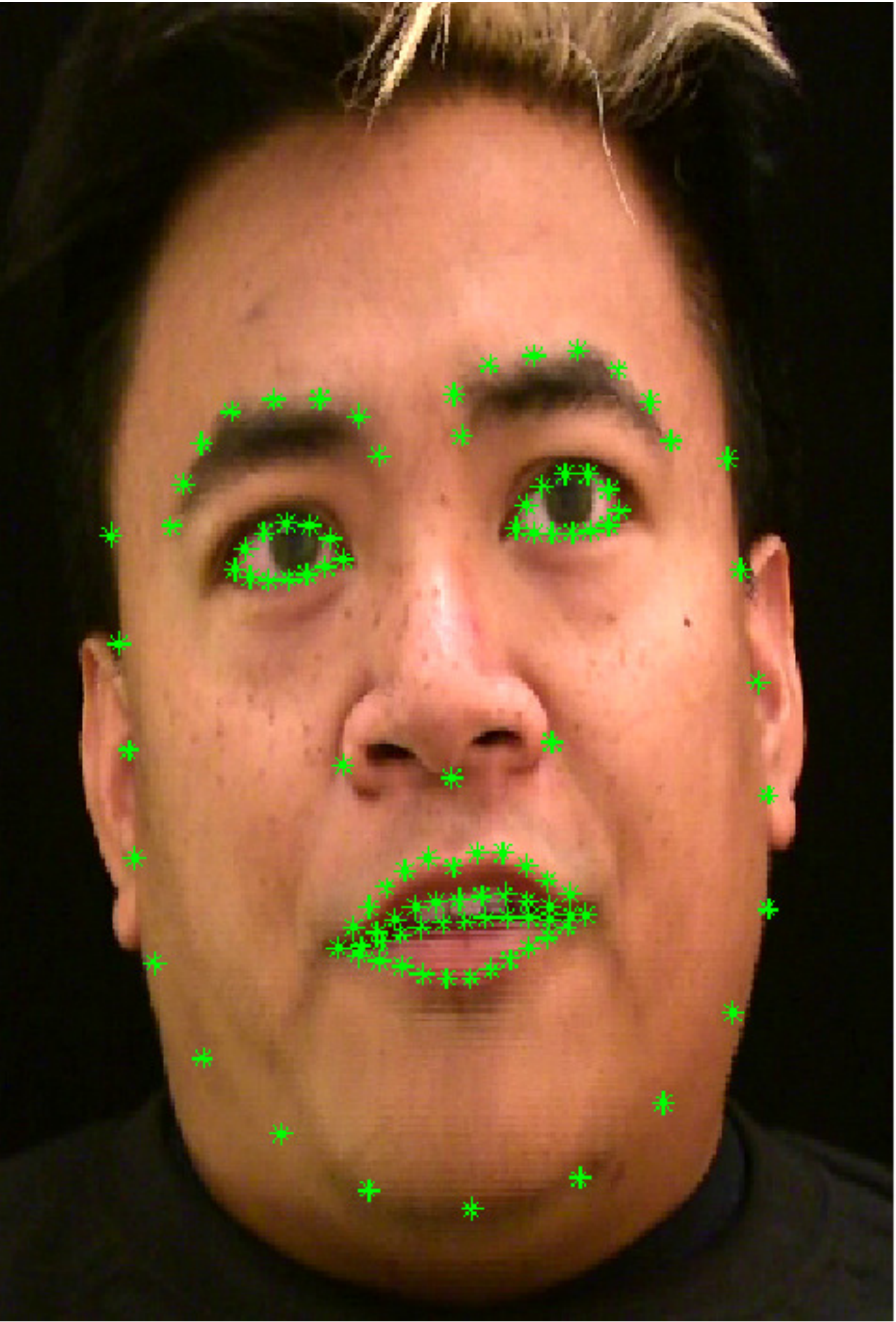} & & \\ 
%viseme row 
%& & &  &\\ 
% phoneme row 
& /n/ & /ay2/ & /t/  \\ 
%word row 
%\multicolumn{5} { c }  {-night} \\ 
%picture row 
dreary & \includegraphics[width=0.17\textwidth]{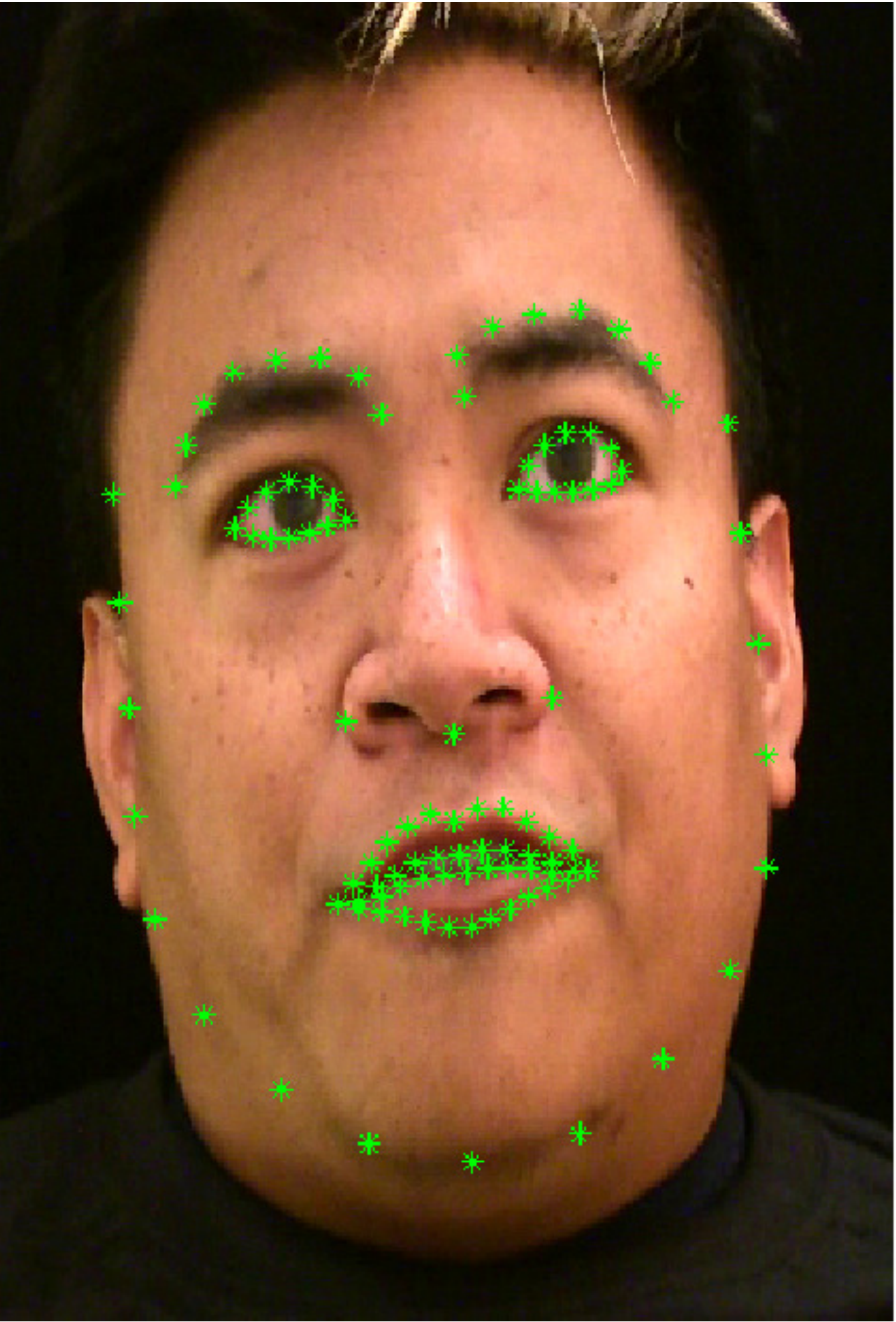} & \includegraphics[width=0.17\textwidth]{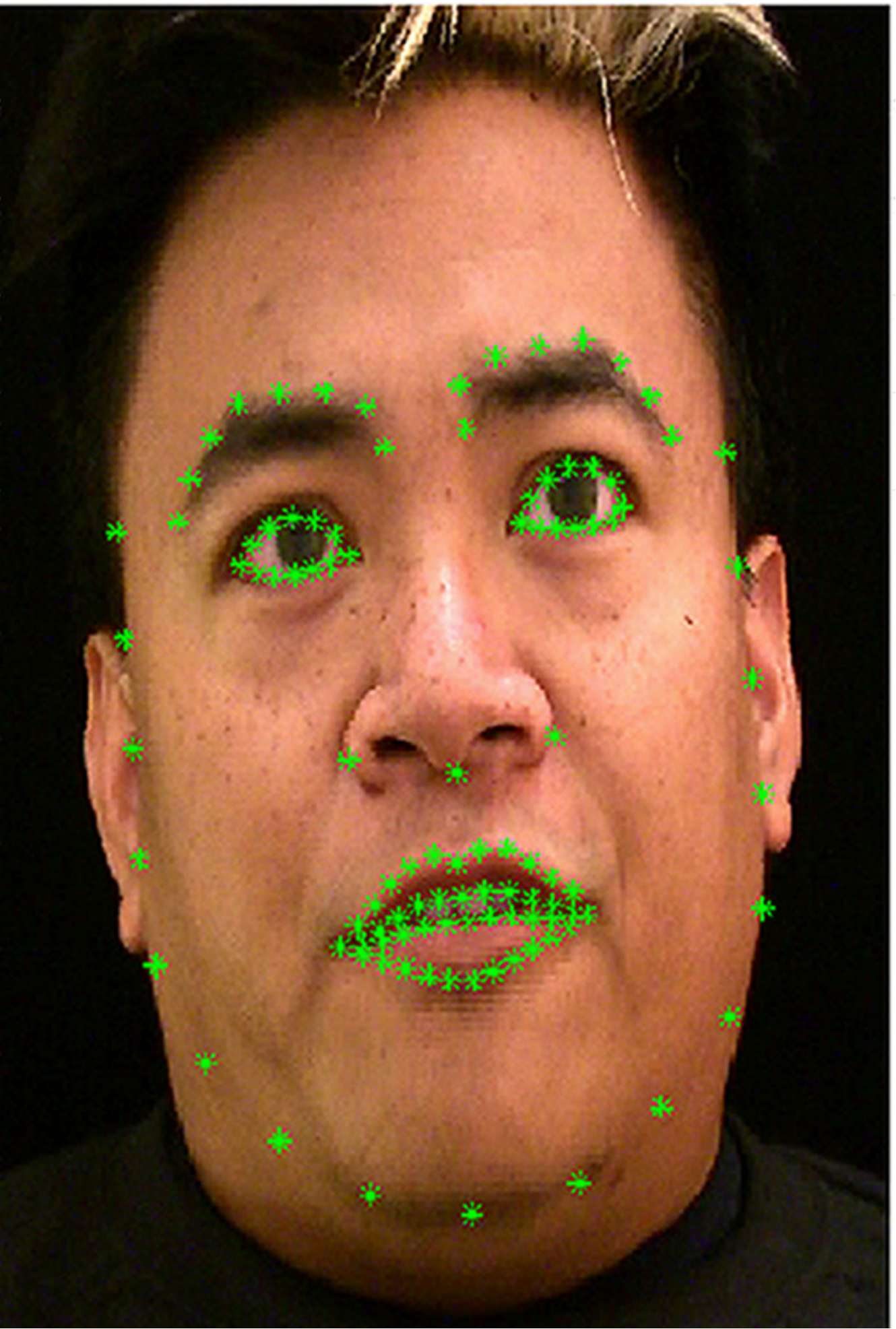} & \includegraphics[width=0.17\textwidth]{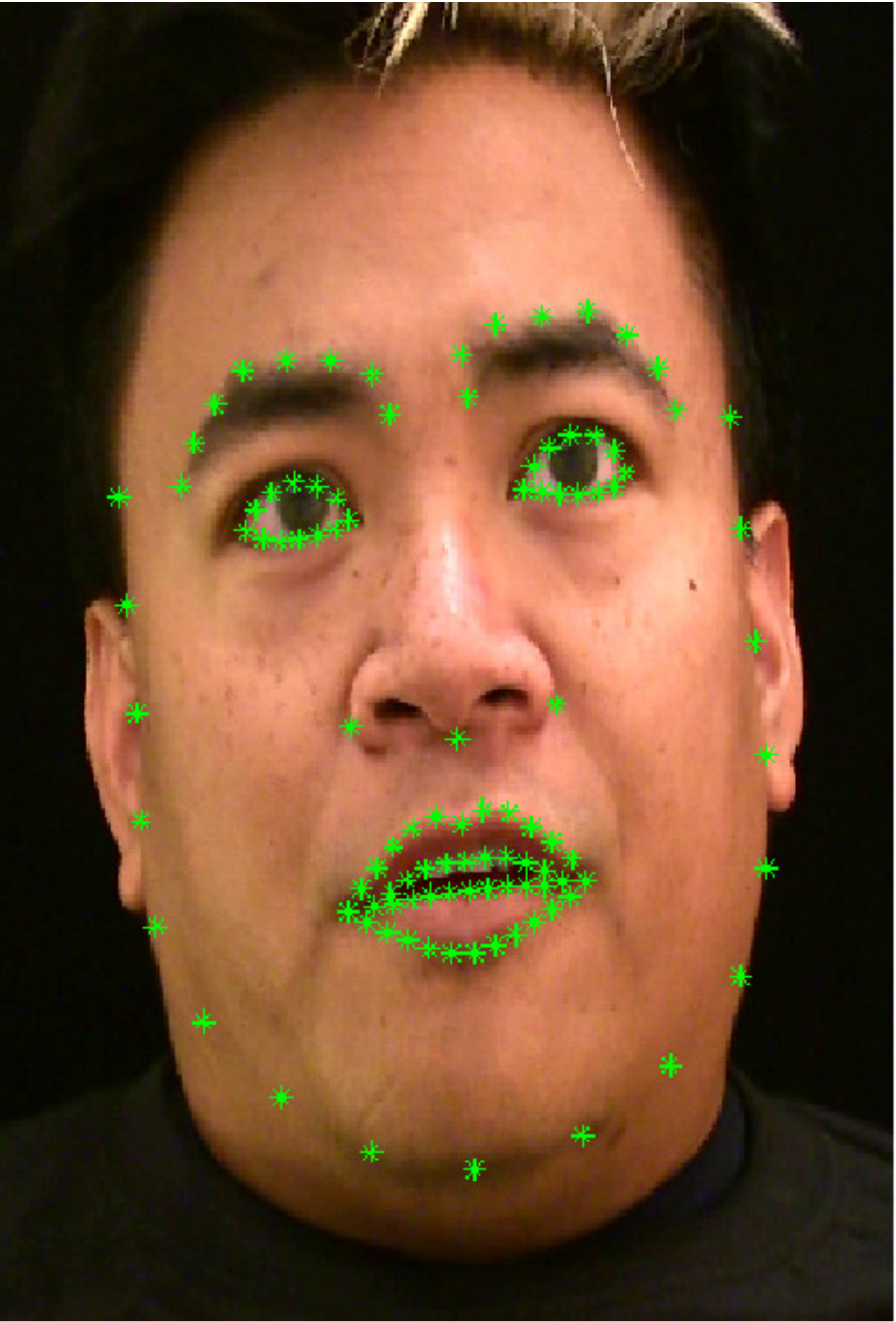} & \includegraphics[width=0.17\textwidth]{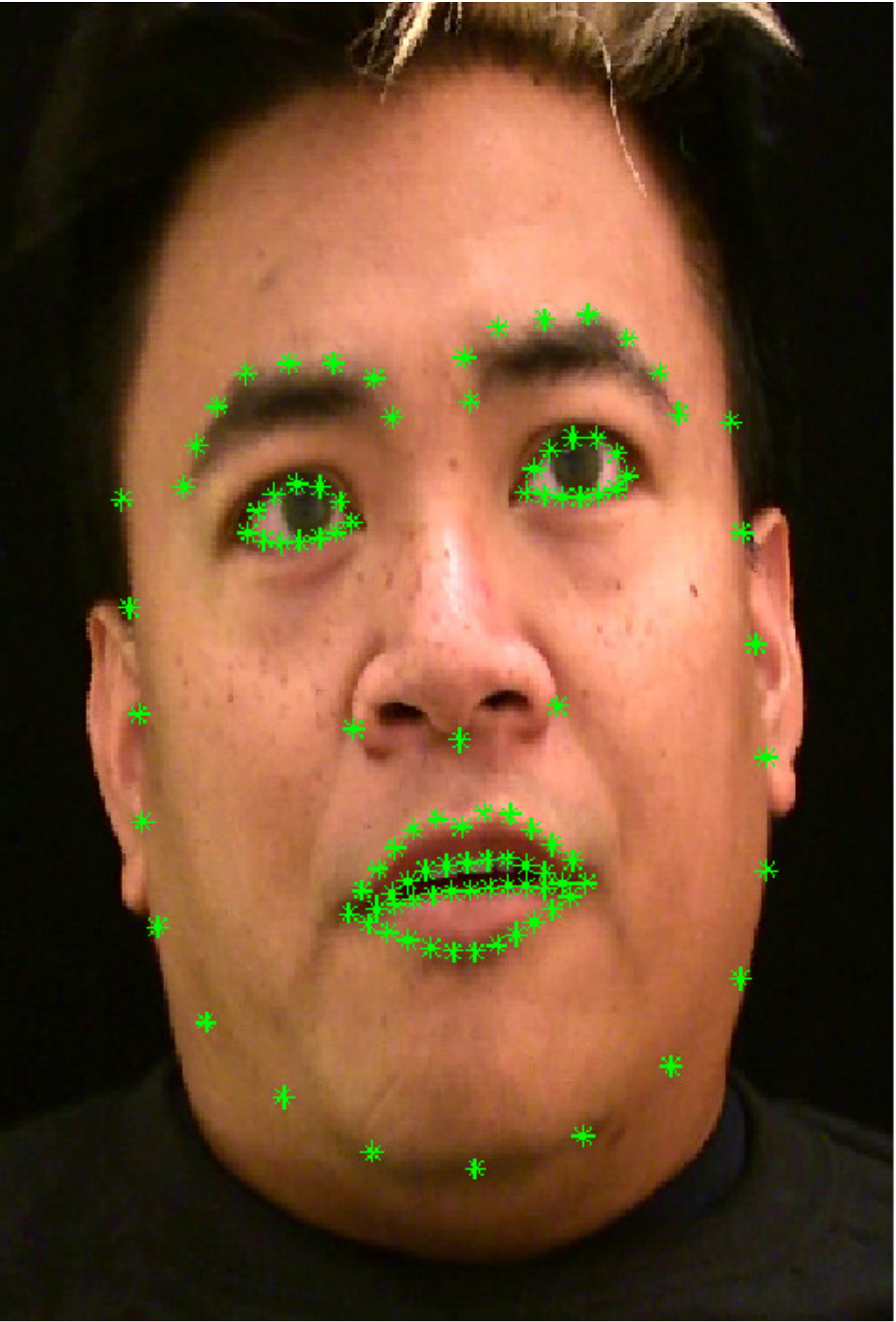}& \includegraphics[width=0.17\textwidth]{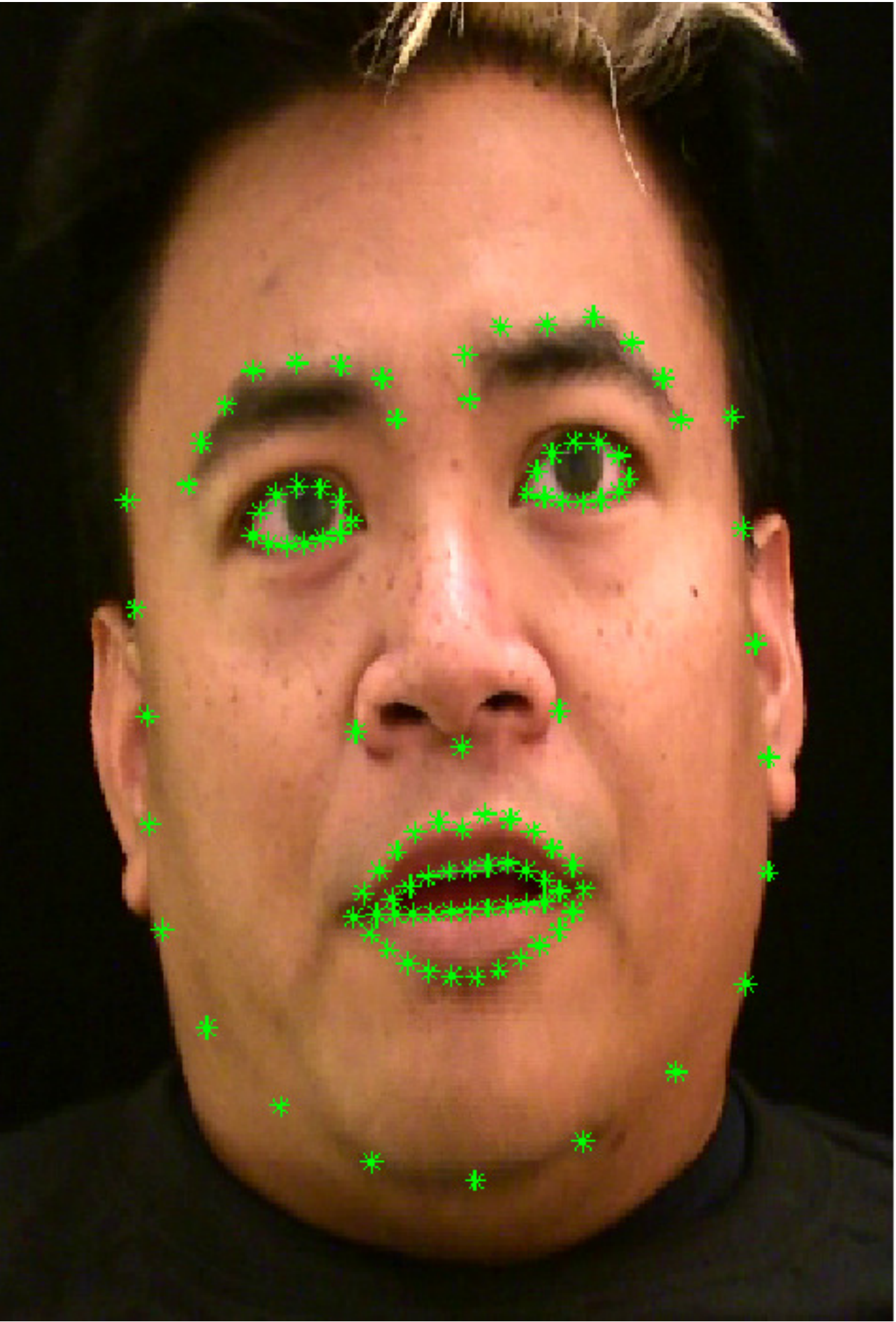}\\ 
%viseme row 
%& & & & \\ 
% phoneme row 
& /d/ & /r/ & /\textsci1/ & /r/ & /iy0/  \\ 
%word row 
%\multicolumn{5} { c }  {dreary.} \\ 
\end{tabular} 
\caption{Tracking a Rosetta Raven speaker saying `Once upon a midnight dreary' with a full-face Active Appearance Model.} 
\label{fig:tracking_imgs} 
\end{figure} 

For this work we use the Rosetta Raven dataset as already described in Section~\ref{sec:rr}. Before feature extraction however, we undertake some image pre-processing. All four videos in the dataset were converted into a set of images (one per frame in PNG format) with ffmpeg \cite{ffmpeg} using image2 encoding at full high-definition resolution ($1440\times1080$). 
 
To build an initial Active Appearance Model for tracking each video, we select the first frame and nine or ten others randomly. These \emph{key frames} are hand-labelled with a model of a face including: facial outline (jaw and hairline, in front of ears), eyebrows, eyes, nose and lips. To track the face, this preliminary AAM is then fitted, via Inverse Composition fitting~\cite{inversecompAlg, Matthews_Baker_2004} to the unlabelled frames (Table~\ref{table:frames} in Chapter~\ref{chap:datasets_dics} gives the numbers of frames for each video). In Figure~\ref{fig:tracking_imgs} we show, for Speaker 1, the tracked full-face AAM mesh (one frame per phoneme), for the first sentence of The Raven ``Once upon a midnight dreary" used in tracking the speaker face. 
 
\begin{figure}[!ht] 
\centering 
\setlength{\tabcolsep}{1pt} 
\resizebox{\columnwidth}{!}{% 
\begin{tabular}{c c c c} 
\includegraphics[width=0.25\textwidth]{figs/full_face_fit.png} &
%\label{fig:mesh2} \\ 
\includegraphics[width=0.25\textwidth]{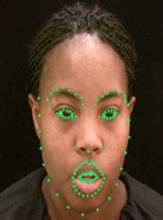} & 
%\label{fig:mesh} \\ 
\includegraphics[width=0.25\textwidth]{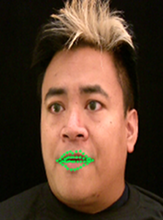} &
%\label{fig:lipmeshT1} & 
\includegraphics[width=0.25\textwidth]{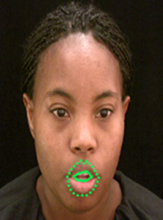} \\
%\label{fig:lip_models} \\ 
(a) S1 face AAM points & (b) S2 face AAM points & (c) S1 lips AAM points & (d) S2 lips AAM points \\ 
\end{tabular} 
}%
\caption{Active Appearance Model shape landmarks for two Rosetta Raven speakers.} 
\label{fig:meshes} 
\end{figure}

At this stage full-face speaker dependent AAMs are tracked and fitted on all full resolution lossless PNG frame images as in Figure~\ref{fig:meshes} (a) and (b) for both speakers in the Rosetta Raven dataset.
 
\begin{figure}[!ht] 
\centering 
\setlength{\tabcolsep}{1pt} 
\begin{tabular}{c c} 
\includegraphics[width=0.44\textwidth]{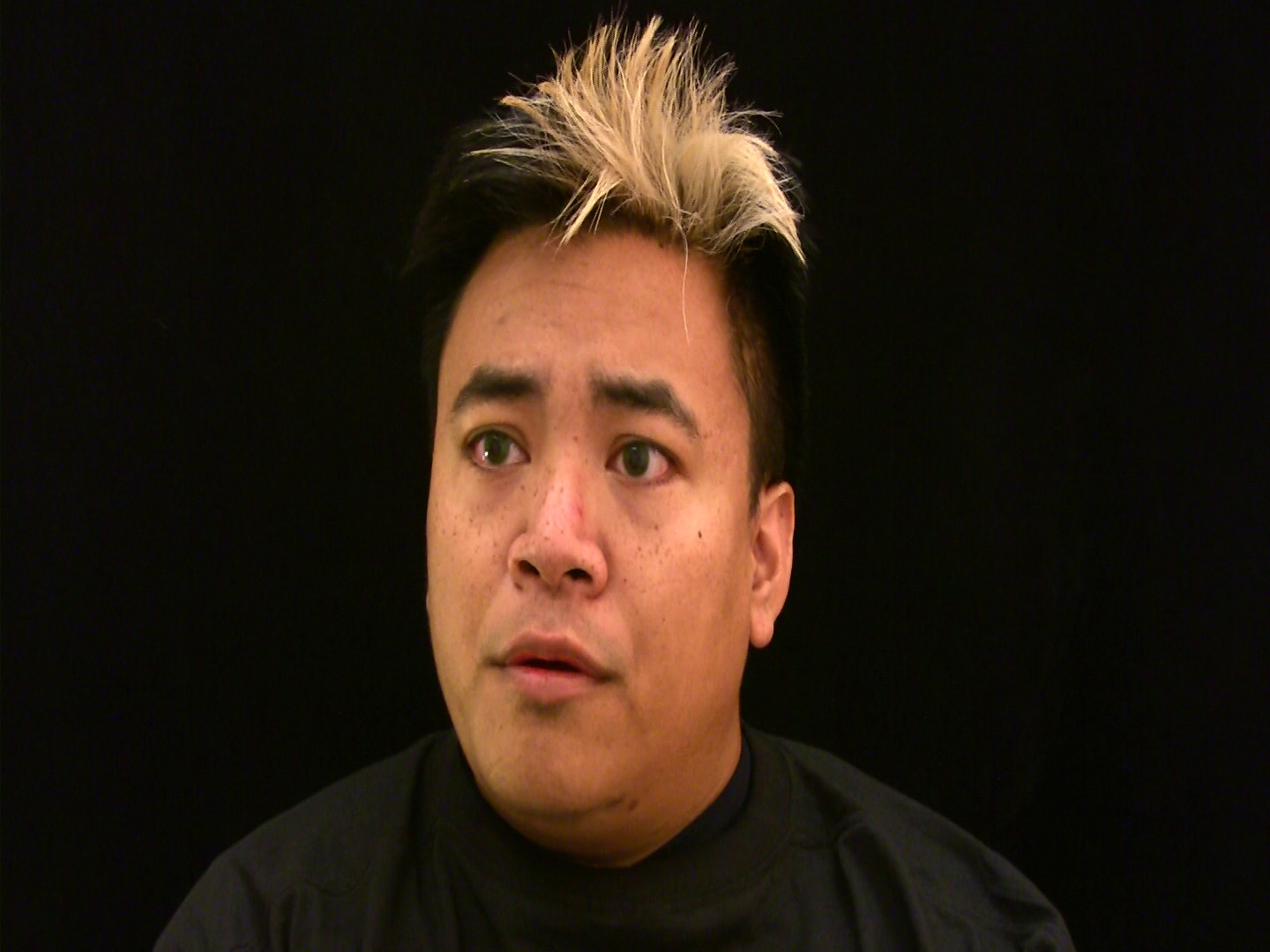} & 
\includegraphics[width=0.44\textwidth]{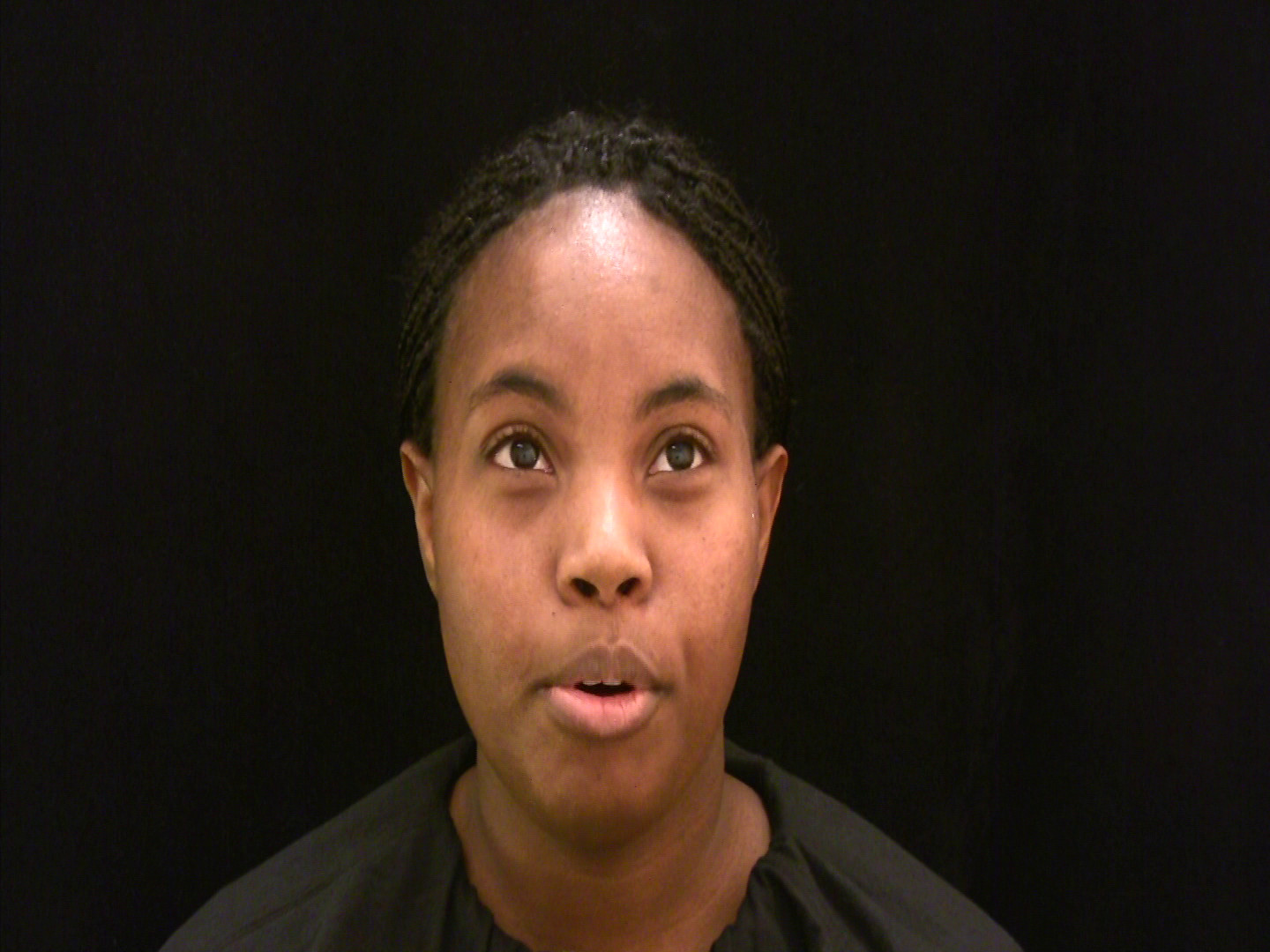} \\
\multicolumn{2}{c}{(a) $1440\times1080$, Original resolution image for S1 \& S2} \\
\includegraphics[width=0.44\textwidth]{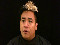} & 
\includegraphics[width=0.44\textwidth]{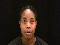} \\
\multicolumn{2}{c}{(b) $60\times45$, S1 \& S2 downsampled }\\
\includegraphics[width=0.44\textwidth]{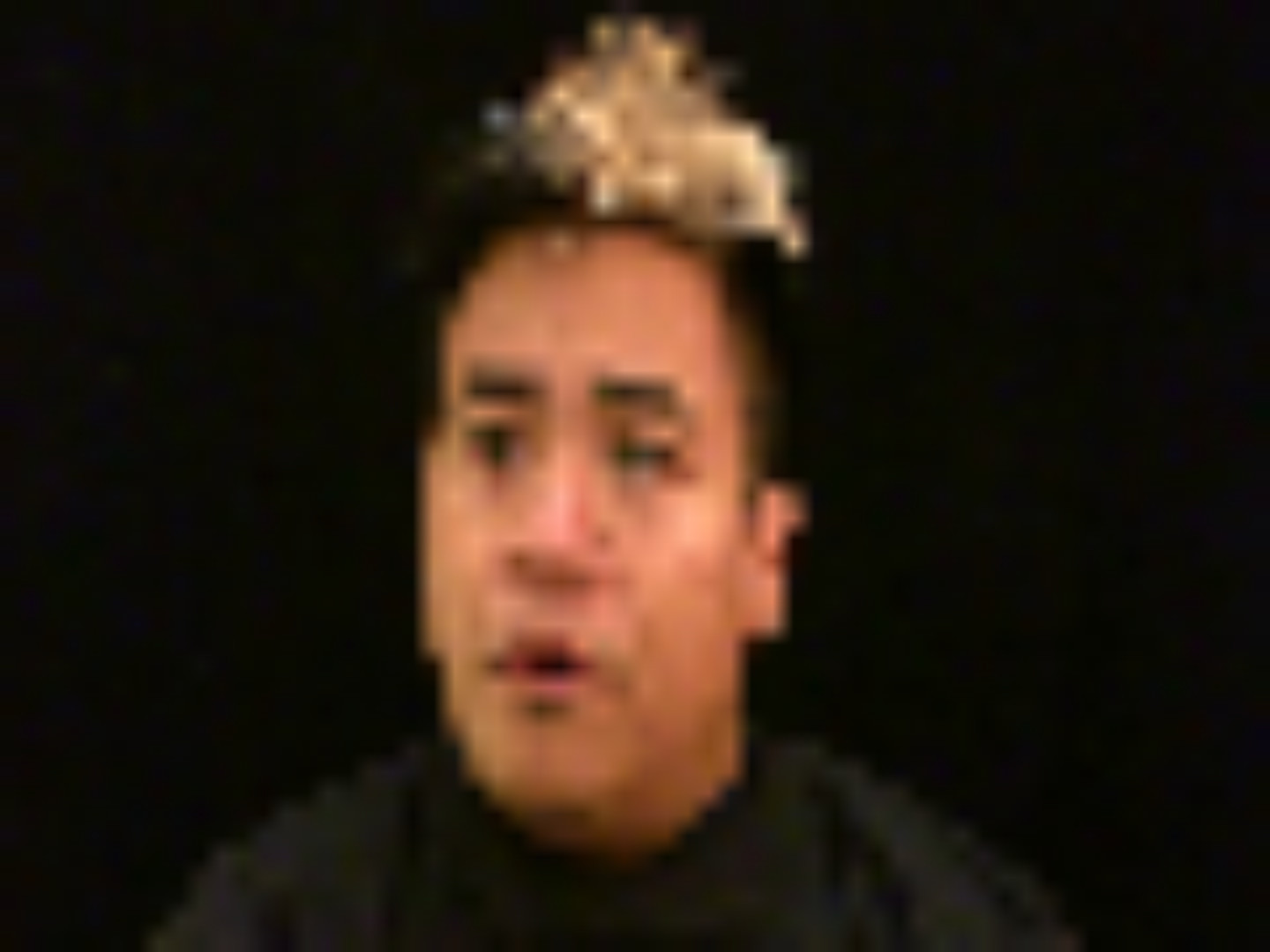}  &
\includegraphics[width=0.44\textwidth]{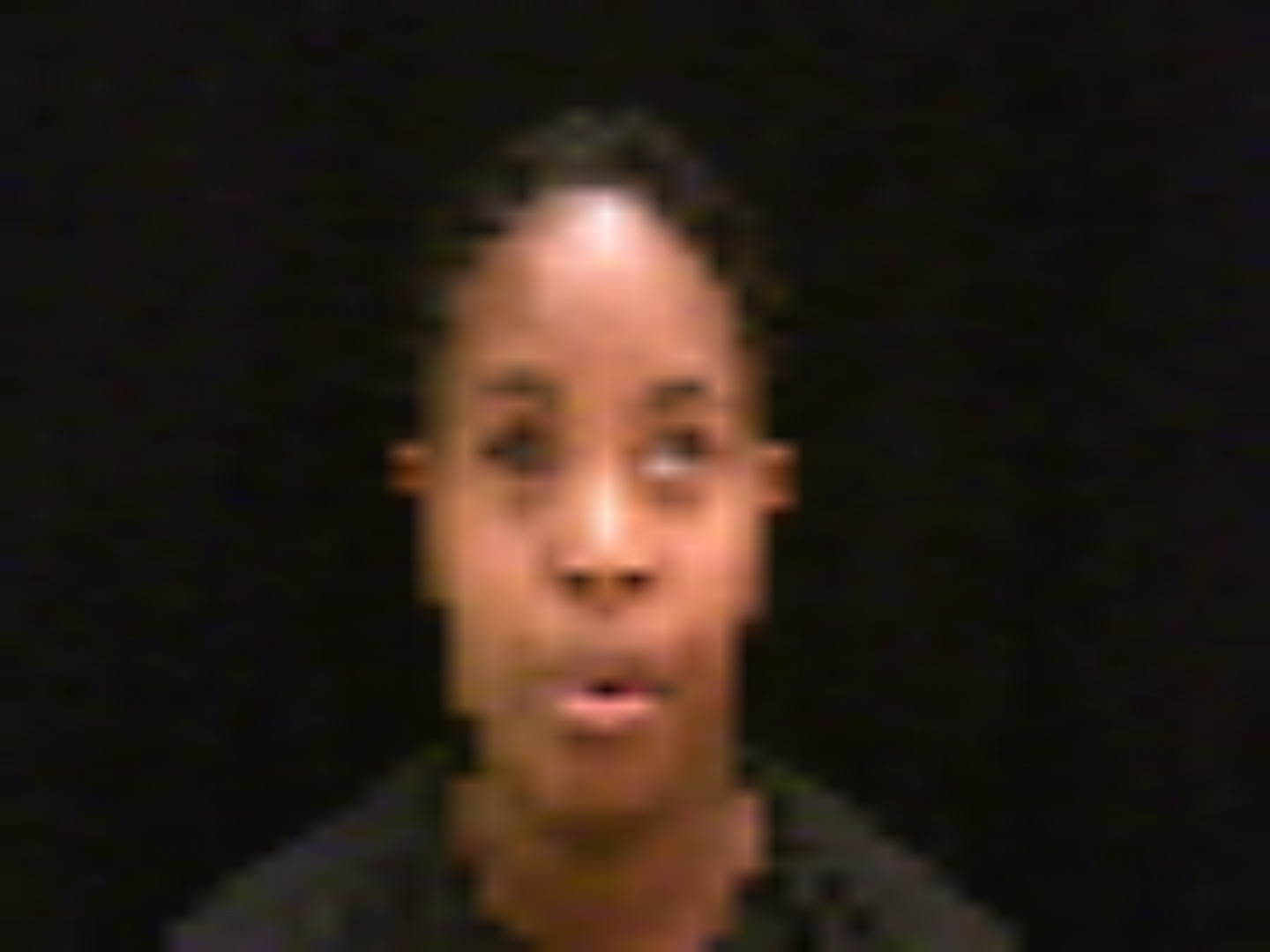}  \\ 
\multicolumn{2}{c}{(c) $1440\times1080$, S1 \& S2 restored} \\
\end{tabular} 
\caption{Downsampling of frame images in PNG format: (a) Original captured images, (b) nearest neighbour down-sampled images and (c) and their bilinear sampled restored pictures without original high definition information.} 
\label{tab:htkacckey} 
\end{figure}

The AAMs used for tracking are now decomposed into sub-models for the eyes, eyebrows, nose, face outline and lips. The purpose of this is to allow us to obtain a robust fit from the full face model but extract features of only the lip information for use during classification. Both speaker lips sub-model can be seen in Figure~\ref{fig:meshes} (c) and (d). There are $24$ landmarks in the outer lip contour and $20$ in the inner lip contour. Next, the video frames used in the high-resolution tracking were down-sampled to each of the required resolutions (listed below) by nearest neighbour sampling (Figure~\ref{tab:htkacckey}(b)) and then up-sampled via bilinear sampling (Figure~\ref{tab:htkacckey}(c)) to provide us with 18 sets of frames per original video. We use a different sampling method to upsample as this provided a more consistent visual degradation of information in the resulting images to show the reduction in resolution with minimum consistent processing artefacts compared to other sampling methods. These new frames are the same physical size as the original ($1440\times1080$) recordings but contain less information due to the downsampling i.e. only the information available at a lower resolution version of the original. 
 
\begin{multicols}{3} 
\begin{enumerate} 
\item $1440\times1080$ 
\item	 $960\times720$ 
\item	  $720\times540$ 
\item	  $ 360\times270$ 
\item	 $240\times180$ 
\item	  $180\times135$ 
\item	   $144\times108$ 
\item	    $120\times90$ 
\item	 $90\times67$ 
\item	  $80\times60$ 
\item	   $72\times54$ 
\item	    $65\times49$ 
\item	$69\times45$ 
\item	 $55\times42$ 
\item	  $51\times39$ 
\item	   $48\times36$ 
\item	 $45\times34$ 
\item	 $42\times32$ 
\end{enumerate} 
\end{multicols} 
 
We remind the reader that our point of interest in this study, is the affect low resolution has on the loss of lip-reading information, rather than the affect it would also have on the AAM tracking process. Some AAM trackers lose track quite easily at low resolutions or on lossy images and we do not wish to be overwhelmed with catastrophic errors caused by tracking issues or artefacts which can often be solved in other ways \cite{5459283}. Accordingly, this is why we have fitted at the original full resolution before the refitting of the lips sub model for feature extraction. Consequently the shape features in this experiment are unaffected by the downsampling process, whereas the appearance features vary. This will turn out to be a useful benchmark.
 
 Our image processing method is specific to our research question, what are the limitations (if any) of resolution in achieving machine lip reading? We have minimised the effects of compression artefacts by using the most successful pair of algorithms for downsampling and upsampling respectively. By using a dataset recorded in laboratory controlled conditions we have no white noise or occlusions.  There are of course other methods available to us, such as simply filling the feature vectors with zeros to represent the loss of data, or not resizing the smaller images back to the original size. But the major advantage of our method is that it encourages good tracking with the AAM and with this good tracking, we can complete a direct A to B comparison of classification outputs from features derived from videos with varying resolution information.
 
For Speaker 1 (S1), six shape and 14 appearance parameters and for Speaker 2 (S2), seven shape and 14 appearance parameters are retained. This number of parameters was chosen to retain 95\% variance in facial information in the usual way \cite{927467}, see Table~\ref{tab:rr_parameters} presented in Chapter~\ref{chap:datasets_dics}. 
 
\section{Classification method} 
 
\begin{table}[!h] 
\centering 
\caption{A phoneme-to-viseme mapping from combining Walden's consonant visemes with Montgomery's vowel visemes.} 
\begin{tabular} {|l|l||l|l|} 
\hline 
vID & Phonemes & vID & Phonemes \\ 
\hline \hline 
v01 & /p/ /b/ /m/ & v10 & /i/ /\textsci/ \\ 
v02 & /f/ /v/ & v11 & /eh/ /ae/ /ey/ /ay/\\ 
v03 & /\textipa{T}/ /\textipa{D}/ & v12 & /\textscripta/ /\textopeno/ /\textturnv/ \\ 
v04 & /t/ /d/ /n/ /k/ /g/ /h/ /j/ & v13 & /\textupsilon/ /\textrevepsilon/ /ax/ \\ 
& /\textipa{N}/ /y/ & & \\ 
v05 & /s/ /z/ & v14 & /u/ /uw/ \\ 
v06 & /l/ & v15 & /\textopeno\textsci/ \\ 
v07 & /r/ & v16 & /iy/ /hh/ \\ 
v08 & /\textipa{S}/ /\textipa{Z}/ /t\textipa{S}/ /d\textipa{Z}/ & v17 & /\textscripta\textupsilon/ /\textschwa\textupsilon/ \\ 
v09 & /w/ & v18 & /sil/ /sp/ \\ 
\hline 
\end{tabular} 
\label{tab:visememapping} 
\end{table} 
 
\begin{figure}[!ht] 
\centering 
\includegraphics[width=0.95\textwidth]{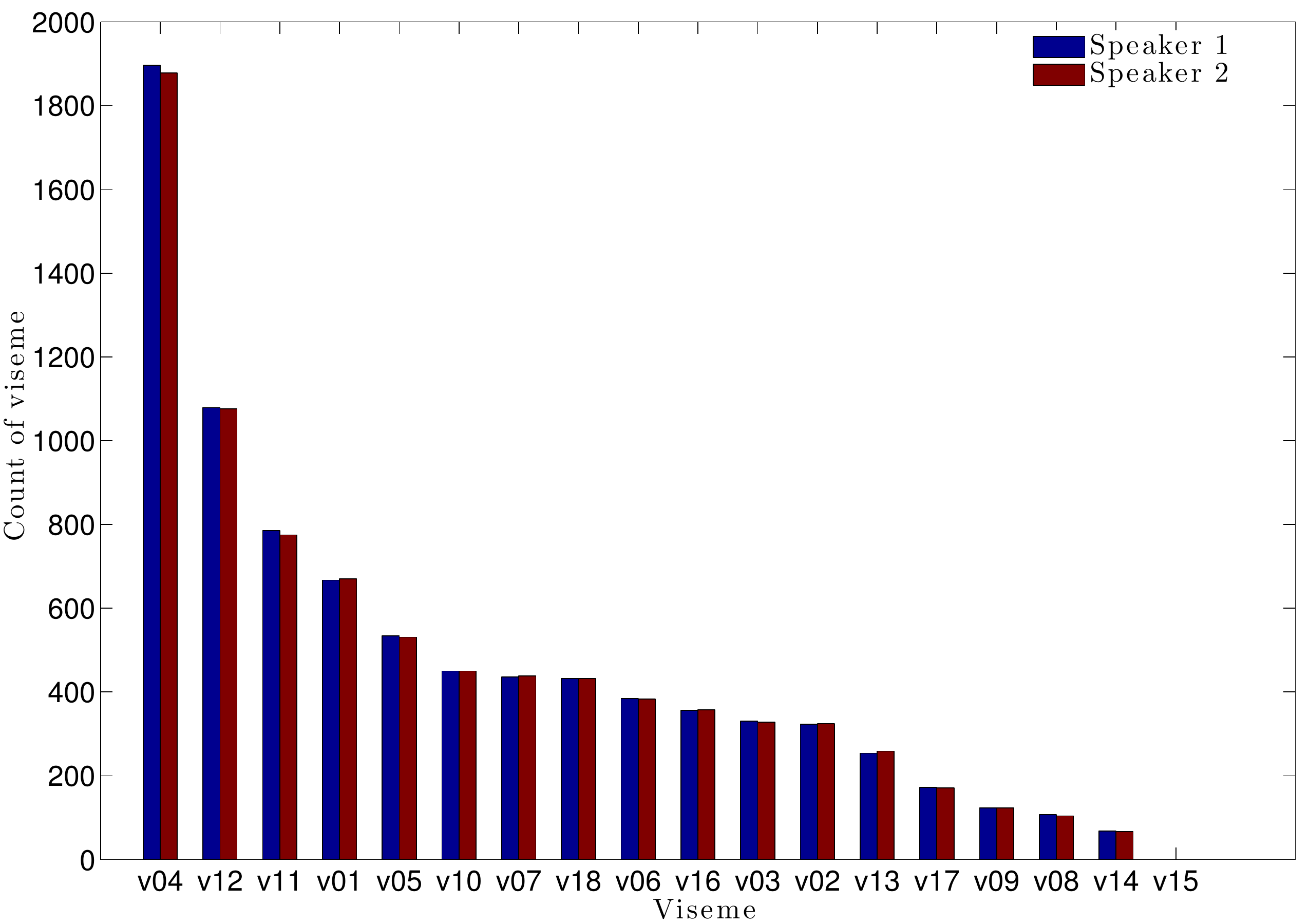} 
\caption{Occurrence frequency of visemes per speaker based upon ground truth transcripts of the Rosetta Raven dataset speakers using Walden's and Montgomery's visemes.} 
\label{fig:viseme_counts} 
\end{figure}

We listened to each recitation of the poem and produced a ground truth text (some recitations of the poem are not word-perfect to the original writing (see Appendix~\ref{The_raven})). This word transcript is converted to an American English phoneme-level transcript using the CMU pronunciation dictionary \cite{cmudict} introduced in Chapter~\ref{chap:datasets_dics}. Then, using the viseme mapping based upon Walden's consonants \cite{walden1977effects} and Montgomery \textit{et al.}'s \cite{massaro98:talking} vowel phoneme-to-viseme mapping (as in Table~\ref{tab:visememapping}), a viseme transcript was created. Thus we have translated each recitation from words, to phonemes, and finally, to visemes. Viseme classification is selected over phonemes as, on a small data set, it has the benefits of reducing the number of classifiers needed and increasing the training data available for each viseme classifier. Note not all visemes are equally represented in the data as is shown by the viseme histogram in Figure~\ref{fig:viseme_counts}, Chapter~\ref{chap:datasets_dics}. Whist the volumes in this Figure are lower than an equivalent histogram for a continuous speech dataset, the distributions are similar. 
 
For each speaker, a test fold is randomly selected as 42 of the 108 lines (20\% of data) in the poem. The remaining lines (80\% of data) are used as the training fold. Repeating this five times gives five-fold cross-validation. Note visemes cannot be equally represented in all folds. 
 
For classification Hidden Markov Models (HMMs) are built with the Hidden Markov Toolkit (HTK)~\cite{young2006htk} already introduced in Section~\ref{sec:htk}. An HMM is initialised using the `flat start' method (using \texttt{HCompV}), with a prototype of five states and five mixture components, and the information in the training samples. Five states and five mixtures are selected based upon the work in \cite{982900}. An HMM is defined for each viseme plus silence and short-pause labels (Table~\ref{tab:visememapping}) and we re-estimate the HMM parameters four times with no pruning. 
 
The HTK tool \texttt{HHEd} ties together the short-pause and silence models between states two and three before re-estimating the HMMs a further two times. Then \texttt{HVite} is used with the \texttt{-m} flag to force-align the data using the word transcript. We create a viseme version of the CMU dictionary for word-to-viseme mapping (whereby the phonemes are replaced with their respective viseme characters from the phoneme-to-viseme map in Table~\ref{tab:visememapping}) and use this viseme CMU dictionary to produce a time-aligned viseme transcription which includes natural breakpoints between words. 
 
The HMMs are now re-estimated twice more. However, now the force-aligned viseme transcript replaces the original viseme transcript used in the previous HMM re-estimations. A word network is needed to complete the classification. \texttt{HLStats} and \texttt{HBuild} used together twice make both a Unigram Word-level Network (UWN) and a Bigram Word-level Network (BWN). Finally, \texttt{HVite} is used with the different network support for the classification task and \texttt{HResults} gives us the correctness and accuracy values. All HTK tools named here are described in Chapter~\ref{sec:htk}.
 
\section{Analysis of resolution affects on classification} 
 
Accuracy, $A$, (Equation~\ref{eq:accuracy}), is selected as a measure rather than correctness, $C$, (Equation~\ref{eq:correctness}) since it accounts for all errors. Including insertion errors is important as they are notoriously common in lip-reading. An insertion error occurs when the recogniser output has extra words/visemes not present in the original transcript \cite{young2006htk}. As an example one could say, ``Once upon a midnight dreary'', 

but the recogniser outputs:

``Once upon upon midnight dreary dreary". 

Here the recogniser has inserted two words which were never present, 

``Once upon \textbf{upon} midnight dreary \textbf{dreary}"

and it has deleted one (`a'). The missing `a' is a deletion error. 

``Once upon \textbf{...} midnight dreary". 
 
\begin{figure} 
\includegraphics[width=0.95\textwidth]{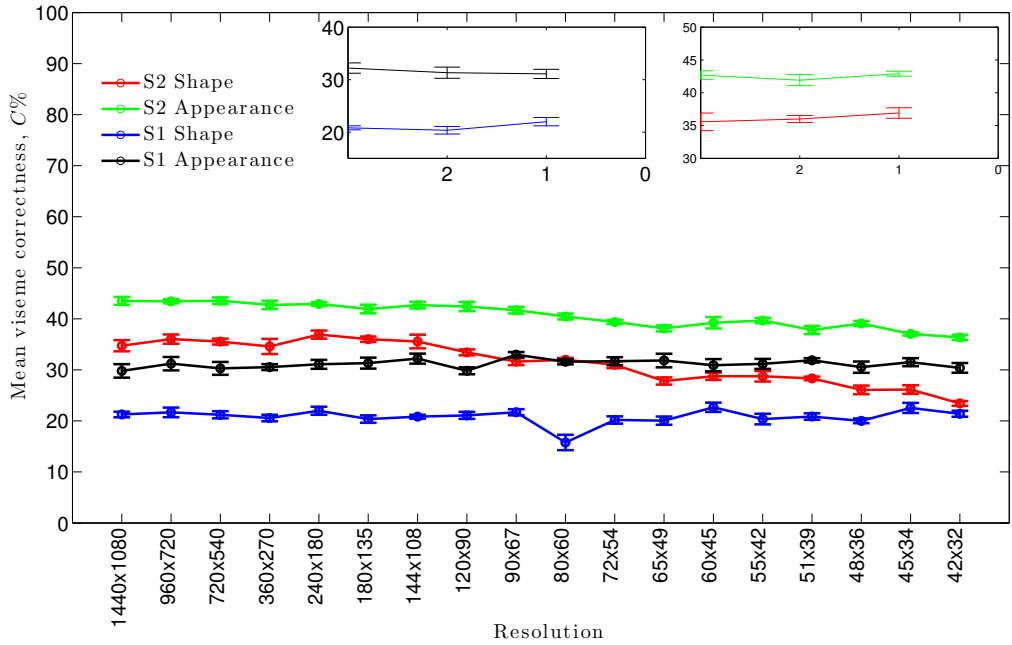} 
\caption{Viseme classification in Correctness, $C\pm1\frac{\sigma}{\sqrt{5}}$, with a unigram word network (on the $y$-axis) at 18 degraded  measured in pixels ($x$-axis).} 
\label{fig:resolutionunigram} 
\end{figure}

\begin{figure}  
\includegraphics[width=0.95\textwidth]{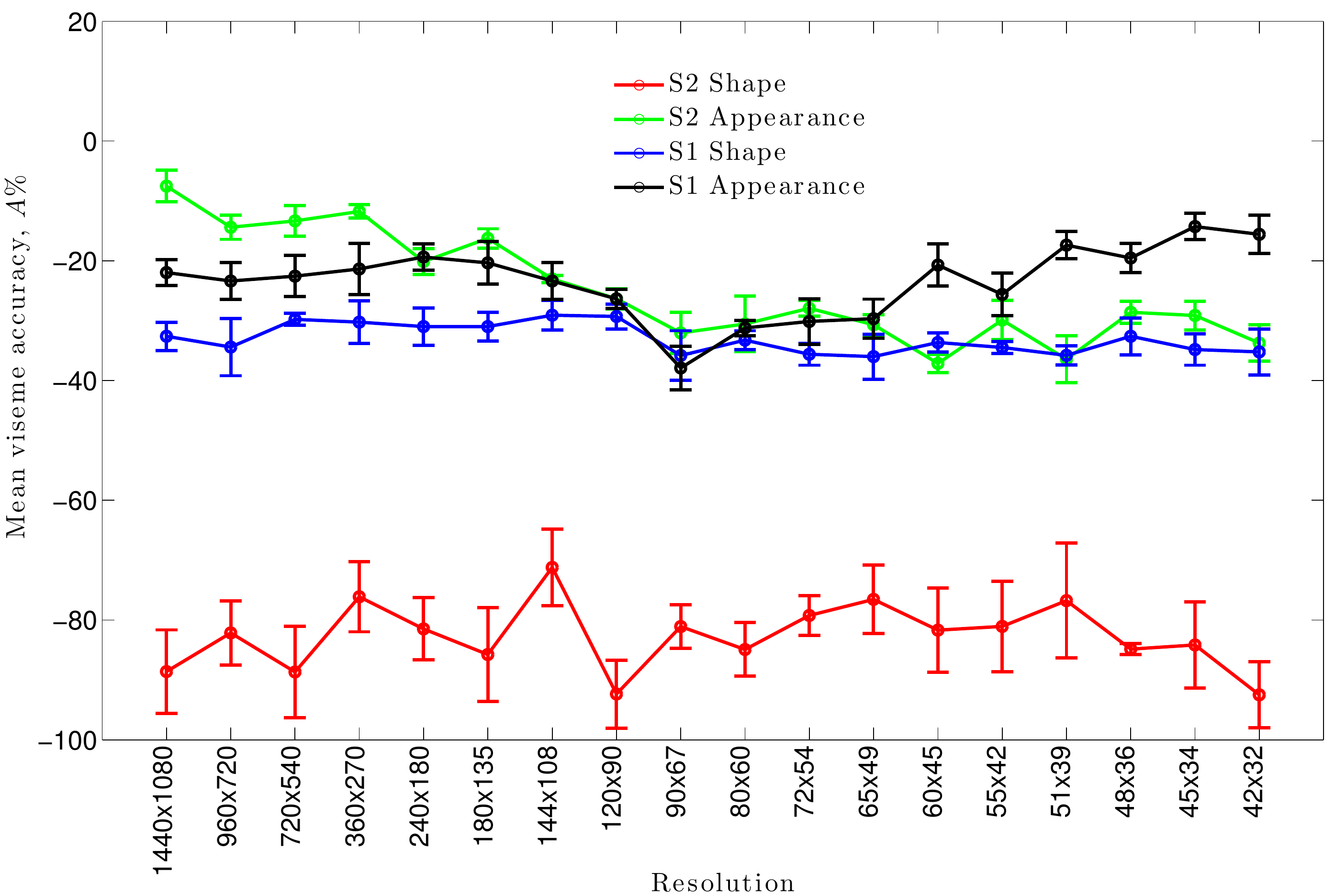} 
\caption{Viseme classification in Accuracy, $A\pm1\frac{\sigma}{\sqrt{5}}$, with a unigram word network (on the $y$-axis) at 18 degraded resolutions in pixels ($x$-axis).} 
\label{fig:resolutionunigram1} 
\end{figure}

\begin{figure} 
\includegraphics[width=0.95\textwidth]{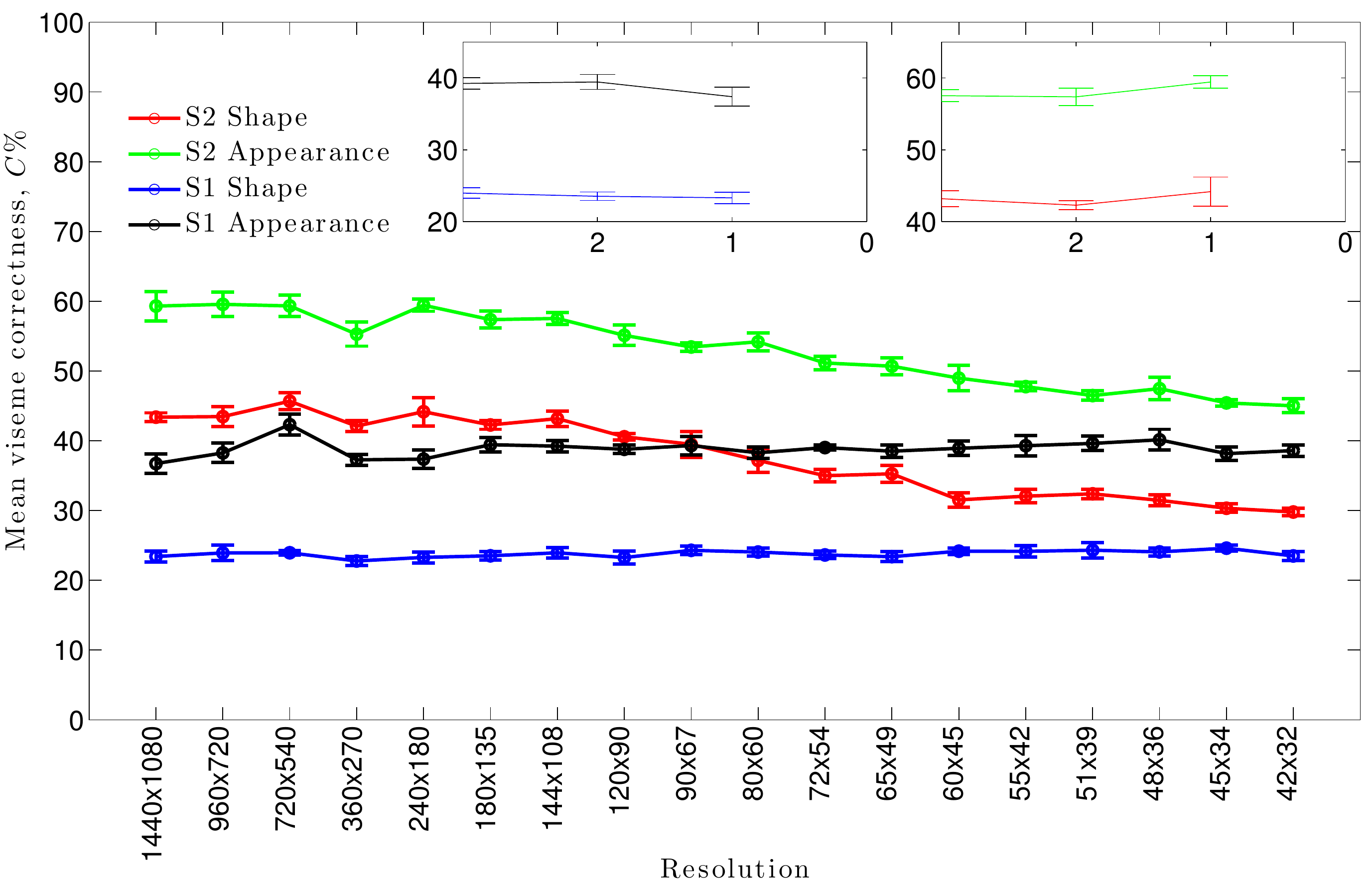} 
\caption{Viseme classification in Correctness, $C\pm1\frac{\sigma}{\sqrt{5}}$, with a bigram word network (on the $y$-axis) at 18 degraded resolutions in pixels ($x$-axis).} 
\label{fig:resolutionbigram} 
\end{figure}

\begin{figure} 
\includegraphics[width=0.95\textwidth]{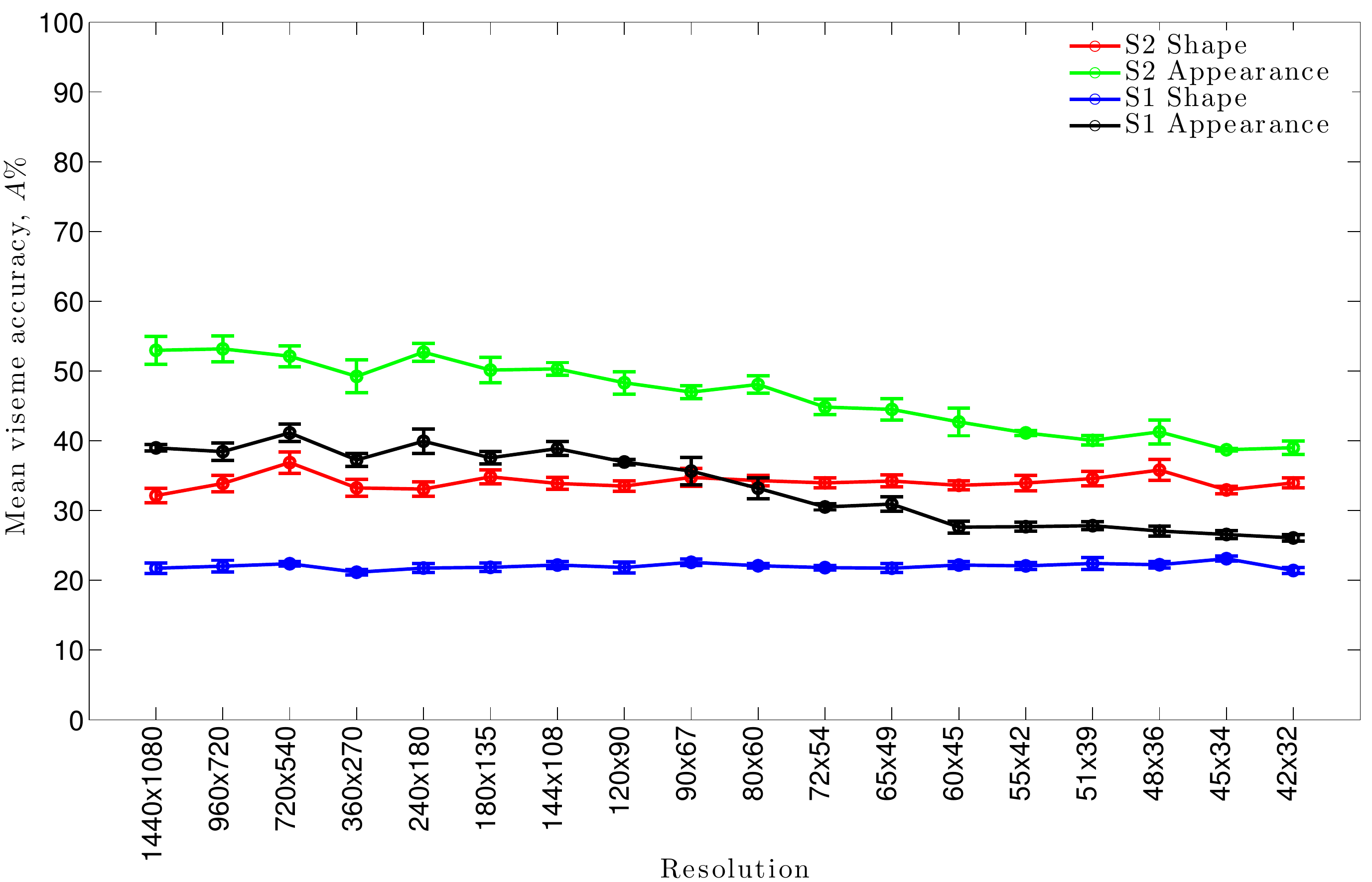} 
\caption{Viseme classification in Accuracy, $A\pm1\frac{\sigma}{\sqrt{5}}$, with a bigram word network (on the $y$-axis) at 18 degraded resolutions in pixels ($x$-axis).} 
\label{fig:resolutionbigram1} 
\end{figure}

%show correctness graphs so we can compare to the accuracy results 
In Figures~\ref{fig:resolutionunigram} and~\ref{fig:resolutionbigram} we have plotted, for our 18 different resolutions along the $x$-axis, the mean viseme correctness on the $y$-axis for each speaker. Supported by a unigram language network and bigram language network respectively. Speaker 1 shape classification is shown in blue and appearance classification in black. Speaker 2 shape and appearance classification is plotted in red and green respectively. The corresponding graphs of mean accuracy classification are shown in Figures~\ref{fig:resolutionunigram1} and~\ref{fig:resolutionbigram1}. All four figures include one standard error over the five folds. 

Figure~\ref{fig:resolutionunigram1} plots viseme accuracy with a unigram network on the $y$-axis and all points are negative values. This is worse than chance and demonstrates the debilitating effect of insertion errors where the language network is not strong enough to sieve them out of the classification output. Viseme correctness supported by a unigram word network is shown in Figure~\ref{fig:resolutionunigram}, where we see a slow but significant decrease in classification as the resolutions decrease in size along the $x$-axis. At no point do the appearance features drop below the shape features. This trend is matched in our BWN experiments in Figures~\ref{fig:resolutionbigram} and ~\ref{fig:resolutionbigram1}. 

\begin{figure}[!ht] 
\centering 
\setlength{\tabcolsep}{1pt} 
\begin{tabular}{c c} 
\includegraphics[width=0.35\textwidth]{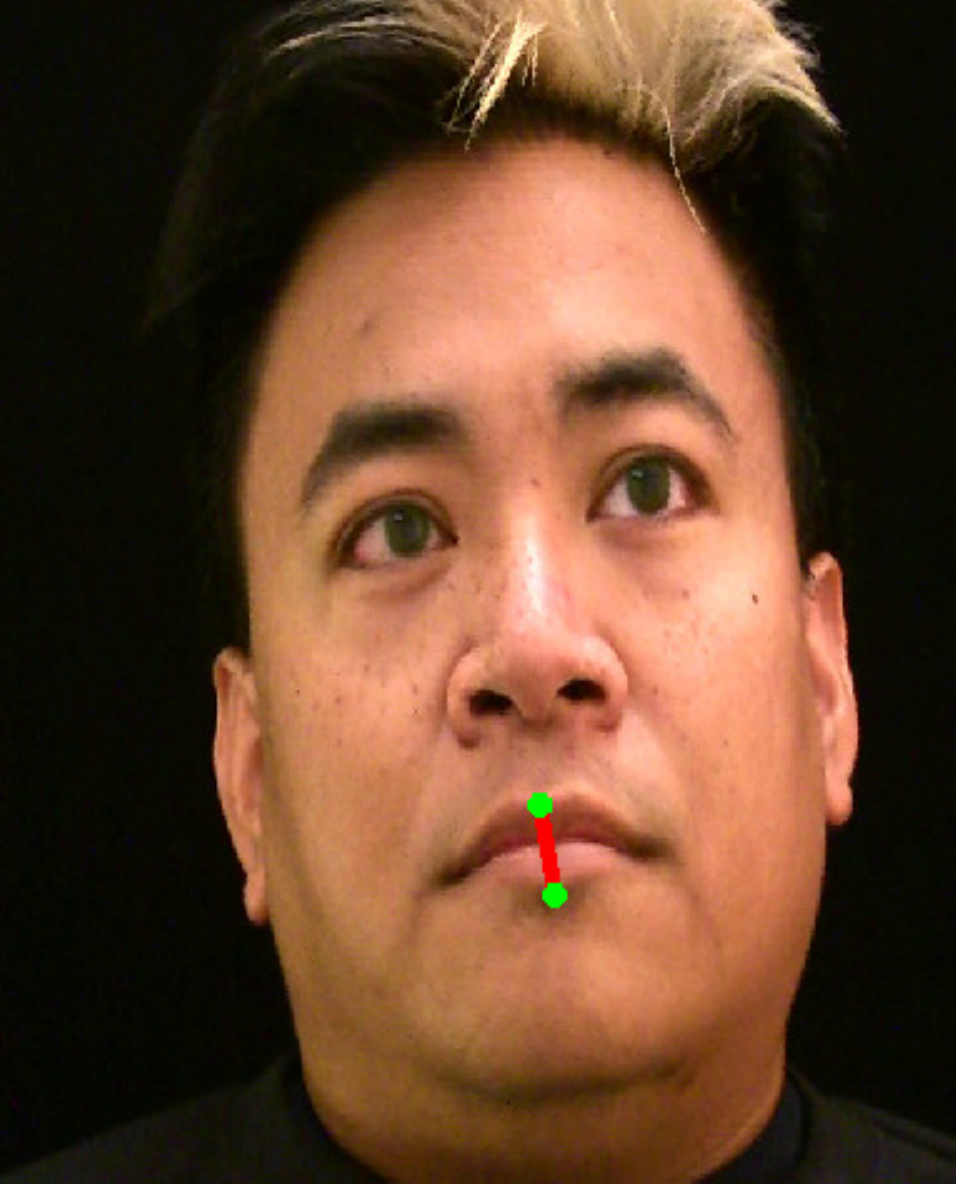} &
\includegraphics[width=0.35\textwidth]{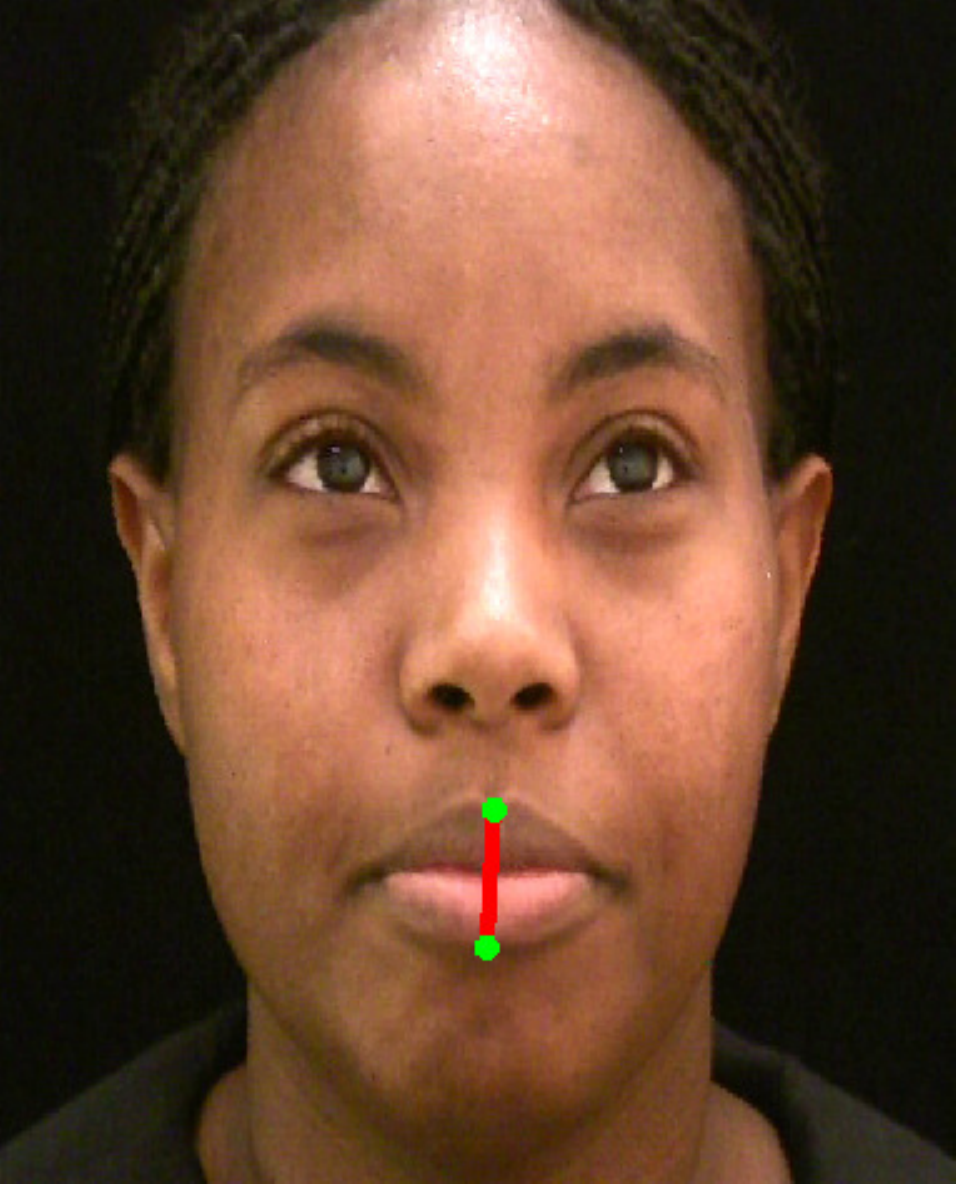} \\ 
(a) S1 & (b) S2 \\ 
\end{tabular} 
\caption{Showing the resting lip-pixel distance measures for two Rosetta Raven speakers.} 
\label{fig:restingpixeldist} 
\end{figure}

\begin{figure} [h]
\includegraphics[width=0.95\textwidth]{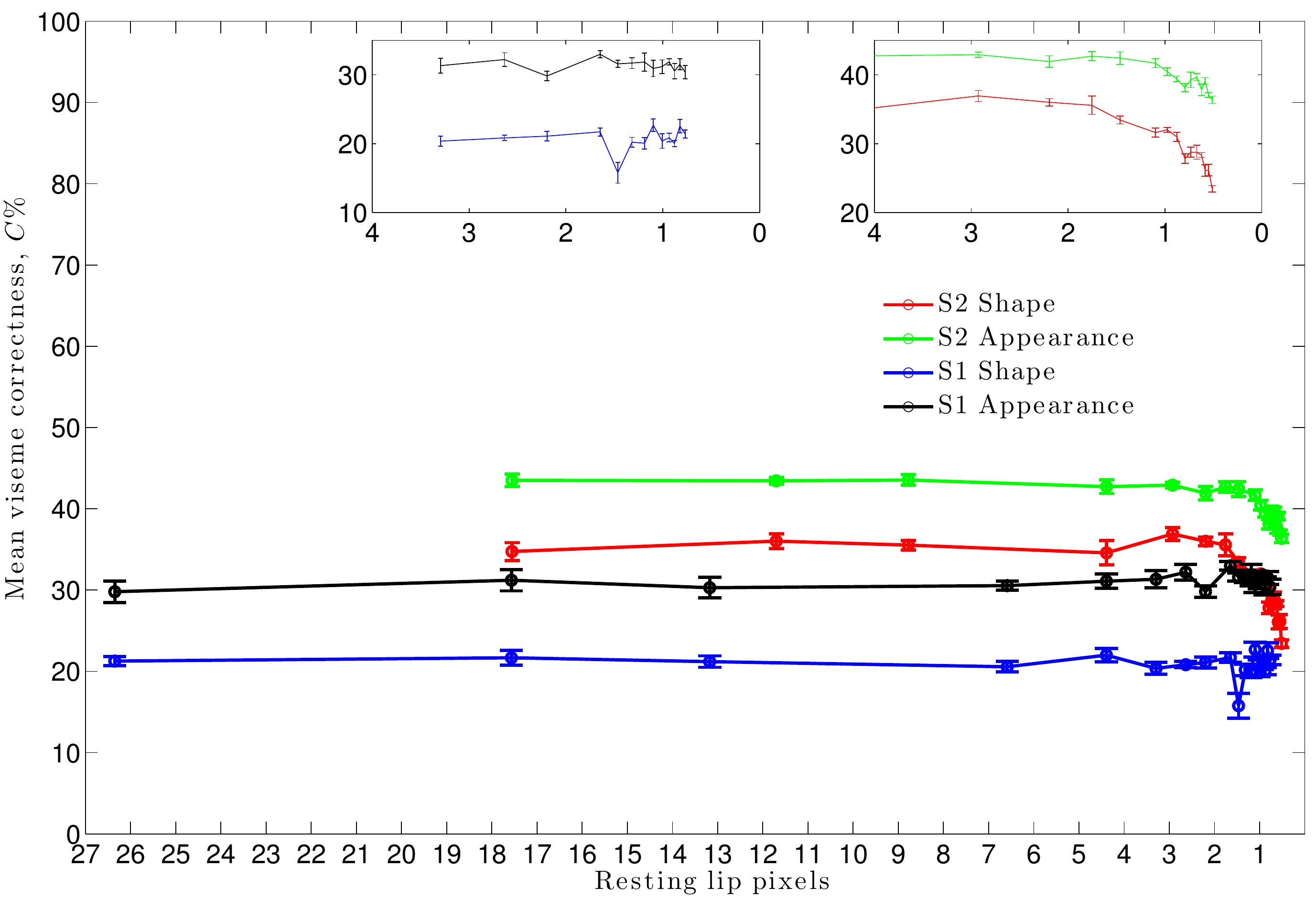} 
\caption{Viseme classification in Correctness, $C\pm1\frac{\sigma}{\sqrt{5}}$, with a unigram word network (on the $y$-axis) by vertical resting lip height in pixels ($x$-axis).} 
\label{fig:HTKAccUEPS} 
\end{figure}

\begin{figure} [h]
\includegraphics[width=0.95\textwidth]{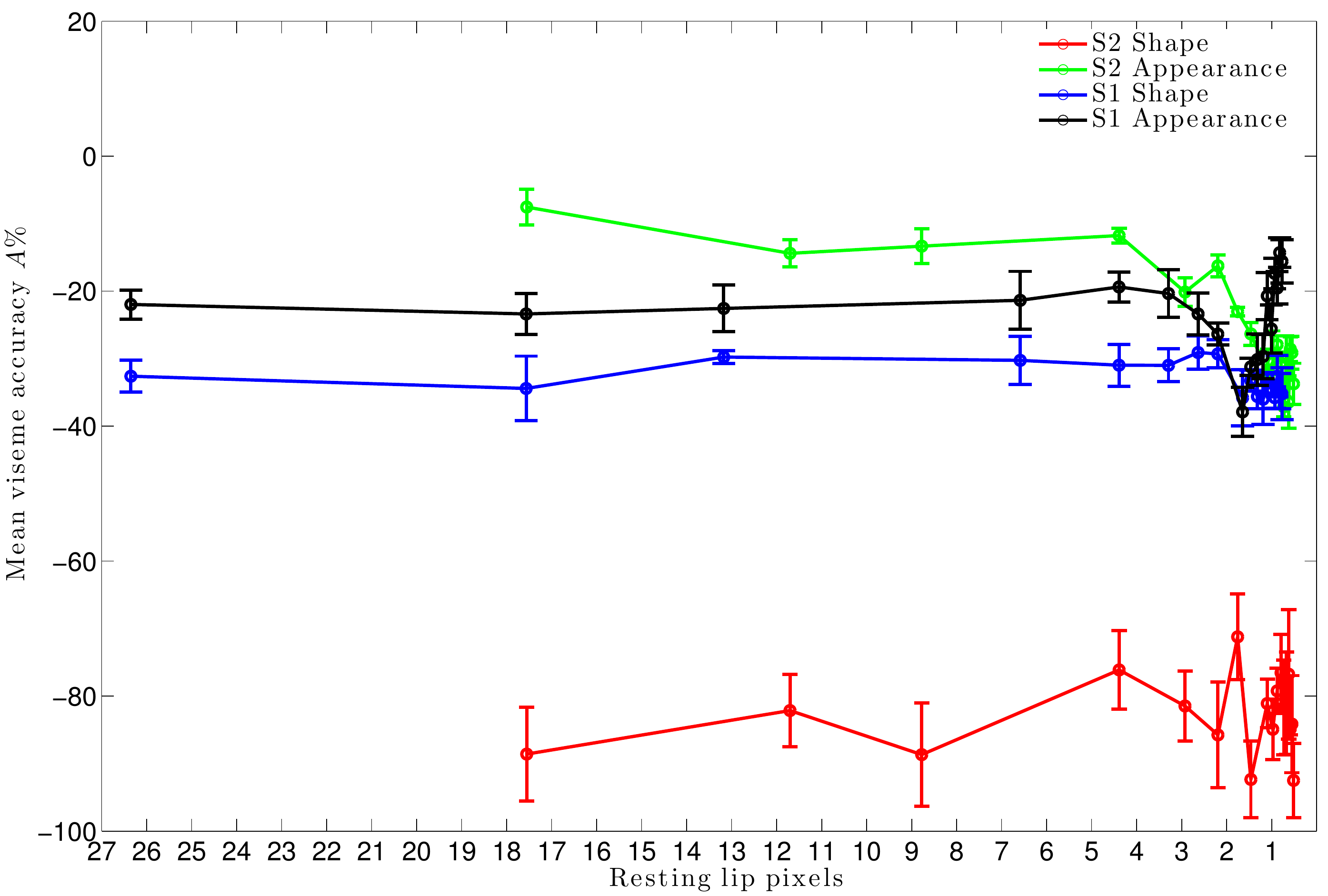}  
\caption{Viseme classification in Accuracy, $A\pm1\frac{\sigma}{\sqrt{5}}$, with a unigram word network (on the $y$-axis) by vertical resting lip height in pixels ($x$-axis).} 
\label{fig:HTKAccUEPS1} 
\end{figure}

\begin{figure} [h]
\includegraphics[width=0.95\textwidth]{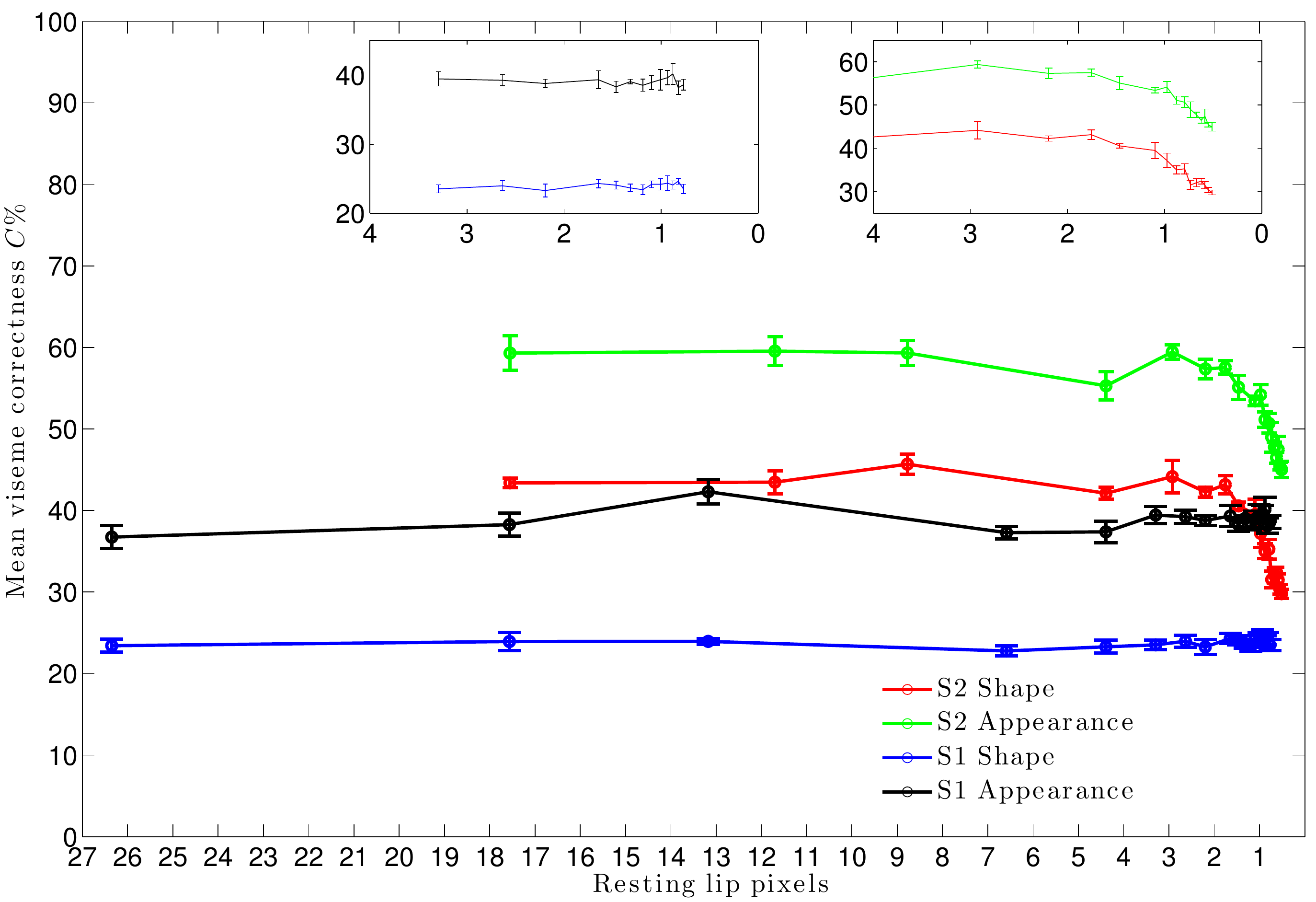} 
\caption{Viseme classification in Correctness, $C\pm1\frac{\sigma}{\sqrt{5}}$, with a bigram word network (on the $y$-axis) by vertical resting lip height in pixels ($x$-axis).} 
\label{fig:HTKAccBEPS} 
\end{figure}

\begin{figure} [h]
\includegraphics[width=0.95\textwidth]{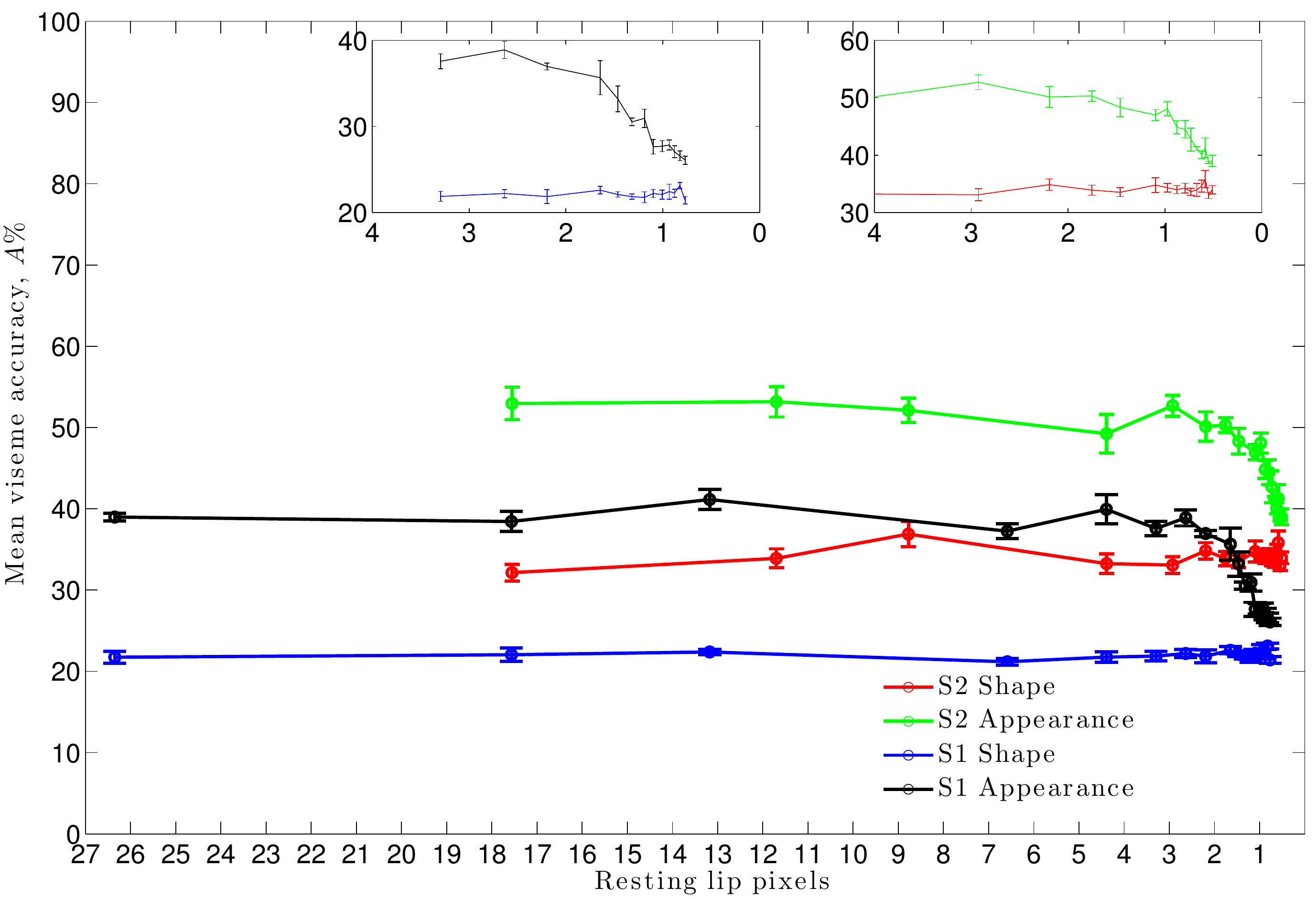} 
\caption{Viseme classification in Accuracy, $A\pm1\frac{\sigma}{\sqrt{5}}$, with a bigram word network (on the $y$-axis) by vertical resting lip height in pixels ($x$-axis).} 
\label{fig:HTKAccBEPS1} 
\end{figure}

These Figures, however, are not normalised to account for the actual differences in information between resolutions. As we can see in our list of resolutions in Section~\ref{sec:resolution}, there is not an equal interval between each size. Therefore we replot these results by measuring the resting lip-pixels which cover the lip-shape. The resting lip pixel distance is shown in Figure~\ref{fig:restingpixeldist} for our two speakers in the first $1080\times1440$ resolution image frame. This means, as there are less pixels per lip we can appropriately plot along our $x$-axis as we have done in Figures~\ref{fig:HTKAccUEPS},~\ref{fig:HTKAccUEPS1},~\ref{fig:HTKAccBEPS} and~\ref{fig:HTKAccBEPS1}.

 

Figure~\ref{fig:HTKAccUEPS1} shows the accuracy, $A$, (on the $y$-axis) versus resolution (on the $x$-axis) for an UWN. The $x$-axis is calibrated by the vertical height of the lips of each speaker in their rest position (Figure~\ref{fig:restingpixeldist}). For example, at the maximum resolution of $1440\times1080$ speaker S1 has a lip-height of approximately 26 pixels in the rest position whereas S2 has a lip-height of approximately 17 pixels. The worst performance is from speaker S2 using shape-only features. The shape features do not vary with resolution so any variation in this curve is due to the cross-fold validation error (all folds do not contain all visemes equally). Nevertheless, the variation is within one standard error, and so not signifiant. This is not a surprise as AAM shape features are scale invariant. The poor performance is, as usual with lip-reading, dominated by insertion errors (hence the negative $A$ values in Figure~\ref{fig:HTKAccUEPS1}). The usual explanation for this effect is shape data contain a few characteristic shapes (which are easily recognised) in a sea of indistinct shapes - it is easier for a classifier to insert garbage symbols than it is to learn the duration of a symbol which has an indistinct start and end shape due to co-articulation. We suggest that speaker S1 has more distinctive shapes so scores better on the shape feature as more distinctive shapes between classification models differentiate more definitively. 
 
However, it is the appearance features which are of more interest since this varies as we downsample. At resolutions lower than four pixels it is difficult to be confident the shape information is effective. However, the basic problem is a very high error rate (shown in Figures~\ref{fig:HTKAccUEPS} and~\ref{fig:HTKAccUEPS1}) therefore a more supportive word model is required \cite{howellPhD}. 
 
Figures~\ref{fig:HTKAccBEPS} and~\ref{fig:HTKAccBEPS1} shows the classification accuracy versus resolution (represented by the same $x$-axis calibration in Figures~\ref{fig:HTKAccUEPS} and~\ref{fig:HTKAccUEPS1}) for a BWN. It also includes two sub-plots which magnify the right-most part of the graph. Again, the shape models perform worse than the appearance models, but looking at the magnified plots, appearance never becomes as poor as shape performance even at very low resolutions. As with the UWN accuracies, there is a clear inflection point at around four pixels (at two pixels per lip), and by two pixels the performance has declined significantly. 
 
\begin{table}[h] 
\centering 
\caption{Insertion, deletion and substitution error counts in classification transcripts at the smallest resolution above (before), and the largest resolution below (after), the minimum required lip pixel height of two pixels per lip. The values are the total sum over all five folds of cross validation.} 
\begin{tabular} {|l|r|r|r|} 
\hline 
& Insertion & Deletion & Substitution \\ 
\hline \hline 
Speaker 1: & & &\\ 
Before & 348 & 3,385 & 1,298 \\ 
After & 305 & 3,646 & 1,355\\ 
\% change & $-12\%$ & $+8\%$ & $+4\%$ \\
\hline 
Speaker 2: & & &\\ 
Before & 571 & 2,339 & 1,423 \\ 
After & 531 & 2,322 & 1,500 \\ 
\% change & $-7\%$ & $-1\%$ & $+5\%$ \\
\hline 
\end{tabular} 
\label{tab:errors} 
\end{table} 
 
In Table~\ref{tab:errors} we have listed the different error types (insertion, deletions and substitutions) which can occur during classification for resolutions just before our identified minimum lip pixel threshold as well as just after. The values are the total errors over all five folds of cross validation. For Speaker 1, both deletion and substitution errors increase when there is no longer have enough pixels to differentiate between the two lips. For Speaker 2, we see only the substitution errors increase but the deletion errors only decrease insignificantly at $-1\%$. 

It is interesting to see there are fewer insertion errors after our minimum lip-pixel threshold. In Chapter~\ref{chap:featuretypes} we saw the difference between Accuracy (Equation~\ref{eq:accuracy}) and Correctness (Equation~\ref{eq:correctness}) were the Insertion errors. Therefore, we can say we may need more visemes within a set to keep insertion errors down as these ensure more minor differences between classifiers are encapsulated within training. %However, as previous Figures~\ref{fig:HTKAccBEPS} a \& b show, these are so badly trained due to lack of training samples, they now reduce the overall classification to impossible. 

\section{The effect of resolution on lip-reading classifiers} 

In Chapter~\ref{sec:current_diff} we discussed the limitations in machine lip-reading. In this chapter we have added to this knowledge with our experiment into resolution. 

Using the new Rosetta Raven data we have shown lip-reading HMM classifiers to have a threshold effect with resolution. We have trained and tested viseme classifiers and measured the effect on classification accuracy as we systematically reduced the resolution information in a video. The best recognition achieved was 59.55\% accuracy with Speaker 2's appearance data with a bigram word level language model, as this is the first time this dataset has been used this is the baseline for future uses. 

Contrary to common assumption and practice, the unexpected observation here is the remarkable resilience to resolution in machine lip-reading. Given modern experiments in lip-reading usually take place with high-resolution video (\cite{bowden2013recent} for example) the disparity between measured performance (shown here) and assumed performance is very remarkable. Our results show for successful lip-reading one needs a minimum of four pixels (two pixels per lip) across the closed lips. 

The realisation of a minimum number of pixels per lip is a new piece of information in the area of machine lip-reading. Previous research in this area \cite{ACP:ACP371, heckmann2003effects} has focused on noisy images and the effect of noise on word error rates in audio-visual speech recognition system. In these experiments, we see corroborating results to support the premise that with less information then lip-reading is negatively affected, but also that there is an lower bound resolution which is essential for good lip-reading. 

It must not be forgotten a higher resolution video is beneficial for the tracking task but, as previous work demonstrates, other factors considered to negatively effect lip-reading classification such as off-axis views~\cite{6298439}, actually have the ability to improve performance and here we see that a lower resolution video is not as detrimental as first assumed. 

We therefore conclude that, for real situations, the limitations on lip-reading are not likely to come from factors to do with the environment. Rather, the poor performance of lip-reading is almost certainly to do with limitations in the signal - the lip-signal is very challenging to decode and what is needed is a better understanding of the visual signal, its components, and how they can be learnt. For this reason, we now turn to the problem of understanding visemes. 
 % ICIP: Resolution work and & (combined models on RR viseme comparison to see if this drops at 6 visemes or 7 (due to appearance data) - i.e. does the shape component of an combined HMM negatively impact recognition or if combined improves?),
%!TEX root = main.tex 
\chapter[A performance evaluation of visemes]{A performance evaluation of visemes} 
\label{chap:contributions} 

This chapter is our first investigation into understanding visemes. Before we undertake complicated experiments and attempt to re-design or augment visemes, it is useful to understand what we can with what we have already tested. Currently we always use a whole set of visemes to include a large number of phonemes. But it would be nice to know:

\begin{itemize}
\item if all visemes contribute equally to the classification? If no, which of the visemes within the set are most useful?
\item Are there any visemes which are not helpful, or in fact, detrimental? And,
\item can we evaluate the performance of each viseme in isolation to understand more about the set of classes as a whole?
\end{itemize}
  
Therefore, this chapter describes an investigation into the difference in the contribution to accuracy of each viseme within a set. An analysis of the confusion matrices produced during viseme classification, obtained by comparing the classification output with the ground truth transcript, both of which are time-aligned, provides us with measurements of viseme contributions to classification. This enables us to compare each viseme within a set to all others and determine which contributes the most for accurate machine lip-reading. 

Additionally the balance between shape and appearance viseme probabilities are reviewed to see which type of feature (shape or appearance) contributes most to classification. We can also compare visual classification to audio using the same viseme classifier labels on audio features (we use MFCCs). This demonstrates a relationship between viseme classification accuracy and the spread of individual viseme contribution to classification. 
 
 \section{Measuring the contribution of individual visemes} 
 
The point of interest in this chapter is in the contribution of each viseme to the classification performance. This work searches for any particular viseme (or subgroup of phonemes) which contributes more to the classification accuracy. 
 
This study continues with the Rosetta Raven features extracted in Section~\ref{sec:resolution}. Short datasets, such as these, may not provide adequate training examples of all visemes. So we group the untrainable visemes into a single garbage viseme. In this case we estimate 150 samples as the minimum threshold (the mean training samples per viseme minus 1.5 standard error) to mitigate the bias caused by variation in training samples per classifier. Thus, visemes $/v08/$, $/v09/$, $/v14/$ and $/v15/$ are grouped giving Table~\ref{tab:newMap}. We have already reviewed the original dataset in Chapter~\ref{chap:datasets_dics}, and Figure~\ref{fig:viseme_counts} shows the occurrence of visemes listed in the original phoneme-to-viseme map (see Table~\ref{tab:visememapping}).
 
\begin{table} [th] 
\vspace{2mm} 
\caption{\label{tab:newMap} {Modified phoneme-to-viseme mapping due to lack of training data per viseme available in the Rosetta Raven dataset.}} 
\centerline{ 
\begin{tabular} {|l|l||l|l|} 
\hline 
vID & Phonemes & vID & Phonemes \\ 
\hline \hline 
v01 & /p/ /b/ /m/ & v11 & /eh/ /ae/ /ey/ /ay/ \\ 
v02 & /f/ /v/ & v12 & /\textscripta/ /\textopeno/ /\textturnv/ \\ 
v03 & /\textipa{T}/ /\textipa{D}/ & v13 & /\textupsilon/ /\textrevepsilon/ /ax/ \\ 
v04 & /t/ /d/ /n/ /k/ /g/ /h/ /j/ & v16 & /iy/ /hh/ \\ 
& /\textipa{N}/ /y/ & v17 & /\textscripta\textupsilon/ /\textschwa\textupsilon/ \\ 
v05 & /s/ /z/ & v18 & silence \\ 
v06 & /l/ & gar & /u/ /uw/ /\textopeno\textsci/ /w/ /\textipa{S}/ \\ 
v07 & /r/ & & /\textipa{Z}/ /t\textipa{S}/ /d\textipa{Z}/ \\ 
v10 & /i/ /\textsci/ & & \\ 
\hline 
\end{tabular}} 
\end{table} 
 
 The classification method used is identical to the method in Chapter~\ref{chap:five}, the methodology varies in the analysis of the classification outputs. 
  
 Values from the \texttt{HResults} confusion matrices are extracted for analysis. For each viseme we have calculated the probability of its classification $\mbox{Pr}\{v|\hat{v}\}$. 

\section{Analysis of viseme contribution} 
 
\begin{figure}[!th] 
\centering 
\includegraphics[width=0.95\textwidth]{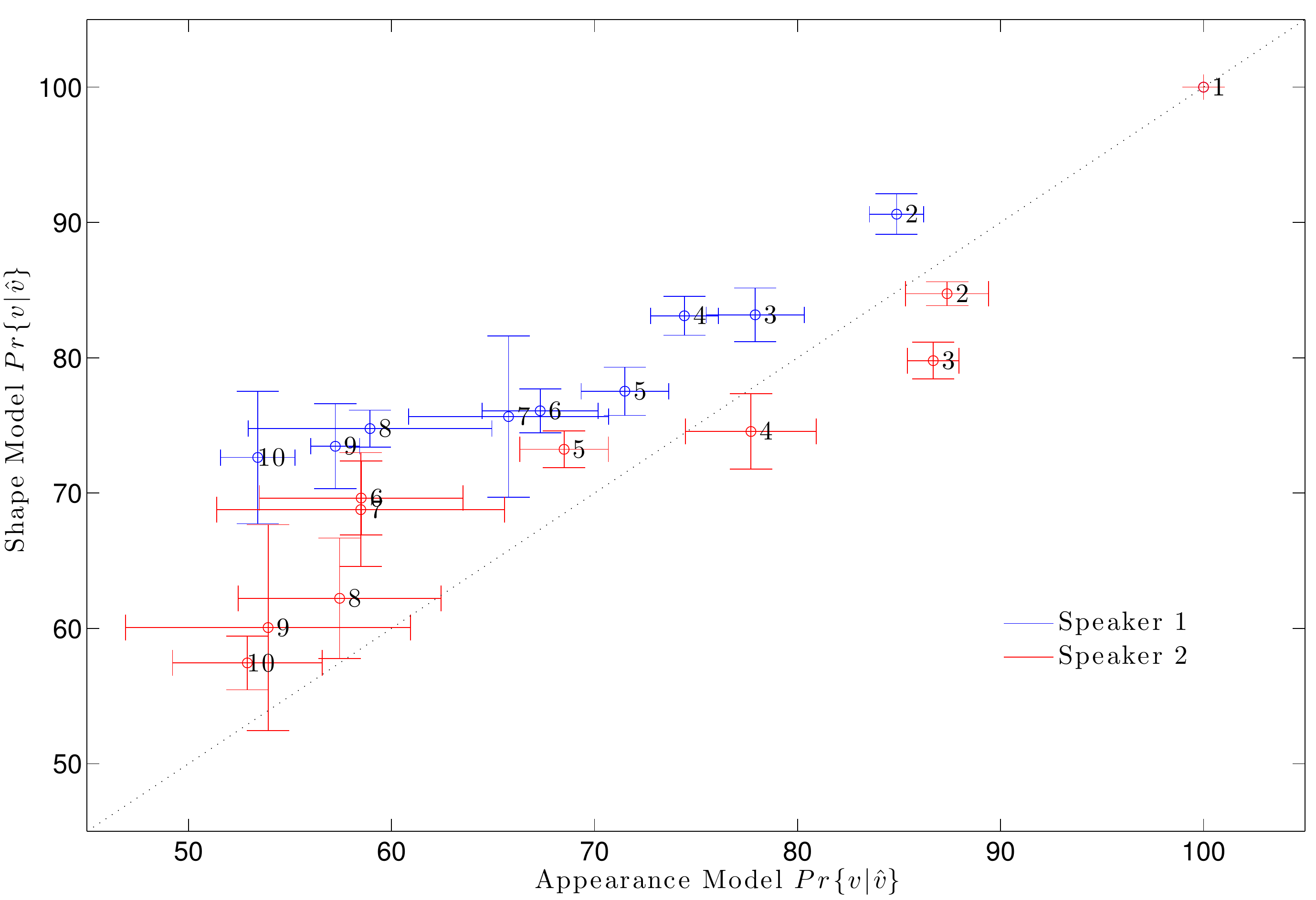} 
\caption{Relationship between shape and appearance model features for Speaker 1 and Speaker 2.} 
\label{fig:sVa} 
\end{figure} 
 
Figure~\ref{fig:sVa} shows the mean Pr$\{p|\hat{p}\}$ for the top 10 visemes over all five folds  $\pm1\frac{\sigma}{\sqrt{5}}$. The $x$-axis is the probability of correct classification when the viseme is trained on an appearance only model, the $y$-axis is the probability of correct classification when the viseme is trained on a shape only model. Red are the results for Speaker 1, and the blue are Speaker 2. As the visemes are plotted by their rank, they do not always match for each speaker. For example, the second position for Speaker 1 is $/v12/$ whereas for Speaker 2 is $/v04/$. All ranked visemes are listed in Table~\ref{tab:rankedVisemes}. The fifth most useful viseme gives superior classification for both speakers. The conventional wisdom is appearance features give the best results but only in studio-type conditions with good tracking, whereas here shape features are more robust than appearance. 

\begin{table}[h]
\centering
\caption{Ranked visemes for separate shape and appearance features for each Rosetta Raven speaker.}
\begin{tabular}{|l|l|l|l|l|}
\hline
 & \multicolumn{2}{c}{Shape} & \multicolumn{2} {c|} {Appearance} \\
Rank & Speaker 1 & Speaker 2 & Speaker 1 & Speaker 2 \\
\hline \hline
1 & $/v18/$ & $/v18/$ & $/v18/$ & $/v18/$ \\
2 & $/v12/$ & $/v04/$ & $/v04/$ & $/v04/$ \\
3 & $/v04/$ & $/v12/$ & $/v12/$ & $/v12/$ \\
4 & $/v11/$ & $/v11/$ & $/v01/$ & $/v01/$ \\
5 & $/v07/$ & $/v01/$ & $/v11/$ & $/v02/$ \\
6 & $/v01/$ & $/v05/$ & $/v07/$ & $/v11/$ \\
7 & $/v06/$ & $/v07/$ & $/v02/$ & $/gar/$ \\
8 & $/v05/$ & $/gar/$ & $/v05/$ & $/v05/$ \\
9 &  $/v02/$ & $/v02/$ & $/gar/$ & $/v10/$ \\
10 & $/gar/$ & $/v10/$ & $/v10/$ & $/v06/$ \\
\hline
\end{tabular}
\label{tab:rankedVisemes} 
\end{table}
 
Note the top right-hand point is the visual silence viseme, $/v18/$, for both Speaker 1 and Speaker 2. In general, visual silence can be quite variable compared to audio silence because speakers breathe and show emotion. However, because the source text is a poem, which has structure and natural pauses within its style, there are well-defined visual silence periods at the start of each line. 
 
\begin{figure}[!ht] 
\centering 
\includegraphics[width=0.95\textwidth]{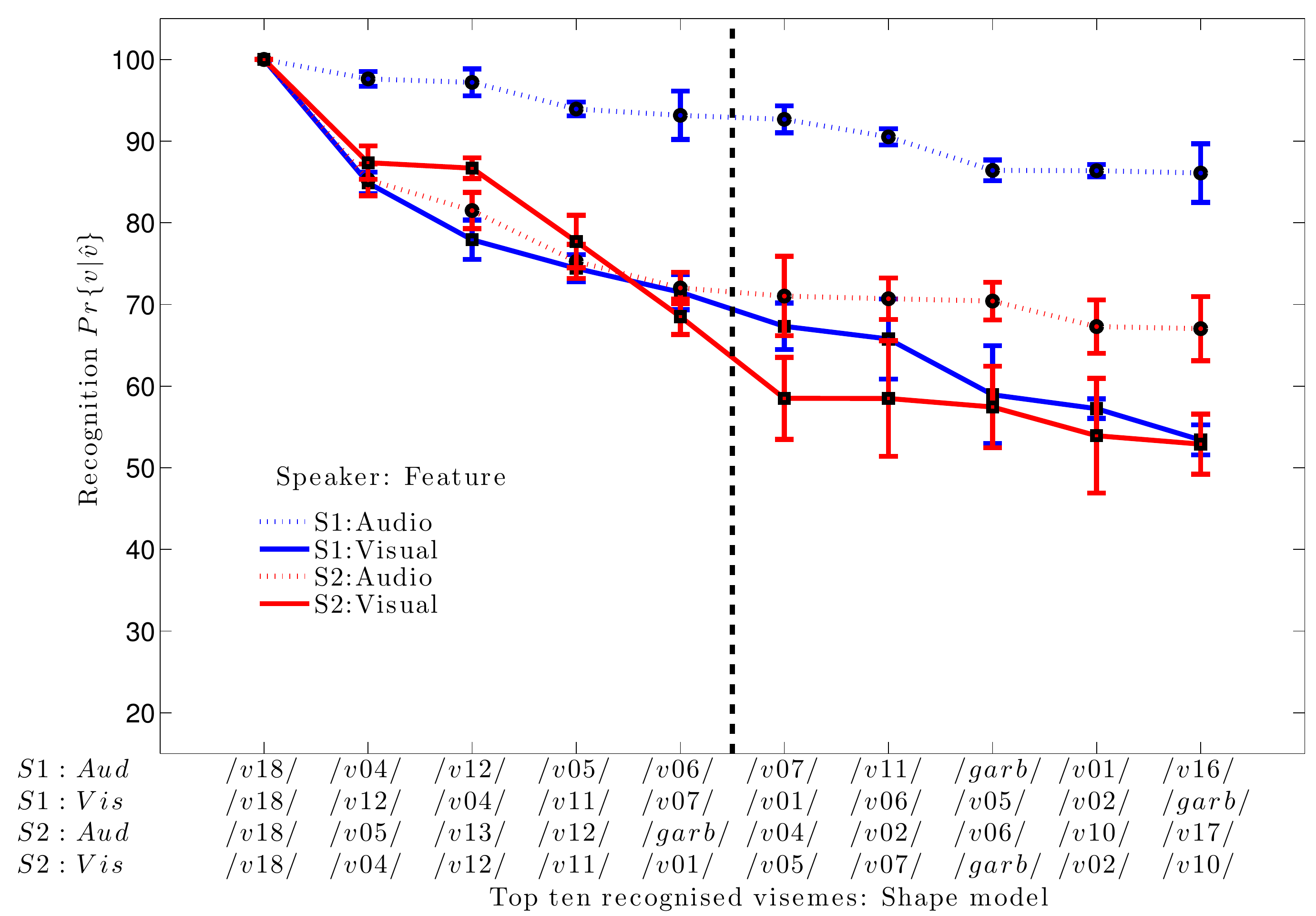} 
\caption{Classification probability Pr$\{p|\hat{p}\}$ with a shape model for the top ten visemes in descending order. A threshold is plotted in a black vertical line to show the point at which the usefulness of each viseme significantly decreases (after five visemes) in the visual channel.} 
\label{fig:shape} 
\end{figure} 
\begin{figure}[!ht] 
\centering 
%[width=\mywidth,keepaspectratio] 
\includegraphics[width=0.95\textwidth]{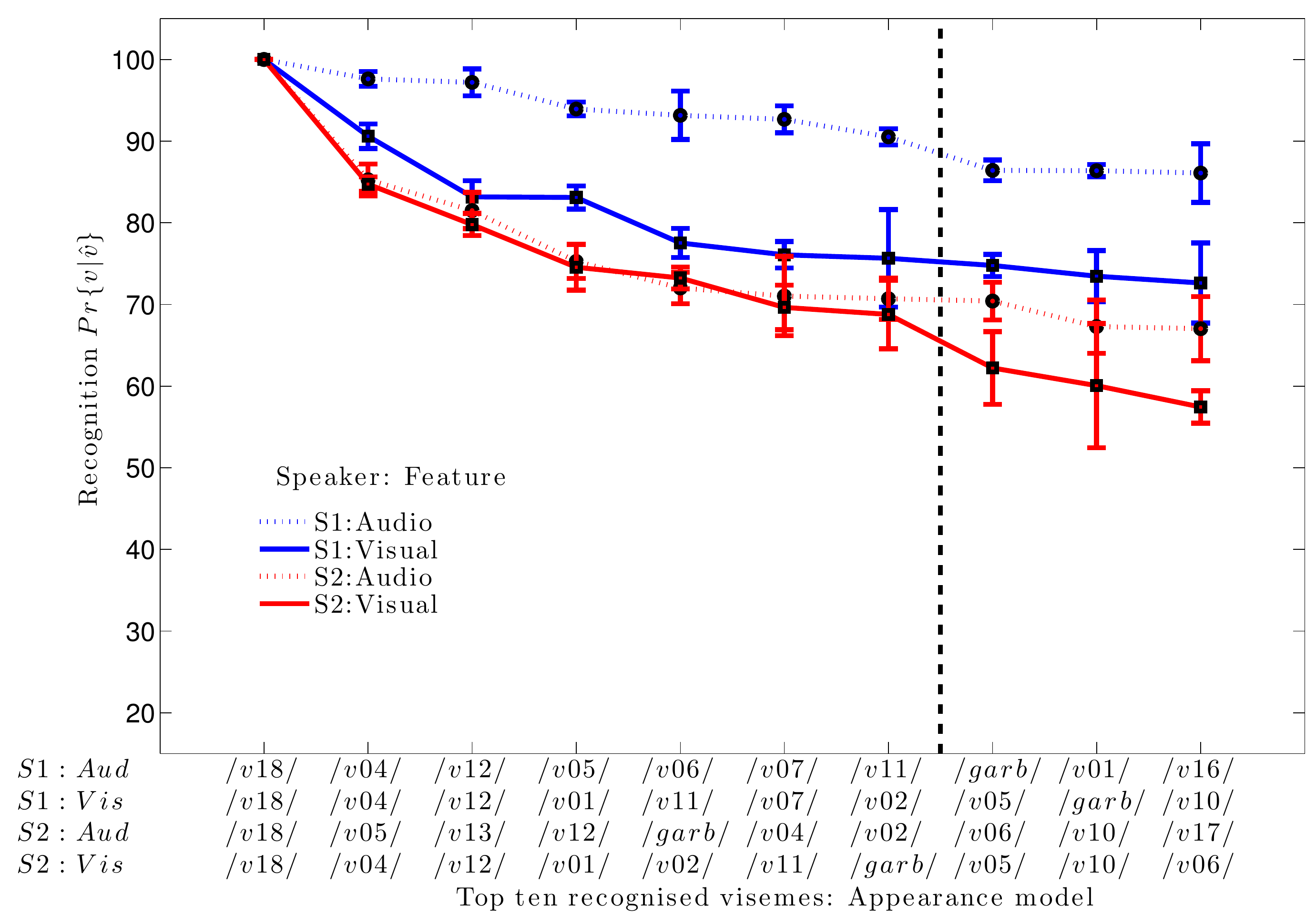} 
\caption{Classification probability Pr$\{p|\hat{p}\}$ with an appearance model for the top ten visemes in descending order. A threshold is plotted in a black vertical line to show the point at which the usefulness of each viseme significantly decreases (after seven visemes) in the visual channel.} 
\label{fig:appearance} 
\end{figure} 
 
\begin{table}[th] 
\centering 
\caption{Ranked mean viseme Pr$\{p|\hat{p}\}$ for shape, appearance, Speaker 1,  Speaker 2 and over all variables.} 
\begin{tabular}{|l|l|l|l|l|} 
\hline 
Shape & Appearance & Speaker 1 & Speaker 2 & Overall \\ 
\hline \hline 
/v18/ 			& /v18/    			& /v18/ 			& /v18/ 			& /v18/  \\ 
\{/v04/ /v12\}  		&  /v04/			& \{/v04/ /v12/\}		& \{/v04/ /v12/\}		& /v04/ \\ 
/v11/ 			& /v12/			& /v11/ 			& /v11/ 			& /v12/ \\ 
/v01/ 			& /v11/ 			& /v01/ 			& /v01/			&  /v01/ \\ 
/v07/ 			& /v01/			& /v07/			& /v07/ 			&  /v11/ \\ 
/v05/  			& /v07/			& \{/v02/ /v05/\} 		& \{/v02/ /v05/\} 	& /v07/ \\ 
\{/v02/ /v06/ 		& \{/v02/ /v05/\}		& /v06/ 			& \{/v06/ /gar/\} 	& \{/v02/ /v05/\} \\ 
/gar/\}			&				&				&				& \\ 
/v10/  			& /v06/			& \{\/v10/ /gar/\} 		& /v10/			& /v19/ \\ 
\{/v03/ /v13/\}  		& /gar/			& /v03/ 			& /v03/ 			& /v06/ \\ 
/v16/  			& /v10/			& /v13/ 			& /v13/ 			& /v10/ \\ 
/v17/  			& /v03/ 			& /v16/ 			& /v16/			& /v13/ \\ 
& \{/v13/ /v16/\} 		& /v17/ 			& /v17/			& /v03/ \\ 
& /v17/ 			&				& 				& /v16/ \\ 
&				&				&				& /v17/ \\ 
\hline 
\end{tabular} 
\label{tab:mean_ranks} 
\end{table} 
Figures~\ref{fig:shape} and~\ref{fig:appearance} show, for the Speaker 1 and Speaker 2 shape and appearance models, the probability of correctly recognising the top ten visemes, $\mbox{Pr}\{v|\hat{v}\}$. They also show the audio (MFCC) performance measured on visemes. The $x$-axis varies by performance, the best performing viseme is on the left hand side which for visual shape and appearance features is silence for all features. The next best viseme varies but is either $/v4/$, $/v5/$ or $/v12/$. $/v4/$ is a phonetically indistinct viseme (it is the biggest cluster of phonemes) so appears as a ``filler'' viseme. 
 
It has been observed in human lip-reading that there are few reliable visual cues and humans use these combined with rich contextual information to interpret or `fill in the gaps' of what a speaker is saying~\cite{erber1975auditory, stork1996speechreading}. Therefore, the hypothesis is that robust audio classification is based upon a large spread of recognised phonemes and the resilience in classification is due to the number of phonemes contributing to the accuracy. Visually, as with human lip-readers, it is anticipated fewer visemes would perform the equivalent classification and, as such, the graph would demonstrate a steeper decline in Pr$\{v|\hat{v}\}$ over the top performing visemes (from left to right along the $x$-axis). 
 
In Figure~\ref{fig:shape} there is a greater decline from left to right over the top ten visemes for visual features than for audio for both speakers. Additionally, the error bars after the $5^{th}$ position viseme increase for Speaker 2 (marginally so for Speaker 1), which provides evidence to support the hypothesis of audio classification is spread over more visemes to be correct. The top visemes (after silence $/v18/$) are $/v04/$, $/v12/$, $/v11/$ and $/v01/$. These are vowels ($/v12/$, $/v11/$) and front-of-mouth consonant visemes ($/v04/$, $/v01/$). 
 
Figure~\ref{fig:appearance}, with appearance features, demonstrates a shallower decline from left to right than the shape graph in Figure~\ref{fig:shape} but still there is a greater decline for visual features than for audio. The error bars here increase after the $7^{th}$ position viseme. Note the order of the audio viseme ordering is identical in both Figures~\ref{fig:shape} and~\ref{fig:appearance} as this is the same experiment. 

The shape of the graph in Figure~\ref{fig:appearance} is similar between audio and video which implies appearance-based classification is similar to noisy acoustic classification for both speakers and hence is less fragile. The top visemes in Figure~\ref{fig:appearance} (not including silence $/v18/$) are: $/v04/$, $/v12/$, $/v11/$, $/v01/$, and $/v7/$ i.e. identical for shape-only in the first six positions. 
 
Where the error bars increase, this may be due to the few data available, which makes classification more unreliable due to less well trained HMM classifiers. This means our estimated threshold for minimum training samples per classifier was not high enough. The impact of this is reduced with the $/gar/$ viseme, but note with Figure~\ref{fig:viseme_counts} there are similarities between our top performing visemes and those with the most training samples. 
 
% Having identified thresholds for the number of useful visemes... 
% 
%  \todo[inline]{ROC Curves with varying thresholds} 
% 
Table~\ref{tab:mean_ranks} lists the mean ranking of visemes of both speakers shape models for all visemes in the tested mapping and both speaker's appearance models. Table~\ref{tab:mean_ranks} also gives mean viseme ranks for each speaker and over all speakers and models. The rankings are similar between all pairings. 
 
\begin{table} [th] 
\caption{\label{tab:t1vt2} {Comparing Speaker 1 and Speaker 2 viseme ordering with Spearman correlation.}} 
\vspace{2mm} 
\centerline{ 
\begin{tabular}{| l | l | r | r |} 
\hline 
Speaker 1 & Speaker 2 & r & p \\ 
\hline \hline 
Audio & Audio & 0.43 & $1.63\times10^{-2}$ \\ 
Shape & Shape & \underline{0.92} & 0.00 \\ 
Appearance & Appearance & \underline{0.93} & 0.00 \\ 
\hline 
\end{tabular}} 
\end{table} 
 
\begin{table} [th] 
\caption{\label{tab:t1} {Speaker 1 Spearman correlations of viseme performance ordering with different features: acoustic, shape, and appearance.}} 
\vspace{2mm} 
\centerline{ 
\begin{tabular}{| l | l | r | r |} 
\hline 
Speaker 1 & Speaker 1 & r & p \\ 
\hline \hline 
Shape & Appearance & \underline{0.90} & 0.00 \\ 
Audio & Shape & \underline{0.85} & $2.39\times 10^{-5}$ \\ 
Audio & Appearance & \underline{0.74} & $9.2\times10^{-3}$ \\ 
\hline 
\end{tabular}} 
\end{table} 
 
\begin{table} [th] 
\caption{\label{tab:t2} {Speaker 2 Spearman correlations of viseme performance ordering with different features: acoustic, shape, and appearance.}} 
\vspace{2mm} 
\centerline{ 
\begin{tabular}{| l | l | r | r |} 
\hline 
Speaker 2 & Speaker 2 & r & p \\ 
\hline \hline 
Shape & Appearance & \underline{0.92} & 0.00 \\ 
Audio & Shape & 0.42 & 0.12 \\ 
Audio & Appearance & 0.48 & 0.07 \\ 
\hline 
\end{tabular}} 
\end{table} 
 
Tables~\ref{tab:t1vt2},~\ref{tab:t1} and~\ref{tab:t2} summarise the similarities between feature types and speakers by using Spearman rank correlation, $r$, \cite{zar1972significance} between the ranked viseme outputs. Those which are significant at the $5\%$ threshold are underlined. This confirms a strong relation between shape-only and appearance-only classification. In lab conditions,  appearance features outperform shape \cite{bear2014resolution} but in real world conditions the shape information is more robust in the absence of non-noisy appearance data \cite{improveVis}. This strong coupling, and previous work, \cite{lan2009comparing}, shows the two modes of information are complimentary and we recommend the use of both, without forgetting that in the real world, artefacts such as motion blur significantly deteriorate appearance information. We also note for Speaker 1 (in Figure~\ref{tab:t1vt2}) the audio ranking is similar to the video ranking although as we have previously noticed there is a more rapid drop-off for video. 
 
\section{Viseme contribution observations} 

To summarise this chapter, we have shown that, with the assumption classifiers are trained with sufficient data, the order of single viseme performances are fairly consistent across the feature modes of audio, shape and appearance. It is also noted the visual classifiers depend more highly on a select few visemes performing well (between five for shape modes and seven for appearance mode out of a possible 15) than the audio classifiers.  

The observation of how fragile machine lip-reading is, is re-enforced by this work. If these critical five or seven visemes cannot be built as sufficiently trained classifiers then lip-reading is impossible. When a human is trained in how to lip-read, many follow the method of recognising a small number of key gestures which we then process using our own sophisticated knowledge of language and context to create a classification output or transcript \cite{comparHumMacLipRead}. 
 
In audio it is surprisingly rare to see this effect measured, even though a good acoustic unit will have accuracies which are at least 10\% higher than an average unit (the mean audio viseme performance on Speaker 2 is 76\% for the all visemes). 
 
We acknowledge most work in this field focuses on improving mean accuracies over the set of all visemes which can conceal the real source of overall performance. A system which achieves a mean viseme accuracy of, say, 53\%, may be one which contains a few supremely accurate viseme classifiers or it maybe a system with a set of a large number of classifiers which all achieve a more modest performance. In our work we have seen a correlation between the spread of viseme contributions to classification and viseme classification performance, so we can now say higher classification is achieved with a set of equally useful visemes rather than a set of visemes where their usefulness ranges from poor to excellent. 
 
This chapter, therefore, suggests two different strategies for improving future lip-reading systems; option one: one makes the select few best viseme classifiers better or, option two: one focuses upon improving the worst, which at this stage do not contribute at all. We can not comment at this time which approach is likely to be more successful but our observations will allow future work to focus attention where it is likely to do the most good. 

This work suggests five of the visemes are largely responsible for accurate classification, whereas for appearance there are seven visemes and for audio there are at least ten. This means there appears to be fewer recognizable shapes than there are distinguishable appearances, and in turn, sounds. This relates to the overall viseme classification of the set where audio results are better than appearance, which in turn are better than shape. 

We suggest that a good threshold of viseme training samples, is not more or less than 1 standard error away from the mean number of training samples for all visemes in a set. This is stricter than the threshold we used and will ensure there is no bias towards any one particular viseme class which could then dominate the classification accuracy of the set. 

Now we have a deeper understanding of visemes and their individual capabilities, we move onto investigating how they relate to phonemes, the acoustic units of speech. We are reminded of our viseme working definition, ``a viseme is the visual equivalent of a phoneme'' so we move on to a review of a number of the phoneme-to-viseme (P2V) mappings which have been presented in literature in order to assess which is optimal for machine lip-reading. %SPIE: individual viseme comparisons; lessons from small datasets. 
%!TEX root = main.tex 

\chapter[Bear speaker-dependent visemes]{Bear speaker-dependent visemes} 
\label{chap:maps} 
 
In computer lip-reading literature there is debate over the mapping of phonemes to visemes. In this chapter the AVLetters2 dataset (Section~\ref{sec:avl2}) is used to train and test classifiers using 120 phoneme-to-viseme (P2V) mappings and the effect on word classification accuracy is measured. This chapter also presents and tests a new data-driven method for devising speaker-dependent phoneme to viseme maps using phoneme confusions. Our method is not influenced by perception bias since our confusions are based on machine observations, and not human perception. We compare word classification achieved with these new maps against the best performing previously published phoneme-to-viseme mapping. We demonstrate that whilst there are differences between each viseme map previously suggested, the best mapping over all speakers is from Lee~\cite{lee2002audio}. This mapping is used as a benchmark to compare the performance of new data-derived speaker-dependent visemes. 
 
%Phonemes are the discriminate sounds of a language~\cite{international1999handbook} and the visual equivalent, although not precisely defined, are the visemes;~\cite{chen1998audio,fisher1968confusions,Hazen1027972}. A good working definition of a viseme is a set of phonemes have identical appearance on the lips. Therefore any phoneme falls into one viseme class but a viseme may represent many phonemes: a many to one mapping. 
 
A summary of published P2V maps is provided in \cite{theobaldPHD} Tables 2.3 and 2.4. This list is not exhaustive and these mappings vary by: a focus on just consonants \cite{binnie1976visual, fisher1968confusions, franks1972confusion, walden1977effects}, are speaker-dependent \cite{kricos1982differences}, or have an ordering \cite{owens1985visemes}. These are useful starting points, but for the purpose of this study we would like the phoneme-to-viseme mappings to include all phonemes in the transcript of the dataset to accurately reflect the range of phonemes used in a full vocabulary. Therefore, some mappings used here are a pairing of two mappings suggested in literature, e.g. one map for the vowels and one map for the consonants. A full list of the mappings used is in Tables~\ref{tab:vowelmappings} and~\ref{tab:consonantmappings}. In total, 15 consonant maps and eight vowel maps are identified here and all of these are paired with each other to provide 120 P2V maps to test. The questions we ask are; does conventional a machine lip-reading system use the correct viseme mappings for machine lip-reading? And, is it possible to find a method for selecting better phoneme-to-viseme mappings? 
 
 \section{Current viseme studies} 
 
There are many viseme classifications present in literature, the most common viseme classifications are: `the Disney 12' \cite{disney}, the `lip-reading 18' by Nichie \cite{lip_reading18}, and Fisher's \cite{fisher1968confusions}. Full phoneme to viseme mappings of these classes can be found in Tables~\ref{tab:disneymap},~\ref{tab:fishermap} and~\ref{tab:nichiemap}. The differences in these classifications are based around different groupings of phonemes, and in the literature we know of a number of recent attempts to compare these, such as \cite{cappelletta2012phoneme} and as part of \cite{theobaldPHD}. In \cite{theobaldPHD} the following list of reasons are given for discrepancies between classifier sets.

\begin{itemize}
\item Variation between speakers - i.e. speaker identity.
\item Variation between viewers - indicating lip-reading ability varies by individuals, those with more practise are better able to identify visemes.
\item The context of the speech presented - context has an influence on how consonants appear on the lips. In real tasks the context will enable easier distinction between indistinguishable phonemes in syllable only tests. 
\item Clustering criteria - the grouping methods vary between authors. For example, `phonemes are said to belong to a viseme if, when clustered, the percent correct identification for the viseme is above some threshold, which is typically between 70 - 75\% correct. A stricter grouping criterion has a higher threshold, so more visemes are identified.'.
\end{itemize}

\begin{table}[!ht]
\centering
\caption{The ``Disney twelve'' phoneme-to-viseme map.}
\begin{tabular}{|l|l|}
 \hline
 Viseme & Phonemes \\
 \hline \hline
/v01/ & /p/ /b/ /m/ \\
/v02/ & /w/ \\
/v03/ & /f/ /v/ \\
/v04/ & /\textipa{T}/ \\
/v05/ & /l/ \\
/v06/ & /d/ /t/ /z/ /s/ /r/ /n/ \\
/v07/ & /s/ /\textipa{S}/ /t\textipa{S}/ /j/ \\
/v08/ & /y/ /g/ /k/ /\textipa{N}/ \\
/v09/ & /\textupsilon/ /\textipa{H}/ \\
/v10/ & /\textipa{E}\textschwa/ /\textsci/ /ai/ /e/ /\textturnv/ \\
/v11/ & /u/ \\
/v12/ & /\textupsilon \textschwa/ /\textopeno/ /\textopeno \textschwa/ \\
 \hline
 \end{tabular}
 \label{tab:disneymap}
 \end{table}
 
  \begin{table}[!ht]
\centering
\caption{Fisher's phoneme-to-viseme map.}
\begin{tabular}{|l|l|}
 \hline
 Viseme & Phonemes \\
 \hline \hline
/v01/ & /k/ /g/ /\textipa{N}/ /m/ \\
/v02/ & /p/ /b/ \\
/v03/ & /f/ /v/ \\
/v04/ & /\textipa{S}/ /\textipa{Z}/ /t\textipa{S}/ /d\textipa{Z}/ \\
/v05/ & /t/ /d/ /n/ /th/ /dh/ /z/ /s/ /r/ /l/ \\
 \hline
 \end{tabular}
 \label{tab:fishermap}
 \end{table}

\begin{table}[!ht]
\centering
\caption{Nichie's ``Lip-reading 18'' phoneme-to-viseme map.}
\begin{tabular}{|l|l|}
 \hline
 Viseme & Phonemes \\
 \hline \hline
/v01/ & /p/ /b/ /m/ \\
/v02/ & /f/ /v/ \\
/v03/ & /\textipa{W}/ /w/ \\
/v04/ & /r/ \\
/v05/ & /s/ /z/ \\
/v06/ & /\textipa{S}/ /\textipa{Z}/ /t\textipa{S}/ /d\textipa{Z}/ \\
/v07/ & /\textipa{D}/ \\
/v08/ & /l/ \\
/v09/ & /t/ /d/ /n/ \\
/v10/ & /y/ \\
/v11/ & /k/ /g/ /\textipa{N}/ \\
/v12/ & /\textipa{H}/ \\
/v13/ & /uw/ \\
/v14/ & /\textupsilon/ /\textschwa \textupsilon/ \\
/v15/ & /\textscripta \textupsilon/ \\
/v16/ & /i/ /ay/ /\textsci/ \\
/v17/ & /u/ \\
/v18/ & /\textturnv/ \\
/v19/ & /iy/ /\textipa{E}/ \\
/v20/ & /e/ /\textsci \textschwa/ \\
/v21/ & /\textschwa/ /ei/ \\
 \hline
 \end{tabular}
 \label{tab:nichiemap}
 \end{table}

\section{Data preparation} 
\label{sec:currentp2vmaps}
The AVLetters2 (AVL2) dataset \cite{cox2008challenge} is used to train and test HMM classifiers based upon our 120 P2V mappings. AAM features are used as they are known to outperform other feature methods in machine lip-reading~\cite{cappelletta2012phoneme}. Tables~\ref{tab:vowelmappings} and~\ref{tab:consonantmappings} show all phonemes in each original P2V map. As each utterance is very short in our data set (each is a one word sentence of a single letter) there is no need to implement $\Delta$s within our features to address co-articulation. 
 
 %where $CF$ is the confusion factor for viseme set $s$, $\#V$ is the number of visemes in $s$ and $\#P$ is the number of phonemes in all $V$. 
 \begin{table}[!ht] 
\centering
\caption{Vowel phoneme-to-viseme maps previously presented in literature.} 
\begin{tabular}{|l|l|} 
\hline 
Classification & Viseme phoneme sets \\ 
\hline \hline 
Bozkurt \cite{bozkurt2007comparison} &  {\footnotesize \{/ei/ /\textturnv/\} \{/ei/ /e/ /\ae/\} \{/\textrevepsilon/\} \{/i/ /\textsci/ /\textschwa/ /y/\} \{/\textscripta\textupsilon/\} } \\ 
& {\footnotesize \{/\textopeno/ /\textscripta/ /\textopeno\textsci/ /\textschwa\textupsilon/\} \{/u/ /\textupsilon/ /w/\} }\\ 
Disney \cite{disney} & {\footnotesize  \{/\textupsilon/ /h/\} \{/\textepsilon\textschwa/ /i/ /ai/ /e/ /\textturnv/\} \{/u/\} \{/\textupsilon\textschwa/ /\textopeno/ /\textopeno\textschwa/\} } \\ 
Hazen \cite{Hazen1027972} & {\footnotesize  \{/\textscripta\textupsilon/ /\textupsilon/ /u/ /\textschwa\textupsilon/ /\textopeno/ /w/ /\textopeno\textsci/\}  \{/\textturnv/ /\textscripta/\} \{/\ae/ /e/ /ai/ /ei/\} } \\ 
& {\footnotesize \{/\textschwa/ /\textsci/ /i/\}  }\\ 
Jeffers \cite{jeffers1971speechreading} &  {\footnotesize \{/\textscripta/ /\ae/ /\textturnv/ /ai/ /e/ /ei/ /\textsci/ /i/ /\textopeno/ /\textschwa/ /\textsci/\} \{/\textopeno\textsci/ /\textopeno/\} \{/\textscripta\textupsilon/\}  }\\ 
& {\footnotesize \{/\textrevepsilon/ /\textschwa\textupsilon/ /\textupsilon/ /u/\} } \\ 
Lee \cite{lee2002audio} & {\footnotesize   \{/i/ /\textsci/\} \{/e/ /ei/ /\ae/\} \{/\textscripta/ /\textscripta\textupsilon/ /ai/ /\textturnv/\} \{/\textopeno/ /\textopeno\textsci/ /\textschwa\textupsilon/\} \{/\textupsilon/ /u/\}   }\\ 
Montgomery \cite{montgomery1983physical} & {\footnotesize  \{/i/ /\textsci/\} \{/e/ /\ae/ /ei/ /ai/\} \{/\textscripta/ /\textopeno/ /\textturnv/\} \{/\textupsilon/ /\textrevepsilon/ /\textschwa/\}\{/\textopeno\textsci/\} } \\ 
& {\footnotesize \{/i/ /hh/\} \{/\textscripta\textupsilon/ /\textschwa\textupsilon/\} \{/u/ /u/\}   }\\ 
Neti \cite{neti2000audio} & {\footnotesize  \{/\textopeno/ /\textturnv/ /\textscripta/ /\textrevepsilon/ /\textopeno\textsci/ /\textscripta\textupsilon/ /\textipa{H}/\} \{/u/ /\textupsilon/ /\textschwa\textupsilon/\} \{/\ae/ /e/ /ei/ /ai/\} } \\ 
& {\footnotesize \{/\textsci/ /i/ /\textschwa/\}  }\\ 
Nichie \cite{lip_reading18} &  {\footnotesize \{/uw/\} \{/\textupsilon/ /\textschwa\textupsilon/\} \{/\textscripta\textupsilon/\} \{/i/ /\textturnv/ /ay/\} \{/\textturnv/\} \{/iy/ /\ae/\} \{/e/ /\textsci\textschwa/\} } \\ 
& {\footnotesize \{/u/\} \{/\textschwa/ /ei/\}    }\\ 
\hline 
\end{tabular} 
\label{tab:vowelmappings} 
\end{table}

\begin{table}[p] 
\centering
\caption{Consonant phoneme-to-viseme maps previously presented in literature.} 
\begin{tabular}{|l|l|} 
\hline 
Classification & Viseme phoneme sets \\ 
\hline \hline 
Binnie \cite{binnie1976visual} &  {\footnotesize \{/p/ /b/ /m/\} \{/f/ /v/\} \{/\textipa{T}/ /\textipa{D}/\}  \{/\textipa{S}/ /\textipa{Z}/\} \{/k/ /g/\} \{/w/\} \{/r/\} }\\ 
& {\footnotesize \{/l/ /n/\} \{/t/ /d/ /s/ /z/\} } \\ 
Bozkurt \cite{bozkurt2007comparison} &  {\footnotesize \{/g/ /\textipa{H}/ /k/ /\textipa{N}/\} \{/l/ /d/ /n/ /t/\} \{/s/ /z/\} \{/t\textipa{S}/ /\textipa{S}/ /d\textipa{Z}/ /\textipa{Z}/\}  \{/\textipa{T}/ /\textipa{D}/\}  }\\ 
&  {\footnotesize \{/r/\} \{/f/ /v/\} \{/p/ /b/ /m/\}  }\\ 
Disney \cite{disney} &  {\footnotesize \{/p/ /b/ /m/\} \{/w/\} \{/f/ /v/\} \{/\textipa{T}/\} \{/l/\} \{/d/ /t/ /z/ /s/ /r/ /n/\} }\\ 
& {\footnotesize \{/\textipa{S}/ /t\textipa{S}/ /j/\} \{/y/ /g/ /k/ /\textipa{N}/\} }\\ 
Finn \cite{finn1988automatic} &  {\footnotesize \{/p/ /b/ /m/\} \{/\textipa{T}/ /\textipa{D}/\} \{/w/ /s/\} \{/k/ /h/ /g/\} \{/\textipa{S}/ /\textipa{Z}/ /t\textipa{S}/ /j/\}  } \\ 
&  {\footnotesize \{/y/\} \{/z/\} }\\ 
& {\footnotesize \{/f/\} \{/v/\} \{/t/ /d/ /n/ /l/ /r/\} } \\ 
Fisher \cite{fisher1968confusions} &  {\footnotesize \{/k/ /g/ /\textipa{N}/ /m/\} \{/p/ /b/\} \{/f/ /v/\} \{/\textipa{S}/ /\textipa{Z}/ /d\textipa{Z}/ /t\textipa{S}/\}  }\\ 
& {\footnotesize \{/t/ /d/ /n/ /\textipa{T}/ /\textipa{D}/ /z/ /s/ /r/ /l/\} } \\ 
Franks \cite{franks1972confusion} &  {\footnotesize \{/p/ /b/ /m/\} \{/f/\} \{/r/ /w/\} \{/\textipa{S}/ /d\textipa{Z}/ /t\textipa{S}/\}  }\\ 
Hazen \cite{Hazen1027972} &  {\footnotesize \{/l/\} \{/r/\} \{/y/\} \{/b/ /p/\} \{m\} \{/s/ /z/ /h/\} \{/t\textipa{S}/ /d\textipa{Z}/ /\textipa{S}/ /\textipa{Z}/\} }\\ 
& {\footnotesize   \{/t/ /d/ /\textipa{T}/ /\textipa{D}/ /g/ /k/\} } \\ 
&   {\footnotesize \{/\textipa{N}/\} \{/f/ /v/\}  }\\ 
Heider \cite{heider1940experimental} &  {\footnotesize \{/p/ /b/ /m/\} \{/f/ /v/\} \{/k/ /g/\} \{/\textipa{S}/ /t\textipa{S}/ /d\textipa{Z}/\} \{/\textipa{T}/\} \{/n/ /t/ /d/\} } \\ 
& {\footnotesize \{/l/\} \{/r/\} } \\ 
Jeffers \cite{jeffers1971speechreading} & {\footnotesize \{/f/ /v/\} \{/r/ /q/ /w/\} \{/p/ /b/ /m/\} \{/\textipa{T}/ /\textipa{D}/\} \{/t\textipa{S}/ /d\textipa{Z}/ /\textipa{S}/ /\textipa{Z}/\} } \\ 
& {\footnotesize \{/s/ /z/\} \{/d/ /l/ /n/ /t/\}  }\\ 
&   {\footnotesize \{/g/ /k/ /\textipa{N}/\}  }\\ 
Kricos \cite{kricos1982differences} &  {\footnotesize \{/p/ /b/ /m/\} \{/f/ /v/\} \{/w/ /r/\} \{/t/ /d/ /s/ /z/\} } \\ 
& {\footnotesize \{/k/ /\textipa{n}/ /j/ /h/ /\textipa{N}/ /g/\} \{/l/\}  \{/\textipa{T}/ /\textipa{D}/\}  } \\ 
&  {\footnotesize \{/\textipa{S}/ /\textipa{Z}/ /t\textipa{S}/ /d\textipa{Z}/\} }\\ 
Lee \cite{lee2002audio} &  {\footnotesize \{/d/ /t/ /s/ /z/ /\textipa{T}/ /\textipa{D}/\} \{/g/ /k/ /n/ /\textipa{N}/ /l/ /y/ /\textipa{H}/\} }\\ 
& {\footnotesize \{/d\textipa{Z}/ /t\textipa{S}/ /\textipa{S}/ /\textipa{Z}/\}  \{/p/ /b/ /m/\} \{/f/ /v/\}  }\\ 
&  {\footnotesize \{/r/ /w/\} }\\ 
Neti \cite{neti2000audio} &  {\footnotesize\{/l/ /r/ /y/\} \{/s/ /z/\} \{/t/ /d/ /n/\} \{/\textipa{S}/ /\textipa{Z}/ /d\textipa{Z}/ /t\textipa{S}/\} \{/p/ /b/ /m/\} } \\ 
& {\footnotesize \{/\textipa{T}/ /\textipa{D}/\} \{/f/ /v/\}  }\\ 
&   {\footnotesize \{/\textipa{N}/ /k/ /g/ /w/\} }\\ 
Nichie \cite{lip_reading18} & {\footnotesize \{/p/ /b/ /m/\} \{/f/ /v/\} \{/\textipa{W}/ /w/\} \{/r/\} \{/s/ /z/\} \{/\textipa{S}/ /\textipa{Z}/ /t\textipa{S}/ /j/\} } \\ 
& {\footnotesize \{/\textipa{T}/\} \{/l/\}  \{/k/ /g/ /\textipa{N}/\} \{/\textipa{H}/\} }\\ 
&   {\footnotesize \{/t/ /d/ /n/\} \{/y/\} }\\ 
Walden \cite{walden1977effects} & {\footnotesize \{/p/ /b/ /m/\} \{/f/ /v/\} \{/\textipa{T} /\textipa{D}/\} \{/\textipa{S}/ /\textipa{Z}/\} \{/w/\} \{/s/ /z/\} \{/r/\} } \\ 
& {\footnotesize \{/l/\} \{/t/ /d/ /n/ /k/ /g/ /j/\} }\\ 
Woodward \cite{woodward1960phoneme} & {\footnotesize \{/p/ /b/ /m/\} \{/f/ /v/\} \{/w /r/ /\textipa{W}/\} } \\ 
& {\footnotesize \{/t/ /d/ /n/ /l/ /\textipa{T}/ /\textipa{D}/ /s/ /z/ /t\textipa{S}/ /d\textipa{Z}/ /\textipa{S}/ /\textipa{Z}/ /j/ /k/ /g/ /h/\}}\\ 
\hline 
\end{tabular} 
\label{tab:consonantmappings} 
\end{table} 
 
  In Table~\ref{tab:lit_visemes_compare} we have described the sources and derivation methods for all of the phoneme-to-viseme maps used in our comparison study. We see the majority are constructed using human perception testing with few test subjects, e.g. Finn \cite{finn1988automatic} only used 1 and Kricos \cite{kricos1982differences} 12. Data-driven methods are most recent, e.g. Lee's \cite{lee2002audio} visemes were presented in 2002 and Hazen's \cite{Hazen1027972} in 2006. The remaining visemes are based around linguistic/phonemic rules. 
  
 As an example, the clustering method of Hazen \cite{Hazen1027972} involved bottom-up clustering using maximum Bhattacharyya distances to measure similarity between the phoneme-labelled models. The models were represented by Gaussian distributions. Before clustering, some phonemes were manually merged, $/em/$ with $/m/$, $/en/$ with $/n/$, and $/\textipa{Z}/$ with $/\textipa{S}/$. 
  
 \begin{table}[!h]
\centering
\caption{A comparison of literature phoneme-to-viseme maps.}
\begin{tabular}{|l|l|l|l|l|}
\hline
Author & Year & Inspiration & Description & Test subjects \\
 \hline \hline
 Binnie & 1976 & Human testing & Confusion patterns & unknown \\
 Bozkurt & 2007 & Subjective linguistics & Common tri-phones & 462\\
 Disney & --- & Speech synthesis &  Observations & unknown \\
 Finn & 1988 & Human perception & Montgomerys visemes & 1 \\ 
 	&		&			& and /\textipa{H}/ &  \\
 Fisher  & 1986 & Human testing & Multiple-choice & 18 \\
 		&		&		&	intelligibility test & \\
 Franks & 1972 & Human perception & Confusions among sounds & unknown\\
 	&		&		&  produced in similar  & \\
	&		&		& articulatory positions & 275\\
 Hazen & 2006 & Data-driven & Bottom-up clustering & 223 \\
 Heider & 1940 & Human perception & Confusions post-training & unknown \\
 Jeffers & 1971 & Linguistics & Sensory and cognitive & unknown \\
 		&		&		&	correlates & \\
 Kricos & 1982 & Human testing & Hierarchical clustering & 12 \\
 Lee & 2002 & Data-driven & Merging of Fisher visemes & unknown \\
 Neti & 2000 & Linguistics & Decision tree clusters & 26 \\
 Nichie & 1912 & Human observations & Human observation of & unknown \\
	&		&				& lip movements & \\
 Walden & 1977 & Human testing & Hierachical clustering & 31 \\
 Woodward & 1960 & Linguistics & Language rules & unknown \\
 		&		&		&	and context & \\
 \hline
 \end{tabular}
 \label{tab:lit_visemes_compare}
 \end{table}

Figure~\ref{fig:histogram} (Chapter~\ref{chap:datasets_dics}) shows the occurrence frequency of the 29 phonemes in AVL2 which details the volume of training samples available. Note, AVL2 does not include all phonemes in the British English phonetic alphabet \cite{international1999handbook}. It is a known problem in visual speech research that one limitation is the lack of sufficiently large datasets available~\cite{cappelletta2012phoneme}. This motivates the drive to find better P2V mappings to potentially avoid the need, and associated cost, in obtaining large audio-visual speech datasets. 
 
A P2V map introduces confusion in machine lip-reading. In an attempt to measure the level of this confusion, a simple ratio metric of the proportion of phonemes to visemes is shown in (Equation \ref{eq:cfequation}), where $CF_s$ is the compression factor for a set of visemes, $s$, $\#V$ is the number of visemes, and $\#P$ is the number of phonemes. The compression factors for the P2V maps are described in Table~\ref{tab:Confusion_Factors}. The ideal ratio is a 1:1 phoneme to viseme mapping as this would mean we are identifying each phoneme uniquely. However, we still need to cluster the phonemes due to the lack of visual distinction between some phonemes. Thus the higher a Compression Factor (CF) (closer to one) the better it is as this means there is less dependency upon the language network for decoding of visemes back to phonemes. Silence and garbage visemes are not included in CFs. 
 
\begin{equation} 
\centering 
CF_s = \frac{\#V}{\#P} 
%\caption(Confusion factor between the visemes and phonemes within a set of viseme classes, $S$) 
\label{eq:cfequation} 
\end{equation} 

\begin{table}[!h] 
\centering 
\caption{Compression factors for viseme maps previously presented in literature.} 
\begin{tabular}{| l | r | r | l | r | r |} 
\hline 
Consonant Map & V:P & CF & Vowel Map & V:P & CF \\ 
\hline \hline 
Woodward  & 4:24 & 0.16 & Jeffers & 3:19 & 0.16\\ 
Disney & 6:22 & 0.18 & Neti & 4:20 & 0.20\\ 
Fisher & 5:21 & 0.23 & Hazen & 4:18 & 0.22 \\ 
Lee & 6:24 & 0.25 & Disney & 4:11 & 0.36 \\ 
Franks & 5:17 & 0.29 & Lee & 5:14 & 0.36\\ 
Kricos & 8:24 & 0.33 & Bozkurt & 7:19 & 0.37\\ 
Jeffers & 8:23 & 0.35 & Montgomery & 8:19 & 0.42\\ 
Neti & 8:23 & 0.35 & Nichie & 9:15 & 0.60\\ 
Bozkurt & 8:22 & 0.36  & - & - & -\\ 
Finn & 10:23 & 0.43 & - & - & -\\ 
Walden & 9:20 & 0.45  & - & - & -\\ 
Binnie & 9:19 & 0.47 & - & - & -\\ 
Hazen & 10:21 & 0.48 & - & - & -\\ 
Heider & 8:16 & 0.50 & - & - & -\\ 
Nichie & 18:33 & 0.54 & - & - & -\\ 
 
\hline 
\end{tabular} 
\label{tab:Confusion_Factors} 
\end{table} 

Deliberate omission of the following phonemes from some mappings is required: $/si/$ (Disney \cite{disney}), $/axr/$ $/en/$ $/el/$ $/em/$ (Bozkirt \cite{bozkurt2007comparison}), $/axr/$ $/em/$ $/epi/$ $/tcl/$ $/dcl/$ $/en/$ $/gcl/$ $kcl/$(Hazen \cite{Hazen1027972}), and $/axr/$ $/em/$ $/el/$ $/nx/$ $/en/$ $/dx/$ $/eng/$ $/ux/$ (Jeffers \cite{jeffers1971speechreading}), because these are American diacritics which are not appropriate to a British English phonetic dataset. Moreover, Kricos provides speaker-dependent visemes \cite{kricos1982differences}. These have been generalised for our tests using the most common mixtures of phonemes as the method is not reproducible. Where a viseme map does not include phonemes present in the ground truth transcript these are grouped into a garbage viseme ($/gar/$) to measure only the performance of the viseme sets previously prescribed in literature. Note that all phonemes in the each P2V map are in the dataset but no mapping includes all 29 phonemes in the AVL2 vocabulary. 
 
\section{Classification method} 
\label{sec:recognition} 
The method for these speaker-dependent classification tests on our combined shape and appearance features uses HMM classifiers built with HTK \cite{htk34}. The features selected are from the AVL2 dataset described in Chapter~\ref{chap:datasets_dics}. The videos are tracked with a full-face AAM and the features extracted consist of only the lip information. The classifiers are based upon viseme labels within each P2V map. A ground truth for measuring correct classification is a viseme transcription produced using the BEEP British English pronunciation dictionary \cite{beep} and a word transcription. The classification output is a viseme level script mapped to sentence (word) level classification. Working in British English the phonetic transcript is converted to a viseme transcript assuming the visemes in the mapping being tested (Tables~\ref{tab:vowelmappings} and~\ref{tab:consonantmappings}). We test using a leave-one-out seven-fold cross validation. Seven folds are selected as we have seven utterances of the alphabet per speaker in AVL2. The HMMs are initialised using `flat start' training and re-estimated eight times and then force-aligned using HTK's \texttt{HVite}. Training is completed by re-estimating the HMMs three more times with the force-aligned transcript. 
 
\section{Comparison of current phoneme to viseme maps} 
\label{sec:comparison} 
In this section, classification performance of the HMMs is measured by correctness, $C$ (Equation \ref{eq:correctness}), as there are no insertion errors to consider \cite{htk34}. It is acknowledged word classification is not as high performing as viseme classification. However, as each viseme set being tested has a different number of phonemes and visemes, a common comparator, here words, are used as they can compare different viseme sets. It is the difference between each set, rather than the individual performance, which is of interest in this investigation. Word level correctness rather than viseme level correctness normalises over all sets for a fair comparison. (Each viseme set has a different number of visemes in it and in turn a varying level of training samples per viseme). 

We compare our values of accuracy to those in the literature, namely \cite{cox2008challenge} \&~\cite{pei2013unsupervised}. In \cite{cox2008challenge} we see that speaker-dependent results with AVL2 are significantly higher than the values we have achieved. However, in this paper the experiments are designed to measure the efficacy of multi-speaker classifiers and thus the authors have permitted different HMM parameters between speakers. For example, the number of HMM states ranges between five and nine. In our work these values are constant to ensure any effects observed are the result of the viseme selection only. 

In \cite{pei2013unsupervised} AVLetters2 data achieves 91.8\% with an unsupervised random forest classification technique. This out performs both \cite{cox2008challenge} and our results here. However, this unsupervised method inhibits the option of knowing the visual units used by the forest. As our priority in this comparison study is to measure the effects of viseme selection rather than optimising a classification method for each individual speaker, we bare the cost to overall classification for the learning gained from the observations by comparing viseme sets.

\begin{figure}[!ht] 
\centering 
\includegraphics[width=0.98\textwidth]{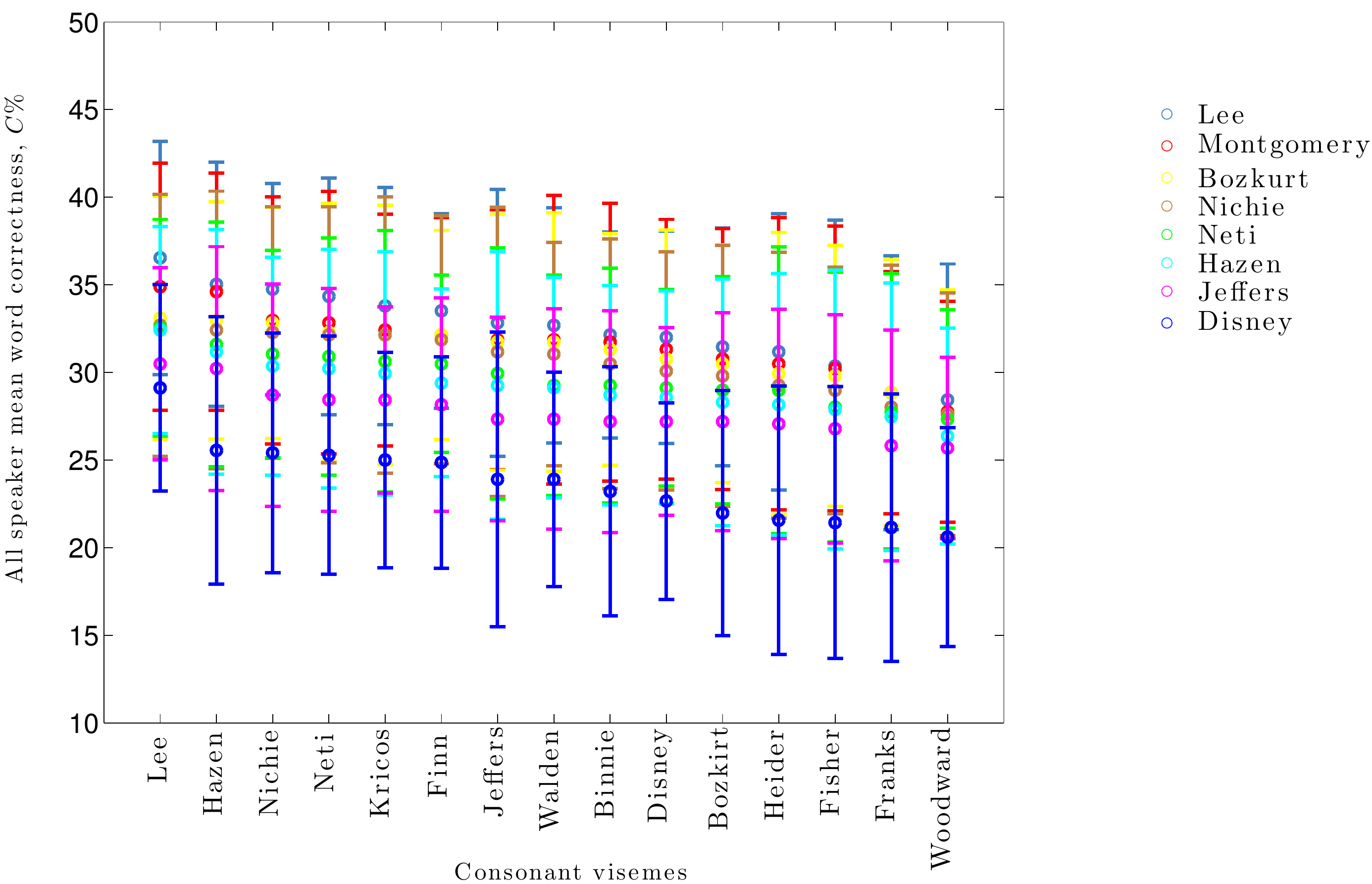} 
\caption{Speaker-dependent all-speaker mean word classification, $C\pm1\frac{\sigma}{\sqrt{7}}$, over all four speakers comparing consonant P2V maps. For a given consonant mapping ($x-$axis) the performance is measured after pairing with all vowel mappings.} 
\label{fig:all_con} 
\end{figure} 

\begin{figure}[!ht] 
\centering 
\includegraphics[width=0.98\textwidth]{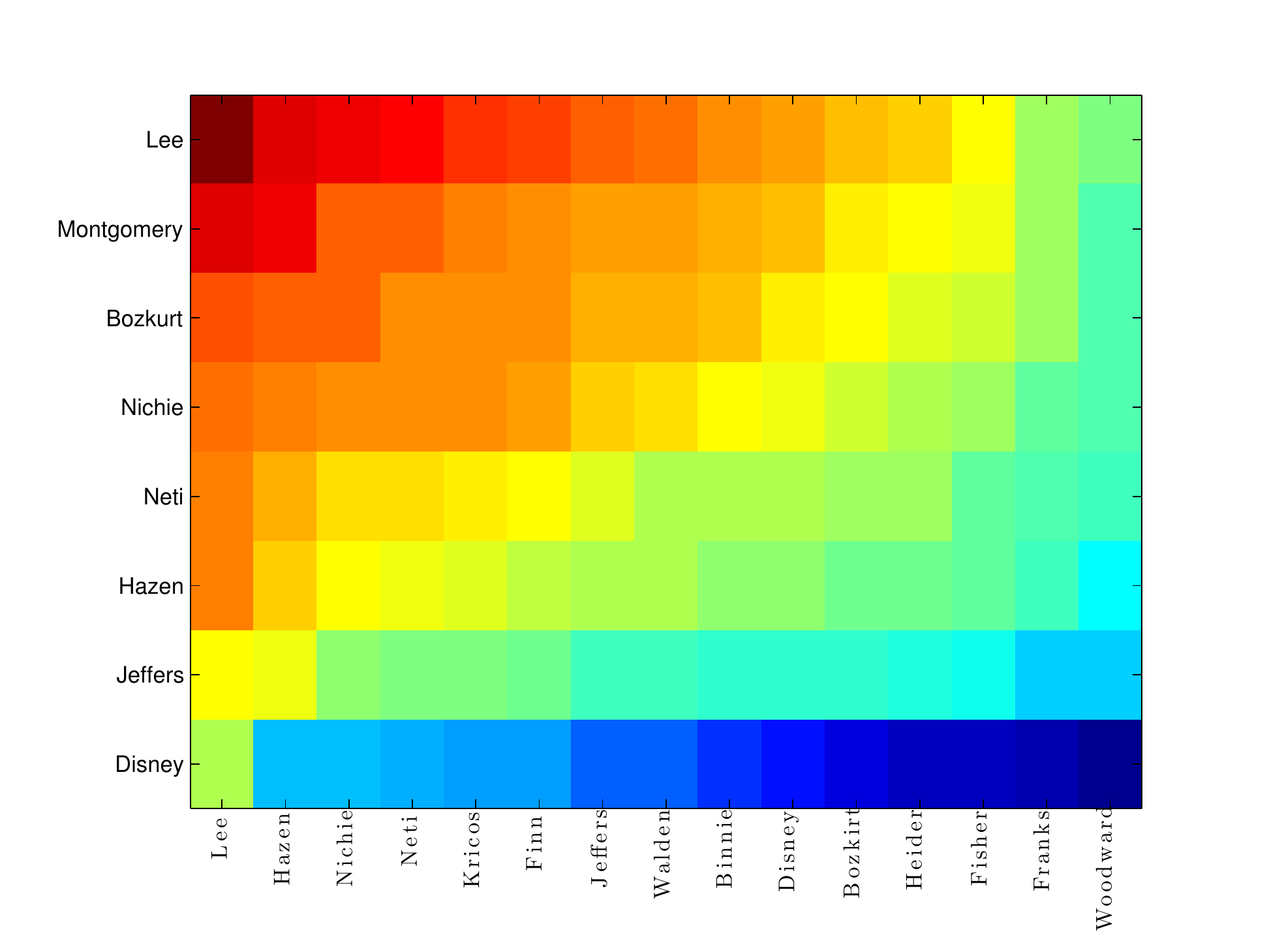} 
\caption{Speaker-dependent all-speaker mean word classification, $C$, heatmap.} 
\label{fig:heat_con} 
\end{figure} 
 
\begin{figure}[H] 
\centering 
\includegraphics[width=0.98\textwidth]{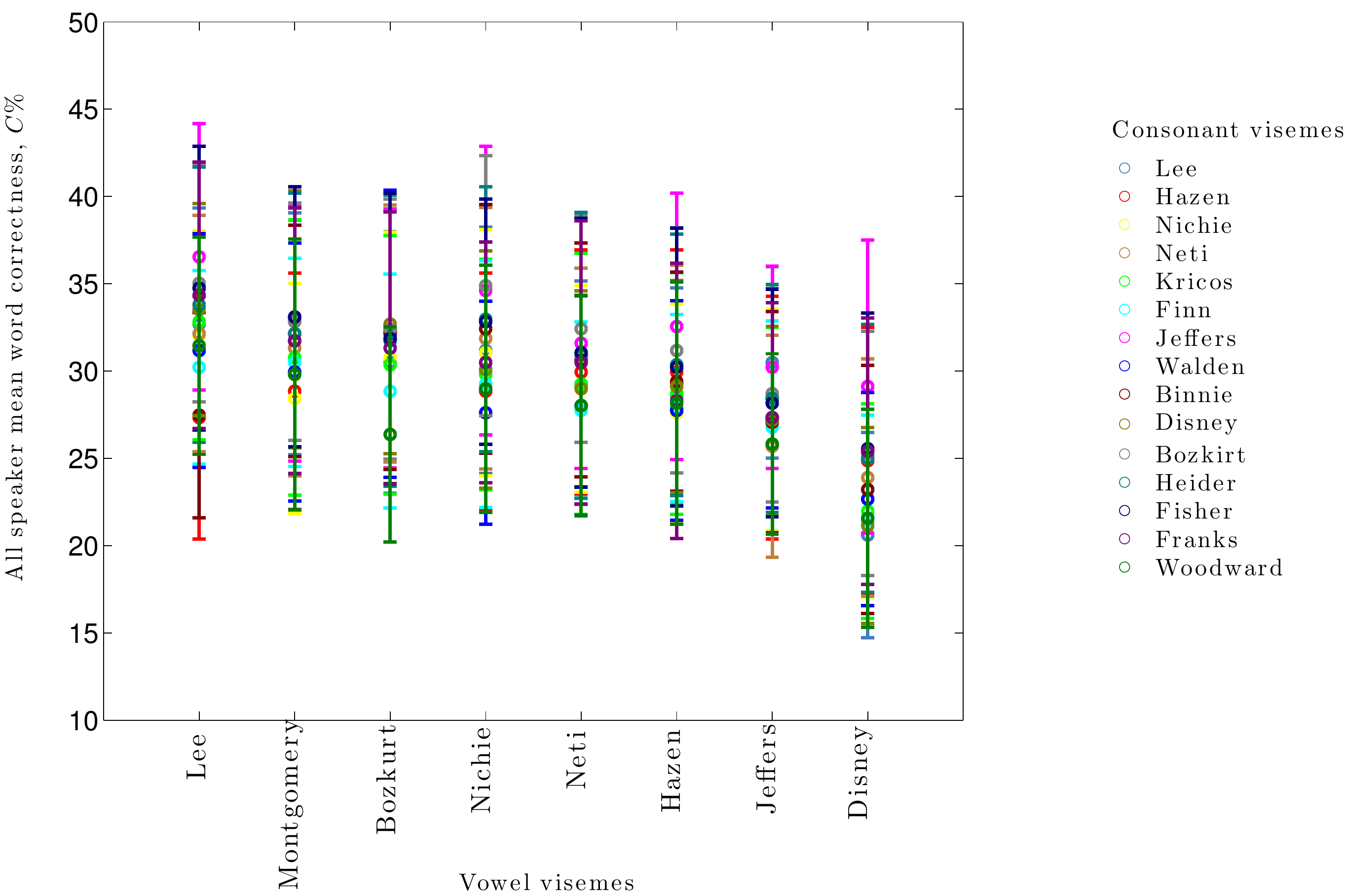} 
\caption{Speaker-dependent all-speaker mean word classification, $C\pm1\frac{\sigma}{\sqrt{7}}$, over all four speakers comparing vowel P2V maps. For a given vowel mapping ($x-$axis) the performance is measured after pairing with all consonant mappings.} 
\label{fig:all_vow} 
\end{figure}

\begin{figure}[h] 
\centering 
\includegraphics[width=0.98\textwidth]{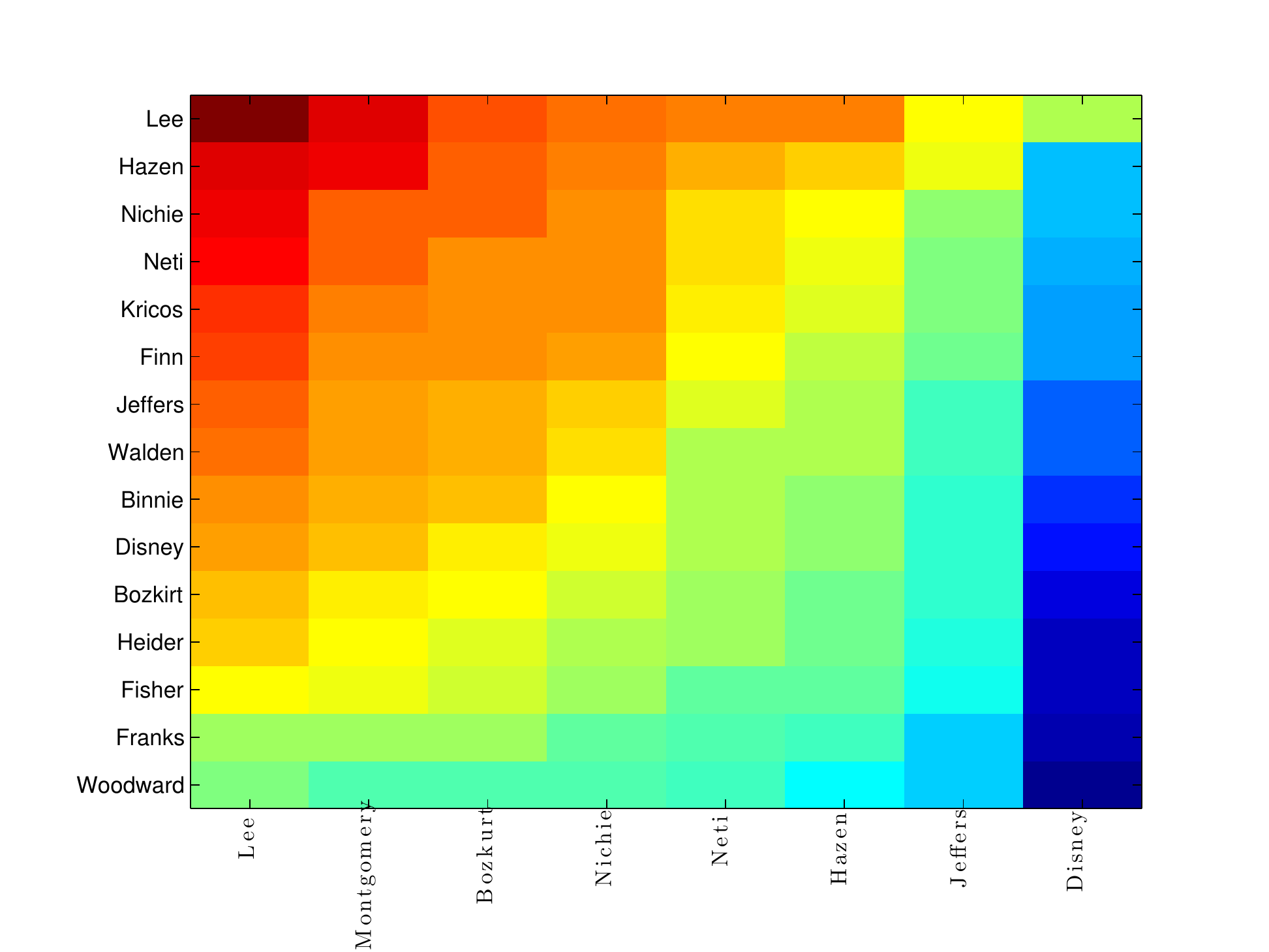} 
\caption{Speaker-dependent all-speaker mean word classification, $C$, heatmap.} 
\label{fig:heat_vow} 
\end{figure}

 Figures~\ref{fig:all_con} and~\ref{fig:all_vow} show the word correctness percentage aggregated over all speakers, $\pm1\frac{\sigma}{\sqrt{7}}$. Respective heat maps for all phoneme-to-viseme maps are in Figures~\ref{fig:heat_con} \&~\ref{fig:heat_vow}. Figure~\ref{fig:all_con} shows all consonant maps along the $x$-axis and, for each consonant map, a pairing with a vowel map has been plotted at the respective consonant map position on the $x$-axis. This shows the differences between each consonant map and the effect of the vowel maps on each consonant map. Figure~\ref{fig:all_vow} is vice versa. The black line is the mean word classification grouped by all paired maps. Both $x-$axies are ordered by the map's mean rank over all speakers. This demonstrates the `best' performing map for both consonants and vowels are from Lee (as this is left-most on the $x-$axis) for all speakers. Therefore, Lee's visemes \cite{lee2002audio} become the benchmark in the next piece of work in this chapter.

Comparing the consonant P2V maps in Figure~\ref{fig:all_con} shows the Disney vowels are significantly worse than all others when paired with all consonant maps. Over the other vowels there is overlap with the majority of error bars suggesting little significant difference over the whole group, although Lee \cite{lee2002audio} and Bozkurt \cite{bozkurt2007comparison} vowels are consistently above the mean and above the upper error bar for Disney \cite{disney}, Jeffers \cite{jeffers1971speechreading} and Hazen \cite{Hazen1027972} vowels. In comparing the vowel P2V maps in Figure~\ref{fig:all_vow} Lee \cite{lee2002audio} and Hazen \cite{Hazen1027972} are the best consonants by a margin above the mean whereas Woodward \cite{woodward1960phoneme} and Franks \cite{franks1972confusion} are the bottom performers. Figures~\ref{fig:all_con} and~\ref{fig:all_vow} show the performance of the viseme maps averaged across speakers, there is a significant difference between the `best' visemes for individual speakers which arises from the unique way in which everyone articulates their speech. 

These observations are confirmed in heatmaps in Figures~\ref{fig:heat_con} \&~\ref{fig:heat_vow}.
 
\begin{figure}[!htbp] 
\centering 
\includegraphics[width=0.85\textwidth]{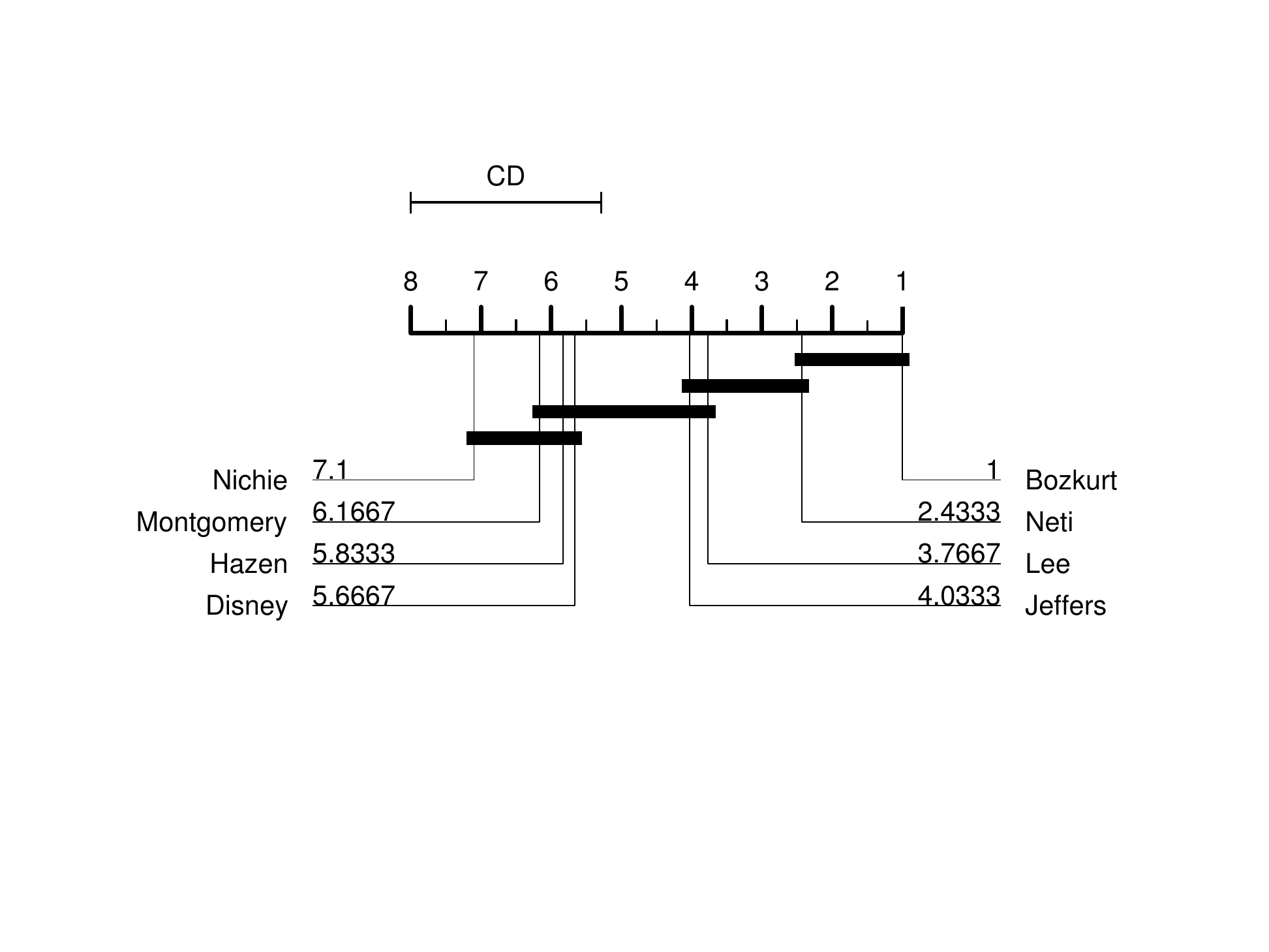} 
\caption{Critical difference of all vowel phoneme-to-viseme maps independent of consonant phoneme-to-viseme map pair partner.} 
\label{fig:con_crit_diff} 
\end{figure}
\begin{figure}[!htbp]
\includegraphics[width=0.85\textwidth]{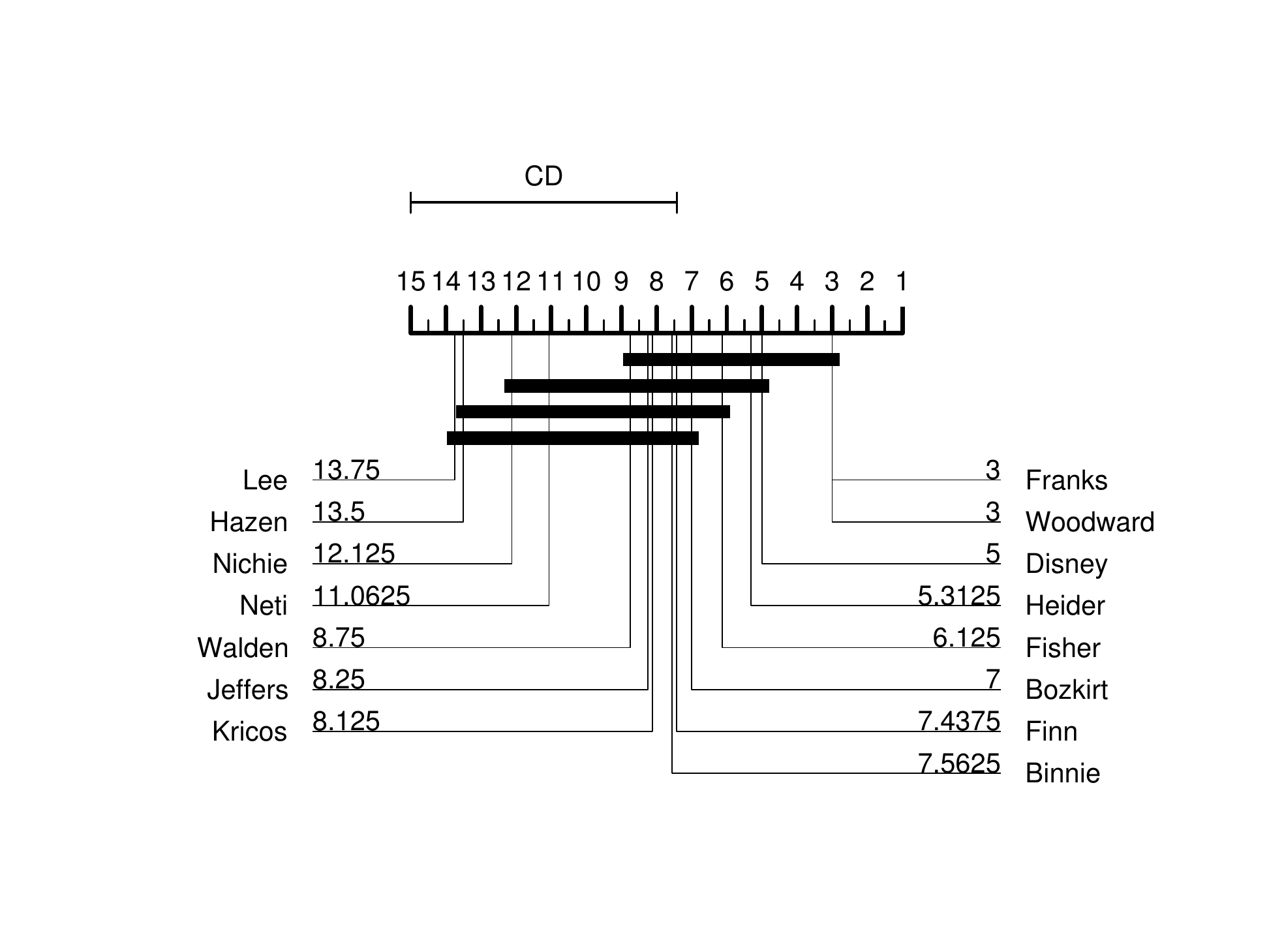} 
\caption{Critical difference of all consonant phoneme-to-viseme maps independent of vowel phoneme-to-viseme pair partner.} 
\label{fig:vow_crit_diff} 
\end{figure} 
 
Figures~\ref{fig:con_crit_diff} and~\ref{fig:vow_crit_diff} are critical difference plots between the viseme class sets based upon their classification performance \cite{criticaldiff}. Critical difference is a measure of our confidence intervals between different machine learning algorithms. Two assumptions within critical difference are: all measured results are `reliable', and all algorithms are evaluated using the same random samples \cite{criticaldiff}. As we use the HTK standard metrics \cite{young2006htk}, and use results with consistent random sampling across folds, these assumptions are not a concern. We have selected critical differences here as these evaluate the performance of multiple classifiers, and previous studies, such as \cite{bouckaert2004evaluating, bengio2004no}, do not consider the applicability of statistics when tested over more than one dataset \cite{criticaldiff}. As our HMM classifiers are speaker dependent, we can safely consider the data of each speaker as an isolated dataset within AVL2. 

Figure~\ref{fig:con_crit_diff} is the comparison of the vowel labelled viseme sets. Starting on the left-hand side of the figure, it shows that Nichie, Montgomery, Hazen, and Disney vowels are not critically different from each other signified by the black horizontal bar crossing their respective lines on the left side of the figure. Likewise, Montgomery, Hazen, Disney, Jeffers, and Lee vowels are also not critically different from each other. These two bars alone demonstrate that Nichie's vowels are critically different from Jeffers, Lee, Neti, and Bozkurt's. On the right hand side of the graph we can see that Bozkurt's vowels are critically different from all bar Neti's vowels. This is interesting as in Figure~\ref{fig:all_vow} they do not appear to perform significantly differently to any other vowel visemes. In fact, whilst Bozkurt and Nichie vowels are the most critically different from each other, they are adjacent in classification performance. This gives us hope that an optimal set of visemes is possible as the effect of clusters of phonemes varies by the specific phonemes being clustered.

Figures~\ref{fig:con_crit_diff} and~\ref{fig:vow_crit_diff}  demonstrate a significant difference between some sub-sets of viseme sets (the bars do not overlap all classifier maps). This is based upon insignificant variation within each sub-set. This suggests there could be dependency between some viseme sets as the groupings align with the derivation method of the P2V mappings. 
  
\begin{figure}[p] 
\centering 
%\setlength{\tabcolsep}{1pt} 
%	\begin{tabular}{c c} 
\includegraphics[width=0.85\textwidth]{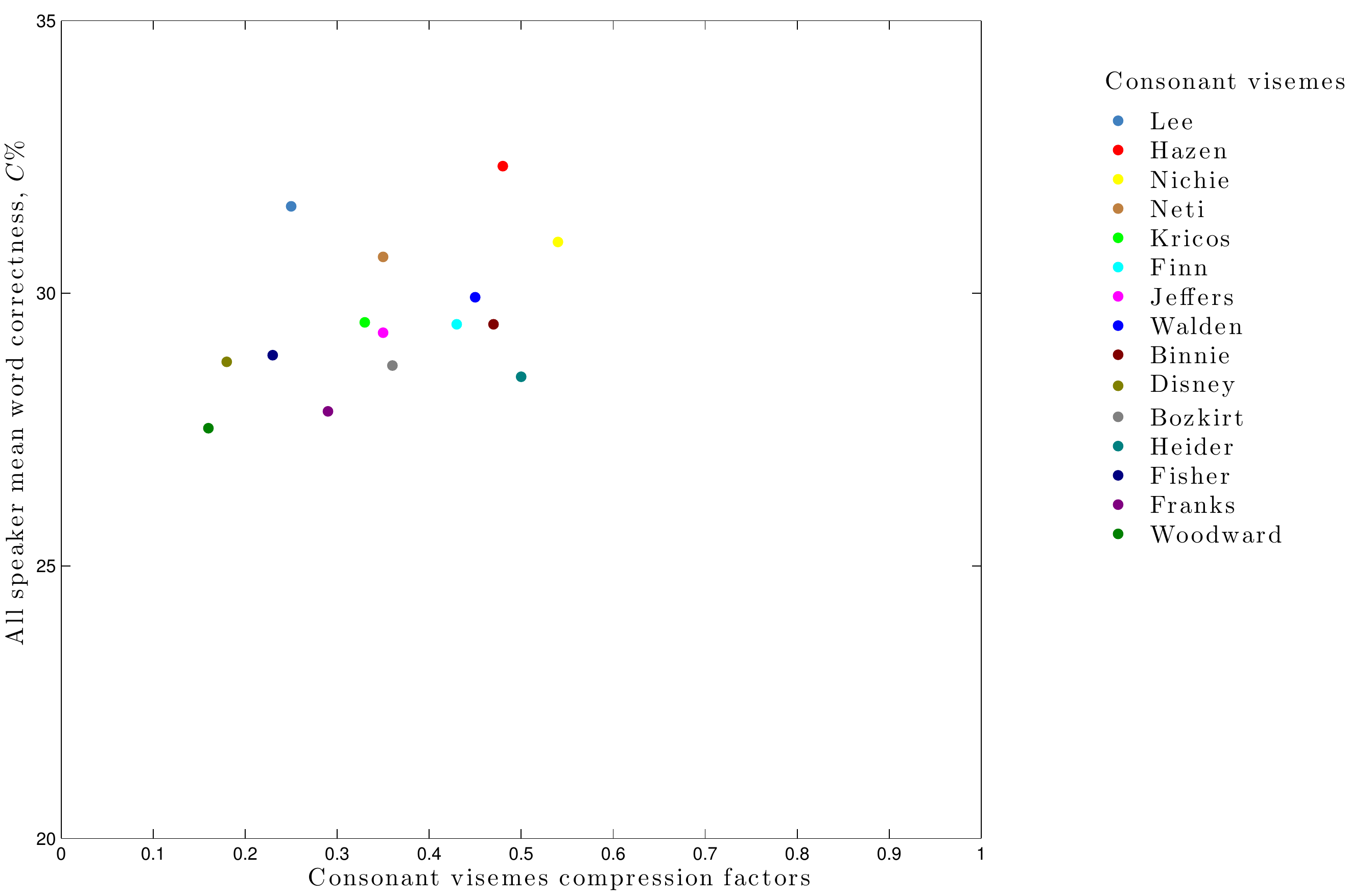} 
\caption{Scatter plot showing the relationship between compression factors and word correctness, $C$, classification with consonant phoneme-to-viseme maps.} 
\label{fig:scatter_con} 
%\end{figure}
%\begin{figure}[!ht]
%\centering
\includegraphics[width=0.85\textwidth]{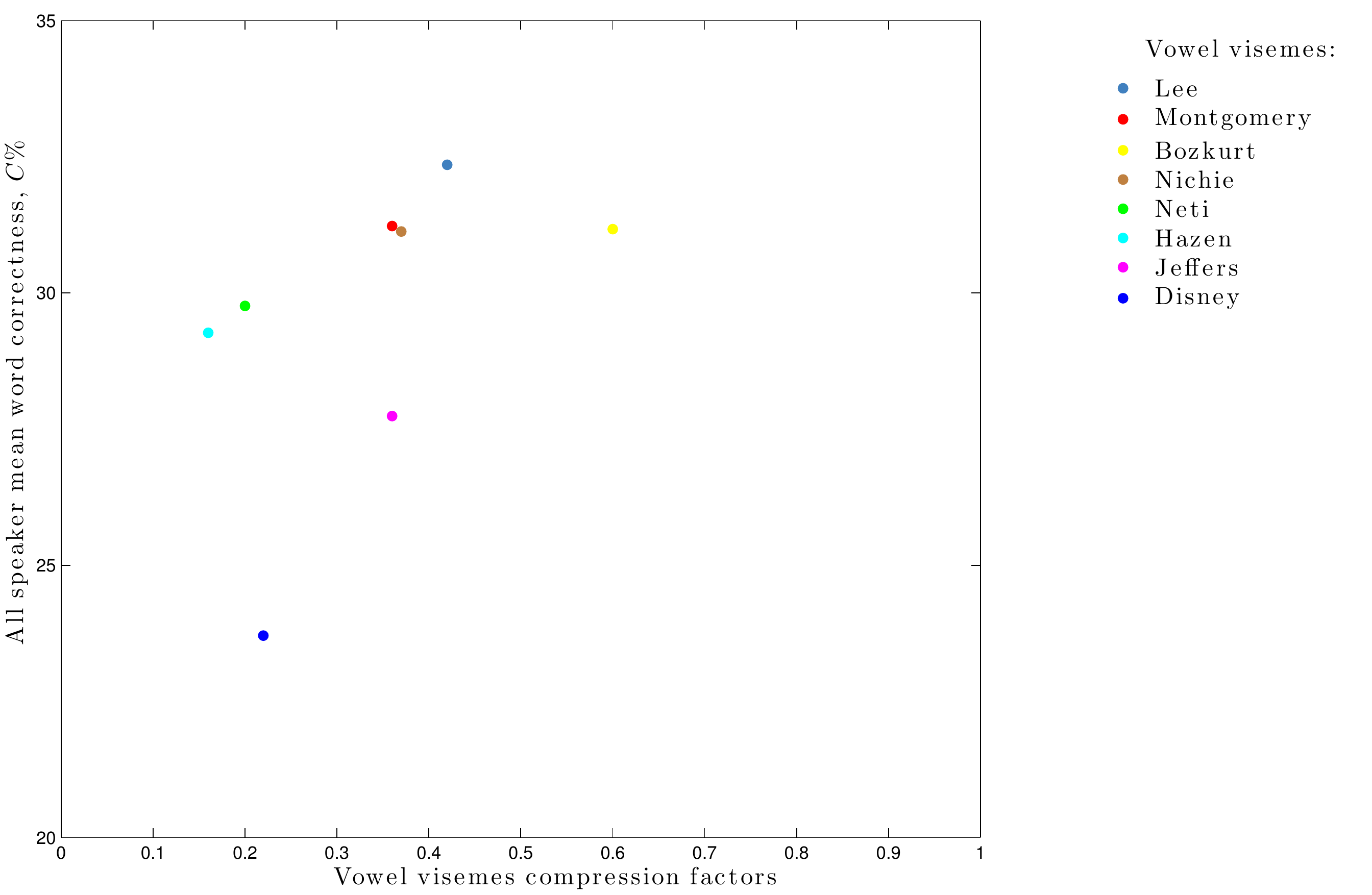} 
%		(a) Consonant Viseme Sets &	(b) Vowel Viseme Sets \\ 
%	\end{tabular} 
\caption{Scatter plot showing the relationship between compression factors and word correctness, $C$, classification with vowel phoneme-to-viseme maps.} 
\label{fig:scatter_vow} 
\end{figure} 
 
\begin{figure}[!ht]
\centering 
\includegraphics[width=0.85\textwidth]{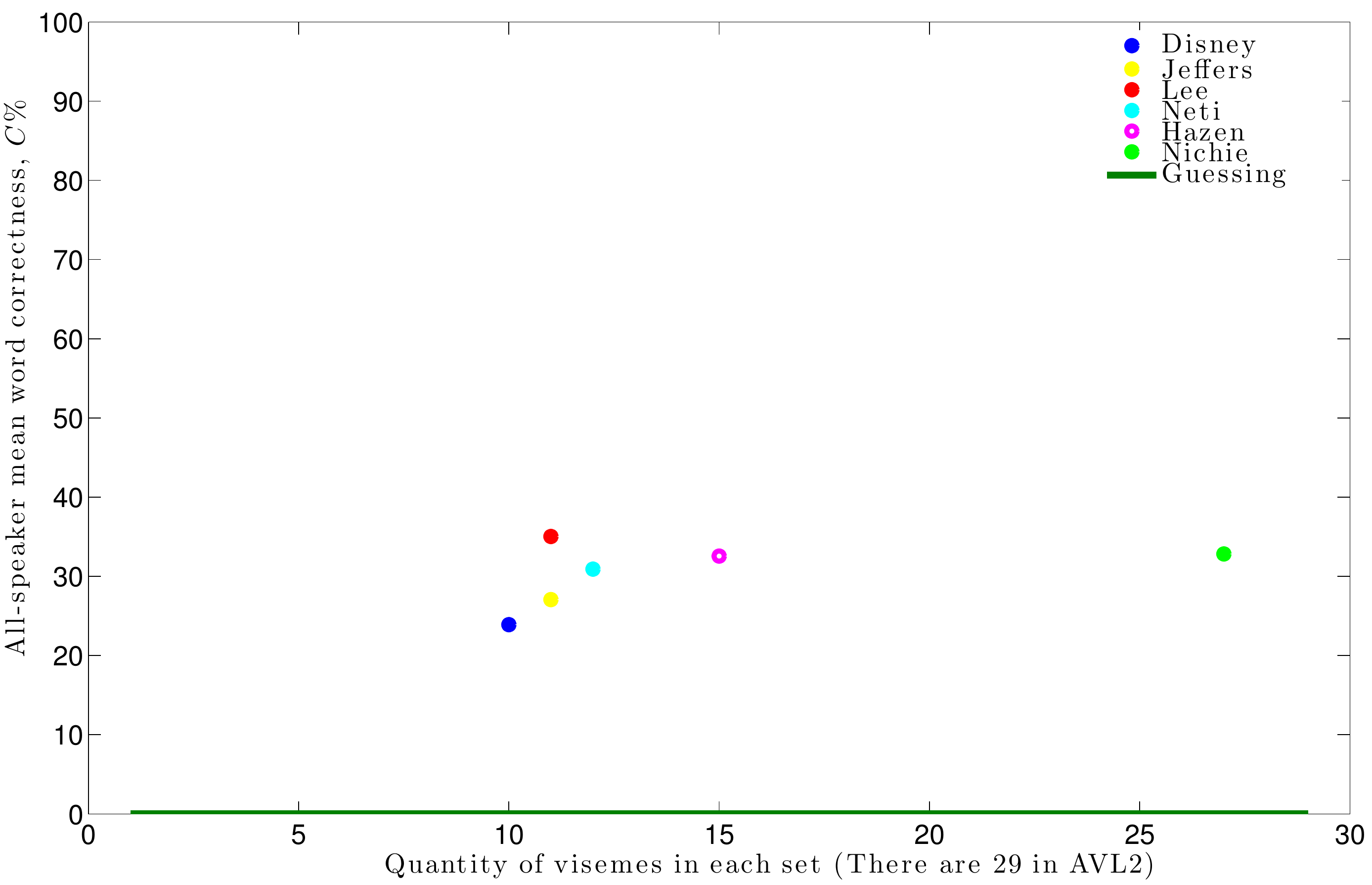} 
\caption{For previously presented phoneme-to-viseme maps which include both vowel and consonant phonemes, word correctness, $C$ is plotted against the count of visemes in each phoneme-to-viseme map.} 
\label{fig:currentSetsVisemeCounts} 
%\caption{Six phoneme-to-viseme mappings from literature with both consonant and vowel phonemes and their viseme count} 
%\begin{tabular}{| l | l |} 
%\hline 
%Viseme Set & Viseme Count \\ 
%\hline 
%Disney & 10 \\ 
%Jeffers & 11 \\ 
%Lee & 11 \\ 
%Neti & 12 \\ 
%Hazen & 14 \\ 
%Nichie & 27 \\ 
%\hline 
%\end{tabular} 
%\label{tab:visemecountsinliterature} 
\end{figure}

The mean word classification for all speakers and all folds for each map is plotted in Figures~\ref{fig:scatter_con} and~\ref{fig:scatter_vow}. Looking at our confusion factors for the best performing P2Vs of each speaker (Figure~\ref{fig:scatter_con} and Figure~\ref{fig:scatter_vow}), this suggests a good preparation of phonemes to visemes is ideally around 0.45 or approximately {\raise.17ex\hbox{$\scriptstyle\sim$}}2 phonemes per viseme. This also is the CF for Lee. Lee has the highest performing word classification map for both consonants and vowels displayed in Figure~\ref{fig:currentSetsVisemeCounts} and interestingly, not the highest number of visemes (the $x$-axis in Figure~\ref{fig:currentSetsVisemeCounts}).
 
\section{New phoneme to viseme maps} 
In the second part of our phoneme-to-viseme mapping study, three approaches are used to find a better method of mapping phonemes to visemes. The first approach uses the most common pairs of phonemes from existing mappings. 

A comparison of previously presented P2V maps shows subgroups of phonemes which are regularly grouped together into visemes \cite{cappelletta2012phoneme, theobaldPHD}. The most popular of these phoneme-subgroups have a high occurrence across sets. Our first new approach uses the number of occurrences and the size of the subgroup as a weighting for grouping together phonemes, i.e. the highest weighted phoneme-subgroup will be grouped into a viseme first, without duplicating phonemes into more than one viseme. The P2V maps used in this clustering process have been devised for different reasons (for example, based upon linguistic rules or upon human lip-reader observations, see Table~\ref{tab:lit_visemes_compare}).  This set helps us to understand that if what we currently assume to be good groups really are the best groups of phonemes for optimal classification. 

The second and third approaches are both speaker-dependent and data-driven from phoneme classification. Two cases are considered: 
\begin{enumerate}
\item a strictly coupled map, where a phoneme can be grouped into a viseme only if it has been confused with \textit{all} the phonemes within the viseme, and
\item a relaxed coupled case, where phonemes can be grouped into a viseme if it has been confused with \textit{any} phoneme within the viseme. 
\end{enumerate}

With all new P2V mappings each phoneme can only be allocated to one viseme class. These new P2V maps are tested on the AVL2 dataset using the same classification method as described in Section~\ref{sec:recognition}. The results from the best performing P2V map from our comparison study (Lee \cite{lee2002audio}) is the benchmark to measure improvements. 
 
\subsection{Common phoneme-pair visemes} 
The first approach for finding a new speaker-independent P2V map uses the most commonly coupled phonemes to build new visemes. In detail, all visemes in the previous maps are searched to make a full dictionary of unique pairs of phonemes. Associated with each dictionary entry is a count of how many times they appear in \textit{any} defined P2V map from those in the comparison study in Section~\ref{sec:comparison} with HTK. This phoneme pair list is sorted by descending occurrence count. On passing through this list the next phoneme pair is assigned to a viseme class based upon matching phonemes (whilst not duplicating the presence of a phoneme within one viseme). A phoneme is not permitted to be added to more than one viseme. Priority is given to the pairings with a higher count. If a particular phoneme was never coupled with other phonemes, that phoneme forms a unique viseme of its own. 
 
\begin{table}[!h] 
\centering 
\caption{Visemes derived using most-common phoneme pairings in previously presented phoneme-to-viseme mappings.} 
\begin{tabular}{| l | l |} 
\hline 
Common-pair Visemes 	& {\footnotesize \{/d/ /l/ /n/ /t/\} \{/b/ /m/ /p/\}  \{/g/ /h/ /\textipa{H}/ /k/ /\textipa{N}/ /y/\} \{/f/ /v/\}  } \\ 
(CF:0.28)			& {\footnotesize \{/\textopeno/ /\textupsilon\textschwa/ /\textopeno\textschwa/\} \{/\textepsilon\textschwa/ /i/\} \{/t\textipa{S}/ /d\textipa{Z}/ /\textipa{S}/ /\textipa{Z}/\}  \{/\textscripta/ /\textturnv/ /ao\}    \{/s/ /z/\}  }\\ 
& {\footnotesize\{/dh/ /\textipa{T}/\} \{/r/ /w/ /\textipa{W}/\} \{/\ae/ /e/ /ei/ /\textsci\textschwa/\} 	} \\ 
& {\footnotesize \{/a/ /ai/ /ai/ /e/ /i/ /\textsci/\} \{/\textscripta\textupsilon/ /\textschwa/ /\textrevepsilon/ /\textschwa\textupsilon/ /u/ /\textupsilon/ /u/\}  } \\ 
\hline 
\end{tabular} 
\label{tab:evolvedvisemes} 
\end{table} 
 
\subsection{Viseme classes with strictly confusable phonemes} 
\label{sec:strict_confuse}
The second and third approaches for identifying visemes are speaker-dependent, data-driven and based on phoneme confusions within the classifier. The first undertaking in this work is to complete classification using phoneme labelled HHM classifiers. The classifiers are built in HTK with flat-started HMMs and force aligned training data for each speaker. The HMMs are re-estimated 11 times  in total over seven folds of leave-one-out cross validation. This overall classification task does not perform well (see Table~\ref{tab:phonemeCorrAVL}) particularly for an isolated word dataset. However, the HTK tool \texttt{HResults} is used to output a confusion matrix for each fold detailing which phoneme labels confuse with others and how often. For both data-driven speaker-dependent approaches, this first step of completing phoneme classification is essential to create the data to derive the P2V maps from. 

\begin{table}[!hb]
\centering
\caption{Mean per speaker Correctness, $C$, of phoneme-labelled HMM classifiers.}
\begin{tabular}{|l|r|r|r|r|}
\hline
 			& Speaker 1 & Speaker 2 & Speaker 3 & Speaker 4 \\
\hline \hline
Phoneme $C$ 	& $24.72$ & $23.63$ & $57.69$ & $43.41$ \\
\hline
\end{tabular}
\label{tab:phonemeCorrAVL}
\end{table}

Now, let us use a smaller seven-unit confusion matrix example to explain our clustering method in full. Our demonstration confusion matrix is in Figure~\ref{fig:demoCMforclustering}.

\begin{figure}[!ht]
\centering
\begin{tabular}{|l||r|r|r|r|r|r|r|}
\hline
& $/p1/$ & $/p2/$ & $/p3/$ & $/p4/$ & $/p5/$ & $/p6/$ & $/p7/$ \\
\hline \hline
$/p1/$ & {\textcolor{red}1} & 0 & 0 & 0 & 0 & 0 &  {\textcolor{blue}4} \\ 
$/p2/$ & 0 &  {\textcolor{red}0} & 0 & {\textcolor{blue}2} & 0 & 0 & 0 \\
$/p3/$ & {\textcolor{blue}1} & 0 &  {\textcolor{red}0} & 0 & 0 & 0 & {\textcolor{blue}1} \\
$/p4/$ & 0 & {\textcolor{blue}2} & {\textcolor{blue}1} &  {\textcolor{red}0} & {\textcolor{blue}2} & 0 & 0 \\
$/p5/$ & {\textcolor{blue}3} & 0 & {\textcolor{blue}1} & {\textcolor{blue}1} &  {\textcolor{red}1} & 0 & 0 \\
$/p6/$ & 0 & 0 & 0 & 0 & 0 &  {\textcolor{red}4} & 0 \\
$/p7/$ & {\textcolor{blue}1} & 0 & {\textcolor{blue}3} & 0 & 0 & 0 &  {\textcolor{red}1} \\
\hline
\end{tabular}
\caption{Demonstration (theoretical) confusion matrix showing confusions between phoneme-labelled classifiers to be used for clustering to create new speaker-dependent visemes. True positive classifications are shown in red, confusions of either false positives and false negatives are shown in blue. The estimated classes are listed horizontally and the real classes are vertical.}
\label{fig:demoCMforclustering}
\end{figure}
 
For the `strictly-confused' viseme set (remember there is one per speaker), the second step of deriving the P2V map is to check for single-phoneme visemes. Any phonemes which have only been correctly recognised as themselves and have no false positive/negative classifications are permitted to be single phoneme visemes. In Figure~\ref{fig:demoCMforclustering} we have highlighted the true positive classifications in red and both false positives and false negative classifications in blue which shows $/p6/$ is the only phoneme to fit our `single-phoneme viseme' definition. $/p6/$ has a true positive value of $+4$ and zero false classifications. Therefore this is our first viseme. $/v1/ = \{/p6/\}$. 

This action is followed by defining all combinations of remaining phonemes which can be grouped into visemes and identifying the grouping that contains the largest number of confusions by ordering all the viseme possibilities by descending size (whole list shown in Figure~\ref{fig:listofpossiblecombs}). 

\begin{figure}[!ht]
\centering
\begin{tabular} {l l l}
$\{/p1/, /p2/, /p3/, /p4/, /p5/, /p7/\}$ & $\{/p1/, /p2/, /p3\}$ & $\{/p1/, /p2/\}$\\
$\{/p1/, /p2/, /p3/, /p4/, /p5/ \}$ & $\{/p1/, /p2/, /p4/\}$ & $\{/p1/, /p3/\}$\\
$\{/p1/, /p2/, /p3/, /p4/, /p7/\}$ & $\{/p1/, /p2, /p5/\}$ & $\{/p1/, /p4/\}$\\
$\{/p1/, /p2/, /p3/, /p5/, /p7/\}$ & $\{/p1/, /p2, /p7/\}$ & $\{/p1/, /p5/\}$ \\
$\{/p1/, /p2/, /p4/, /p5/, /p7/\}$ & $\{/p2/, /p3, /p4/\}$ & $\{/p1/, /p7/\}$\\
$\{/p1/, /p3/, /p4/, /p5/, /p7/\}$ & $\{/p2/, /p3, /p5/\}$ & $\{/p2/, /p3/\}$ \\
$\{/p2/, /p3/, /p4/, /p5/, /p7/\}$ & $\{/p2/, /p3, /p7/\}$ & $\{/p2/, /p4/\}$\\
$\{/p1/, /p2/, /p3/, /p4/ \}$ & $\{/p3/, /p4/, /p5/\}$ & $\{/p2/, /p5/\}$ \\
$\{/p1/, /p2/, /p3/, /p5/ \}$ & $\{/p3/, /p4/, /p7/\}$ & $\{/p2/, /p7/\}$ \\
$\{/p1/, /p2/, /p3/, /p7/ \}$ & $\{/p1/, /p3/, /p4/\}$ & $\{/p3/, /p4/\}$ \\
$\{/p2/, /p3/, /p4/, /p5/ \}$ & $\{/p4/, /p5/, /p7/\}$ & $\{/p3/, /p5/\}$ \\
$\{/p2/, /p3/, /p4/, /p7/ \}$ & $\{/p1/, /p4/, /p5/\}$ & $\{/p3/, /p7/\}$ \\
$\{/p3/, /p4/, /p5/, /p7/ \}$ & $\{/p2/, /p4/, /p5/\}$ & $\{/p4/, /p5/\}$ \\
$\{/p1/, /p3/, /p4/, /p5/ \}$ & $\{/p1/, /p5/, /p7/\}$ & $\{/p4/, /p7/\}$  \\
$\{/p1/, /p4/, /p5/, /p7/ \}$ & $\{/p2/, /p5/, /p7/\}$ &$\{/p5/, /p7/\}$  \\
$\{/p2/, /p4/, /p5/, /p7/ \}$ & $\{/p3/, /p5/, /p7/\}$ & \\
& $\{/p1/, /p3/, /p5/\}$  & \\
& $\{/p1/, /p3/, /p7/\}$  & \\
& $\{/p1/, /p4/, /p7/\}$  & \\
& $\{/p2/, /p4/, /p7/\}$  & \\
\end{tabular}
\caption{List of all possible subgroups of phonemes with an example set of seven phonemes}
\label{fig:listofpossiblecombs}
\end{figure}

Our grouping rule states that phonemes can be grouped into a viseme class only if all of the phonemes within the candidate group are mutually confusable. This means each pair of phonemes within a viseme must have a total false positive and false negative classification greater than zero. Once a phoneme has been assigned to a viseme class it can no longer be considered for grouping, and so any possible phoneme combinations that include this viseme are discarded. This ensures phonemes can belong to only a single viseme. 

By iterating though our list of all possibilities in order, we check if all the phonemes are mutually confused. This means all phonemes have a positive confusion value (a blue value in Figure~\ref{fig:demoCMforclustering}) with all others. 

The first phoneme possibility in our list where this is true is $\{/p1/, /p3/, /p7/\}$. This is confirmed by the Figure~\ref{fig:demoCMforclustering} values: \\%row + col
Pr$\{/p1/|/p3/\} + $Pr$\{/p3/|/p1/\} = 0 + 1 = 1$ which is $> 0 $ \\
also,
Pr$\{/p1/|/p7/\} + $Pr$\{/p7/|/p1/\} = 4 + 1 = 5 $ which is $ > 0 $ \\
and
Pr$\{/p3/|/p7/\} + $Pr$\{/p7/|/p3/\} = 1 + 3 = 4 $ which is $ > 0 $.

This becomes our second viseme and thus our current viseme list looks like Table~\ref{tab:thirdpass}.

\begin{table}[!ht]
\centering
\caption{Demonstration example 1: first-iteration of clustering, a phoneme-to-viseme map for strictly-confused phonemes.}
	\begin{tabular} {|l|l|}
	\hline
	Viseme & Phonemes \\
	\hline \hline
	$/v1/$ & $\{/p6/\} $ \\
	$/v2/$ & $\{/p1/, /p3/, /p7/\}$ \\
	\hline
	\end{tabular}
\label{tab:thirdpass}
\end{table}
We now only have three remaining phonemes to cluster, $p2, p4$ and $p5$. This reduces our list of possible combinations substantially, see Figure~\ref{fig:listofpossiblecombs1}.\\
\begin{figure}
\centering
\begin{tabular} {l}
$\{/p2/, /p4/, /p5/\}$ \\
$\{/p2/, /p4/\} $ \\
$\{/p2/, /p5/\} $ \\
$\{/p4/, /p5/\} $ \\
\end{tabular}
\caption{List of all possible subgroups of phonemes with an example set of seven phonemes after the first viseme is formed.}
\label{fig:listofpossiblecombs1}
\end{figure}

The next iteration of our clustering algorithm identifies the combination of remaining phonemes which correspond to the next largest number of confusions, and so on, until no phonemes can be merged. This leaves us with the final visemes in Table~\ref{tab:firstpass}.
\begin{table}[!ht]
\centering
\caption{Demonstration example 2: final phoneme-to-viseme map for strictly-confused phonemes.}
	\begin{tabular} {|l|l|}
	\hline
	Viseme & Phonemes \\
	\hline \hline
	$/v1/$ & $\{/p6/\} $ \\
	$/v2/$ & $\{/p1/, /p3/, /p7/\}$ \\
	$/v3/$ & $\{/p2/, /p4/\}$ \\
	$/v4/$ & $\{/p5/\} $ \\
	\hline
	\end{tabular}
\label{tab:firstpass}
\end{table}

Our original phoneme classification has produced confusion matrices which permit confusions between vowel and consonant phonemes. We can see in Section~\ref{sec:currentp2vmaps} (Tables~\ref{tab:vowelmappings}~and~\ref{tab:consonantmappings}), previously presented P2V maps that vowel and consonant phonemes are not commonly mixed within visemes. Therefore, we make two types of P2V maps: one which permits vowels and consonant phonemes to be mixed within the same viseme, and a second which restricts visemes to be vowel or consonant only by putting an extra condition in when checking for confusions greater than zero. 
 
It should be remembered that not all phonemes present in the ground truth transcripts will have been recognised and included in the phoneme confusion matrix. Any of the remaining phonemes which have not been assigned to a viseme are grouped into a single garbage $/gar/$ viseme. This approach ensures any phonemes which have been confused are grouped into a viseme and we do not lose any of the `rarer', and less common visual phonemes. For example, $/ea/$, $/oh/$, $/ao/$, and $/r/$ are not in the original transcript and so can be placed into $/gar/$. But for Speaker 2, $/gar/$ also contains $/ay/$ and $/p/$, and for Speaker 4 $/gar/$ also contains $/p/$ and $/z/$, as these do not show up in the speaker's phoneme classification outputs. This task has been undertaken for all four speakers in our dataset. The final P2V maps are shown in Table~\ref{tab:tcvisemes_split}. 
 
\begin{table}[!ht] 
\centering
\caption{Strictly-confused phoneme speaker-dependent visemes. The score in brackets is the ratio of visemes to phonemes.} 
\begin{tabular} {|l|l|} 
\hline 
Classification & P2V mapping - permitting mixing of vowels and consonants\\ 
\hline \hline 
Speaker1	& {\footnotesize \{/\textturnv/ /ai/ /i/ /n/ /\textschwa\textupsilon/\} \{/b/ /e/ /ei/ /y/ \} \{/d/ /s/\} \{/t\textipa{S}/ /l/\} \{/\textschwa/ /v/\}} \\ 
(CF:0.48)	& {\footnotesize  \{/w/\} \{/f/\} \{/k/\}  \{/\textschwa/ /v/\} \{/d\textipa{Z}/ /z/\}  \{/\textscripta/ /u/\} 	 \{/t/\}  }\\ 
Speaker2 	& {\footnotesize\{/\textschwa/ /ai/ /ei/ /i/ /s/\} \{/e/ /v/ /w/ /y/\} \{/l/ /m/ /n/\} \{/b/ /d/ /p/\} }\\ 
(CF: 0.44)	&  {\footnotesize \{/z/\} \{t\textipa{S}/\} \{/t/\} \{/\textscripta/\}  \{/d\textipa{Z}/ /k/\} \{/\textturnv/ /f/\} \{/\textschwa\textupsilon/ /u/\}}	\\ 
Speaker3 	& {\footnotesize \{/ei/ /f/ /n/\} \{/d/ /t/ /p/\} \{/b/ /s/\} \{/l/ /m/\} \{/\textschwa/ /e/\} \{/i/\} \{/u/\}  }\\ 
(CF: 0.68)	& {\footnotesize \{/\textscripta/\} \{/d\textipa{Z}/\} \{/\textschwa\textupsilon/\} \{/z/\} \{/y/\} \{/t\textipa{S}\}/ \{/ai/\} \{/\textturnv/\} \{/\textscripta/\} \{/d\textipa{Z}/\}  \{/\textschwa\textupsilon/\}	} \\ 
& {\footnotesize \{/k/ /w/\} \{/v/\}  \{/z/\}   }\\ 
Speaker4	& {\footnotesize \{/\textturnv/ /ai/ /i/ /ei/ \} \{/m/ /n/\} \{/\textschwa/ /e/ /p/\} \{/k/ /w/\} \{/d/ /s/\} \{/d\textipa{Z}/ /t/\}  } \\ 
(CF: 0.64)	& {\footnotesize \{/f/\} \{/v/\} \{/\textscripta/\} \{/z/\} \{/t\textipa{S}/\} \{/b/\} \{/\textschwa\textupsilon/\}	  \{/\textschwa\textupsilon/\} \{/l/\} \{/u/\} \{/b/\} } \\ 
%	Multi-Speaker & \{/\textscripta/ /\textturnv/ /ai/ /u/ /m/ /n/ /\textschwa\textupsilon/ /ei/ /i/ /z/\} \{/b/ /d/ /e/ /p/ /y/\} 	\\ 
%	(CF: 0.36)	& \{/f/ /l/ /s/\} \{/k/ /v/ /w/\} \{/d\textipa{Z}/ /t/\} \{/\textschwa/ /t\textipa{S}/\}\\ 
\hline 
\hline 
Classification & P2V mapping - restricting mixing of vowels and consonants \\ 
\hline \hline 
Speaker1	& {\footnotesize \{/\textturnv/ /i/ /\textschwa\textupsilon/ /u/\} \{/\textscripta/ /ei/\} \{/\textschwa/ /e/ /ei/\} \{/d/ /s/ /t/ \} \{/t\textipa{S}/ /l/ \}  \{/k/\}  }\\ 
(CF:0.50)	& {\footnotesize \{/z/\} \{/w/\} \{/f/\} \{/m/ /n/\} \{/d\textipa{Z}/ /v/\} \{/b/ /y/\} 	}\\ 
Speaker2 	& {\footnotesize \{/ai/ /ei/ /i/ /u/\} \{/\textschwa\textupsilon/\} \{/\textschwa/\} \{/e/\} \{/\textturnv/\} \{/\textscripta/\} \{/v/ /w/\} \{/d\textipa{Z}/ /p/ /y/\}}\\ 
(CF: 0.58)	& {\footnotesize  \{/d/ /b/\} \{/t/\} \{/k/\} \{/t\textipa{S}/\}  \{/l/ /m/ /n/\}  \{/f/ /s/\}  }\\ 
Speaker3 	& {\footnotesize \{/ei/ /i/\} \{/ai/\} \{/\textschwa/ /e/\} \{/\textturnv/\} \{/d/ /p/ /t/\} \{/l/ /m/\} \{/k/ /w/\} \{/v/\}  }\\ 
(CF: 0.68)	& {\footnotesize \{/t\textipa{S}/\} \{/\textschwa\textupsilon/\} \{/y/\} \{/u/\} \{/\textscripta/\} \{/z/\}  \{/f/ /n/\} \{/b/ /s/\} \{/d\textipa{Z}/\}   }\\ 
Speaker4 	& {\footnotesize \{/\textturnv/ /ai/ /i/ /ei/\} \{/\textschwa/ /e/\} \{/m/ /n/\} \{/k/ /l/\} \{/d\textipa{Z}/ /t/\} \{/d/ /s/\} \{/t\textipa{S}/\} }\\ 
(CF: 0.65)	& {\footnotesize  \{/\textschwa\textupsilon/\} \{/y/\} \{/u/\} \{/\textscripta/\}  \{/w/\} \{/f/\} \{/v/\} \{/b/\} }\\ 
%	Multi-Speaker & \{/\textturnv/ /ai/ /u/ /\textschwa\textupsilon/ /ei/ /i/\} \{/\textscripta/\} \{/\textschwa/ /e/\} \{/f/ /l/ /n/\} \{/d/ /t/ /s/ /v/\} 	\\ 
%	(CF: 0.36)	& \{/b/ /w/ /y/\} \{/d\textipa{Z}/\} \{/z/\} \{/p/\} \{/m/\} \{/k/\} \{/t\textipa{S}/\}\\ 
\hline 
\end{tabular} 
\label{tab:tcvisemes_split} 
\end{table} 
 
\subsection{Viseme classes with relaxed confusions between phonemes} 
\label{sec:loose_confuse}
A disadvantage of the strictly confusable viseme set is that it contains some spurious single-phoneme visemes where the phoneme cannot be grouped because it is not confused with \emph{all} other phonemes in the viseme. These types of phonemes are likely to be either: borderline cases at the extremes of a viseme cluster, i.e. they have subtle visual similarities to more than one phoneme cluster, or they do not occur frequently enough in the training data to be differentiated from other phonemes. 

To address this we complete a second pass-through of the strictly-confused visemes listed in Table~\ref{tab:firstpass}. We begin with the visemes as they currently stand (in our demonstration example containing four classes) and relax the condition requiring confusion with all of the phonemes. Now any single phoneme viseme (in our demonstration, $/v4/$) can be allocated to a previously existing viseme if it has been confused with any phoneme in the viseme. In Figure~\ref{fig:demoCMforclustering} we see $/p5/$ was confused with $/p1/$, $/p3/$, and $/p4/$. Because $/p4/$ is not in the same viseme as $/p1/$ and $/p3/$ we use the value of confusion to decide which to allocate it to as follows. \\
Pr$\{/p1/|/p5/\} + $Pr$\{/p5/|/p1/\} = 0 + 3 = 3$ \\
Pr$\{/p3/|/p5/\} + $Pr$\{/p5/|/p3/\} = 0 + 1 = 1$ \\
Pr$\{/p4/|/p5/\} + $Pr$\{/p5/|/p4/\} = 2 + 1 = 3$ \\
Therefore; for $p5$ the total confusion with $/v2/$ is $3+1=4$, whereas the total confusion with $/v3/$ is $3$. We select the viseme with most confusion to incorporate the unallocated phoneme $/p5/$. This reduces the number of viseme classes by merging single-phoneme visemes from Table~\ref{tab:firstpass} to form a second set shown in Table~\ref{tab:secondpass}. This has the added benefit that we have also increased the number of training samples for each classifier. 

\begin{table}[!ht]
\centering
\caption{Demonstration example 3: final phoneme-to-viseme map for relaxed-confused phonemes.}
	\begin{tabular} {|l|l|}
	\hline
	Viseme & Phonemes \\
	\hline \hline
	$/v1/$ & $\{/p6/\} $ \\
	$/v2/$ & $\{/p1/, /p3/, /p5/, /p7/\}$  \\
	$/v3/$ & $\{/p2/, /p4/\}$ \\
	\hline
	\end{tabular}
\label{tab:secondpass}
\end{table}

Remember, as we have two versions of Table~\ref{tab:firstpass} - one with mixed vowel and consonant phonemes and a second with divided vowels and consonant phonemes - the same still applies to our relaxed-confused visemes sets. This means we end up with four types of speaker-dependent phoneme-to-viseme maps, described in Table~\ref{fig:sdtypes}. For our strictly-confused P2V maps in Table~\ref{tab:tcvisemes_split}, these become the relaxed P2V maps in Table~\ref{tab:lcvisemes_split}. 

 \begin{table}[h]
 \centering
 \begin{tabular}{|l|l|}
 \hline
 Mixed vowels and consonants & Split vowels and consonants \\
 \multicolumn{1}{|c|}{$+$} & \multicolumn{1}{c|}{$+$}  \\
 Strict-confusion of phonemes & Strict-confusion of phonemes \\
 \hline 
 Mixed vowels and consonants & Split vowels and consonants \\
 \multicolumn{1}{|c|}{$+$} & \multicolumn{1}{c|}{$+$}  \\
 Relaxed-confusion of phonemes & Relaxed-confusion of phonemes\\
 \hline
 \end{tabular}
 \caption{The four variations on speaker-dependent phoneme-to-viseme maps derived from phoneme confusion in phoneme classification.}
 \label{fig:sdtypes}
 \end{table}
  
\begin{table}[!ht] 
\centering
\caption{Relaxed-confused phoneme speaker-dependent visemes. The score in brackets is the ratio of visemes to phonemes.} 
\begin{tabular} {|l|l|} 
\hline 
Classification & P2V mapping - permitting mixing of vowels and consonants \\ 
\hline \hline 
Speaker1 	& {\footnotesize \{/b/ /e/ /ei/ /p/ /w/ /y/ /k/\} \{/\textturnv/ /ai/ /f/ /i/ /m/ /n/ /\textschwa\textupsilon/\}}\\ 
(CF:0.28)	& {\footnotesize \{/d\textipa{Z}/ /z/\} \{/\textscripta/ /u/\}  \{/d/ /s/ /t/\} \{/t\textipa{S}/ /l/\}  \{/\textschwa/ /v/\}\{/\textschwa/ /v/\}  	}\\ 
Speaker2 	& {\footnotesize \{/\textscripta/ /\textschwa/ /ai/ /ei/ /i/ /s/ /t\textipa{S}/\} \{/e/ /t/ /v/ /w/ /y/\} \{/l/ /m/ /n/\} }\\ 
(CF: 0.32)	& {\footnotesize  \{/\textturnv/ /f/\} \{/z/\}  \{/b/ /d/ /p/\} \{/\textschwa\textupsilon/ /u/\} \{/d\textipa{Z}/ /k/\} }\\ 
Speaker3	& {\footnotesize \{/\textturnv/ /ai/ /ei/ /f/ /i/ /n/\} \{/\textschwa/ /e/ /y/ /t\textipa{S}/\} \{/b/ /s/ /v/\} \{/l/ /m/ /u/\} 	}\\ 
(CF: 0.40)	& {\footnotesize  \{/d\textipa{Z}/\} \{/\textschwa\textupsilon/\} \{/z/\} \{/d/ /p/ /t/\} \{/k/ /w/\} \{/\textscripta/\} }\\ 
Speaker4 	& {\footnotesize \{/\textturnv/ /ai/ /t\textipa{S}/ /i/ /ei/ \} \{/\textscripta/ /m/ /u/ /n/\} \{/\textschwa/ /e/ /p/ /v/ /y/\} }\\ 
(CF: 0.32)	& {\footnotesize \{/d\textipa{Z}/ /t/\}  \{/k/ /l/ /w/\} \{/\textschwa\textupsilon/\} \{/d/ /f/ /s/\} \{/b/\}  }\\ 
%	Multi-Speaker & \{/\textscripta/ /\textturnv/ /ai/ /u/ /m/ /n/ /\textschwa\textupsilon/ /ei/ /i/ /z/\} \{/b/ /d/ /e/ /p/ /y/\} 	\\ 
%	(CF: 0.24)	& \{/f/ /l/ /s/\} \{/k/ /v/ /w/\} \{/d\textipa{Z}/ /t/\} \{/\textschwa/ /t\textipa{S}/\}\\ 
\hline 
%	\end{tabular} 
%	\caption{Loosely confused phoneme multi-speaker and speaker-dependent visemes } 
%	\label{tab:lcvisemes} 
%\end{table*} 
% 
%\begin{table*}[!ht] 
%	\begin{tabular} {|l|l|} 
\hline 
Classification & P2V mapping - restricting mixing of vowels and consonants\\ 
\hline \hline 
Speaker1 	& {\footnotesize \{/\textturnv/ /i/ /\textschwa\textupsilon/ /u/\} \{/\textscripta/ /ai/\} \{/\textschwa/ /e/ /ei/\} \{/b/ /w/ /y/\} \{/d/ /f/ /s/ /t/\} }\\ 
(CF:0.47)	& {\footnotesize \{/k/\} \{/z/\} \{/m/\} \{/l/\}  \{/t\textipa{S}/\}   \{/d\textipa{Z}/ /k/ /v/ /z/\} }\\ 
Speaker2 	& {\footnotesize \{/\textscripta/ /\textturnv/ /\textschwa/ /ai/ /ei/ /i/ /\textschwa\textupsilon/ /u/\} \{/k/ /t/ /v/ /w/\} \{/t\textipa{S}/ /l/ /m/ /n/\}  }\\ 
(CF: 0.29)	& {\footnotesize \{/f/ /s/\} \{/d\textipa{Z}/ /p/ /y/\}  \{/b/ /d/\} \{/z/\}	}\\ 
Speaker3 	& {\footnotesize \{/\textturnv/ /ai/ /i/ /ei/\} \{/\textschwa/ /e/\} \{/b/ /s/ /v/\} \{/d/ /p/ /t/\} \{/l/ /m/\} }\\ 
(CF: 0.56)	& {\footnotesize \{/y/\} \{/d\textipa{Z}/\} \{/\textschwa\textupsilon/\} \{/z/\} \{/u/\} \{/\textschwa/ /e/\} \{/k/ /w/\} \{/f/ /n/\}  \{/\textscripta/\}   \{/t\textipa{S}/\}  }	\\ 
Speaker4 	& {\footnotesize \{/\textturnv/ /ai/ /i/ /ei/\}  \{/t\textipa{S}/ /k/ /l/ /w/\} \{/d/ /f/ /s/ /v/\} \{/m/ /n/\}  }\\ 
(CF: 0.50)	& {\footnotesize \{/f/\} \{/\textscripta/\} 	\{/d\textipa{Z}/ /t/\} \{/\textschwa\textupsilon/\} \{/u/\} \{/y/\} \{/b/\} }\\ 
%	Multi-Speaker & \{/\textscripta/ /\textturnv/ /ai/ /u/ /\textschwa\textupsilon/ /ei/ /i/\} \{/\textschwa/ /e/\} \{/b/ /p/ /w/ /y/\} 	\\ 
%	(CF: 0.24)	& \{/t\textipa{S}/ /f/ /k/ /l/ /m/ /n/\} \{/d/ /d\textipa{Z}/ /s/ /t/ /v/\} \{/z/\} \{/w/\} \{/k\}\\ 
\hline 
\end{tabular} 
\label{tab:lcvisemes_split} 
\end{table} 
 
Now, and this is why these visemes are defined as relaxed, any remaining phonemes which have confusions, but are so far not assigned to a viseme, the phoneme-pair confusions are used to map the remaining phonemes to an appropriate viseme, even though it does not confuse with all phonemes already in it. Any remaining phonemes which are not assigned to a viseme are grouped into a new garbage /gar/ viseme. This approach ensures any phonemes which have been confused with any other are grouped into a viseme.

\section{Bear speaker-dependent visemes} 

\begin{figure}[h] 
\centering 
\includegraphics[width=0.85\textwidth]{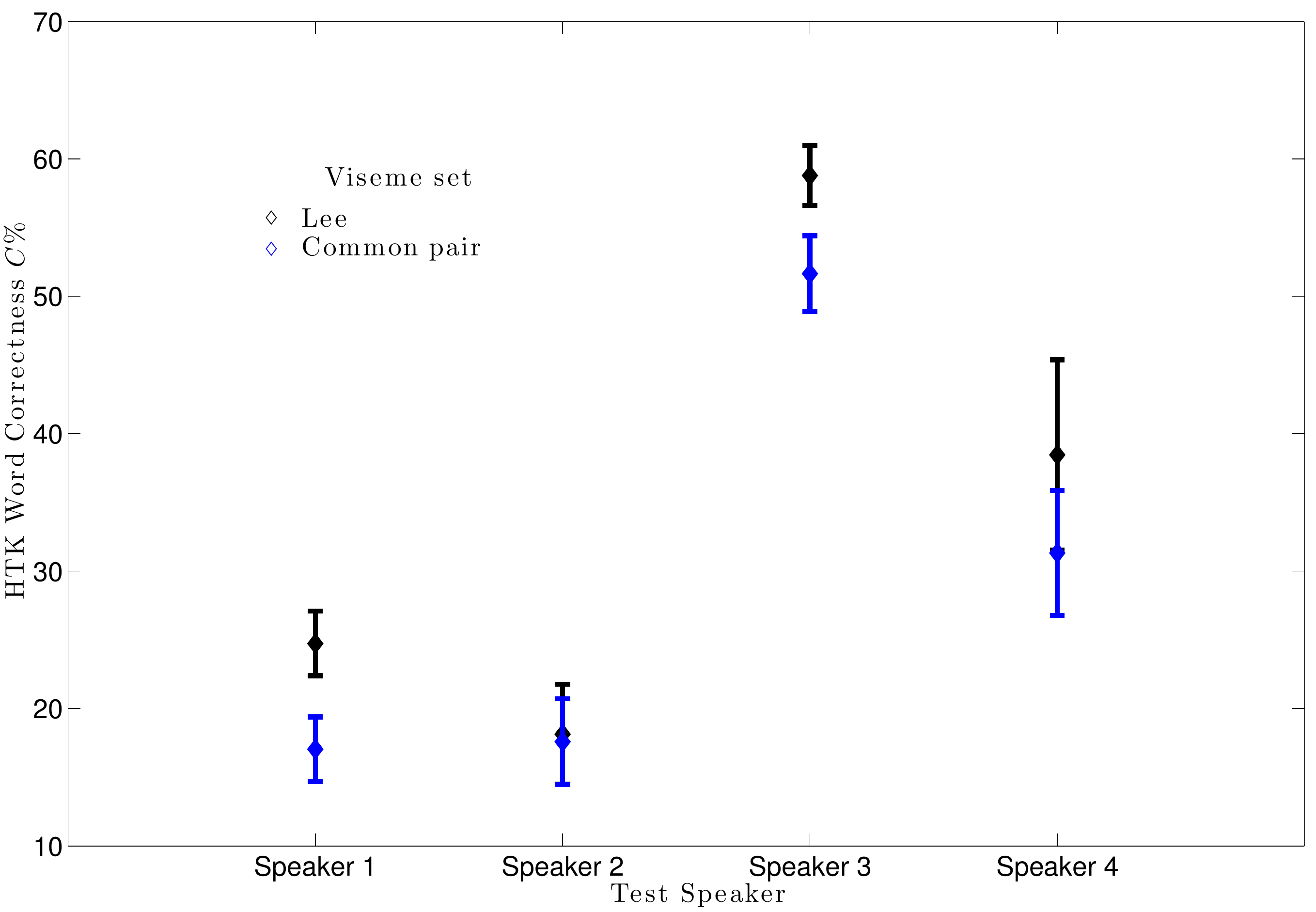} 
\caption{Word classification correctness $C \pm1\frac{\sigma}{\sqrt{7}}$, using the common phoneme-pairs phoneme-to-viseme map. Lees benchmark is in black.} 
\label{fig:evolvedresults} 
\end{figure}

 Figure~\ref{fig:evolvedresults} shows word correctness of the common phoneme-pair visemes against Lee's benchmark. It is no surprise the common-pair visemes are all worse than Lee, as Lee gave the maximum performance of the original P2V mappings used to deduce the new map. However, the overlap in error bars shows that for two speakers this is not a significant reduction. Unfortunately, no particular viseme, or group of visemes, particularly contribute to the set correctness.

\begin{figure}[h] 
\centering 
\includegraphics[width=0.85\textwidth]{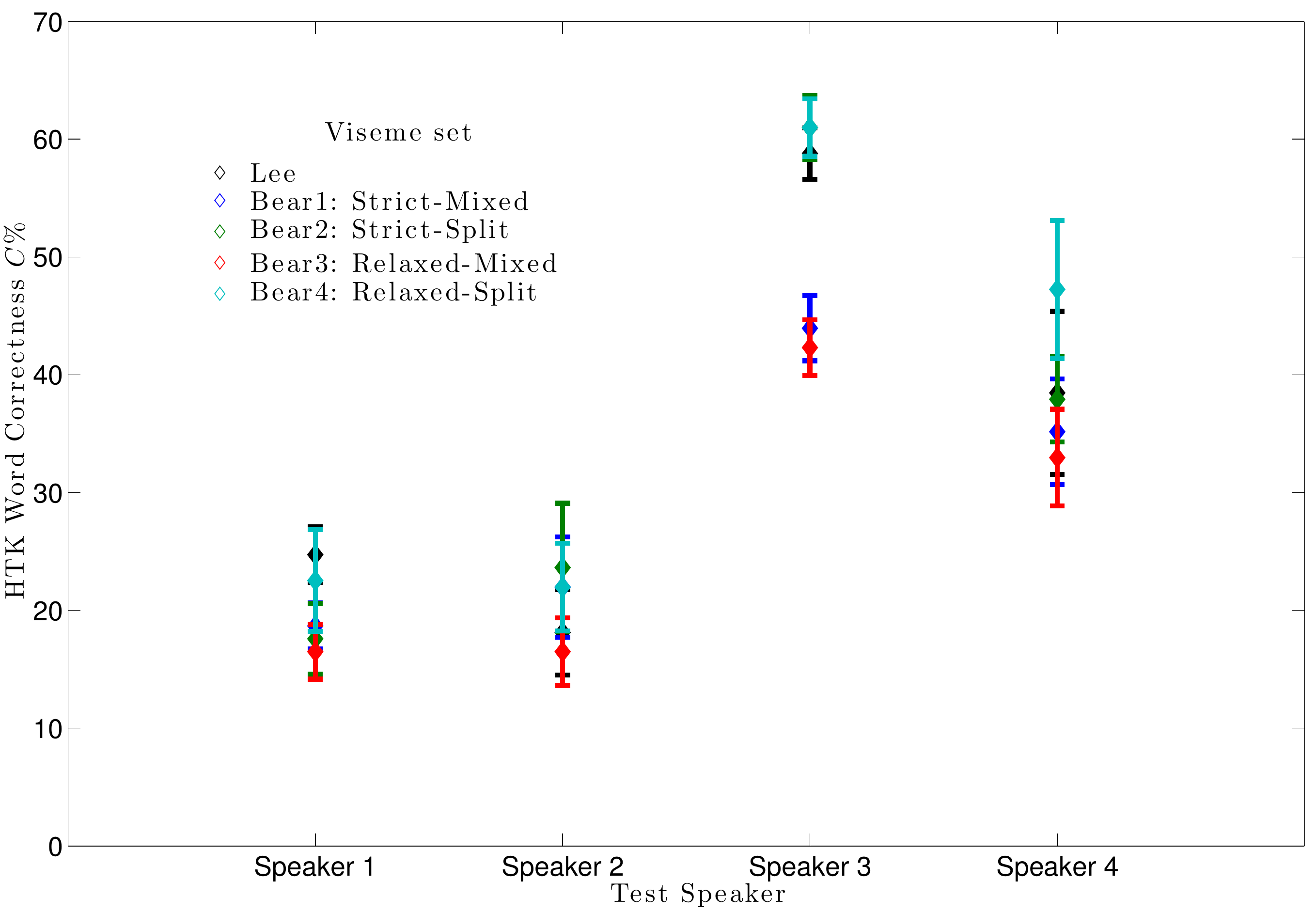} 
\caption{Word classification correctness $C \pm1\frac{\sigma}{\sqrt{7}}$, using all four new methods of deriving speaker dependent visemes. Lees benchmark is in black.} 
\label{fig:q1results} 
\end{figure} 
 
 \begin{figure}[h] 
\centering 
\includegraphics[width=0.85\textwidth]{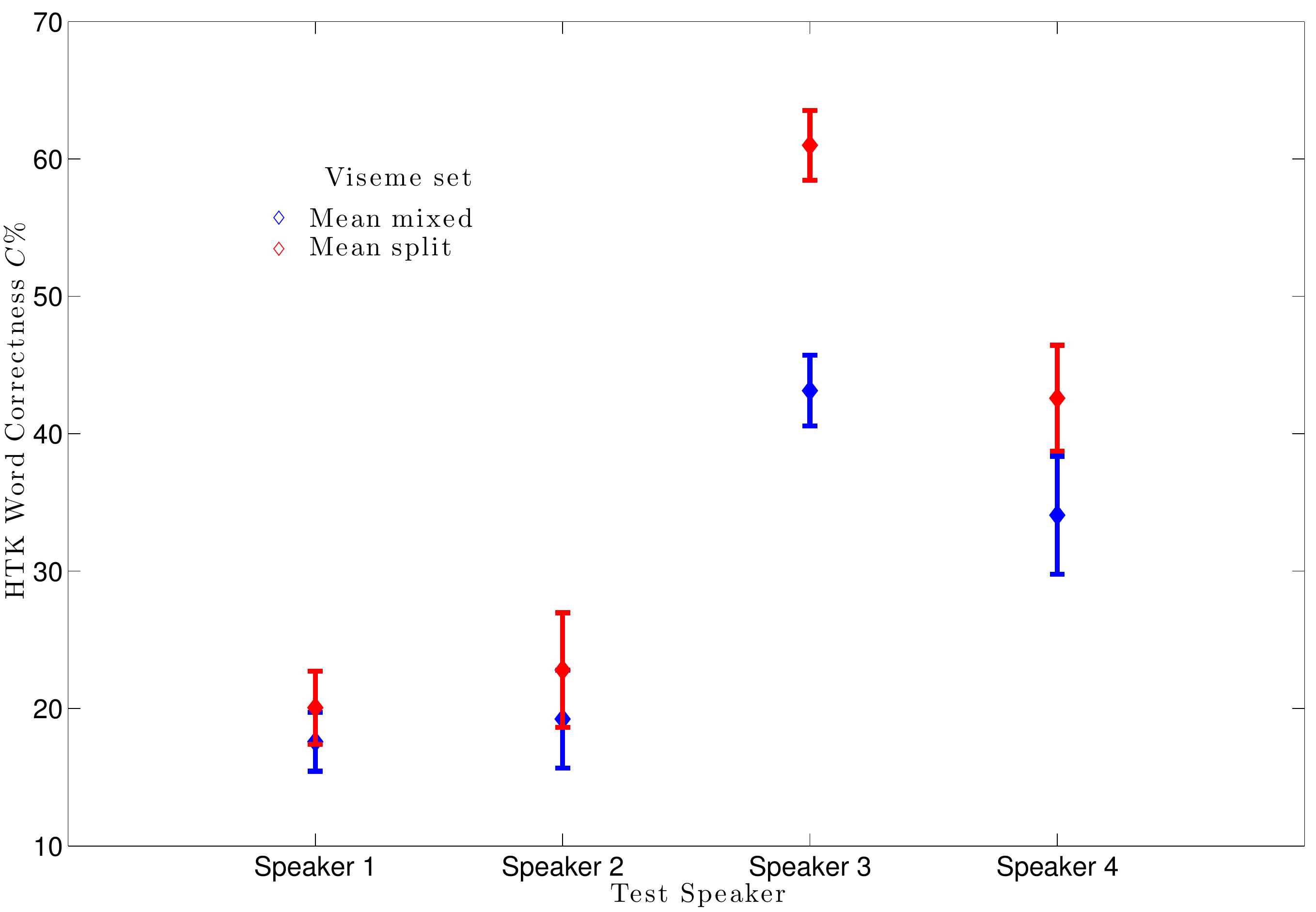} 
\caption{A comparison of the split vowel and consonant phoneme visemes and the mixed vowel and consonant phoneme visemes with AVLetters2 speakers.} 
\label{fig:splitVmixed} 
\end{figure} 

\clearpage %layout hack
\begin{figure}[h] 
\centering 
\includegraphics[width=0.85\textwidth]{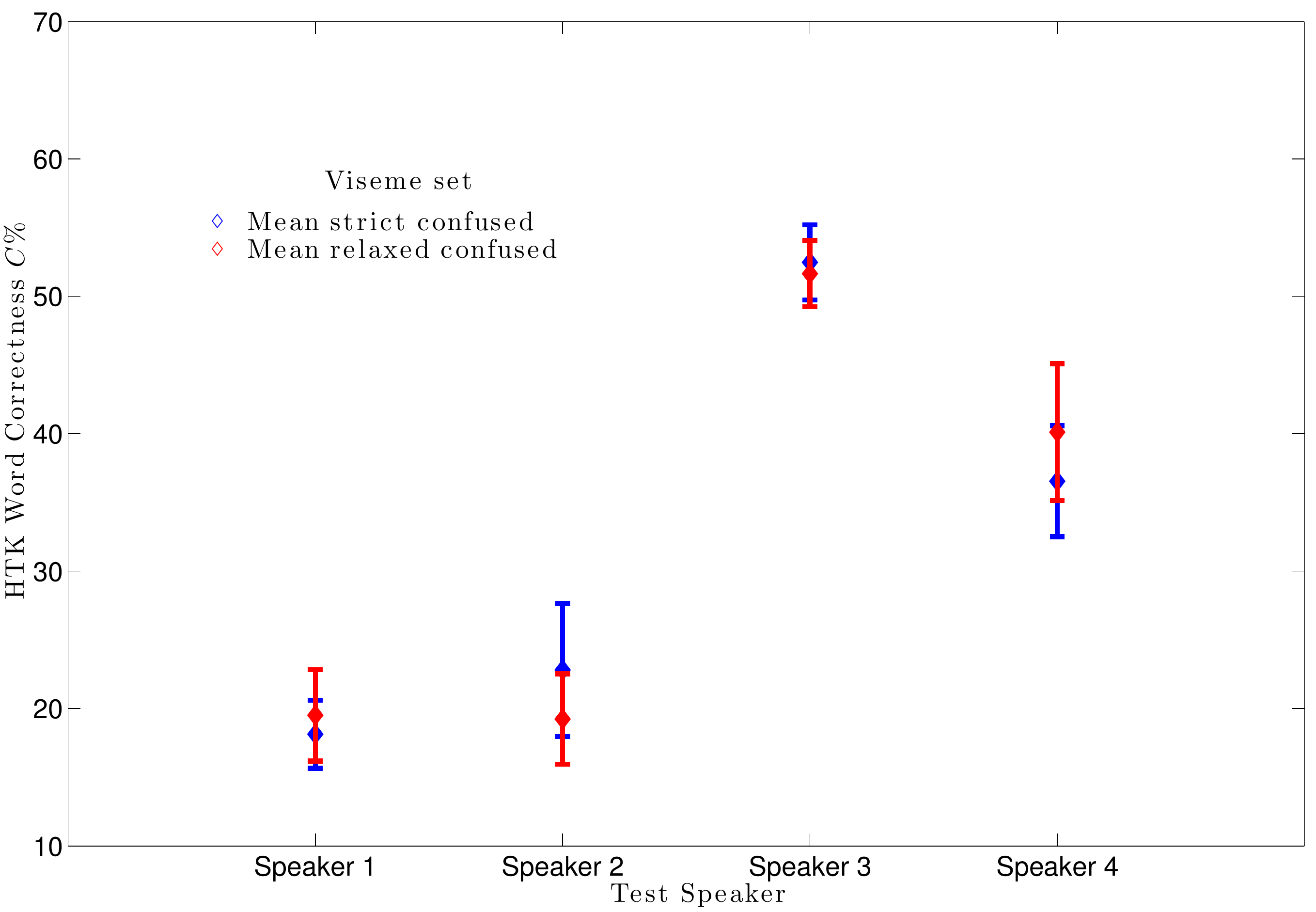} 
\caption{A comparison of the strict mutually confusable phoneme viseme classes and the relaxed confused phoneme visemes with AVLetters2 speakers.} 
\label{fig:types_questioned} 
\end{figure} 
%\clearpage

In Figure~\ref{fig:q1results} all four speaker-dependent maps tested on each speaker are plotted on the $x$-axis to compare the difference in word classification (shown on the $y$-axis). The benchmark from the comparison study, Lee, is in black. For Speaker 1 and Speaker 3, no new viseme map significantly improves upon Lee's performance although we do see improvements for both Speaker 2 and Speaker 4. The strictly-confused and split viseme map improves upon Lee's previous best word classification. 

Figure~\ref{fig:splitVmixed} compares the mixed consonant and vowel maps against split consonant and vowel maps, also measured in word correctness, $C$, on the $y$-axis. The split P2V maps are always better than mixed for all speakers. Figure~\ref{fig:types_questioned} shows the comparison of strictly-confused and loosely confused viseme classes. The strict confusions are better for two out of four speakers. These are speakers with the highest ratio of phonemes to visemes (Tables~\ref{tab:lcvisemes_split} and~\ref{tab:tcvisemes_split}).

\begin{sidewaysfigure}[h] 
\setlength{\tabcolsep}{1pt} 
\caption{How Correctness varies with quantity of visemes in each set. All four variants on a speaker-dependent data-driven approach to finding visemes plotted against the count of visemes within each set.} 
\begin{tabular}{l | c  c}
& Mixed Vowels and Consonant phonemes & Split Vowels and Consonant Phonemes \\ 
\hline 
 &  \multirow{15}{*}{\includegraphics[width=0.4\textwidth]{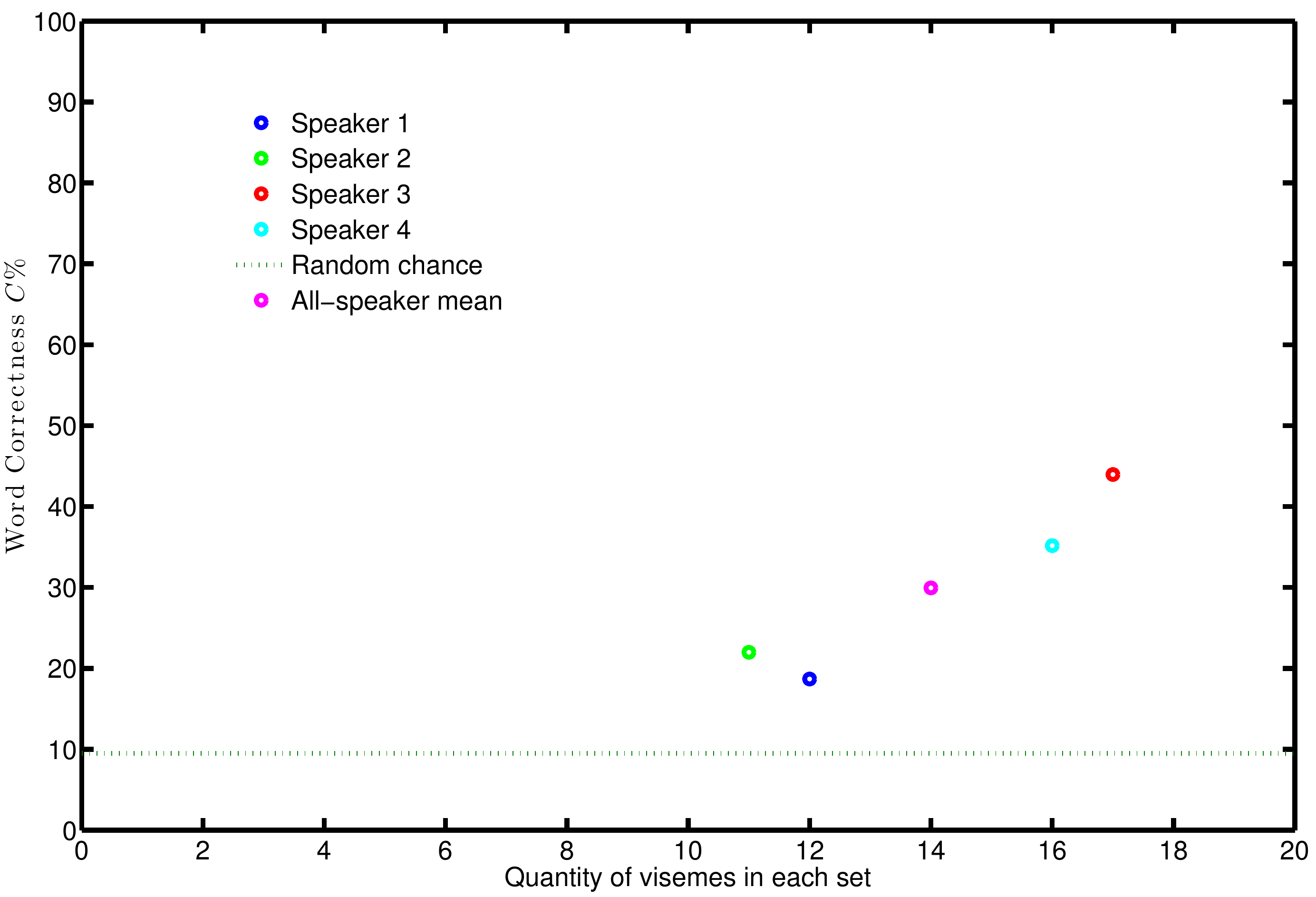}} & \multirow{15}{*}{\includegraphics[width=0.4\textwidth]{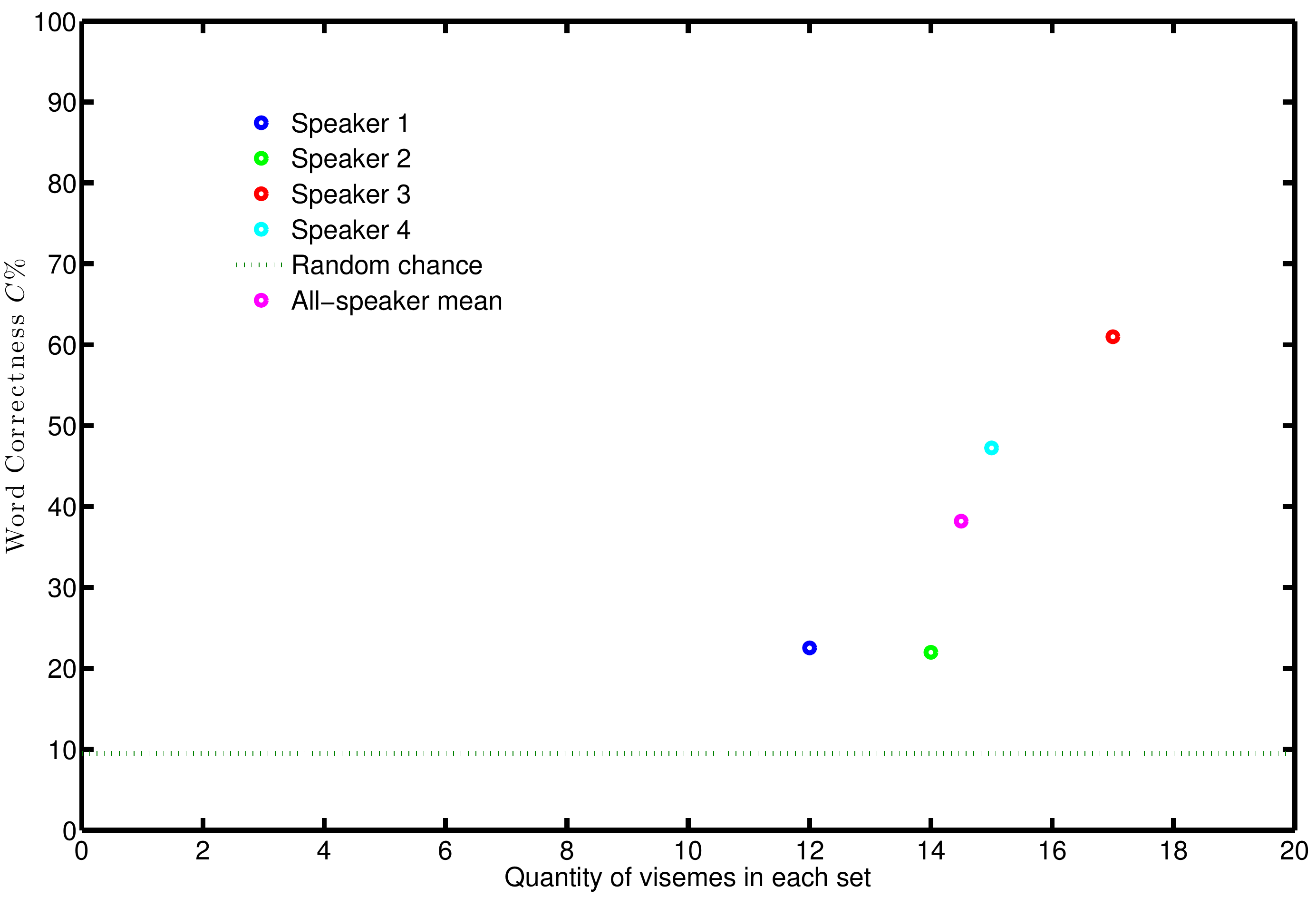}}\\
& & \\ & & \\ & & \\ & & \\ & & \\ & & \\ Strictly confused &   &   \\ 
& & \\ & & \\ & & \\ & & \\ & & \\ & & \\
 &  \multirow{15}{*}{\includegraphics[width=0.4\textwidth]{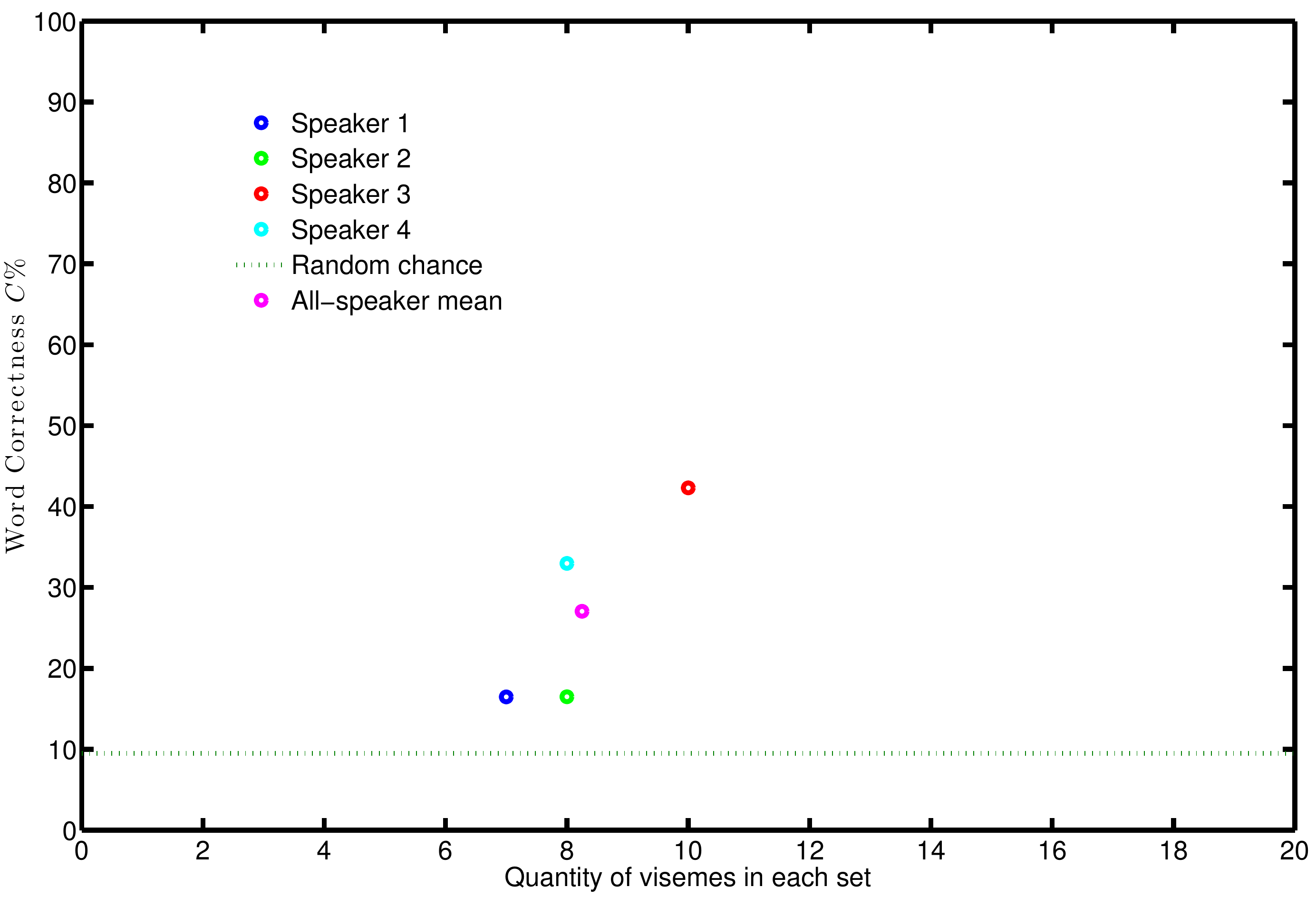}} & \multirow{15}{*}{\includegraphics[width=0.4\textwidth]{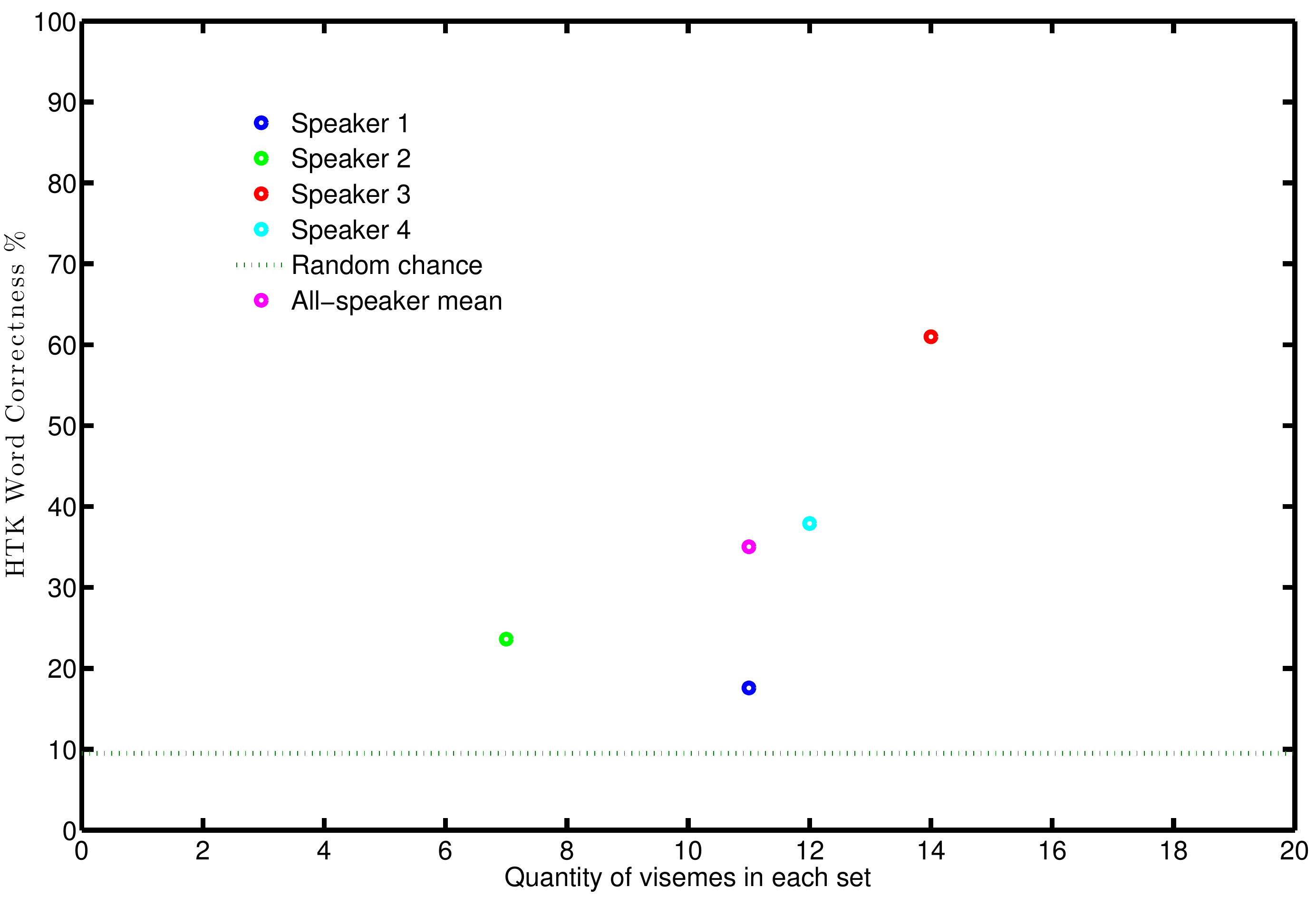}}\\
& & \\ & & \\ & & \\ & & \\ & & \\ & & \\ Relaxed confused & &  \\ 
& & \\ & & \\ & & \\ & & \\ & & \\ & & \\
\end{tabular} 
\label{figs:bearvisemeclassvolume} 
\end{sidewaysfigure}

In Figure~\ref{figs:bearvisemeclassvolume}, all four variants of our new P2V maps are plotted for each speaker and an all-speaker mean against the number of visemes in each set. Splitting vowel and consonant phonemes gives a greater number of classifiers, which reduces the number of training samples per class, but results in higher correctness for all speakers. This shows that having the \textit{right} training samples is more important than having simply `more data'. Whilst showing a smaller effect, the two graphs on the left hand side of Figure~\ref{figs:bearvisemeclassvolume} shows the relaxing of confusable phonemes has a negative influence, even though this reduces the number of visemes and increases training samples per class, they are not good training samples to include for the class.

\begin{figure}[h] 
\centering 
\includegraphics[width=0.85\textwidth]{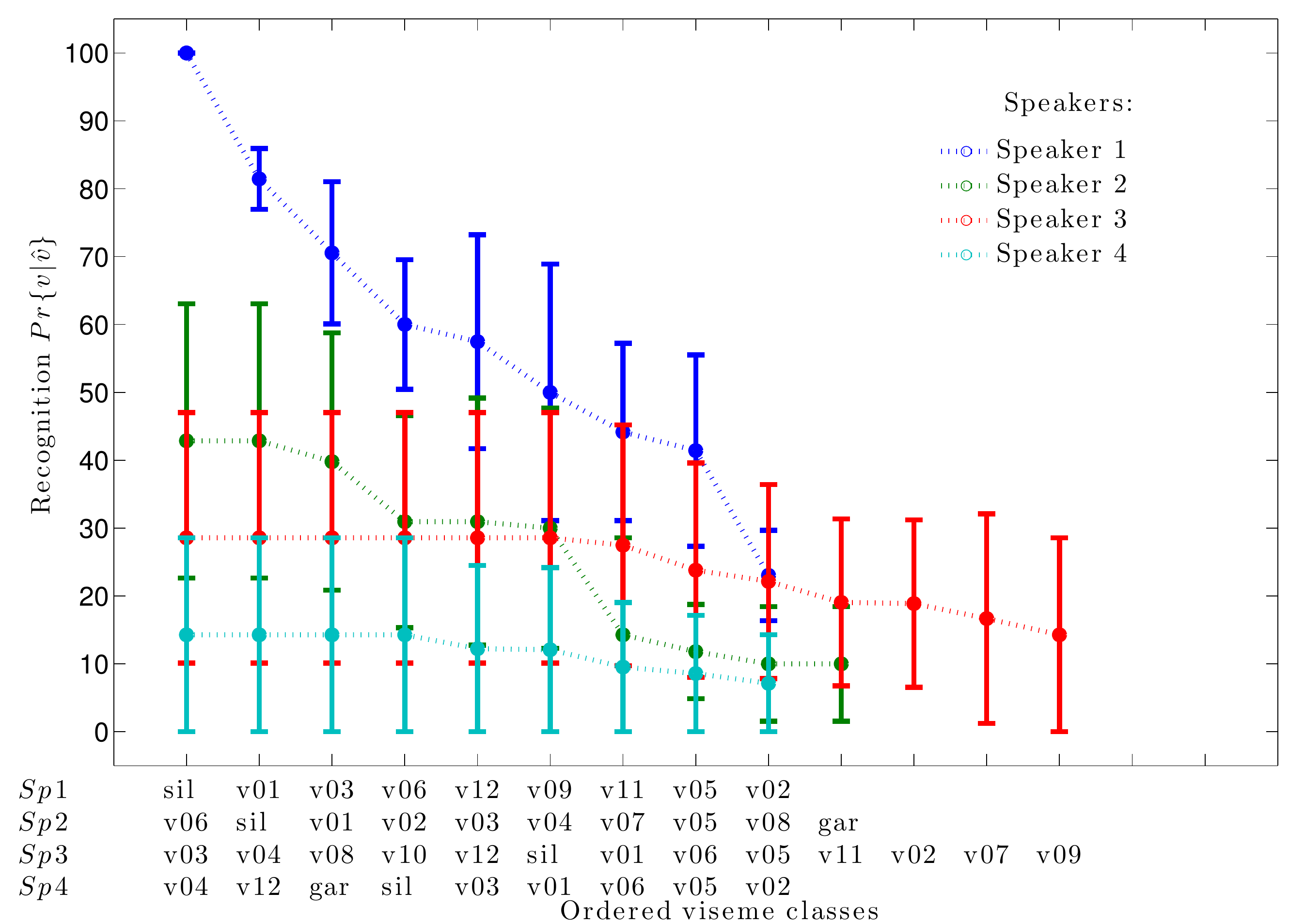} \\ 
\caption{Individual viseme classification, Pr$\{v|\hat{v}\}$ with the relaxed, mixed vowels and consonant Bear visemes.} 
\label{fig:contrib_rm} 
%\end{figure}
%
%\begin{figure}[h] 
%\centering 
\includegraphics[width=0.85\textwidth]{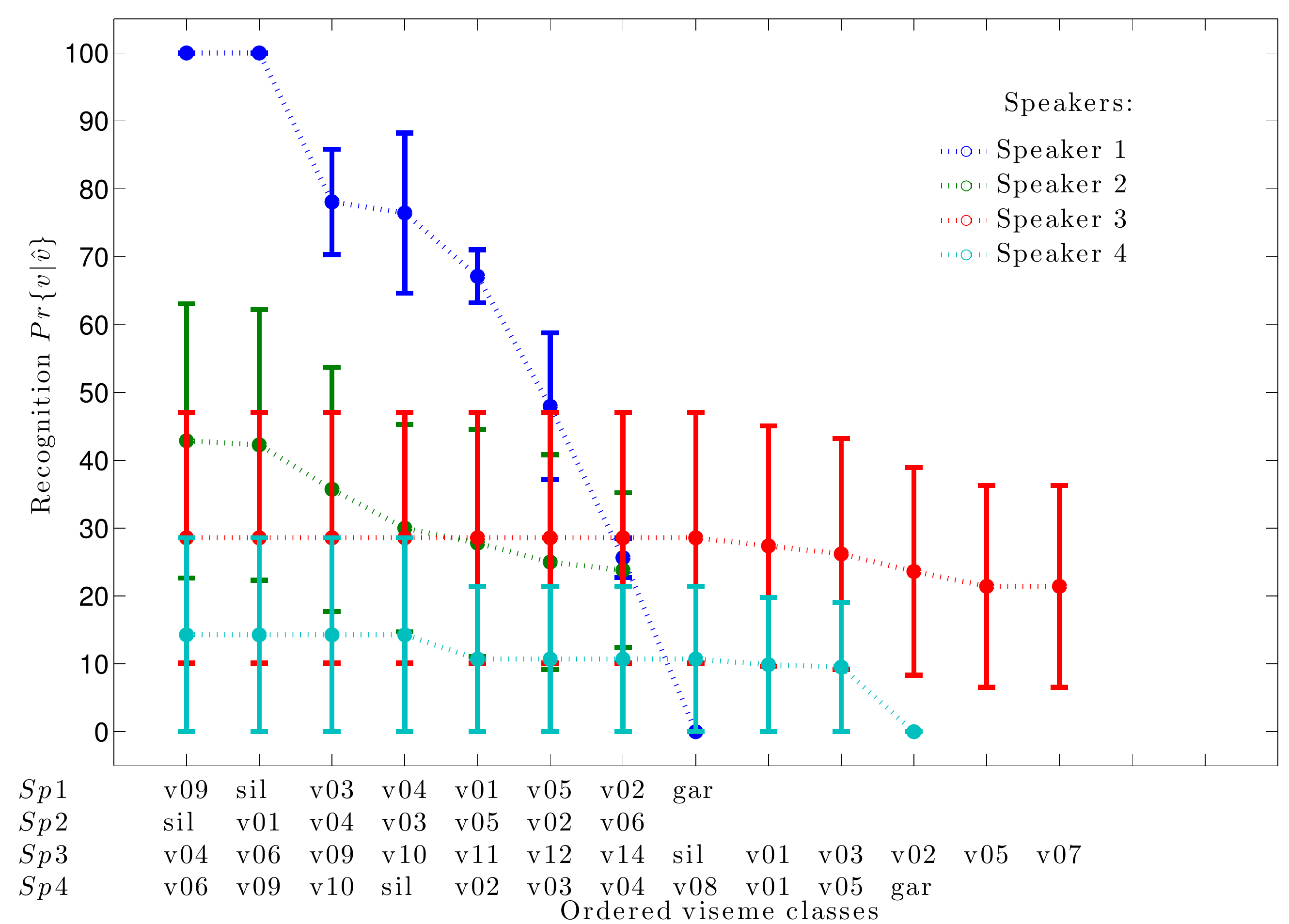} \\ 
\caption{Individual viseme classification, Pr$\{v|\hat{v}\}$ with the relaxed, split vowels and consonant Bear visemes.} 
\label{fig:contrib_rs} 
\end{figure}

\begin{figure}[h] 
\centering 
\includegraphics[width=0.85\textwidth]{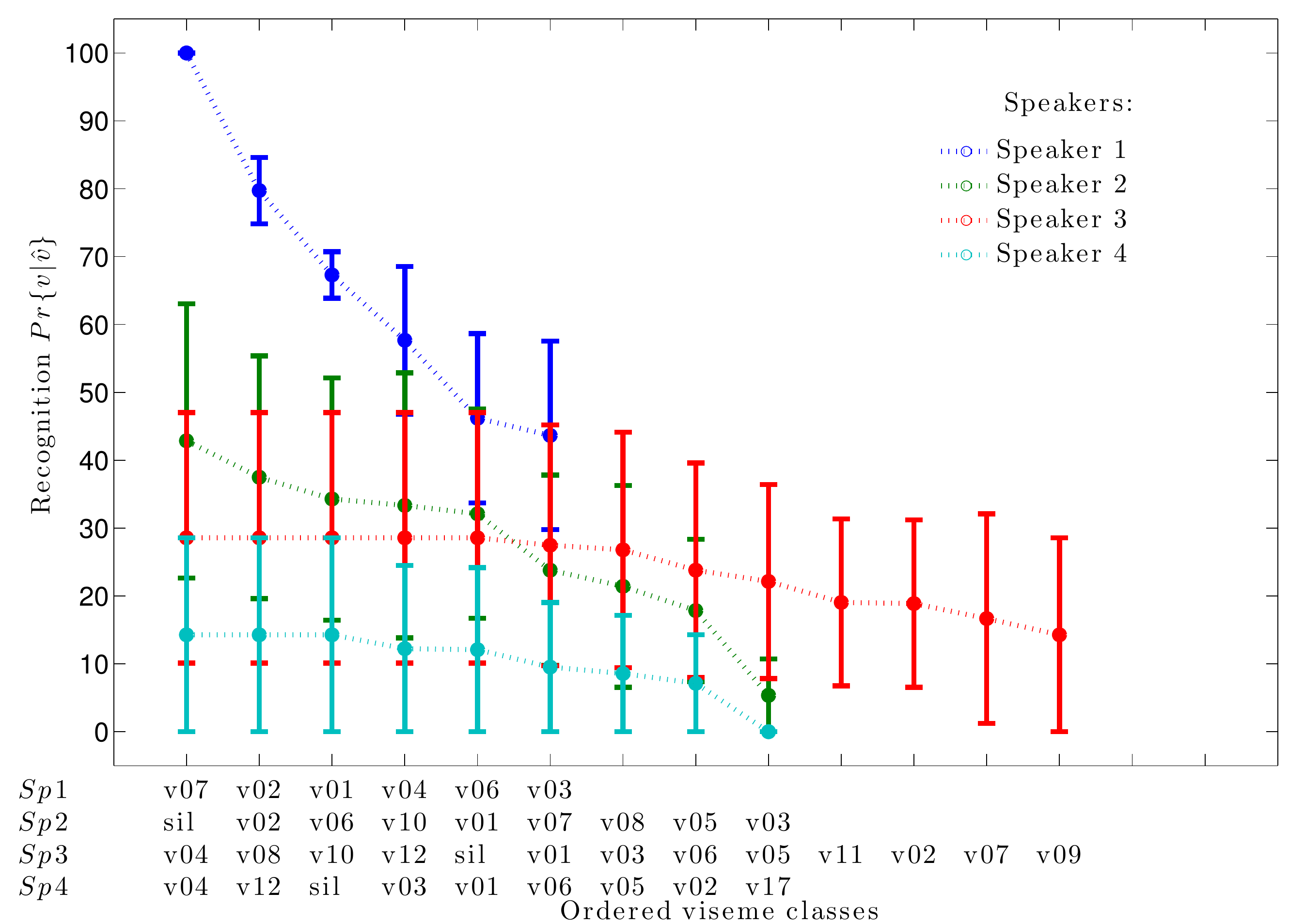} \\ 
\caption{Individual viseme classification, Pr$\{v|\hat{v}\}$ with the strictly confused, mixed vowels and consonant Bear visemes.} 
\label{fig:contrib_sm} 
%\end{figure}
%
%\begin{figure}[h] 
%\centering 
\includegraphics[width=0.85\textwidth]{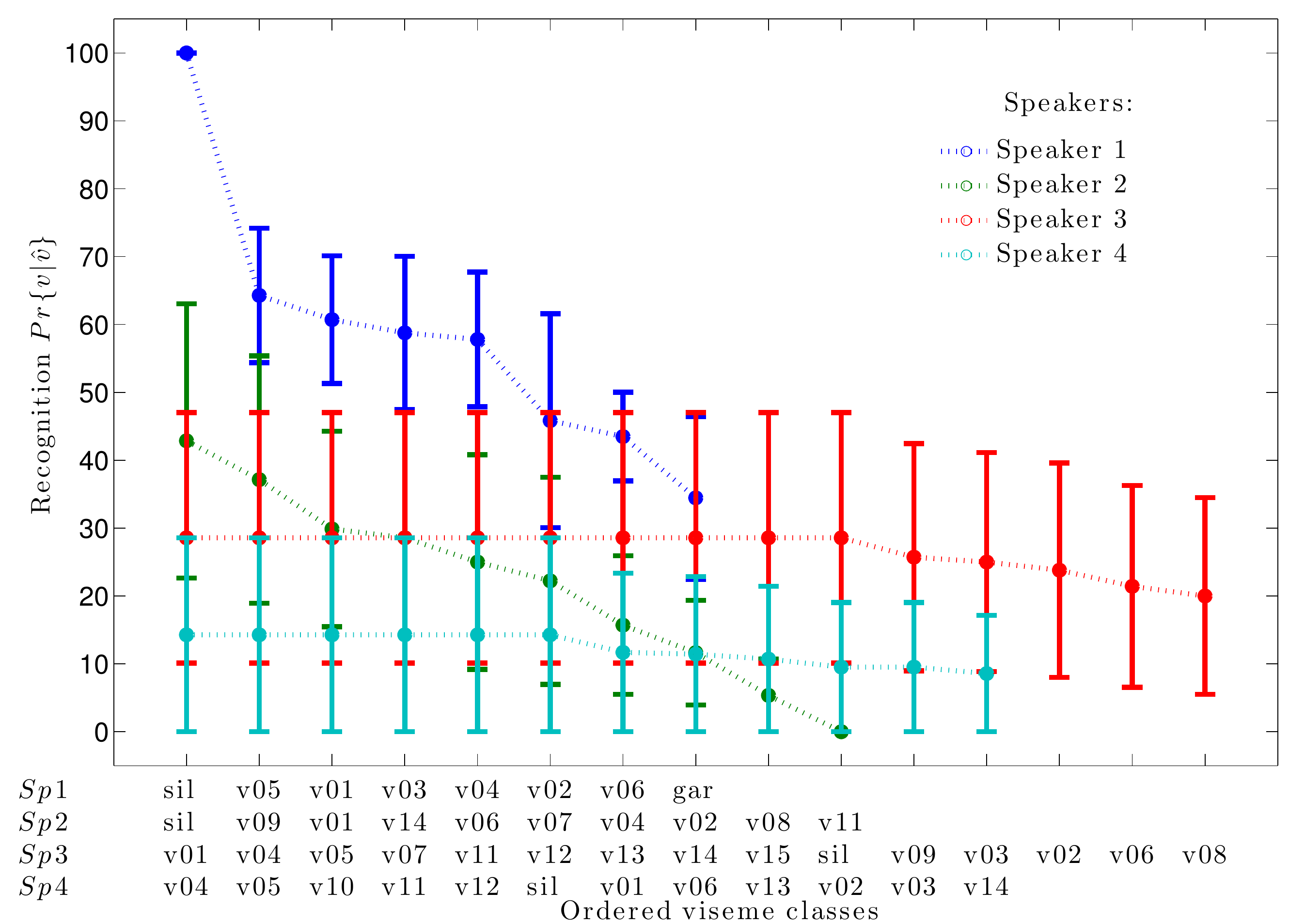} \\ 
\caption{Individual viseme classification, Pr$\{v|\hat{v}\}$ with the strictly confused, split vowels and consonant Bear visemes.} 
\label{fig:contrib_ss} 
\end{figure}

\clearpage
 In Figures~\ref{fig:contrib_rm},~\ref{fig:contrib_rs},~\ref{fig:contrib_sm}, and~\ref{fig:contrib_ss}, the contribution of each viseme has been listed in descending order along the $x-$axis for each speaker in AVL2. The contribution of each viseme is measured as the inverse probability of each class, Pr$\{v|\hat{v}\}$. These values have been calculated from the \texttt{HResults} confusion matrices. There is no significant step when a viseme contribution is no longer needed, that is in speaker-dependent visemes we need all class labels within a set. This analysis of visemes within a set is also used in \cite{bear2014some}, which proposes a threshold subject to the information in the features. Using combined shape and appearance features here removes the threshold as these figures show irrespective of which method of phoneme-clustering is used for devising visemes, the greater the number of visemes in a set, the higher the overall classification. More important to see is the overall classification $C$ is higher when there is less range between individual viseme Pr$\{v|\hat{v}\}$ values within a set of visemes. The difference values between the highest and least contributing visemes for each method and speaker are listed in Table~\ref{tab:differencesbetweenvisemes}.

\begin{table}[h]
\centering
\caption{Viseme variation in Pr$\{v|\hat{v}\}$ showing the best and worst classifiers within each set of visemes for each derivation method per speaker.}
\begin{tabular}{|l|r|r|r|r|}
\hline
Method & Speaker 1 & Speaker 2 & Speaker 3 & Speaker 4 \\
\hline \hline
Relaxed: mixed 	& 76.97 	& 32.86		& 14.29 	& 7.14 \\
Relaxed: split 	& 100.00	& 19.05	& 7.14	& 14.29 \\
Strict: mixed	& 56.35	& 37.50	& 14.26	& 14.29 \\
Strict: split		& 65.54		& 42.86	& 8.57	& 5.71  \\
 \hline
 \end{tabular}
 \label{tab:differencesbetweenvisemes}
 \end{table}

\clearpage %interesting layout for layout purposes 

\section{Improving lip-reading with speaker-dependent phoneme-to-viseme maps} 
\begin{figure}[h]
\centering
\includegraphics[width=0.95\textwidth]{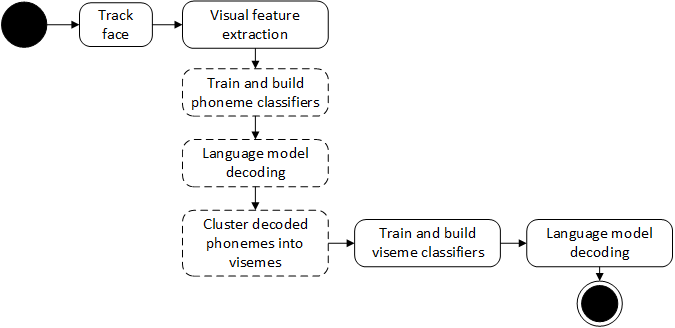}
\caption{First augmentation to the conventional lip-reading system to include speaker-dependent visemes.}
\label{fig:augment1}
\end{figure}

This chapter has described a comprehensive study of previously suggested P2V maps and shown Lee's \cite{lee2002audio} is the best of the previously published P2V maps. The new data-driven approach respects speaker individuality in speech and uses this to demonstrate our second data-driven method tested, a strictly-confused viseme derivation with split vowel and consonant phonemes, can improve word classification. We call these speaker-dependent visemes `Bear visemes' after the author's surname and show how these fit into the conventional lip-reading system in Figure~\ref{fig:augment1}, our new steps are highlighted with dash-edged boxes.

For phoneme confusion driven visemes, it is possible AVL2 contains insufficient samples to fairly identify confusion. So whilst improving performance, the classifiers still need more data for training. The reduction in word correctness by the data-driven confused mixed visemes is attributed to the mixing of vowels and consonants as this work shows when keeping these separate an improvement is possible. 
 
The ratio of phonemes to visemes is useful, but secondary to confusions between phonemes, and does not help to discriminate phonemes within visemes for improved word classification. To discriminate between words which are visually similar we still need to be able to reverse any P2V mapping. 
 
This work highlights bad training samples are worse than less training samples and the boundary between good and bad samples is blurred. We have designed and implemented a new method of producing speaker-dependent visemes, in doing so showing speaker identity is important for good machine lip-reading classification. As speaker dependence is prevalent in machine lip-reading systems, we need to cast our eyes towards how difficult a task speaker-independent classification is. This is what our next chapter investigates. 
 % ISVC: Viseme Set comparison & split viseme study, speaker-dependent visemes,  
%!TEX root = main.tex 
\chapter[Speaker-independence in phoneme-to-viseme maps]{Speaker-independence in phoneme-to-viseme maps} 
\label{chap:seven} 
 
More than in audio speech, in machine lip-reading speaker identity is important for accurate classification \cite{cox2008challenge}. We know a major difficulty in visual speech is the labelling of classifier units so we need to address the questions; to what extent such maps are independent of the speaker? And if so, how might speaker independent P2V maps be examined? Alongside of this, it would be useful to understand the interactions between the model training data and the classes. Therefore in this chapter we will use both the the AVL2 dataset \cite{cox2008challenge} and the RMAV dataset to train and test classifiers based upon a series of P2V mappings. 
 
 \section{Speaker independence} 
At the current time, good machine lip-reading performances are achieved with speaker dependent classification models, this means the test speaker must be included within the classifier training data. Speaker independent machine lip-reading is less successful \cite{cox2008challenge}.  Only a few large scale investigations have shown speaker independence to be viable. Neti \textit{et al.} in \cite{neti2000audio} state that they created multi-speaker classifiers as contingency should speaker independent models fail to generalise well to unseen speakers. After preliminary experiments these multi-speaker classifiers were considered not needed. However, this is achieved with state-of-the-art modelling, a permitted increase in word error rate and with a lot of speakers (IBM's via voice has 290 speakers \cite{ibmviavoicebookref}) (a currently unavailable dataset) which implies that with enough data and speakers that the speaker independence obstacle is surmountable by achieving generalisation on a large scale. In the majority of papers referenced in this thesis for example, speaker-dependent experiments are still used for greater results as speaker independence is rare and difficult to achieve. 

On the continuous speech datasets, it is interesting to note that most still use speaker-dependent tests \cite{lan2012insights, Hazen1027972, Wong20111503, cappelletta2012phoneme}. We note that some are single speaker-dependent, others multi-speaker dependent, the crux of the point is that test speaker samples are included in the training data. In contrast only AVICAR \cite{kleinschmidt2008continuous} and IBM's LVCSR \cite{neti2000audio} achieve speaker-independent success. The former is a specific AV dataset for in car speech, and the latter is not available \cite{cappelletta2012phoneme} so we suffice with the best datasets we have available to us.  

Thus we understand speaker independence in visual speech to be the ability to classify a speaker who is not involved in the classifier training. This is a difficult, and as yet, unsolved problem. From this we are confident that, in visual speech, the identification of the person speaking is important. One could wonder if, with a large enough dataset with a significant number of speakers, then it could be sufficient to train classifiers which are generalised to cover a whole population including independent speakers. But we still struggle without a dataset of the size needed to test this theory, particularly as we do not know how much is `enough' data or speakers.

An example of a study into speaker independence in machine lip-reading is \cite{cox2008challenge}, here the authors use AVL2 and compare single speaker, multi-speaker and speaker independent classification using two types of classifiers (HMMs \& Sieves \cite{bangham1996nonlinear}). However, this investigation uses word labels for classifiers and we are interested to know if the results could be improved using either phonemes or speaker-dependent visemes. 

\section{Method overview} 
 
We use the phoneme clustering approach described in Chapter~\ref{chap:maps} (or \cite{bear2014phoneme}) to produce a series of speaker-dependent P2V maps. This series of maps is made up of the following:  
\begin{enumerate} 
\item a speaker-dependent P2V map for each speaker; 
\item a multi-speaker P2V map using {\em all} speakers' phoneme confusions; 
\item a speaker-independent P2V map for each speaker using confusions of all {\em other} speakers in the data. 
\end{enumerate} 

So we have nine phoneme-to-viseme maps for AVL2 (four speaker maps for map types one and three, and one multi-speaker map) and 25 for RMAV (12 speaker maps for map types one and three, and one multi-speaker map). AVL2 P2V maps are constructed using separate training and test data over seven fold cross-validation \cite{efron1983leisurely}. RMAV maps from ten fold cross-validation. The variation in folds is due to the volume of data in each dataset. 

With the HTK toolkit \cite{htk34} HMM classifiers are built with the viseme classes in each P2V map. HMMs are flat-started with \texttt{HCompV}, re-estimated 11 times over (\texttt{HERest}) with forced alignment between seventh and eighth re-estimates. The final steps are classification using \texttt{HVite} and output of results with \texttt{HResults}. The models are three state HMMs each having an associated Gaussian mixture of five components to keep our results comparable to previous work. 

To measure our performance of AVL2 speakers we note the classification network restricts the output to be one of the 26 letters of the alphabet. Therefore, our simplified measure of accuracy is;
% \begin{equation}
%  \dfrac{\#letters correct}{\#letters classified}
% \end{equation}
% 
 \mbox{$  \dfrac{\#letters correct}{\#letters classified} $}.

For RMAV a bigram word network is built with \texttt{HBuild} and \texttt{HLStats}, and classification is measured as Correctness (Equation~\ref{eq:correctness}). The BEEP pronunciation dictionary used throughout these experiments is in British English  \cite{beep} for all speakers.

\section{Experiment design}  
The P2V maps formed in these experiments are designated as: 
\begin{equation} 
M_n(p,q)\quad 
\label{eq2} 
\end{equation} 
This means the P2V map is derived from speaker $n$, but trained using visual speech data from speaker $p$ and tested using visual speech data from speaker $q$. For example, $M_1(2,3)$ would designate the result of testing a P2V map constructed from Speaker 1, using data from Speaker 2 to train the viseme models, and testing on Speaker 3's data. 
 
\subsection{Baseline: Same Speaker-Dependent (SSD) maps} 
\label{sec:expSetup} 
For our experiments we need a baseline for comparison. We select our same speaker-dependent P2V maps as based on previous literature \cite{bear2014phoneme}, these provide the best results. The baseline tests involved are: $M_1(1,1)$, $M_2(2,2)$, $M_3(3,3)$ and $M_4(4,4)$ (for the four speakers in AVL2), additional tests for RMAV are: $M_5(5,5)$, $M_6(6,6)$, $M_7(7,7)$ and $M_8(8,8)$, $M_9(9,9)$, $M_{10}(10,10)$, $M_{11}(11,11)$ and $M_{12}(12,12)$. Remember, we now have AVL2 speakers 1 to 4, and RMAV speakers 1 to 12. Speakers 1 to 4 are not the same in AVL2 and RMAV. These tests are Same Speaker-Dependent (SSD) because the same speaker is used to create the map, to train the models and for the testing data. Tables~\ref{tab:sd} \&~\ref{tab:sd_l} depict how these tests are constructed.

\begin{table}[!ht] 
\centering 
\caption{Same Speaker-Dependent (SSD) experiments for AVLetters2 speakers. The results from these tests will be used as a baseline.} 
\begin{tabular}{| l | l | l | l |} 
\hline 
\multicolumn{4}{| c |}{Same speaker-dependent (SD)}  \\ 
Mapping ($M_n$) & Training data ($p$) & Test speaker ($q$) & $M_n(p,q)$ \\ 
\hline \hline 
Sp1 & Sp1 & Sp1 & $M_1(1,1)$ \\ 
Sp2 & Sp2 & Sp2 & $M_2(2,2)$ \\ 
Sp3 & Sp3 & Sp3 & $M_3(3,3)$ \\ 
Sp4 & Sp4 & Sp4 & $M_4(4,4)$ \\ 
\hline 
\end{tabular} 
\label{tab:sd} 
\end{table} 

\begin{table}[!ht] 
\centering 
\caption{Same Speaker-Dependent (SSD) experiments for RMAV speakers. The results from these tests will be used as a baseline.} 
\begin{tabular}{| l | l | l | l |} 
\hline 
\multicolumn{4}{| c |}{Same speaker-dependent (SD)}  \\ 
Mapping ($M_n$) & Training data ($p$) & Test speaker ($q$) & $M_n(p,q)$ \\ 
\hline \hline 
Sp1 & Sp1 & Sp1 & $M_1(1,1)$ \\ 
Sp2 & Sp2 & Sp2 & $M_2(2,2)$ \\ 
Sp3 & Sp3 & Sp3 & $M_3(3,3)$ \\ 
Sp4 & Sp4 & Sp4 & $M_4(4,4)$ \\ 
Sp5 & Sp5 & Sp5 & $M_5(5,5)$ \\ 
Sp6 & Sp6 & Sp6 & $M_6(6,6)$ \\ 
Sp7 & Sp7 & Sp7 & $M_7(7,7)$ \\ 
Sp8 & Sp8 & Sp8 & $M_8(8,8)$ \\ 
Sp9 & Sp9 & Sp9 & $M_9(9,9)$ \\ 
Sp10 & Sp10 & Sp10 & $M_{10}(10,10)$ \\ 
Sp11 & Sp11 & Sp11 & $M_{11}(11,11)$ \\ 
Sp12 & Sp12 & Sp12 & $M_{12}(12,12)$ \\ 
\hline 
\end{tabular} 
\label{tab:sd_l} 
\end{table} 

\begin{table}[!ht] 
\centering 
\caption{Speaker-dependent phoneme-to-viseme mapping derived from phoneme classification confusions for each speaker in AVLetters2.} 
\begin{tabular} {| l | l || l | l |} 
\hline 
\multicolumn{2}{| c ||}{Speaker 1 $M_1$} & \multicolumn{2}{ c |}{Speaker 2 $M_2$} \\ 
\multicolumn{1}{| c |}{Viseme} & \multicolumn{1}{ c ||}{Phonemes}  		& \multicolumn{1}{ c |}{Viseme}  & \multicolumn{1}{ c |}{Phonemes}  	\\ 
\hline \hline 
/v01/	& /\textturnv/ /iy/ /\textschwa\textupsilon/ /uw/ 	& /v01/	& /ay/ /ey/ /iy/ /uw/		\\ 
/v02/	& /\textschwa/ /eh/ /ey/		& /v02/	& /\textschwa\textupsilon/			\\ 
/v03/	& /\textscripta/ /ay/			& /v03/	& /\textschwa/			\\ 
/v04/	& /d/ /s/ /t/			& /v04/	& /eh/			\\ 
/v05/	& /t\textipa{S}/ /l/  			& /v05/	& /\textturnv/ 			\\ 
/v06/	& /m/ /n/			& /v06/ 	& /\textschwa/			\\ 
/v07/	& /d\textipa{Z}/ /v/			& /v07/	& /d\textipa{Z}/ /p/ /y/ 		\\ 
/v08/	& /b/ /y/ 			& /v08/	& /l/ /m/ /n/ 		\\ 
/v09/	& /k/				& /v09/	& /v/ /w/			\\ 
/v10/	& /z/ 				& /v10/	& /d/ /b/			\\ 
/v11/	& /w/				& /v11/	& /f/ /s/			\\ 
/v12/	& /f/				& /v12/	& /t/ 				\\ 
&				& /v13/	& /k/				\\ 
&				& /v14/	& /t\textipa{S}/ 			\\ 
/sil/ 	& /sil/			& /sil/	& /sil/			\\ 
/garb/& /\textipa{E}/ /\textturnscripta/ /\textopeno/ /r/ /p/	& /garb/	& /\textipa{E}/ /\textturnscripta/ /\textopeno/ /r/ /z/	\\ 
\hline \hline 
\multicolumn{2}{| c ||}{Speaker 3 $M_3$} & \multicolumn{2}{ c  |}{Speaker 4 $M_4$} \\ 
Viseme & Phonemes			& Viseme & Phonemes \\ 
\hline \hline 
/v01/	& /ey/ /iy/			& /v01/	& /\textturnv/ /ay/ /ey/ /iy/	\\ 
/v02/	& /\textschwa/ /eh/			& /v02/	& /\textschwa/ /eh/			\\ 
/v03/	& /ay/			& /v03/	& /\textschwa/			\\ 
/v04/	& /\textturnv/			& /v04/	& /\textschwa\textupsilon/			\\ 
/v05/	& /\textschwa/			& /v05/	& /uw/			\\ 
/v06/	& /\textschwa\textupsilon/			& /v06/	& /m/ /n/			\\ 
/v07/	& /uw/			& /v07/	& /k/ /l/			\\ 
/v08/	& /d/ /p/ /t/			& /v08/	& /d\textipa{Z}/ /t/			\\ 
/v09/	& /l/ /m/			& /v09/	& /d/ /s/			\\ 
/v10/	& /k/ /w/			& /v10/	& /w/				\\ 
/v11/	& /f/ /n/			& /v11/	& /f/				\\ 
/v12/	& /b/ /s/			& /v12/	& /v/				\\ 
/v13/	& /v/				& /v13/	& /t\textipa{S}/ 		\\ 
/v14/	& /d\textipa{Z}/ 		& /v14/	& /b/				\\ 
/v15/	& /t\textipa{S}/ 		& /v15/	& /y/				\\ 
/v16/	& /y/				&		&				\\ 
/v17/	& /z/				& 		&				\\ 
/sil/		& /sil/			& /sil/	& /sil/			\\ 
/garb/	& /\textipa{E}/ /\textturnscripta/ /\textopeno/ /r/	& /garb/	& /\textipa{E}/ /\textturnscripta/ /\textopeno/ /r/ /p/ /z/ \\ 
\hline 
\end{tabular} 
\label{tab:td_v} 
\end{table} 
  
 The resulting AVL2 four speakers SSD P2V maps are listed in Table~\ref{tab:td_v}, Tables~\ref{tab:lilirvamps1-2},~\ref{tab:lilirvamps3-4},~\ref{tab:lilirvamps5-6},~\ref{tab:lilirvamps7-8},~\ref{tab:lilirvamps9-10} \&~\ref{tab:lilirvamps11-12}. We also permit a garbage, $/garb/$, viseme which is a cluster of phonemes in the ground truth which did not appear at all in the output from the phoneme classifier. Every viseme is listed with its associated mutually-confused phonemes e.g. for AVL2 Speaker 1, $M_1$, we see $/v01/$ is made up of phonemes \{/\textturnv/, /iy/, /\textschwa\textupsilon/, /uw/\}. We know from our clustering method in Chapter~\ref{chap:maps} this means in the phoneme classification, all four phonemes \{/\textturnv/, /iy/, /\textschwa\textupsilon/, /uw/\} were confused with the other three in the viseme. We are using the `strictly-confused' method from Chapter~\ref{chap:maps} with split vowel and consonant groupings as these achieved the most accurate classification. 

 \subsection{Different Speaker-Dependent maps \& Data (DSD\&D)} 
The second set of tests within this experiment start to look at using P2V maps with different test speakers. This means the HMM classifiers trained on each single speaker are used to recognise data from alternative speakers. 

\begin{table}[!h] 
\centering 
\caption{Different Speaker-Dependent maps and Data (DSD\&D) experiments with the four AVLetters2 speakers.} 
\begin{tabular}{| l | l | l | l |} 
\hline 
\multicolumn{4}{| c |}{Different Speaker-Dependent maps \& Data (DSD\&D)} \\ 
Mapping ($M_n$) & Training data ($p$) & Test speaker ($q$) & $M_n(p,q)$ \\ 
\hline \hline 
Sp2 & Sp2 & Sp1 & $M_2(2,1)$ \\ 
Sp3 & Sp3 & Sp1 & $M_3(3,1)$ \\ 
Sp4 & Sp4 & Sp1 & $M_4(4,1)$ \\ 
Sp1 & Sp1 & Sp2 & $M_1(1,2)$ \\ 
Sp3 & Sp3 & Sp2 & $M_3(3,2)$ \\ 
Sp4 & Sp4 & Sp2 & $M_4(4,2)$ \\ 
Sp1 & Sp1 & Sp3 & $M_1(1,3)$ \\ 
Sp2 & Sp2 & Sp3 & $M_2(2,3)$ \\ 
Sp4 & Sp4 & Sp3 & $M_4(4,3)$ \\ 
Sp1 & Sp1 & Sp4 & $M_1(1,4)$ \\ 
Sp2 & Sp2 & Sp4 & $M_2(2,4)$ \\ 
Sp3 & Sp3 & Sp4 & $M_3(3,4)$ \\ 
\hline 
\end{tabular} 
\label{tab:sid} 
\end{table} 

Within AVL2 this is completed for all four speakers using the P2V maps of the other speakers, and the data from the other speakers. Hence for Speaker 1 we construct $M_2(2,1)$, $M_3(3,1)$ and $M_4(4,1)$ and so on for the other speakers, this is depicted in Table~\ref{tab:sid}. 

\begin{table}[!h] 
\centering 
\caption{Different Speaker-Dependent maps and Data (DSD\&D) experiments for one of the 12 RMAV speakers (speaker one).} 
\begin{tabular}{| l | l | l | l |} 
\hline 
\multicolumn{4}{| c |}{Different Speaker-Dependent maps \& Data (DSD\&D)} \\ 
Mapping ($M_n$) & Training data ($p$) & Test speaker ($q$) & $M_n(p,q)$ \\ 
\hline \hline 
Sp2 & Sp2 & Sp1 & $M_2(2,1)$ \\ 
Sp3 & Sp3 & Sp1 & $M_3(3,1)$ \\ 
Sp4 & Sp4 & Sp1 & $M_4(4,1)$ \\ 
Sp5 & Sp5 & Sp1 & $M_5(5,1)$ \\ 
Sp6 & Sp6 & Sp1 & $M_6(6,1)$ \\ 
Sp7 & Sp7 & Sp1 & $M_7(7,1)$ \\ 
Sp8 & Sp8 & Sp1 & $M_8(8,1)$ \\ 
Sp9 & Sp9 & Sp1 & $M_9(9,1)$ \\ 
Sp10 & Sp10 & Sp1 & $M_{10}(10,1)$ \\ 
Sp11 & Sp11 & Sp1 & $M_{11}(11,1)$ \\ 
Sp12 & Sp12 & Sp1 & $M_{12}(12,1)$ \\ 
\hline 
\end{tabular} 
\label{tab:sid_l} 
\end{table} 
 
 For the RMAV speakers, we undertake this for all 12 speakers using the maps of the 11 others. We show the tests for a single speaker (Speaker 1) in Table~\ref{tab:sid_l} as an example.
 
\subsection{Different Speaker-Dependent maps (DSD)} 
Now we wish to isolate the effects of the HMM classifier from the effect of using different viseme P2V by training the classifiers on single speakers with the labels of the alternative speaker P2V maps. E.g. for AVL2 Speaker 1, the tests are: $M_2(1,1)$, $M_3(1,1)$ and $M_4(1,1)$. (All tests are listed in Table~\ref{tab:si}).  

\begin{table}[!h] 
\centering 
\caption{Different Speaker-Dependent maps (DSD) experiments for AVLetters2 speakers.} 
\begin{tabular}{| l | l | l | l |} 
\hline 
\multicolumn{4}{| c |}{Different Speaker-Dependent maps (DSD)} \\ 
Mapping ($M_n$) & Training data ($p$) & Test speaker ($q$) & $M_n(p,q)$ \\ 
\hline \hline 
Sp2 & Sp1 & Sp1& $M_2(1,1)$ \\ 
Sp3 & Sp1 & Sp1 & $M_3(1,1)$ \\ 
Sp4 & Sp1 & Sp1 & $M_4(1,1)$ \\ 
Sp1 & Sp2 & Sp2 & $M_1(2,2)$ \\ 
Sp3 & Sp2 & Sp2 & $M_3(2,2)$ \\ 
Sp4 & Sp2 & Sp2 & $M_4(2,2)$ \\ 
Sp1 & Sp3 & Sp3 & $M_1(3,3)$ \\ 
Sp2 & Sp3 & Sp3 & $M_2(3,3)$ \\ 
Sp4 & Sp3 & Sp3 & $M_4(3,3)$ \\ 
Sp1 & Sp4 & Sp4 & $M_1(4,4)$ \\ 
Sp2 & Sp4 & Sp4 & $M_2(4,4)$ \\ 
Sp3 & Sp4 & Sp4 & $M_3(4,4)$ \\ 
\hline 
\end{tabular} 
\label{tab:si} 
\end{table} 

These are the same P2V maps as in Table~\ref{tab:td_v} but trained and tested differently. In Table~\ref{tab:si_l} we show the equivalent DSD tests for Speaker 1 of RMAV as an example.

\begin{table}[!ht] 
\centering 
\caption{Different Speaker-Dependent maps (DSD) for one of the 12 RMAV speakers (Speaker one).} 
\begin{tabular}{| l | l | l | l |} 
\hline 
\multicolumn{4}{| c |}{Different Speaker-Dependent maps (DSD)} \\ 
Mapping ($M_n$) & Training data ($p$) & Test speaker ($q$) & $M_n(p,q)$ \\ 
\hline \hline 
Sp2 & Sp1 & Sp1& $M_2(1,1)$ \\ 
Sp3 & Sp1 & Sp1 & $M_3(1,1)$ \\ 
Sp4 & Sp1 & Sp1 & $M_4(1,1)$ \\ 
Sp5 & Sp1 & Sp1 & $M_5(1,1)$ \\ 
Sp6 & Sp1 & Sp1 & $M_6(1,1)$ \\ 
Sp7 & Sp1 & Sp1 & $M_7(1,1)$ \\ 
Sp8 & Sp1 & Sp1 & $M_8(1,1)$ \\ 
Sp9 & Sp1 & Sp1 & $M_9(1,1)$ \\ 
Sp10 & Sp1 & Sp1 & $M_{10}(1,1)$ \\ 
Sp11 & Sp1 & Sp1 & $M_{11}(1,1)$ \\ 
Sp12 & Sp1 & Sp1 & $M_{12}(1,1)$ \\ 
\hline 
\end{tabular} 
\label{tab:si_l} 
\end{table}

%\clearpage
\subsection{Multi-Speaker maps (MS)} 
A multi-speaker (MS) P2V map forms the viseme classifier labels in our third set of experiments. This map is constructed using phoneme confusions produced by {\em all} speakers in each data set and is shown in Table~\ref{tab:mt_v}, for the four AVL2 speakers, and Table~\ref{tab:lilirMSmap} for the 12 RMAV speakers. 

\begin{table}[!ht] 
\centering 
\caption{Multi-Speaker (MS) phoneme-to-viseme mapping for AVLetters2 speakers.} 
\begin{tabular}{| l | l |} 
\hline 
%\multicolumn{2}{| c |}{Multi-Speaker $M_{1234}$} \\ 
Viseme & Phonemes \\ 
\hline \hline 
/v01/ & /\textturnv/ /ay/ /ey/ /iy/ /\textschwa\textupsilon/ /uw/ \\ 
/v02/ & /\textschwa/ /eh/ \\ 
/v03/ & /\textscripta/ \\ 
/v04/ & /d/ /s/ /t/ /v/ \\ 
/v05/ & /f/ /l/ /n/ \\ 
/v06/ & /b/ /w/ /y/ \\ 
/v07/ & /d\textipa{Z}/ \\ 
/v08/ & /z/ \\ 
/v09/ & /p/ \\ 
/v10/ & /m/ \\ 
/v11/ & /k/ \\ 
/v12/ & /t\textipa{S}/ \\ 
/sil/ & /sil/ \\ 
/gar/ & /\textipa{E}/ /\textturnscripta/ /\textopeno/ /r/ \\	\hline 
\end{tabular} 
\label{tab:mt_v} 
\end{table} 

 \begin{table}[!ht] 
 \centering 
 	\caption{Multi-Speaker (MS) phoneme-to-viseme mapping for RMAV speakers.} 
 	\begin{tabular} {| l | l |} 
 	\hline  Viseme & Phonemes  \\ 
 	\hline \hline 
/v01/ & /\textscripta/ /\ae/ /\textturnv/ /\textopeno/ /\textschwa/ /ay/ /\textipa{E}/ /eh/ \\
&   /\textrevepsilon/ /ey/ /\textsci \textschwa/ /\textsci/ /iy/ /\textturnscripta/ /\textschwa \textupsilon/   \\ 
/v02/ & /\textopeno\textschwa/ /\textupsilon/ /\textopeno\textschwa/  \\ 
 /v03/ & /\textscripta\textupsilon/  \\ 
 /v04/ & /\textopeno\textsci/  \\ 
 /v05/ & /\textschwa/  \\ 
 /v06/ & /b/ /t\textipa{S}/ /d/ /\textipa{D}/ /f/ /g/ /\textipa{H}/ /d\textipa{Z}/ \\
 &   /k/ /l/ /m/ /n/ /\textipa{N}/ /p/ /r/ /s/ \\
 &  /\textipa{S}/ /t/ /\textipa{T}/ /v/ /w/ /y/ /z/   \\ 
/sil/ & /sil/  \\ 
 /sp/ & /sp/  \\ 
 /gar/ & /\textipa{Z}/ /c/  \\ 
  \hline 
 	 \end{tabular} 
 \label{tab:lilirMSmap} 
 \end{table}  

For our multi-speaker experiment notation, we substitute in the word `all' in place of a list of all the speakers for ease of reading. Therefore, the AVL2 MS map is tested as follows: $M_{[all]}(1,1)$, $M_{[all]}(2,2)$, $M_{[all]}(3,3)$ and $M_{[all]}(4,4)$: this is explained in Table~\ref{tab:ms} and the RMAV MS map is tested as: $M_{[all]}(1,1)$, $M_{[all]}(2,2)$, $M_{[all]}(3,3)$, $M_{[all]}(4,4)$, $M_{all]}(5,5)$, $M_{[all]}(6,6)$, $M_{[all]}(7,7)$, $M_{[all]}(8,8)$, $M_{[all]}(9,9)$, $M_{[all]}(10,10)$, $M_{[all]}(11,11)$, $M_{[all]}(12,12)$, as shown in Table~\ref{tab:ms_l}. 

\begin{table}[!ht] 
\centering 
\caption{Multi-Speaker (MS) experiments for AVLetters2 speakers.} 
\begin{tabular}{| l | l | l | l |} 
\hline 
\multicolumn{4}{| c |}{Multi-Speaker (MS)} \\ 
\multicolumn{1}{| c }{Mapping ($M_n$)} & \multicolumn{1}{ c }{Training data ($p$)} & \multicolumn{1}{ c }{Test speaker ($q$)} & \multicolumn{1}{ c |}{$M_n(p,q)$}\\ 
\hline \hline 
Sp[all] & Sp1 & Sp1 & $M_{[all]}(1,1)$ \\ 
Sp[all] & Sp2 & Sp2 & $M_{[all]}(2,2)$ \\ 
Sp[all] & Sp3 & Sp3 & $M_{[all]}(3,3)$ \\ 
Sp[all] & Sp4 & Sp4 & $M_{[all]}(4,4)$ \\ 
\hline 
\end{tabular} 
\label{tab:ms} 
\end{table} 

\begin{table}[!ht] 
\centering 
\caption{Multi-Speaker (MS) experiments for RMAV speakers.} 
\begin{tabular}{| l | l | l | l |} 
\hline 
\multicolumn{4}{| c |}{Multi-Speaker (MS)} \\ 
\multicolumn{1}{| c }{Mapping ($M_n$)} & \multicolumn{1}{ c }{Training data ($p$)} & \multicolumn{1}{ c }{Test speaker ($q$)} & \multicolumn{1}{ c |}{$M_n(p,q)$}\\ 
\hline \hline 
Sp[all] & Sp1 & Sp1 & $M_{all}(1,1)$ \\ 
Sp[all] & Sp2 & Sp2 & $M_{all}(2,2)$ \\ 
Sp[all] & Sp3 & Sp3 & $M_{all}(3,3)$ \\ 
Sp[all] & Sp4 & Sp4 & $M_{all}(4,4)$ \\
Sp[all] & Sp5 & Sp5 & $M_{all}(5,5)$ \\
Sp[all] & Sp6 & Sp6 & $M_{all}(6,6)$ \\
Sp[all] & Sp7 & Sp7 & $M_{all}(7,7)$ \\
Sp[all] & Sp8 & Sp8 & $M_{all}(8,8)$ \\
Sp[all] & Sp9 & Sp9 & $M_{all}(9,9)$ \\
Sp[all] & Sp10 & Sp10 & $M_{all}(10,10)$ \\
Sp[all] & Sp11 & Sp11 & $M_{all}(11,11)$ \\
Sp[all] & Sp12 & Sp12 & $M_{all}(12,12)$ \\
\hline 
\end{tabular} 
\label{tab:ms_l} 
\end{table}

\subsection{Speaker-Independent maps (SI)} 
Finally, our last set of tests looks at speaker independence in P2V maps themselves. Here we use maps which are derived using all speakers confusions bar the test speaker. This time we substitute the symbol `!$x$' in place of a list of speaker identifying numbers, meaning `not including speaker $x$'. The tests for these maps are as follows $M_{!1}(1,1)$, $M_{!2}(2,2)$, $M_{!3}(3,3)$ and $M_{!4}(4,4)$ as shown in Tables~\ref{tab:sim} \&~\ref{tab:sim_l} for AVL2 and RMAV speakers respectively. Speaker independent P2V maps for AVL2 speakers are shown in Table~\ref{tab:l1o_v}. 

\begin{table}[!pht] 
\centering 
\caption{Phoneme-to-viseme mapping derived from phoneme classification confusions of the three other speakers in AVLetters2.} 
\begin{tabular}{| l | l || l | l || l | l || l | l | } 
\hline 
\multicolumn{2}{| c ||}{Speaker 1 $M_{234}$} & \multicolumn{2}{ c ||}{Speaker 2 $M_{134}$ } \\ 
Viseme & Phonemes 		& Viseme & Phonemes 		\\ 
\hline \hline 
/v01/ 	& 	/\textturnv/ /\textschwa/ /ay/		& /v01/	& 	/\textturnv/ /ay/ /ey/ \\ 
		&	 /ey/ /iy/			&		&	/iy/			\\ 
/v02/ 	& 	/\textschwa\textupsilon/ /uw/ 			& /v02/ 	& 	/\textscripta/ /\textschwa\textupsilon/ /uw/	\\ 
/v03/ 	&	/eh/ 						& /v03/	& 	/\textschwa/ /eh/		\\ 
/v04/ 	& 	/\textscripta/ 				& /v04/ 	& 	/d/ /s/ /t/		\\ 
/v05/ 	& 	/d/ /s/ /t/ /v/ 				& /v05/	& 	/t\textipa{S}/ /l/	\\ 
/v06/ 	& 	/l/ /m/ /n/ 					& /v06/ 	& 	/b/ /d\textipa{Z}/		\\ 
/v07/ 	& 	/d\textipa{Z}/ /p/ /y/ 			& /v07/	& 	/v/ /y/			\\ 
/v08/ 	& 	/k/ /w/ 					& /v08/	& 	/k/ /w/ 		\\ 
/v09/ 	& 	/f/ 						& /v09/ 	& 	/p/			\\ 
/v10/ 	& 	/t\textipa{S}/ 						& /v10/	& 	/z/ 			\\ 
/v11/ 	& 	/b/ 						& /v11/	& 	/m/			\\ 
		&									&		&				\\ 
/sil/ 		& 	/sil/ 						& /sil/	& 	/sil/		 \\ 
/garb/ 	& 	/\textipa{E}/ /\textturnscripta/ /\textopeno/ /r/ /z/ 	& /garb/ 	& /\textipa{E}/ /\textturnscripta/ /\textopeno/ /r/ /f/ /n/	\\ 
\hline \hline 
\multicolumn{2}{| c ||}{Speaker 3 $M_{124}$ } & \multicolumn{2}{ c ||}{Speaker 4 $M_{123}$ } \\ 
Viseme & Phonemes			& Viseme & Phonemes \\ 
\hline \hline 
/v01/ 	&	/\textturnv/ /ay/ /ey/ 		& /v01/ 	& /\textturnv/ /ay/ /ey/ \\ 
&	/iy/ /\textschwa\textupsilon/ /uw/ 	&		&  /iy/ /\textschwa\textupsilon/ /uw/ \\ 
/v02/	& 	/\textscripta/				& /v02/	& /\textscripta/ \\ 
/v03/ 	& 	/\textschwa/ /eh/		& /v03/	& /\textschwa/ /eh/ \\ 
/v04/	& 	/d/ /s/ /t/ /v/				& /v04/	& /d\textipa{Z}/ /s/ /t/ /v/\\ 
/v05/ 	& 	/l/ /m/ /n/ 			& /v05/	& /f/ /l/ /n/ \\ 
/v06/ 	& 	/b/ /w/ /y/ 			& /v06/ 	& /b/ /d/ /p/ \\ 
/v07/	& 	/d\textipa{Z}/			& /v07/	& /w/ /y/ \\ 
/v08/	& 	/z/					& /v08/	& /z/ \\ 
/v09/ 	&	/p/ 				& /v09/ 	& /m/ \\ 
/v10/ 	& 	/k/ 				& /v10/ 	& /k/ \\ 
/v11/	& 	/f/ 					& /v11/ 	& /t\textipa{S}/ \\ 
/v12/	& 	/t\textipa{S}/			&  		&  \\ 
/sil/ 		& 	/sil/				& /sil/ 	& /sil/ \\ 
/garb/	& 	/\textipa{E}/ /\textturnscripta/ /\textopeno/ /r/ /iy/	& /garb/ 	& ea/ /\textturnscripta/ /\textopeno/ /r/ \\ 
\hline 
\end{tabular} 
\label{tab:l1o_v} 
\end{table} 

\begin{table}[!ht] 
\centering 
\caption{Speaker-Independent (SI) experiments with AVLetters2 speakers.} 
\begin{tabular}{| l | l | l | l |} 
\hline 
\multicolumn{4}{| c |}{Speaker-Independent (SI)} \\ 
Mapping ($M_n$) & Training data ($p$) & Test speaker ($q$) & $M_n(p,q)$\\ 
\hline \hline 
Sp[!1] & Sp1 & Sp1 & $M_{!1}(1,1)$ \\ 
Sp[!2] & Sp2 & Sp2 & $M_{!2}(2,2)$ \\ 
Sp[!3] & Sp3 & Sp3 & $M_{!3}(3,3)$ \\ 
Sp[!4] & Sp4 & Sp4 & $M_{!4}(4,4)$ \\ 
\hline 
\end{tabular} 
\label{tab:sim} 
\end{table} 
 
 \begin{table}[!ht] 
\centering 
\caption{Speaker-Independent (SI) experiments with RMAV speakers.} 
\begin{tabular}{| l | l | l | l |} 
\hline 
\multicolumn{4}{| c |}{Speaker-Independent (SI)} \\ 
Mapping ($M_n$) & Training data ($p$) & Test speaker ($q$) & $M_n(p,q)$\\ 
\hline \hline 
Sp[!1] & Sp1 & Sp1 & $M_{!1}(1,1)$ \\ 
Sp[!2] & Sp2 & Sp2 & $M_{!2}(2,2)$ \\ 
Sp[!3] & Sp3 & Sp3 & $M_{!3}(3,3)$ \\ 
Sp[!4] & Sp4 & Sp4 & $M_{!4}(4,4)$ \\ 
Sp[!5] & Sp5 & Sp5 & $M_{!5}(5,5)$ \\ 
Sp[!6] & Sp6 & Sp6 & $M_{!6}(6,6)$ \\ 
Sp[!7] & Sp7 & Sp7 & $M_{!7}(7,7)$ \\ 
Sp[!8] & Sp8 & Sp8 & $M_{!8}(8,8)$ \\ 
Sp[!9] & Sp9 & Sp9 & $M_{!9}(9,9)$ \\ 
Sp[!10] & Sp10 & Sp10 & $M_{[10}(10,10)$ \\ 
Sp[!11] & Sp11 & Sp11 & $M_{[11}(11,11)$ \\ 
Sp[!12] & Sp12 & Sp12 & $M_{!12}(12,12)$ \\ 
\hline 
\end{tabular} 
\label{tab:sim_l} 
\end{table} 

\section{The homophone risk factor} 
P2V maps are a many-to-one mapping. This creates the possibility of creating visual homophones when translating a phonetic transcript into a viseme transcript. For example, in the AVL2 data (isolated words are the letters of the alphabet) the phonetic realisation of the word `B' is `$/b/ /iy/$' and of `D' is `$/d/ /iy/$'. Using $M_2(2,2)$ to translate these into visemes they are identical `$/v08/ /v01/$' .
 
\begin{table}[!ht] 
\centering 
\caption{Count of visual homophones by each phoneme-to-viseme map, allowing for variation in pronunciation in AVLetters2 speakers.} %, and the percentage of the total words in AVL2 these make up} 
\begin{tabular}{| l | r | } %r | 
\hline 
Map 	&	Tokens $T$ \\ % & 	$T\%$ of $N$ \\ 
\hline \hline 
$M_1$	&	19		\\ % &	0.84	\\ 
$M_2$	&	19		\\ % & 	0.88	\\ 
$M_3$	& 	24		\\ % 	& 	1.07	\\ 
$M_4$	& 	24		\\ % 	& 	1.00	\\ 
\hdashline 
$M_{[all]}$	&	14	\\ % 	& 	0.69	\\ 
\hdashline 
$M_{!1}$	&	17	\\ % & 	0.69	\\ 
$M_{!2}$	&	18	\\ % &	0.80 \\ 
$M_{!3}$	&	20	\\ % &	0.96 \\ 
$M_{!4} $	&	15	\\ % &	0.73 \\ 
\hline 
\end{tabular} 
\label{tab:homophones} 
\end{table} 

Permitting variations in pronunciation, the total tokens ($T$) for each map after each word has been translated to visemes are listed in Table~\ref{tab:homophones}. More homophones means a greater the chance of substitution errors and a reduced correct classification. 
 
\section{Measuring similarity between phoneme-to-viseme maps} 
 
In Table~\ref{tab:avl_similarity} and~\ref{tab:lilir_similarity} we present a similarity score for comparing each pair of phoneme-to-viseme maps for AVL2 speakers and RMAV speakers respectively. 

\begin{table}[!ht]
         \centering
         \caption{Similarity scores between all AVLetters2 phoneme-to-viseme maps.}
         \begin{tabular}{|l||r|r|r|r|r|r|r|r|r|}
        \hline& $M_1$ & $M_2$ & $M_3$ & $M_4$ & $M_{[all]}$ & $M_{!1}$ & $M_{!2}$ & $M_{!3}$ & $M_{!4}$ \\
        \hline\hline
 $M_1$ & 0.000 & 0.327 & 0.322 & 0.247 & 0.199 & 0.244 & 0.048 & 0.112 & 0.222  \\
 $M_2$ & 0.327 & 0.000 & 0.410 & 0.303 & 0.333 & 0.266 & 0.256 & 0.254 & 0.253  \\
 $M_3$ & 0.322 & 0.410 & 0.000 & 0.157 & 0.465 & 0.400 & 0.394 & 0.398 & 0.396  \\
 $M_4$ & 0.247 & 0.303 & 0.157 & 0.000 & 0.301 & 0.298 & 0.172 & 0.246 & 0.378  \\
 $M_{[all]}$ & 0.199 & 0.333 & 0.465 & 0.301 & 0.000 & 0.311 & 0.220 & 0.098 & 0.136  \\
 $M_{!1}$ & 0.244 & 0.266 & 0.400 & 0.298 & 0.311 & 0.000 & 0.086 & 0.160 & 0.218  \\
  $M_{!2}$ & 0.048 & 0.256 & 0.394 & 0.172 & 0.220 & 0.086 & 0.000 & 0.155 & 0.160  \\
 $M_{!3}$ & 0.112 & 0.254 & 0.398 & 0.246 & 0.098 & 0.160 & 0.155 & 0.000 & 0.222  \\
 $M_{!4}$ & 0.222 & 0.253 & 0.396 & 0.378 & 0.136 & 0.218 & 0.160 & 0.222 & 0.000  \\
        \hline
        \end{tabular}
\label{tab:avl_similarity}
\end{table}

 \begin{sidewaystable}[!ht] 
 	 \centering 
 	 \caption{Similarity scores between all RMAV phoneme-to-viseme maps.}
 	 \resizebox{\columnwidth}{!}{% 
 	 \begin{tabular}{|l||r|r|r|r|r|r|r|r|r|r|r|r||r||r|r|r|r|r|r|r|r|r|r|r|r|}
	\hline& $M_1$ & $M_2$ & $M_3$ & $M_4$ & $M_5$ & $M_6$ & $M_7$ & $M_8$ & $M_9$ & $M_{10}$ & $M_{11}$ & $M_{12}$ & $M_{all}$ & $M_{!1}$ & $M_{!2}$ & $M_{!3}$ & $M_{!4}$ & $M_{!5}$ & $M_{!6}$ & $M_{!7}$ & $M_{!8}$ & $M_{!9}$ & $M_{!10}$ & $M_{!11}$ & $M_{!12}$ \\ 
	\hline\hline
 $M_1$ & 0.000 & 0.174 & 0.289 & 0.099 & 0.009 & 0.313 & 0.201 & 0.237 & 0.095 & 0.256 & 0.006 & 0.258 & 0.653 & 0.564 & 0.564 & 0.660 & 0.564 & 0.571 & 0.564 & 0.658 & 0.564 & 0.660 & 0.564 & 0.471 & 0.564  \\
 $M_2$ & 0.174 & 0.000 & 0.196 & 0.031 & 0.159 & 0.235 & 0.194 & 0.175 & 0.194 & 0.192 & 0.026 & 0.304 & 0.544 & 0.449 & 0.449 & 0.550 & 0.449 & 0.456 & 0.449 & 0.550 & 0.449 & 0.550 & 0.449 & 0.347 & 0.449  \\
 $M_3$ & 0.289 & 0.196 & 0.000 & 0.128 & 0.044 & 0.357 & 0.228 & 0.333 & 0.271 & 0.188 & 0.024 & 0.260 & 0.633 & 0.543 & 0.543 & 0.645 & 0.543 & 0.566 & 0.543 & 0.655 & 0.543 & 0.645 & 0.543 & 0.572 & 0.543  \\
 $M_4$ & 0.099 & 0.031 & 0.128 & 0.000 & 0.072 & 0.133 & 0.126 & 0.264 & 0.151 & 0.185 & 0.172 & 0.108 & 0.598 & 0.505 & 0.505 & 0.605 & 0.505 & 0.512 & 0.505 & 0.605 & 0.505 & 0.605 & 0.505 & 0.417 & 0.505  \\
 $M_5$ & 0.009 & 0.159 & 0.044 & 0.072 & 0.000 & 0.256 & 0.046 & 0.213 & 0.130 & 0.237 & 0.086 & 0.301 & 0.738 & 0.649 & 0.649 & 0.746 & 0.649 & 0.658 & 0.649 & 0.746 & 0.649 & 0.746 & 0.649 & 0.556 & 0.649  \\
 $M_6$ & 0.313 & 0.235 & 0.357 & 0.133 & 0.256 & 0.000 & 0.213 & 0.279 & 0.340 & 0.357 & 0.291 & 0.220 & 0.715 & 0.621 & 0.621 & 0.751 & 0.621 & 0.679 & 0.621 & 0.714 & 0.621 & 0.751 & 0.621 & 0.572 & 0.621  \\
 $M_7$ & 0.201 & 0.194 & {0.228} & 0.126 & 0.046 & 0.213 & 0.067 & 0.334 & 0.263 & 0.327 & 0.379 & 0.324 & 0.631 & 0.544 & 0.544 & 0.638 & 0.544 & 0.550 & 0.544 & 0.637 & 0.544 & 0.638 & 0.544 & 0.466 & 0.544  \\
 $M_8$ & 0.237 & 0.175 & 0.333 & 0.264 & 0.213 & 0.279 & 0.334 & 0.000 & 0.229 & 0.340 & 0.230 & 0.310 & 0.667 & 0.569 & 0.569 & 0.635 & 0.569 & 0.575 & 0.569 & 0.671 & 0.569 & 0.635 & 0.569 & 0.475 & 0.569  \\
 $M_9$ & 0.095 & 0.194 & 0.271 & 0.151 & 0.130 & 0.340 & 0.263 & 0.229 & 0.000 & 0.276 & 0.209 & 0.324 & 0.741 & 0.656 & 0.656 & 0.748 & 0.656 & 0.672 & 0.656 & 0.790 & 0.656 & 0.748 & 0.656 & 0.570 & 0.656  \\
 $M_{10}$ & 0.256 & 0.192 & 0.188 & 0.185 & 0.237 & 0.357 & 0.327 & 0.340 & 0.276 & 0.000 & 0.181 & 0.272 & 0.694 & 0.598 & 0.598 & 0.626 & 0.598 & 0.544 & 0.598 & 0.636 & 0.598 & 0.626 & 0.598 & 0.463 & 0.598  \\
 $M_{11}$ & 0.006 & 0.026 & 0.024 & 0.172 & 0.086 & 0.291 & 0.379 & 0.230 & 0.209 & 0.181 & 0.000 & 0.167 & 0.610 & 0.517 & 0.517 & 0.618 & 0.517 & 0.548 & 0.517 & 0.642 & 0.517 & 0.618 & 0.517 & 0.515 & 0.517  \\
 $M_{12}$ & 0.258 & 0.304 & 0.260 & 0.108 & 0.301 & 0.220 & 0.324 & 0.310 & 0.324 & 0.272 & 0.167 & 0.008 & 0.664 & 0.561 & 0.561 & 0.589 & 0.561 & 0.528 & 0.561 & 0.561 & 0.561 & 0.589 & 0.561 & 0.496 & 0.561  \\ \hline
 $M_{all}$ & 0.653 & 0.544 & 0.633 & 0.598 & 0.738 & 0.715 & 0.631 & 0.667 & 0.741 & 0.694 & 0.610 & 0.664 & 0.000 & 0.000 & 0.000 & 0.005 & 0.000 & 0.035 & 0.000 & 0.076 & 0.000 & 0.005 & 0.000 & 0.070 & 0.000  \\ \hline
 $M_{!1}$ & 0.564 & 0.449 & 0.543 & 0.505 & 0.649 & 0.621 & 0.544 & 0.569 & 0.656 & 0.598 & 0.517 & 0.561 & 0.000 & 0.000 & 0.000 & 0.057 & 0.000 & 0.033 & 0.000 & 0.020 & 0.000 & 0.057 & 0.000 & 0.067 & 0.000  \\
  $M_{!2}$ & 0.564 & 0.449 & 0.543 & 0.505 & 0.649 & 0.621 & 0.544 & 0.569 & 0.656 & 0.598 & 0.517 & 0.561 & 0.000 & 0.000 & 0.000 & 0.057 & 0.000 & 0.033 & 0.000 & 0.020 & 0.000 & 0.057 & 0.000 & 0.067 & 0.000  \\
 $M_{!3}$ & 0.660 & 0.550 & 0.645 & 0.605 & 0.746 & 0.751 & 0.638 & 0.635 & 0.748 & 0.626 & 0.618 & 0.589 & 0.005 & 0.057 & 0.057 & 0.000 & 0.057 & 0.070 & 0.057 & 0.030 & 0.057 & 0.000 & 0.057 & 0.018 & 0.057  \\
 $M_{!4}$ & 0.564 & 0.449 & 0.543 & 0.505 & 0.649 & 0.621 & 0.544 & 0.569 & 0.656 & 0.598 & 0.517 & 0.561 & 0.000 & 0.000 & 0.000 & 0.057 & 0.000 & 0.033 & 0.000 & 0.020 & 0.000 & 0.057 & 0.000 & 0.067 & 0.000  \\
 $M_{!5}$ & 0.571 & 0.456 & 0.566 & 0.512 & 0.658 & 0.679 & 0.550 & 0.575 & 0.672 & 0.544 & 0.548 & 0.528 & 0.035 & 0.033 & 0.033 & 0.070 & 0.033 & 0.000 & 0.033 & 0.099 & 0.033 & 0.070 & 0.033 & 0.100 & 0.033  \\
 $M_{!6}$ & 0.564 & 0.449 & 0.543 & 0.505 & 0.649 & 0.621 & 0.544 & 0.569 & 0.656 & 0.598 & 0.517 & 0.561 & 0.000 & 0.000 & 0.000 & 0.057 & 0.000 & 0.033 & 0.000 & 0.020 & 0.000 & 0.057 & 0.000 & 0.067 & 0.000  \\
 $M_{!7}$ & 0.658 & 0.550 & 0.655 & 0.605 & 0.746 & 0.714 & 0.637 & 0.671 & 0.790 & 0.636 & 0.642 & 0.561 & 0.076 & 0.020 & 0.020 & 0.030 & 0.020 & 0.099 & 0.020 & 0.000 & 0.020 & 0.030 & 0.020 & 0.018 & 0.020  \\
 $M_{!8}$ & 0.564 & 0.449 & 0.543 & 0.505 & 0.649 & 0.621 & 0.544 & 0.569 & 0.656 & 0.598 & 0.517 & 0.561 & 0.000 & 0.000 & 0.000 & 0.057 & 0.000 & 0.033 & 0.000 & 0.020 & 0.000 & 0.057 & 0.000 & 0.067 & 0.000  \\
 $M_{!9}$ & 0.660 & 0.550 & 0.645 & 0.605 & 0.746 & 0.751 & 0.638 & 0.635 & 0.748 & 0.626 & 0.618 & 0.589 & 0.005 & 0.057 & 0.057 & 0.000 & 0.057 & 0.070 & 0.057 & 0.030 & 0.057 & 0.000 & 0.057 & 0.018 & 0.057  \\
 $M_{!10}$ & 0.564 & 0.449 & 0.543 & 0.505 & 0.649 & 0.621 & 0.544 & 0.569 & 0.656 & 0.598 & 0.517 & 0.561 & 0.000 & 0.000 & 0.000 & 0.057 & 0.000 & 0.033 & 0.000 & 0.020 & 0.000 & 0.057 & 0.000 & 0.067 & 0.000  \\
 $M_{!11}$ & 0.471 & 0.347 & 0.572 & 0.417 & 0.556 & 0.572 & 0.466 & 0.475 & 0.570 & 0.463 & 0.515 & 0.496 & 0.070 & 0.067 & 0.067 & 0.018 & 0.067 & 0.100 & 0.067 & 0.018 & 0.067 & 0.018 & 0.067 & 0.000 & 0.067  \\
 $M_{!12}$ & 0.564 & 0.449 & 0.543 & 0.505 & 0.649 & 0.621 & 0.544 & 0.569 & 0.656 & 0.598 & 0.517 & 0.561 & 0.000 & 0.000 & 0.000 & 0.057 & 0.000 & 0.033 & 0.000 & 0.020 & 0.000 & 0.057 & 0.000 & 0.067 & 0.000  \\
	\hline
	\end{tabular} % 
	}
\label{tab:lilir_similarity} 
\end{sidewaystable}
\clearpage

The score addresses the phonemes within each viseme, the total number of phonemes clustered, the number of visemes within each set and ignores the ordering of the visemes within the set. As an example to explain our similarity algorithm, imagine we have the two phoneme-to-viseme maps shown in Figure~\ref{tab:examplemaps}.

\begin{figure}[!ht]
\centering
\begin{tabular}{|l|l|l|}
\hline
Map &Viseme & Phonemes \\
\hline \hline
& /v01/ & \{/p1/ /p2/ /p3/\} \\
Map 1 & /v02/ & \{/p4/ /p5/\} \\
& /v03/ & \{/p6/\} \\
& /v04/ & \{/p7/ /p8/\} \\
\hline
& /v01/ & \{/p1/ /p3/\} \\
& /v02/ & \{/p2/ /p4/\} \\
Map 2 & /v03/ & \{/p5/\} \\
& /v04/ & \{/p6/\} \\
& /v05/ & \{/p7/ /p8/ /p9/\} \\
\hline
\end{tabular}
\caption{Similarity algorithm: example phoneme-to-viseme maps.}
\label{tab:examplemaps}
\end{figure}

Our first step is to attribute a weight to each phoneme within each viseme. This is; $\frac{1}{\#phonemes}$ and is shown in Figure~\ref{tab:examplemaps1}.
\begin{figure}[!ht]
\centering
\begin{tabular}{|l|l|l|l|}
\hline
& Viseme & Phonemes & Phoneme weight  \\
\hline \hline
& /v01/ & \{/p1/ /p2/ /p3/\} & 0.3r \\
Map 1 & /v02/ & \{/p4/ /p5/\} & 0.5 \\
& /v03/ & \{/p6/\} & 1.0 \\
& /v04/ & \{/p7/ /p8/\} & 0.5 \\
\hline
& /v01/ & \{/p1/ /p3/\} & 0.5\\
& /v02/ & \{/p2/ /p4/\} & 0.5 \\
Map 2 & /v03/ & \{/p5/\} & 1.0 \\
& /v04/ & \{/p6/\} & 1.0 \\
& /v05/ & \{/p7/ /p8/ /p9/\} & 0.3r  \\
\hline
\end{tabular}
\caption{Phoneme-to-viseme map similarity algorithm step 1: Example phoneme-to-viseme maps with weighted phonemes.}
\label{tab:examplemaps1}
\end{figure}

Now we use these values to compare all visemes of one map with another. 

\begin{figure}[H]
\centering
\begin{tabular}{|l|l|r|r|r|r|r|r|}
\hline
\multicolumn{1}{| c |}{} & \multicolumn{1}{ c |}{} & \multicolumn{5}{ c |}{Map 2} \\
\hline
\multicolumn{1}{| c |}{} & \multicolumn{1}{ c |}{} & /v1/ & /v2/ & /v3/ & /v4/ & /v5/ \\
\hline \hline
\multirow{4}{*}{Map 1}&/v1/ & $/p1/, /p3/$ 	& $/p2/$ 	& - 		& - 		& - \\
&/v2/ & - 			& $/p4/$ 	& $/p5/$ 	& - 		& - \\
&/v3/ & - 			& - 		& - 		& $/p6/$ 	& - \\
&/v4/ & - 			& - 		& - 		& - 		& $/p7/, /p8/$\\
\hline
\end{tabular}
\caption{Phoneme-to-viseme map similarity algorithm step 2: phoneme in viseme matches.}
\label{fig:phonematches}
\end{figure}

Where two visemes $V_i$ and $V_j$ contain the same phonemes (see Figure~\ref{fig:phonematches}), the $V_{ij}$ score is the sum of the matched phoneme weights (Figure~\ref{fig:simscoreeg}). The final values are in Figure~\ref{fig:simscores}.

\begin{figure}[H]
\centering
\resizebox{\columnwidth}{!}{% 
\begin{tabular}{|l|r|r|r|r|r|r|}
\hline
\multicolumn{1}{| c |}{} & \multicolumn{1}{ c |}{} & \multicolumn{5}{ c |}{Map 2} \\
\hline
\multicolumn{1}{| c |}{} & \multicolumn{1}{ c |}{} & /v1/ 			& /v2/			& /v3/ 			& /v4/ 			& /v5/ \\
\hline \hline
\multirow{4}{*}{Map 1}& /v1/ 	& $/p1/=0.3r+0.5$ 	& $/p2/=0.3r+0.3r$ 	& 0 				& 0 				& 0 \\
&	& $/p3/=0.3r+0.5$	& -				& -				& -				& -\\
& /v2/ 	& 0				& $/p4/=0.5+0.5$	& $/p5/=0.5+1.0$ 	& 0 				& 0 \\
& /v3/ 	& 0 				& 0 				& 0 				& $/p6/=1.0+1.0$ 	& 0 \\
& /v4/ 	& 0 				& 0 				& 0 				& 0				& $/p7/=0.5+0.3r$  \\
&	& -				& - 				& -				& 0 				& $/p8/=0.5+0.3r$ \\
\hline
\end{tabular} %
}
\caption{Phoneme-to-viseme map similarity algorithm step 3: summing the phoneme weights.}
\label{fig:simscoreeg}
\end{figure}

\begin{figure}[H]
\centering
\begin{tabular}{|l|r|r|r|r|r|r|}
\hline
\multicolumn{1}{| c |}{} & \multicolumn{1}{ c |}{} & \multicolumn{5}{ c |}{Map 2} \\
\hline
\multicolumn{1}{| c |}{} & \multicolumn{1}{ c |}{} & /v1/ & /v2/ & /v3/ & /v4/ & /v5/ \\
\hline \hline
\multirow{4}{*}{Map 1}&/v1/ & $1.6r$ & $0.6r$ & 0 & 0 & 0 \\
&/v2/ & 0 & $1.0$ & $1.5$ & 0 & 0 \\
&/v3/ &0 & 0 & 0 & $2.0$ & 0 \\
&/v4/ & 0 & 0 & 0 & 0 & $1.6r$ \\
\hline
\end{tabular}
\caption{Phoneme-to-viseme map similarity algorithm step 4: total phoneme weights.}
\label{fig:simscores}
\end{figure}

Finally we need to sum of all values in the upper triangle $U \forall_{ij}i>j$, minus the sum of all values in the lower triangle $L \forall_{ij}i<j$, normalised by dividing by the total number of matched phonemes, $N_p$, (in our example, eight) to give the value $0.73'$ (\ref{equation:simval}). $S$ is the similarity score.  

\begin{equation}
S = U-L
\label{equation:simval}
\end{equation}

This similarity measure is calculated to compare all the P2V maps used in our experiments in pairs and the results are shown in Tables~\ref{tab:avl_similarity} and~\ref{tab:lilir_similarity}. The values closest to zero show the most similar maps, thus the closer to 1, the more different the maps are. We have not compared the maps between datasets due to biased effects caused by the disparity between word content and data size. Unsurprisingly, with the RMAV dataset, the MS and SI P2V maps are all very similar because of the volume of speakers and folds of phoneme classification, there is more chance of unique phonemes being confused. There is at most 3 phonemes different between them all. 

If we compare all the P2V maps in Tables~\ref{tab:mt_v} \&~\ref{tab:l1o_v}, there are similarities. Mostly because there is only one speaker at a time removed from within SI P2V maps. However, if these are compared to the speaker-dependent maps in Table~\ref{tab:td_v}, a different picture can be seen. Speaker 4 is significantly affected by the introduction of /\textschwa\textupsilon/ and /uw/ into viseme $/v01/$. Where Speaker 1 has these in $M_1(1,1)$, his SD word classification of $15.9\%$ is less than half of Speaker 4's $38.4\% $ (Figure~\ref{fig:correctnessDSD}). 
 
\section{Analysis of speaker independence in phoneme-to-viseme maps} 
%All experiments have been conducted with both AVL2 (isolated words) and the RMAV (continuous speech) datasets. 
\begin{figure}[!ht] 
\centering 
\includegraphics[width=0.8\linewidth]{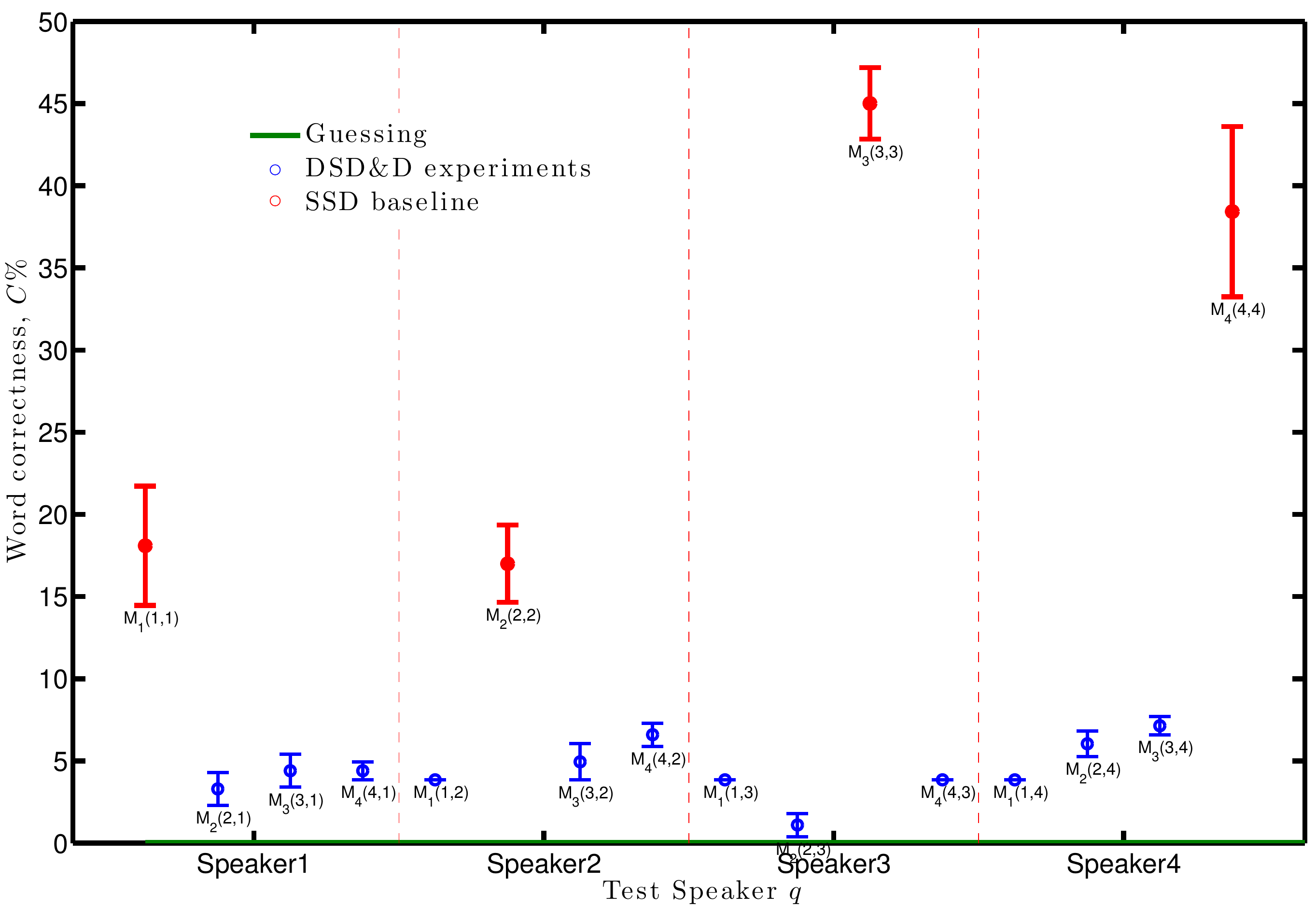} 
\caption{Word classification correctness, $C\pm1\frac{\sigma}{\sqrt{7}}$, of the DSD\&D tests where HMM classifiers are tested on all three other speakers in AVLetters2. Baseline is the SSD maps.} 
\label{fig:indep_Corr} 
\end{figure} 
 
\begin{figure}[!ht] 
\centering 
\includegraphics[width=0.8\linewidth]{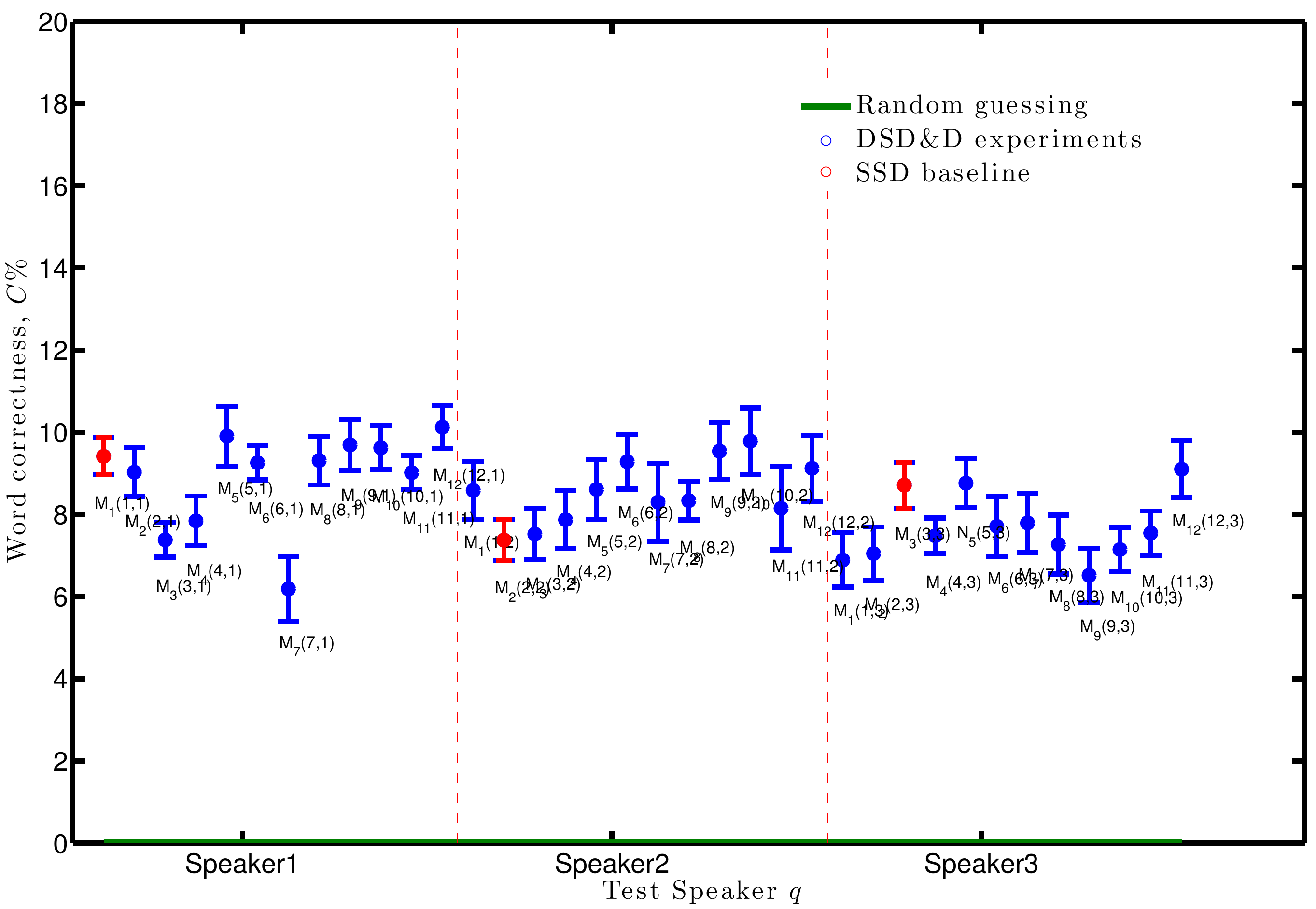} 
\caption{Word classification correctness, $C\pm1\frac{\sigma}{\sqrt{10}}$, of the DSD\&D tests where HMM classifiers are tested on all eleven other speakers in RMAV. Baseline is SSD maps (red) - Speakers 1-3.} 
\label{fig:indep_CorrL1} 
\end{figure} 
\begin{figure}[!ht] 
\centering 
\includegraphics[width=0.8\linewidth]{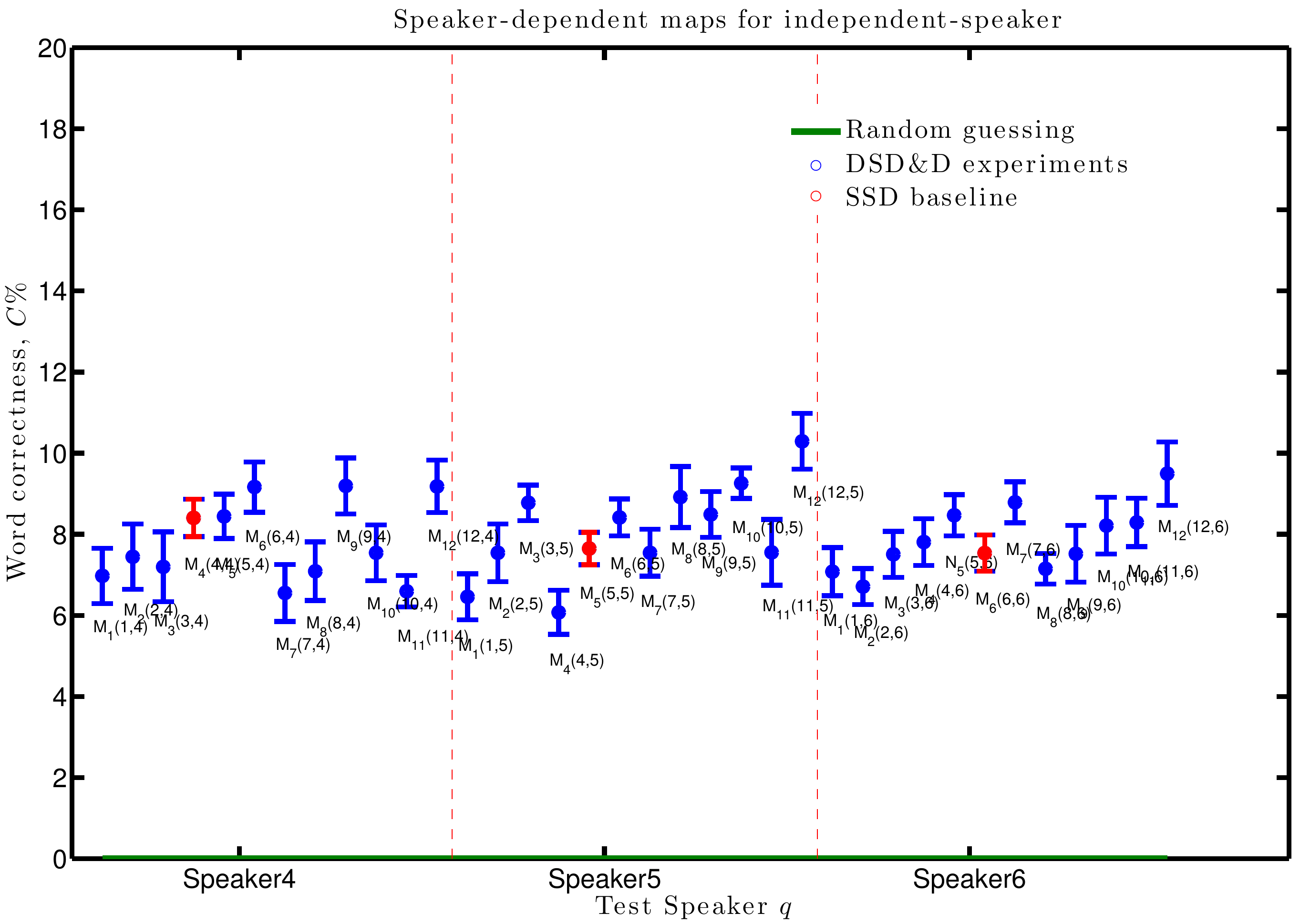} 
\caption{Word classification correctness, $C\pm1\frac{\sigma}{\sqrt{10}}$, of the DSD\&D tests where HMM classifiers are tested on all eleven other speakers in RMAV. Baseline is SSD maps (red) - Speakers 4-6.} 
\label{fig:indep_CorrL4} 
\end{figure} 
\begin{figure}[!ht] 
\centering 
\includegraphics[width=0.8\linewidth]{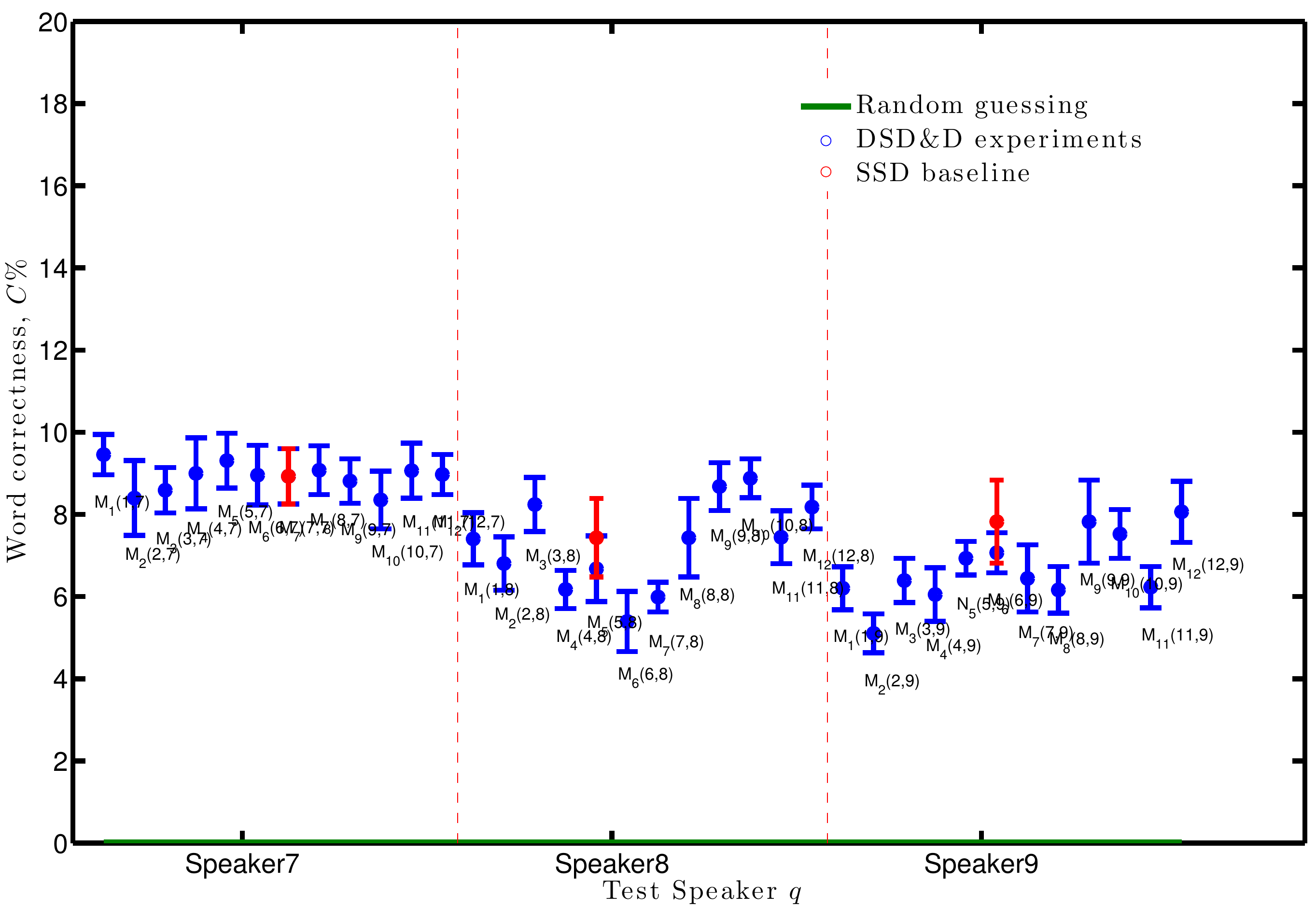} 
\caption{Word classification correctness, $C\pm1\frac{\sigma}{\sqrt{10}}$, of the DSD\&D tests where HMM classifiers are tested on all eleven other speakers in RMAV. Baseline is SSD maps (red) - Speakers 7-9.} 
\label{fig:indep_CorrL7} 
\end{figure} 
\begin{figure}[!ht] 
\centering 
\includegraphics[width=0.8\linewidth]{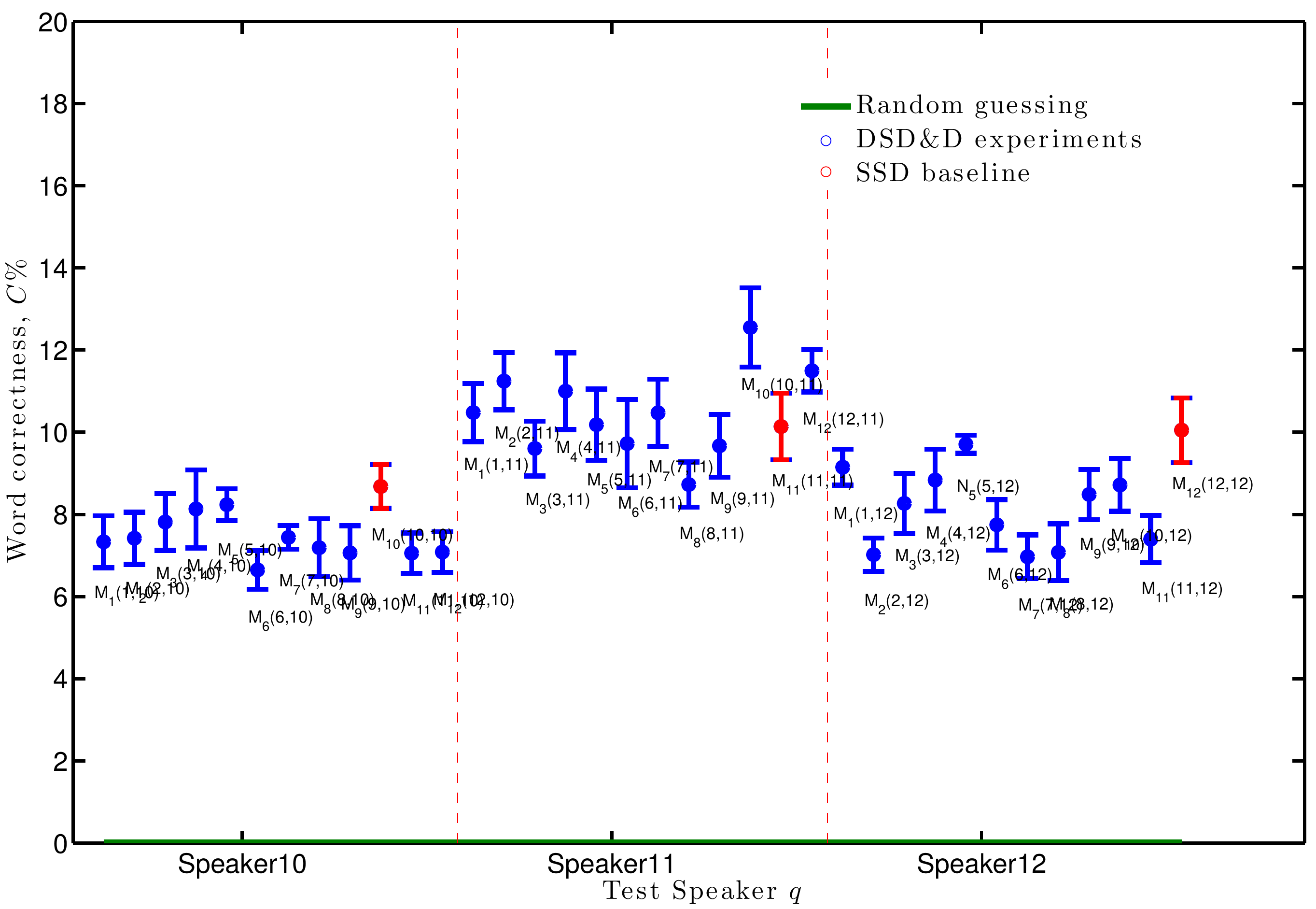} 
\caption{Word classification correctness, $C\pm1\frac{\sigma}{\sqrt{10}}$, of the DSD\&D tests where HMM classifiers are tested on all eleven other speakers in RMAV. Baseline is SSD maps (red) - Speakers 10-12.} 
\label{fig:indep_CorrL10} 
\end{figure} 
 
Figure~\ref{fig:indep_Corr} shows the word correctness of AVL2 speaker-dependent viseme classes on the $y$-axis. In this figure, the baseline is $n=p=q$ for all $M$. These are compared to the DSD\&D tests: $M_2(2,1)$, $M_3(3,1)$, $M_4(4,1)$ for Speaker 1, $M_1(1,2)$, $M_3(3,2)$, $M_4(4,2)$ for Speaker 2, $M_1(1,3)$, $M_2(2,3)$, $M_4(4,3)$ for Speaker 3 and $M_1(1,4)$, $M_2(2,4)$, $M_3(3,4)$ for Speaker 4 as in Table~\ref{tab:sid}. We also plot guessing (calculated as $1/N$, where $N$ is the total number of words in the dataset. For AVL2 this is 26, for RMAV speaker this ranges between 1362 and 1802). DSD HMM classifiers are significantly worse than SSD HMMs, as all results where $p$ is not the same speaker as $q$ are around the equivalent performance of  guessing. This correlates with similar tests of independent HMM's in \cite{cox2008challenge}. This gap is attributed to two possible effects, either - the visual units are incorrect, or they are trained on the incorrect speaker. 

Figures~\ref{fig:indep_CorrL1}, ~\ref{fig:indep_CorrL4}, ~\ref{fig:indep_CorrL7}, \&~\ref{fig:indep_CorrL10} show the same tests but on the continuous speech data. It is reassuring to see some speakers significantly deteriorate the classification rates when the speaker used to train the classifier is not the same as the test speaker. As an example we look at Speaker 1 on the leftmost side of Figure~\ref{fig:indep_CorrL1}. Here the test speaker is Speaker 1. The speaker-dependent maps for all 12 speakers have been used to build HMMs classifiers. But when tested on Speaker 1, only maps and models for speakers 3, 7 and 12 show a significant reduction in word correctness. All eight other speakers are within one standard error. 

Figure~\ref{fig:indep_CorrL4}, for the RMAV speakers four to six, we see a similar trend with Speaker 4 showing the most variation of these three speakers. To lip-read Speaker 4 we actually see a significant improvement by using the map and model of Speaker 6 and less significant improvements by speakers 3, 5 and 11. In Figure~\ref{fig:indep_CorrL7} we see Speaker 11's SD map and models majorly improve the classification of Speaker 8. However, whilst these are all signs of possibly making strides towards speaker independent classification, Speaker 12 in Figure~\ref{fig:indep_CorrL10} shows the most common trend is there is a lot of overlap between our continuous speech speakers and this natural variation is attributed to the speaker identity.

\begin{figure}[!ht] 
\centering 
\includegraphics[width=0.8\linewidth]{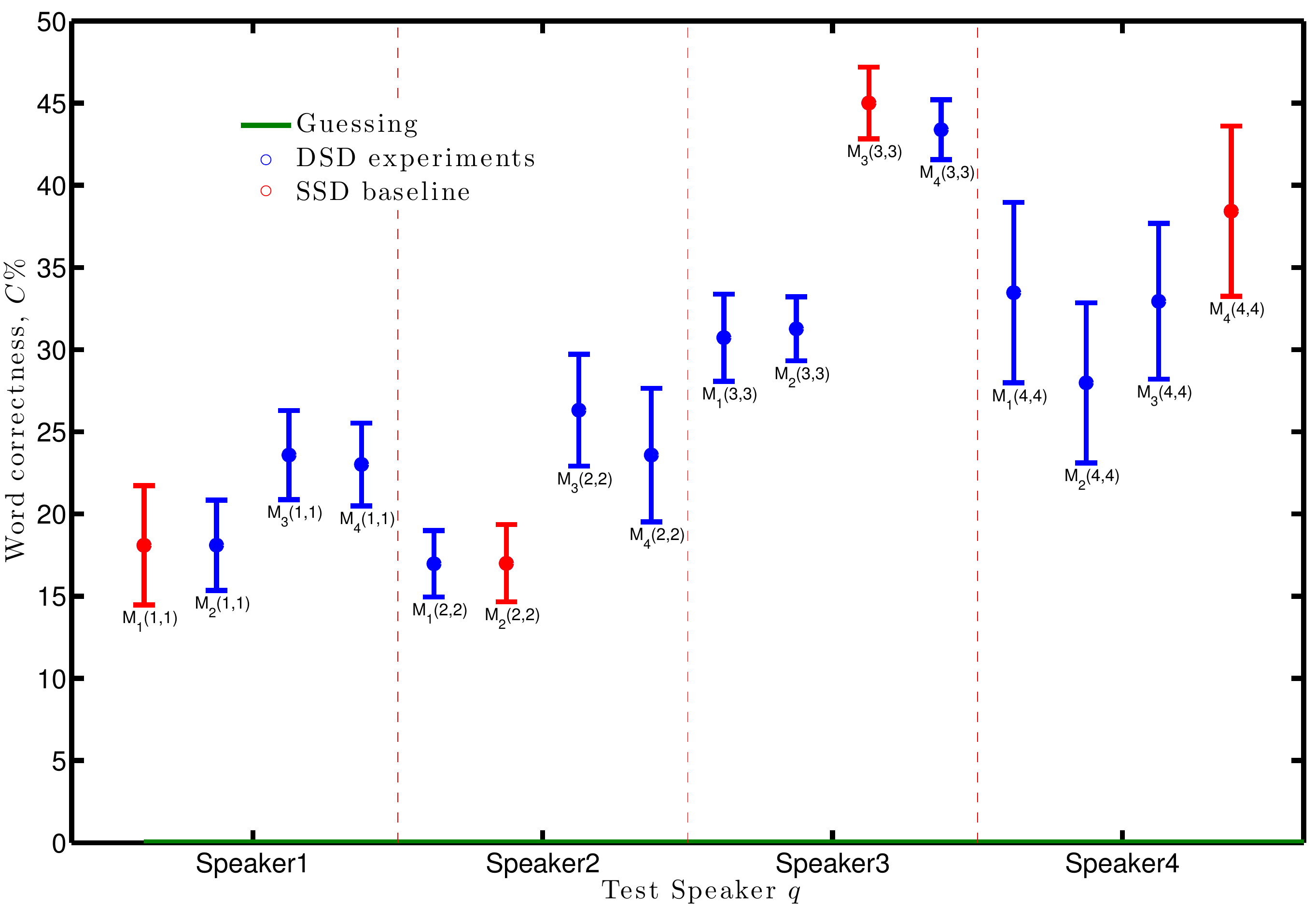} 
\caption{Word classification correctness, $C\pm1\frac{\sigma}{\sqrt{7}}$, of the DSD tests where HMM classifiers are constructed with single-speaker dependent phoneme-to-viseme maps for all four speakers in AVLetters2. Baseline is the SSD maps.} 
\label{fig:correctnessDSD} 
\end{figure} 
 
Figure~\ref{fig:correctnessDSD} shows our AVL2 DSD experiments from Table~\ref{tab:si}. Our results in word correctness, $C$, are plotted on the $y$-axis and we also plot the same benchmark as in Figure~\ref{fig:indep_Corr} ($n=p=q$). In our DSD tests, the HMM is allowed to be trained on the relevant speaker, so the other tests are: $M_2(1,1)$, $M_3(1,1)$, $M_4(1,1)$ for Speaker 1, $M_1(2,2)$, $M_3(2,2)$, $M_4(2,2)$ for Speaker 2, $M_1(3,3)$, $M_2(3,3)$, $M_4(3,3)$ for Speaker 3 and finally $M_1(4,4)$, $M_2(4,4)$, $M_3(4,4)$ for Speaker 4. Now the word correctness has improved substantially which implies the previous poor performance in Figure~\ref{fig:indep_Corr} was not due to the choice of visemes but rather, the badly trained HMMs. 

\begin{figure}[!ht] 
\centering 
\includegraphics[width=0.8\linewidth]{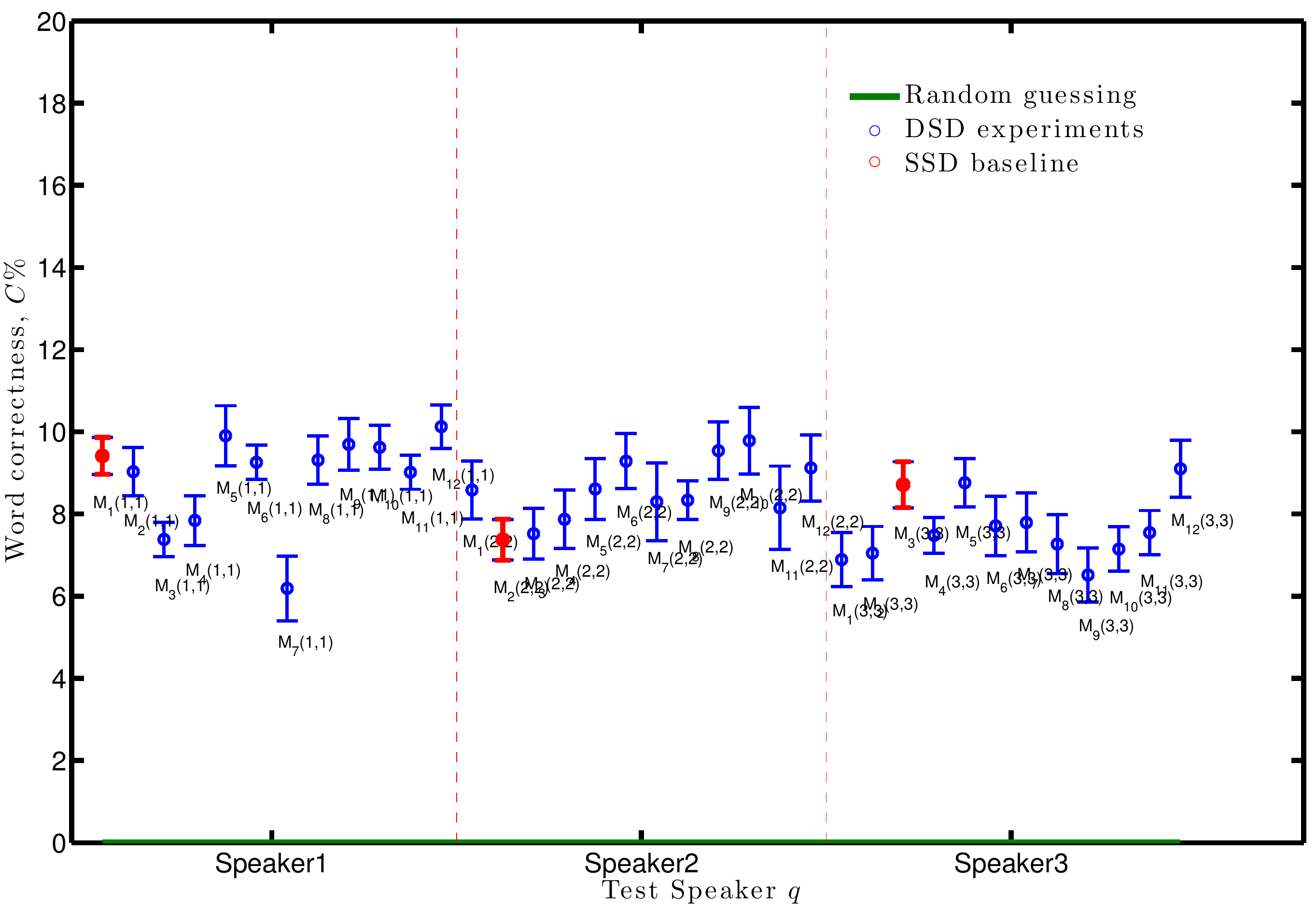} 
\caption{Word classification correctness, $C\pm1\frac{\sigma}{\sqrt{10}}$, of the DSD tests where HMM classifiers are constructed with single-speaker dependent phoneme-to-viseme maps for all speakers in RMAV and tested on others. Baseline is SSD maps (red), results shown for HMMs trained on speakers 1-3.} 
\label{fig:correctness1} 
\end{figure} 
 \begin{figure}[!ht] 
\centering 
\includegraphics[width=0.8\linewidth]{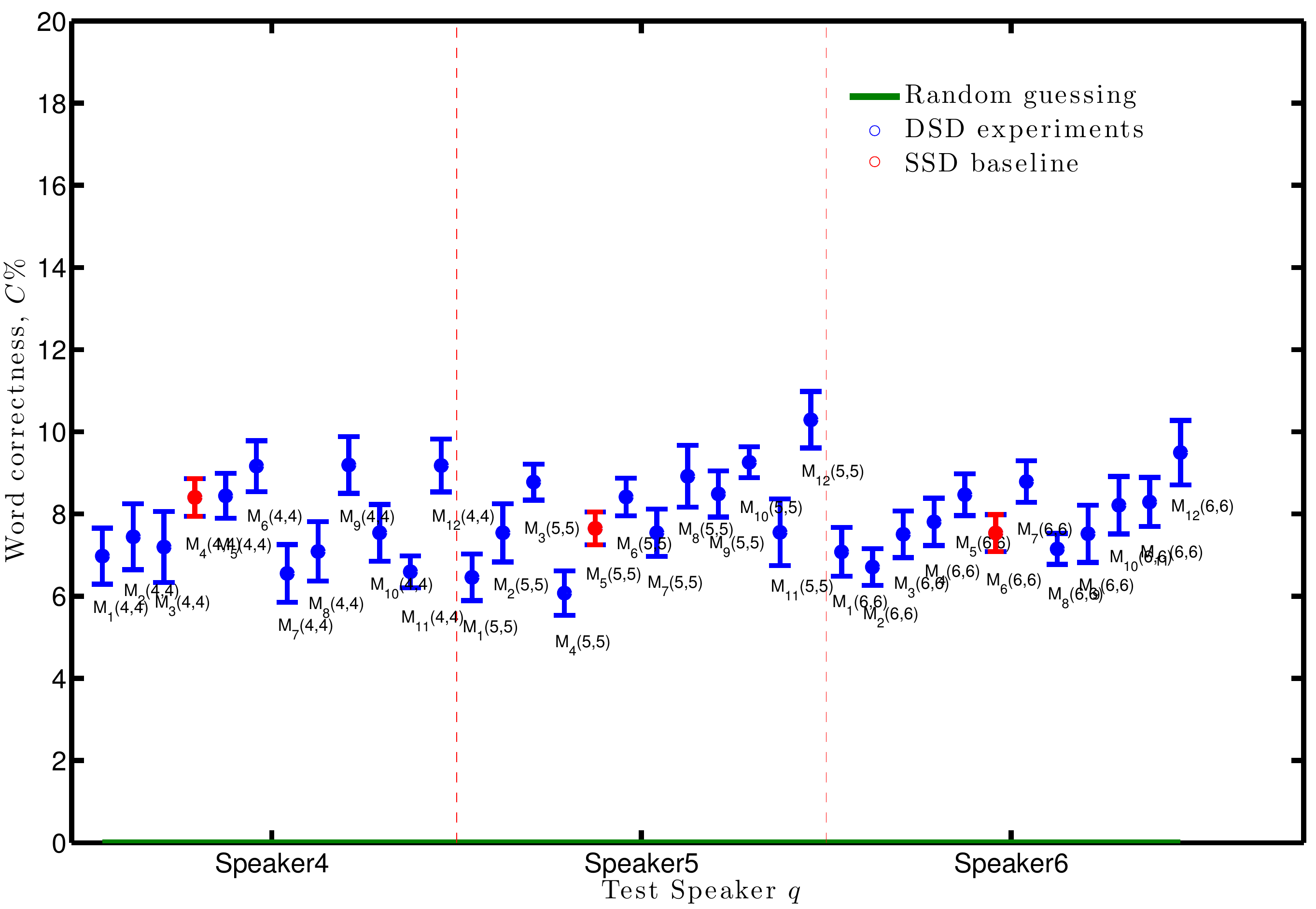} 
\caption{Word classification correctness, $C\pm1\frac{\sigma}{\sqrt{10}}$, of the DSD tests where HMM classifiers are constructed with single-speaker dependent phoneme-to-viseme maps for all speakers in RMAV and tested on others. Baseline is SSD maps (red), results shown for HMMs trained on speakers 4-6.} 
\label{fig:correctness2} 
\end{figure} 
\begin{figure}[!ht] 
\centering 
\includegraphics[width=0.8\linewidth]{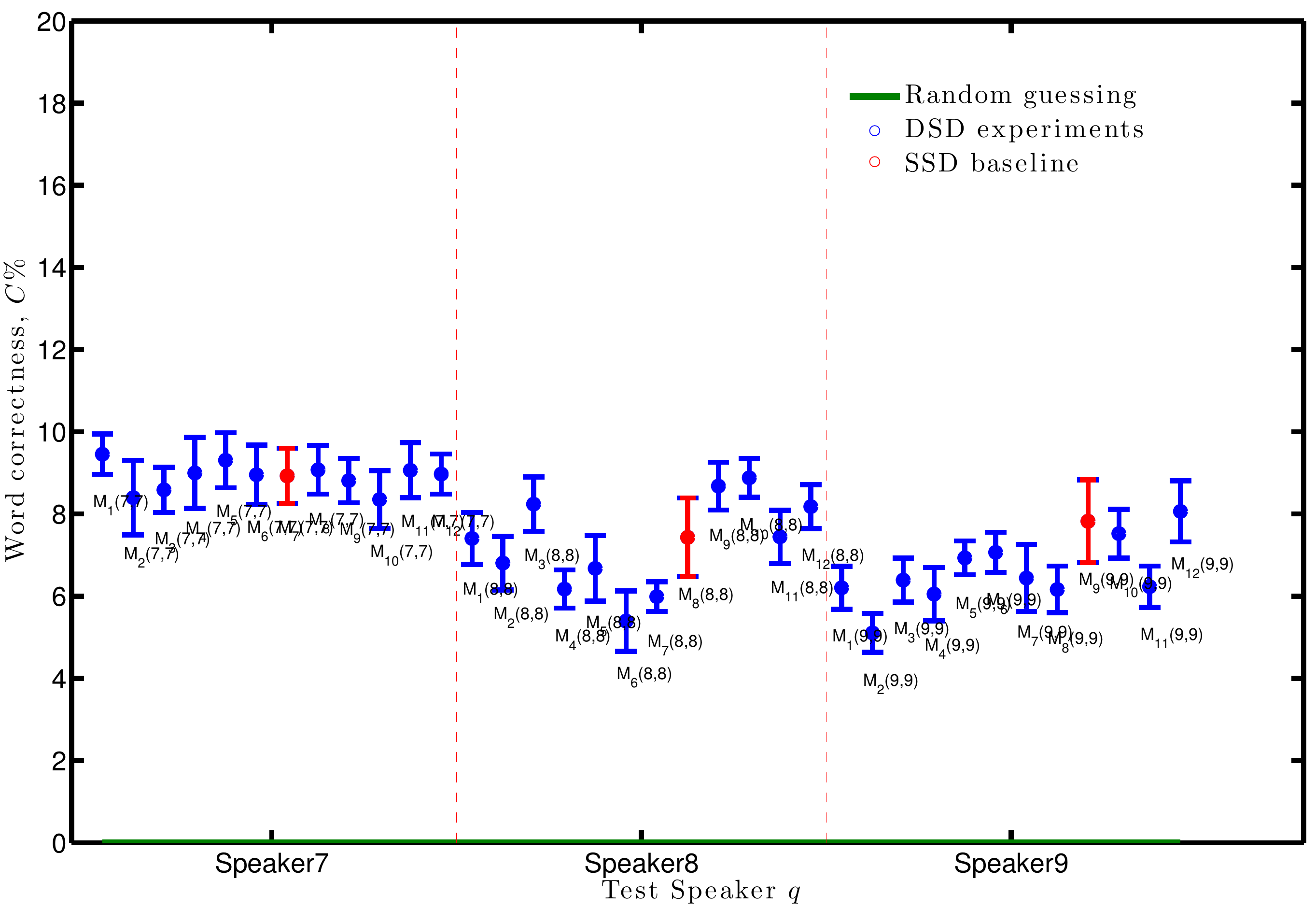} 
\caption{Word classification correctness, $C\pm1\frac{\sigma}{\sqrt{10}}$, of the DSD tests where HMM classifiers are constructed with single-speaker dependent phoneme-to-viseme maps for all speakers in RMAV and tested on others. Baseline is SSD maps (red), results shown for HMMs trained on speakers 7-9.} 
\label{fig:correctness3} 
\end{figure}  \clearpage
\begin{figure}[!ht] 
\centering 
\includegraphics[width=0.8\linewidth]{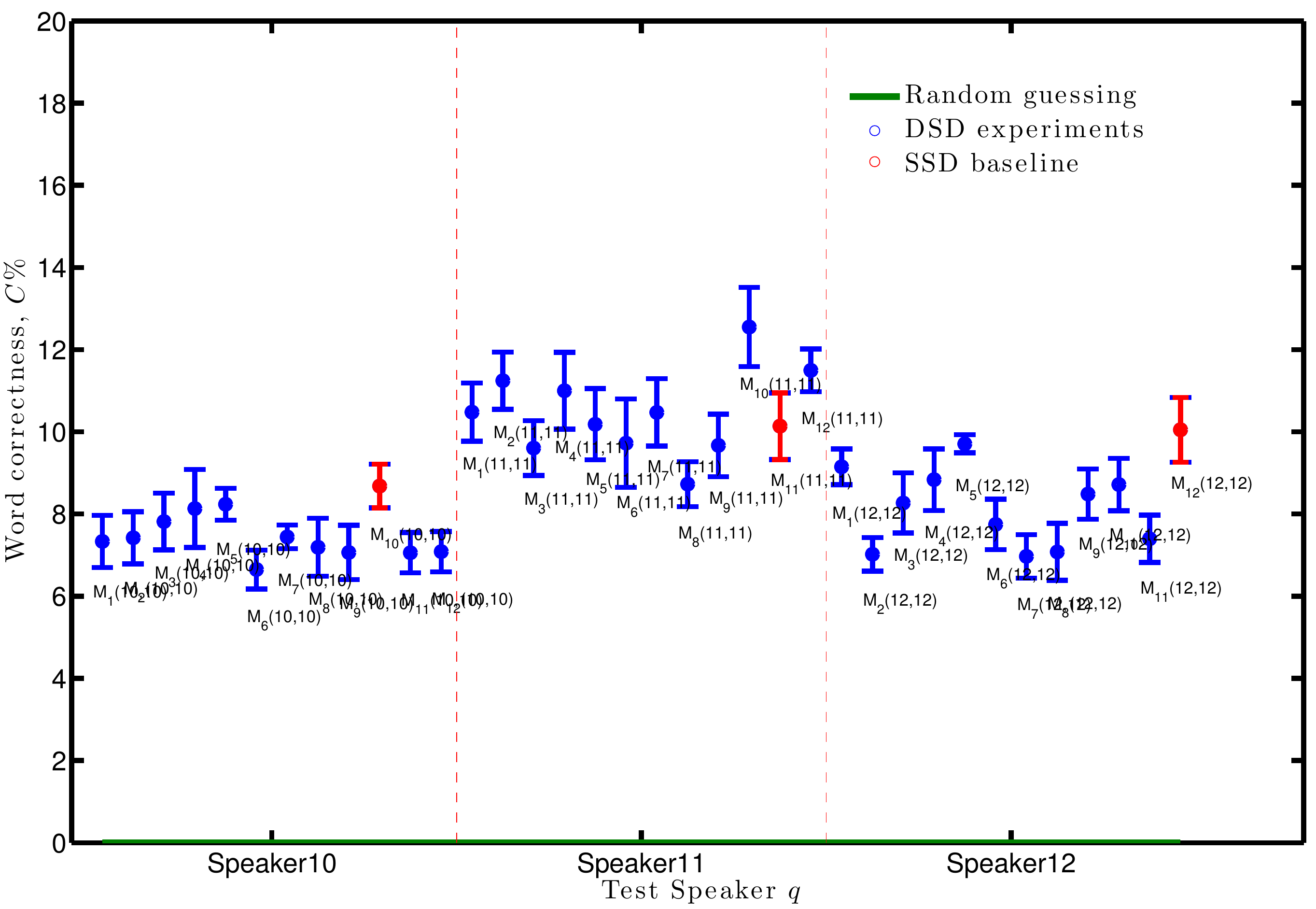} 
\caption{Word classification correctness, $C\pm1\frac{\sigma}{\sqrt{10}}$, of the DSD tests where HMM classifiers are constructed with single-speaker dependent phoneme-to-viseme maps for all speakers in RMAV and tested on others. Baseline is SSD maps (red), results shown for HMMs trained on speakers 10-12.} 
\label{fig:correctness4} 
\end{figure} 
 
The equivalent graphs for the 12 RMAV speakers are in Figures~\ref{fig:correctness1},~\ref{fig:correctness2},~\ref{fig:correctness3} and~\ref{fig:correctness4}. Now we can see the effects of the unit selection. Using Speaker 1 for example, in Figure~\ref{fig:correctness1} the three maps $M_3, M_7$ and $M_{12}$ all significantly reduce the correctness for Speaker 1. In contrast, for Speaker 2 there are no significantly reducing maps but maps 1, 4, 5, 6, 9 and 11 all significantly improve the classification of Speaker 2. This suggests its not just the speakers identity which is important for good classification but how it is used. Some individuals may simply be easier to lip read (for reasons as yet unknown) or there are similarities between certain speakers which when learned properly on one speaker are able to better classify the rarer visual distinctions between phonemes on similar other speakers.
 
In Figure~\ref{fig:correctness3} we see Speaker 7 is particularly robust to visual unit selection for the classifier labels. Conversely Speakers 5 (Figure~\ref{fig:correctness2}) and 12 (Figure~\ref{fig:correctness4}) are really affected by the visemes (or phoneme clusters). Its interesting to note this is a variability not previously considered, some speakers may be dependent on good visual classifiers and the mapping back to acoustics utterances, but others not so much. Again, the number of visual classifiers really does vary subject to the speaker identity. 
 
\begin{figure}[h] 
\centering 
\includegraphics[width=0.8\linewidth]{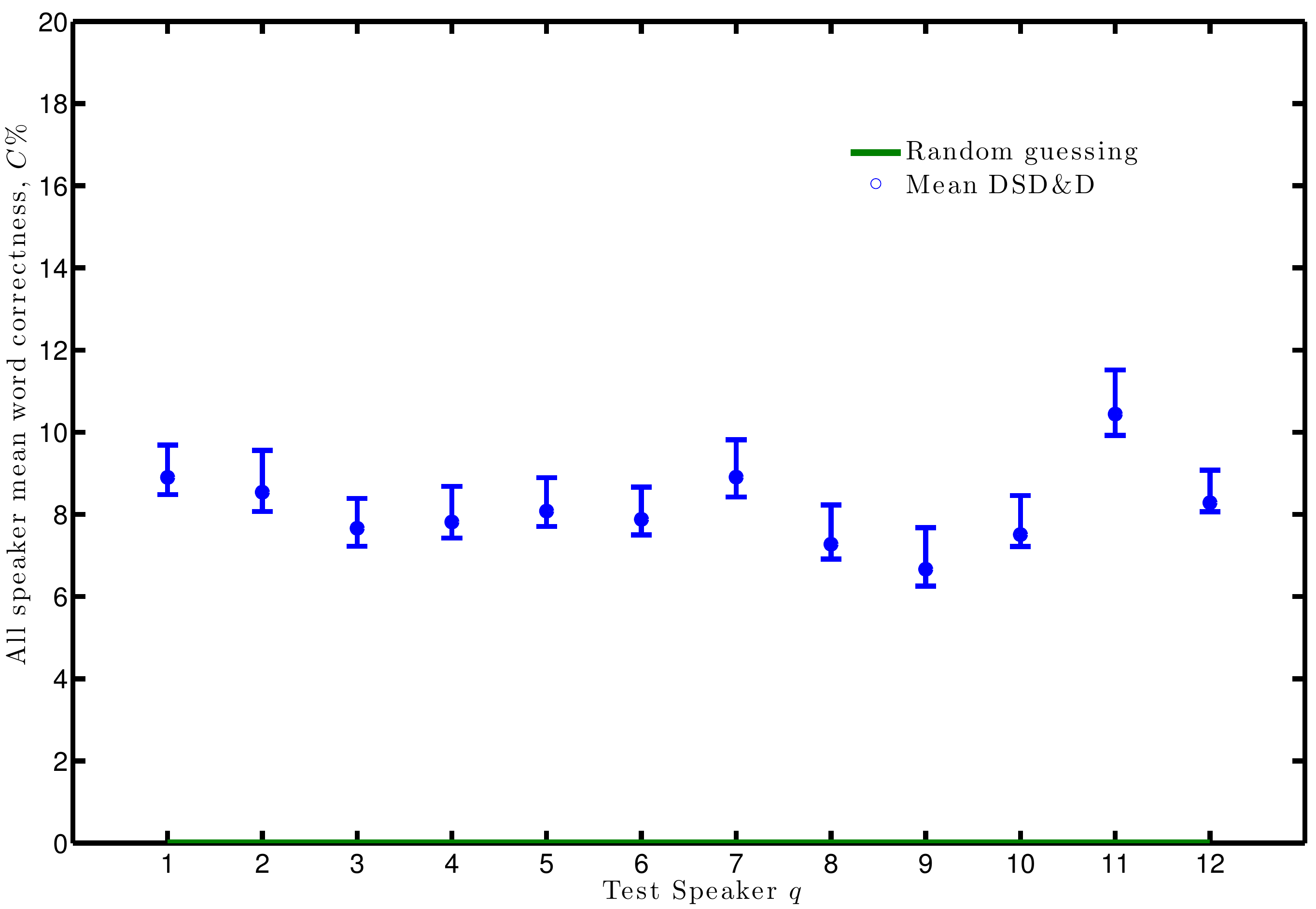} 
\caption{All-speaker mean word classification correctness, $C$, of the DSD classifiers constructed with single-speaker dependent phoneme-to-viseme maps for twelve speakers in RMAV and tested on others. Baseline is SSD maps (red) and error bars show $\pm1\frac{\sigma}{\sqrt{10}}$.} 
\label{fig:correctnessagg} 
\end{figure} 
 
Figure~\ref{fig:correctnessagg} shows the mean word correctness of the DSD classifiers per speaker in RMAV. The $y$-axis shows the \% word correctness and the $x$-axis is a speaker per point. We have also plotted random guessing and one standard error over the ten folds. Speaker 11 is the best performing speaker irrespective of the P2V selected. All speakers have a similar standard error but a low mean within this bound. This suggests subject to speaker similarity, there is more possibility to improve classification correctness with another speakers visemes (if they include the original speakers visual cues) than to use weaker self-clustered visemes. 
 
The performance of each viseme set is ranked by speaker by weighting the effect of the DSD tests. Each map scores as in Table~\ref{tab:weighting}. If a map increases on SSD performance within error bar range this scores $+1$ or outside error bar range scores $+2$. If a map decreases classification on SSD performance, these values are negative. 
\begin{table} [h]
\centering 
\caption{Weighted ranking scores from comparing the use of speaker-dependent maps for \emph{other} speaker lip-reading in isolated word speech (AVLetters2 speakers).} 
\begin{tabular}{|l|r|r|r|r|} 
\hline 
& $M_1$ & $M_2$ & $M_3$ & $M_4$ \\ 
\hline \hline 
Sp01 & $0$ 		& $+1$		& $+2$		& $+2$ \\ 
Sp02 & $-1$		& $0$		& $+2$		& $+1$ \\ 
Sp03 & $-2$		& $-2$		& $0$		& $-1$ \\ 
Sp04 & $-1$		& $+1$		& $-1$		& $0$ \\ 
\hline
Total	 & $-4$		& $0$		& $\textbf{+3}$		& $\textbf{+2}$ \\ 
\hline 
\end{tabular} 
\label{tab:weighting} 
\end{table} 

Therefore these values show $M_3$ is the best of the four AVL2 SSD maps, followed by $M_4$, $M_2$ and finally $M_1$ is the most susceptible to speaker identity in AVL2. Note this order matches a decreasing order of quantity of visemes in the speaker-dependent viseme sets i.e. the more similar to phoneme classes visemes are, then the better the classification performance. This ties in with Table~\ref{tab:homophones}, where the larger P2V maps create less homophones. 

In Table~\ref{tab:td_v}, which lists our AVL2 speaker-dependent P2V maps, the phoneme pairs \{/\textschwa/, $/eh/$\}, \{$/m/$, $/n/$\} and \{$/ey/$, $/iy/$\} are present for three speakers and \{/\textturnv/, $/iy/$\} and \{$/l/$, $/m/$\} are pairs for two speakers. Of the single-phoneme visemes, \{/t\textipa{S}/\} is present three times, \{$/f/$\}, \{$/k/$\}, \{$/w/$\} and \{$/z/$\} twice. The lesson from Figure~\ref{fig:correctnessDSD}, is the selection of incorrect units, whilst detrimental, is not as devastating as training classification classes on alternative speakers. 

\begin{table} [h]
\centering 
\caption{Weighted scores from comparing the use of speaker-dependent maps for \emph{other} speaker lip-reading in continuous speech (RMAV speakers).} 
\begin{tabular}{|l|r|r|r|r|r|r|r|r|r|r|r|r|} 
\hline 
& $M_1$ & $M_2$ & $M_3$ & $M_4$ & $M_5$ & $M_6$ & $M_7$ & $M_8$ & $M_9$ & $M_{10}$ & $M_{11}$ & $M_{12}$ \\ 
\hline \hline 
Sp01 & $0$ 		& $-1$		& $-2$		& $-2$		& $+1$		& $-1$		& $-1$		& $-1$		& $+1$		& $+1$		& $-1$		& $+1$ \\ 
Sp02 & $+2$ 		& $0$		& $+1$		& $+1$		& $+2$		& $+2$		& $+1$		& $+1$		& $+2$		& $+2$		& $+1$		& $+2$ \\ 
Sp03 & $-2$ 		& $-2$		& $0$		& $-2$		& $+1$		& $-1$		& $-1$		& $-2$		& $-2$		& $-2$		& $-2$		& $+1$ \\ 
Sp04 & $-2$ 		& $-1$		& $-1$		& $0$		& $+1$		& $+1$		& $-2$		& $-2$		& $+1$		& $-1$		& $-2$		& $+1$ \\ 
Sp05 & $-2$ 		& $-1$		& $+2$		& $-2$		& $0$		& $+1$		& $-1$		& $+2$		& $+1$		& $+2$		& $-1$		& $+2$ \\ 
Sp06 & $-1$ 		& $-1$		& $-1$		& $+1$		& $+2$		& $0$		& $+2$		& $-1$		& $-1$		& $+1$		& $+1$		& $+2$ \\ 
Sp07 & $+1$ 		& $-1$		& $-1$		& $+1$		& $+1$		& $+1$		& $0$		& $+1$		& $-1$		& $-1$		& $+1$		& $+1$ \\ 
Sp08 & $-1$ 		& $-1$		& $+1$		& $-1$		& $-1$		& $-2$		& $-2$		& $0$		& $+1$		& $+2$		& $+1$		& $+1$ \\ 
Sp09 & $-2$ 		& $-2$		& $-1$		& $-2$		& $-1$		& $-1$		& $-1$		& $-2$		& $0$		& $-1$		& $-2$		& $+1$ \\ 
Sp10 & $-2$ 		& $-2$		& $-1$		& $-1$		& $-1$		& $-2$		& $-2$		& $-2$		& $-2$		& $0$		& $-2$		& $-2$ \\ 
Sp11 & $-1$ 		& $+1$		& $-1$		& $+1$		& $+1$		& $-1$		& $+1$		& $-1$		& $-1$		& $+2$		& $0$		& $+2$ \\ 
Sp12 & $-1$ 		& $-2$		& $-2$		& $-1$		& $-1$		& $-2$		& $-2$		& $-2$		& $-2$		& $-1$		& $-2$		& $0$ \\ 
\hline
Total	 & $-9$		& $-11$		& $-6$		& $-7$ 		& $\textbf{+3}$ 		& $-5$ 		& $-8$ 		& $-9$ 		& $-3$ 		& $-4$ 		& $-8$ 		& $\textbf{+12}$ \\ 
\hline 
\end{tabular} 
\label{tab:weighting_lilir} 
\end{table} 
%\clearpage

The same measure has been listed in Table~\ref{tab:weighting_lilir} for our 12 RMAV speakers. The key observation in this table is Speaker 12 on the far right column. The speaker dependent map of Speaker 12 is one of only two ($M_{12}$ and $M_5$) which make an overall improvement on other speakers classification (they have positive values in the total row at the bottom of Table~\ref{tab:weighting_lilir}), and crucially, $M_{12}$ only has one speaker (Speaker 10) for whom the visemes in $M_{12}$ does not make an improvement in classification. The one other speaker P2V map which improves over other speakers is $M_5$. All others show a negative effect, this reinforces our assertion visual speech is dependent upon the individual but we also now have evidence there are exceptions to the rule. In order the RMAV P2Vs are:   
\begin{multicols}{2} 
\begin{enumerate}
 \setlength\itemsep{0.001em}
\item $M_{12}$
\item $M_5$
\item $M_9$
\item $M_{10}$
\item $M_6$
\item $M_3$
\item $M_4$
\item $M_7$ and $M_{11}$
\item $M_1$§ and $M_8$
\item $M_2$
\end{enumerate}
\end{multicols}
 
%Here Speaker 4 visemes, $M_4$, are most suited for Speakers 1, 2 \& 4 showing the highest word classification. Only Speaker 3s SD P2V map, $M_3(3,3)$ ever out performs Speaker 4's when trained on Speaker 3. Interestingly, all three other-speaker P2V maps improve the performance of Speaker 1, $M_1$. Speakers 3 \& 4 both suffer a reduction in word classification using other-speaker P2V maps so overall we say $M_3$ \& $M_4$ have the most useful visemes. 

\begin{figure}[h] 
\centering 
\includegraphics[width=0.8\linewidth]{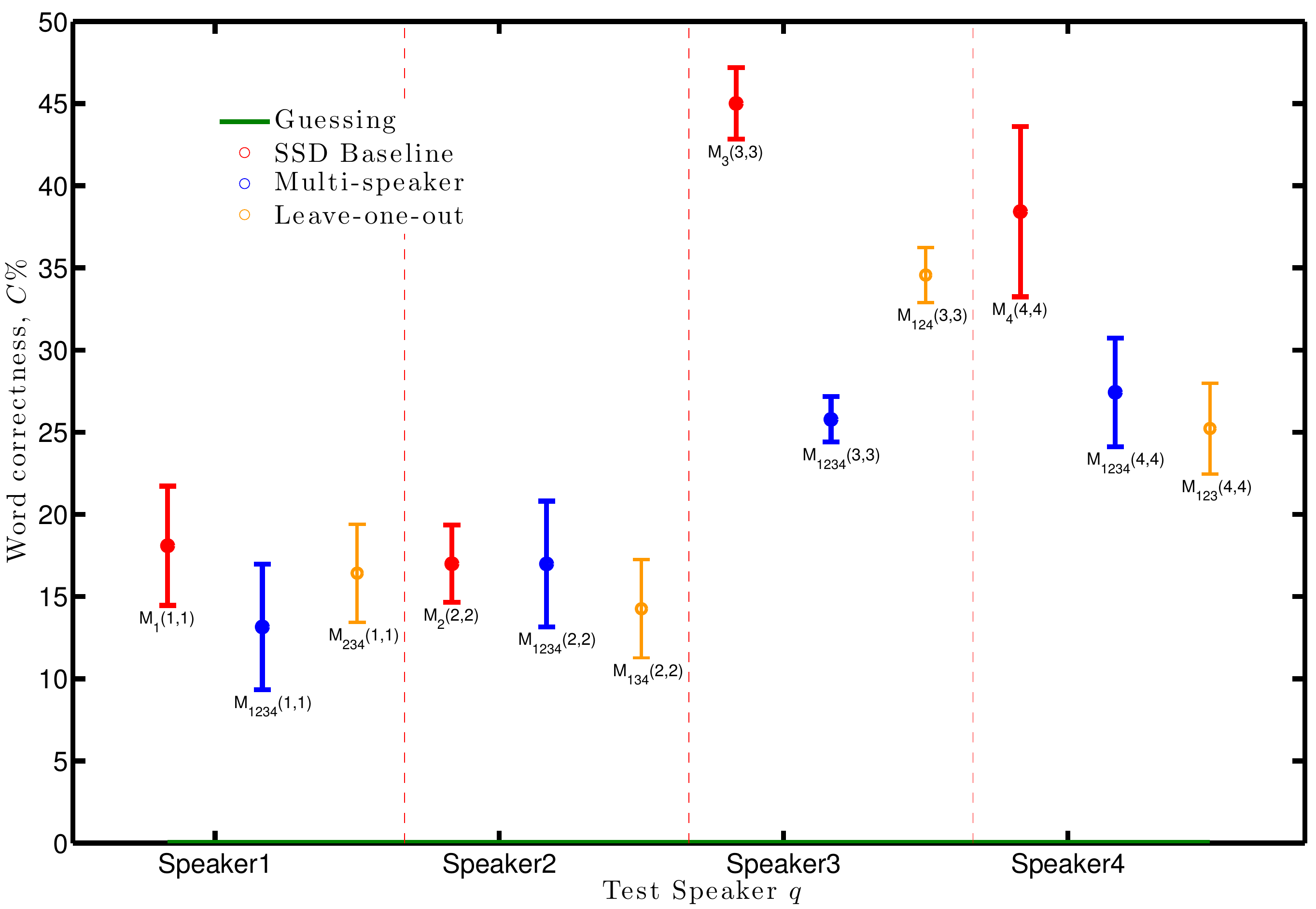} 
\caption{Word classification correctness, $C\pm1\frac{\sigma}{\sqrt{7}}$, of the classifiers using MS and SI phoneme-to-viseme maps on AVLetters2 speakers. Baseline is SSD maps (red).} 
\label{fig:accuracy} 
\end{figure} 
 
Figure~\ref{fig:accuracy} shows the correctness of both the MS viseme class set and the SI tests (Tables~\ref{tab:ms} and~\ref{tab:sim}) against our SSD baseline for AVL2 speakers. Word correctness, $C$ is plotted on the $y$-axis. For the multi-speaker classifiers, these are all built on the same map $M_{all}$, and tested on the same speaker so, $p=q$. Therefore the tests are: $M_{all}(1,1)$, $M_{all}(2,2)$, $M_{all}(3,3)$, $M_{all}(4,4)$. To test the SI maps, we plot  $M_{!1}(1,1)$, $M_{!2}(2,2)$, $M_{!3}(3,3)$ and $M_{!4}(4,4)$. Again the same baseline is repeated where $n=p=q$ for reference.

There is no significant difference on Speaker 2, and while Speaker 3 word classification is reduced, it is not eradicated. It is interesting for Speaker 3, for whom their speaker-dependent classification was the best of all speakers, the SI map ($M_{!3}$) out performs the multi-speaker viseme classes ($M_{all}$) significantly. This maybe due to Speaker 3 having a unique visual talking style which reduces similarities with Speakers 1, 2 \& 4. But more likely, we see the $/iy/$, phoneme is not classified into a viseme in $M_3$, whereas it is in $M_1$, $M_2$ \& $M_4$ and so re-appears in $M_{all}$. Phoneme $/iy/$ is the most common phoneme in the AVL2 data. This suggests it may be best to avoid high volume phonemes for speaker-dependent visemes as we are trying to maximise on the speaker individuality to make better viseme classes. 

\begin{figure}[h] 
\centering 
\includegraphics[width=0.8\linewidth]{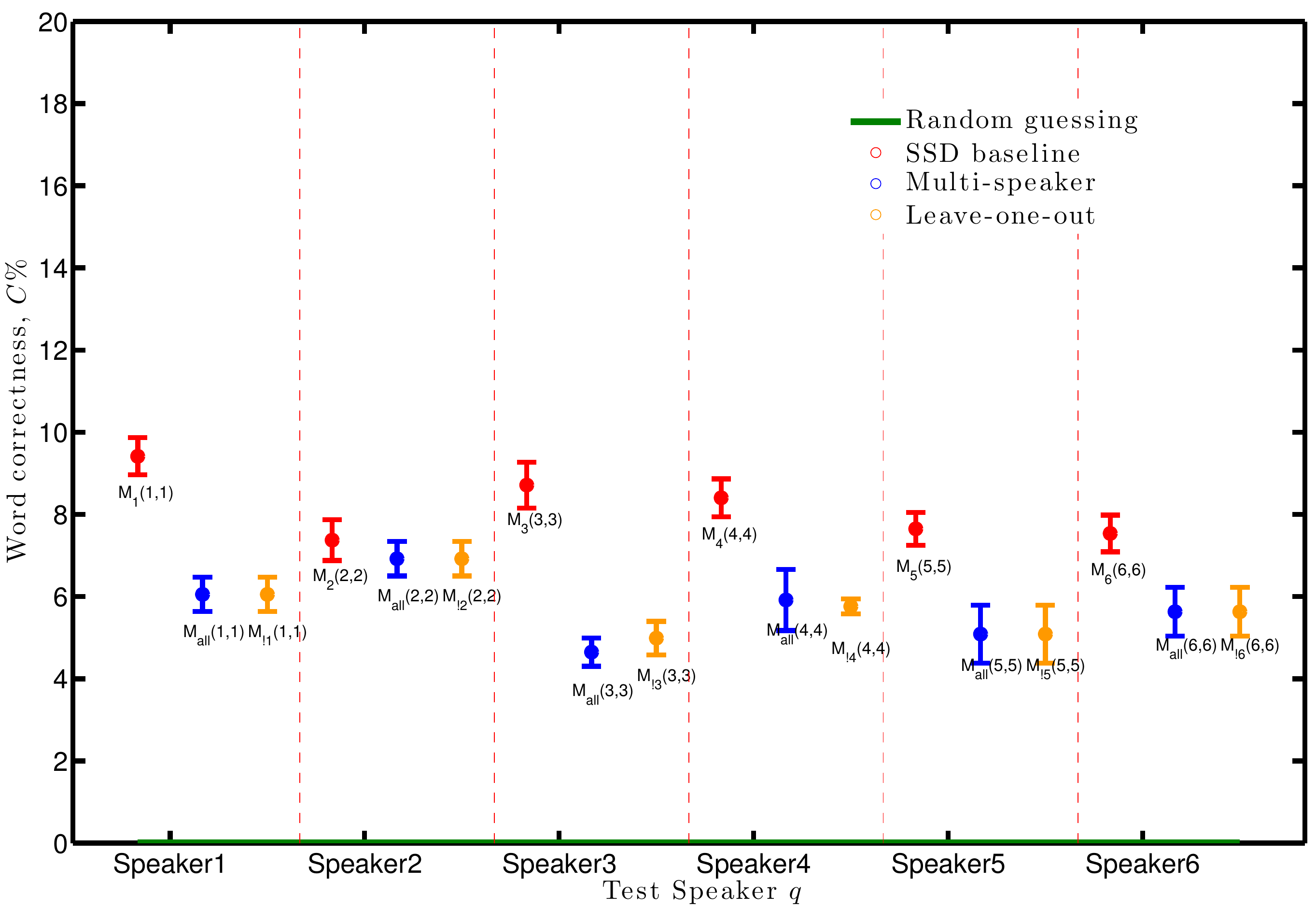} 
\caption{Mean word correctness, $C\pm1\frac{\sigma}{\sqrt{10}}$, of the classifiers using MS and SI phoneme-to-viseme maps on RMAV speakers. Baseline is SSD maps (red) - Speakers 1-6.} 
\label{fig:accuracy1} 
\end{figure} 
\begin{figure}[h] 
\centering 
\includegraphics[width=0.8\linewidth]{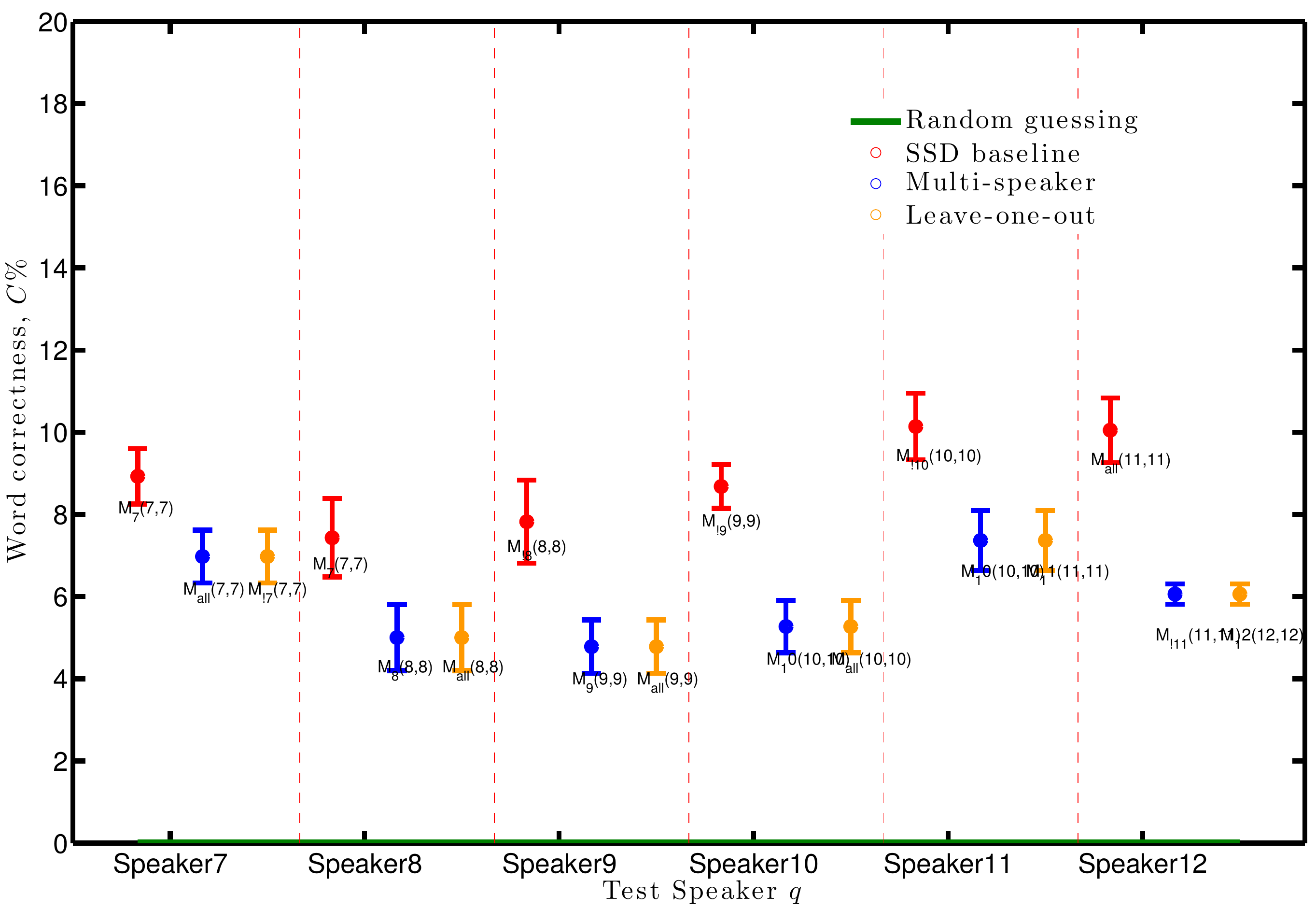} 
\caption{Mean word correctness, $C\pm1\frac{\sigma}{\sqrt{10}}$, of the classifiers using MS and SI phoneme-to-viseme maps on RMAV speakers. Baseline is SSD maps (red) - Speakers 7-12.} 
\label{fig:accuracy2} 
\end{figure} 
 
We have plotted the same MS \& SI experiments on RMAV speakers in Figures~\ref{fig:accuracy1} and~\ref{fig:accuracy2} (six speakers in each figure). In continuous speech, all bar Speaker 2 are significantly negatively affected by using generalised multi-speaker visemes, whether the visemes include the test speakers phoneme confusions or not. This reminds us of the dependency on speaker identity in machine lip-reading but we do see the scale of this effect depends on which two speakers are being compared. For our exception speaker (Speaker 2 in Figure~\ref{fig:accuracy1}) there is only a insignificant decrease in correctness when using MS and SI visemes. Therefore it could be possible with making multi-speaker visemes based upon groupings of visually similar speakers, even better visemes could be created. The challenge remains in knowing which speakers should be grouped together before undertaking P2V map derivation. 
 
\section{Speaker independence between sets of visemes} 
 
For isolated word classification our main conclusion of this chapter is shown by comparing Figures~\ref{fig:correctnessDSD} \&~\ref{fig:accuracy} with Figure~\ref{fig:indep_Corr}. The reduction in performance in Figure~\ref{fig:indep_Corr} is when the system classification models are trained on a speaker who is not the test speaker. This raised the question if this degradation was due to the wrong choice of P2V map or speaker identity mismatch between the training and test data samples. We have concluded that, whilst the wrong unit labels are not conducive for good lip-reading, is it not the choice of phoneme-to-viseme map which causes significant degradation to accurate classification, but rather the speaker identity. This regain of performance is irrespective of whether the map is chosen for a different speaker, multi-speaker or independently of the speaker. 
 
This observation is important as it tells us the repertoire of visual units across speakers does not vary significantly. This is comforting since the prospect of classification using a symbol alphabet which varies by speaker is daunting. This is further reinforced by Tables~\ref{tab:td_v},~\ref{tab:mt_v} \&~\ref{tab:l1o_v}. There are differences between speakers, but not significant ones. However, we have seen some exceptions within our continuous speech speakers whereby the effect of the P2V map selection is more prominent and where sharing HMMs trained on non-test speakers has not been completely detrimental. This gives some hope with similar visual speakers, and with more `good' training data speaker independence, whether by classifier or viseme selection, might be possible. 
 
To provide an analogy; in acoustic speech we could ask if an accented Norfolk speaker requires a different set of phonemes to a standard British talker? The answer is no. They are represented by the same set of phonemes; but due to their individuality they use these phonemes in a different way. 
 
Comparing our multi-speaker and SI maps, there are 11-12 visemes per set whereas in our single-speaker-dependent maps we have a range of 12 to 17. It is $M_3$ with 17 visemes, which out performs all other P2V maps. So we can conclude, there is a high risk of over-generalising a speaker-dependent P2V map when attempting multi-speaker or speaker-independent phoneme-to-viseme mappings. This is something we have seen with our RMAV experiments. 
 
Therefore we must consider it is not just the speaker-dependency which varies but also the contribution of each viseme within the set which also contributes to the word classification performance, an idea first shown in~\cite{bear2014some}. Here we have highlighted some phonemes which are a good subset of potentially independent visemes \{/\textschwa/, $/eh/$\}, \{$/m/$, $/n/$\} and \{$/ey/$, $/iy/$\}, and what these results present, is a combination of certain phoneme groups combined with some speaker-dependent visemes, where the latter provide a lower contribution to the overall classification would improve speaker-independent maps with speaker-dependent visual classifiers. 

We compare our speaker independent results to the AAM results of Neti \textit{et al.} \cite{neti2000audio} and we see our results are inferior overall. Neti \textit{et al.} achieved a $w.e.r$ of 64\% compared to our accuracy of around 5\% (with AVL2) and between 6-10\% with RMAV. We attribute this to the training data volumes in each dataset. The IBM via voice dataset \cite{matthews2001comparison} used in \cite{neti2000audio} is not publicly available but as it has 290 speakers and 10,500 word vocabulary, compared to RMAV which has 12 speakers and ~1000 words per speaker. 
 
It is often said in machine lip-reading there is high variability between speakers. This should now be clarified to state there is not a high variability of visual cues given a language, but there is high variability in trajectory between visual cues of an individual speakers with the same ground truth.  In continuous speech we have seen how not just speaker identity affects the visemes (phoneme clusters) but also how the robustness of each speakers classification varies in response to changes in this. This implies a dependency upon the number of visemes within each set for individuals so this is what we investigate in the next chapter.

 % AVSP - AVL2, talker dependent maps
%!TEX root = main.tex 
\chapter[Finding phonemes]{Finding phonemes} 
\label{chap:eight} 
 
Due to the many-to-one relationship in traditional mappings of phonemes to visemes, any resulting set of visemes will always be smaller than the set of phonemes. We know a benefit of this is more training samples per class which compensates for the limited data in currently available datasets but the disadvantage is generalisation between different articulated sounds. To find an optimal set of viseme classes, we need to minimise the generalisation to maintain good classification but also to maximise the training data available. 

In Chapter~\ref{chap:maps} we have shown how P2V maps can be derived automatically from phoneme confusions. A by-product of clustering phonemes from classification data is the option to control how many visemes a set contains within the phoneme clustering algorithm. This allows precision when answering questions about the optimal number of visemes. We ask how many visemes is the optimum number? And does this optimum vary by speaker in visual speech? 
  
\begin{figure}[t] 
\centering 
\includegraphics[width=\linewidth]{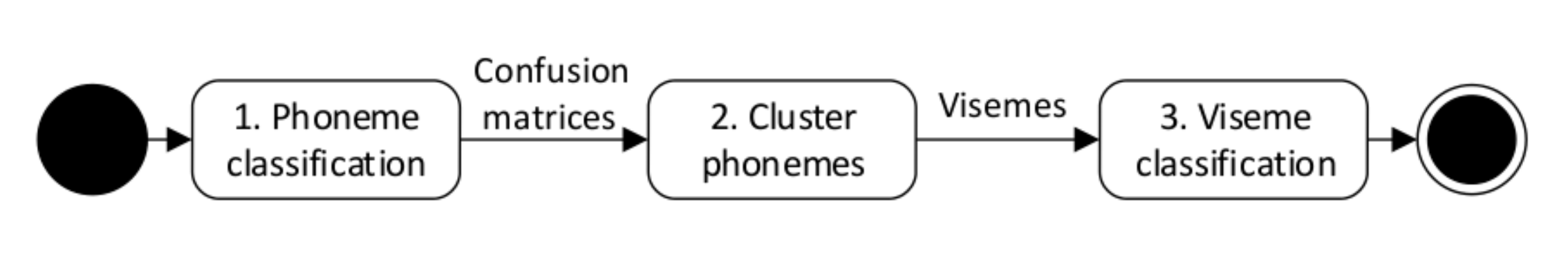} 
\caption{Three-step high-level process for viseme classification where the visemes are derived from phoneme confusions.} 
\label{fig:process} 
\end{figure}

For this work we use the RMAV dataset \cite{improveVis} and BEEP pronunciation dictionary \cite{beep}. Figure~\ref{fig:process} shows a high level overview of the experiment. It begins by performing classification using phoneme-labelled classifiers. This provides a set of speaker-dependent confusion matrices which are used to cluster together single phonemes (monophones) into subgroups, or as we call them, visemes. 
 
This time around, we adopt a different phoneme clustering process (described in subsection~\ref{sec:two}). By this process, a new P2V mapping is derived for every time a pair of classes is re-classified in to a new class grouping. There are a maximum of 45 phonemes in the phonetic transcript of the RMAV speakers. This means we can create up to 45 P2V maps per speaker. The actual number of maps produced is subject to the number of phonemes matched during the phoneme classification (step 1 of Figure~\ref{fig:process}). This first step produces the phoneme confusion matrices from which we create new phoneme clusters into visemes. If a phoneme has not been classified,  either incorrectly or correctly, then it is not included in the resulting confusion matrix from which our visemes are created. Thus, we now have up to 45 sets of viseme labels to use for labelling our HMMs when repeating the word classification task. 
 
We continue with analysing the word classification rather than visemes as we do not wish our results to be affected by the variance in training samples for each set of classifiers. It is not the performance itself which is relevant here, rather it is any improvement a variance in classes can provide. It is important the reader remember the presentation of this new method is \textit{not} a suggestion this particular clustering algorithm will deliver the optimum visemes, but rather address the need in this case for a method to enable a controlled comparison of the phoneme to viseme distributions as the number of classes reduces. 
 
\section{Step One: phoneme classification} 
\label{sec:one} 
 
Step 1 implements 10-fold cross-validation with replacement \cite{efron1983leisurely}, of 200 sentences per speaker, 20 are randomly selected as test samples and are not included in the training folds. Using the HTK toolkit \cite{htk34} to implement HMM classifiers, the HMMs are initialised by the flat-start method, and re-estimated 11 times with forced alignment between seventh and eighth estimates. The prototype HMM is based upon a Gaussian mixture of five components and three state HMMs. Included is a single-state tied short-pause, or `sp' HMM for short silences between words in the sentence utterances. A bigram word network is used to support classification. %There are a maximum of 46 phonemes within our phoneme classification results, but not all speakers used all phonemes within their speech utterances. 
 
\section{Step Two: phoneme clustering} 
\label{sec:two} 
 
The phonemes are clustered into new viseme classes for each speaker as follows; step 1 produces ten confusion matrices for each speaker (one from each fold), these are summed together to form one confusion matrix representing all confusions for that speaker. Clustering begins with this phoneme confusion matrix: 
 
\begin{equation} 
[K_{m}]_{ij} = N (\phat{p}_j | p_i)\quad 
\label{eq3} 
\end{equation} 
 
where the $ij^{th}$ element is the count of the number of times phoneme $i$ is classified as phoneme $j$. This algorithm works with the column normalised version, 
 
\begin{equation} 
[P_m]_{ij} = Pr\{p_i | \phat{p}_j \} \quad 
\label{eq4} 
\end{equation} 
 
the probability that, given a classification of $p_j$ that the phoneme really was $p_i$. The subscript $m$ in $K_m$ and $P_m$ indicates $K_m$ and $P_m$ have $m^{2}$ elements ($m$ phonemes). Merging of phonemes is done by looking for the two most confused phonemes and hence create a new class with confusions $K_{m-1}, P_{m-1}$. 
 
Specifically for each possible merged pair, $Pr,Ps$ score is calculated by: 
\begin{equation} 
q = [P_{m}]_{rs} + [P_{m}]_{sr} \quad \\ 
= Pr\{\phat{P}r | Ps \} + Pr\{\phat{P}s | Pr\} 
\label{eq5} 
\end{equation} 
 
Phonemes are assigned to one of two classes, $V \& C$, vowels and consonants. Vowels and consonants can not be mixed. The pair with the highest $q$ is merged. Equal scores are broken randomly. This process is repeated until $m = 2$. Each intermediate step, $M = 45,44,43 ... 2$ forms a possible set of visual units. 
 
This is a more controlled approach than the method used in Chapter~\ref{chap:maps} and \cite{bear2014phoneme}, and incorporates our conclusions vowel and consonant phonemes should not be clustered together when devising phoneme-to-viseme mappings. An example P2V mapping is shown in Table~\ref{tab:example}. 
 
\begin{table} 
\centering 
\caption{An example phoneme-to-viseme map, this is the phoneme-to-viseme map for RMAV Speaker 1 with ten visemes.} 
\begin{tabular}{|l|l|} 
\hline 
Viseme & Phonemes \\ 
\hline \hline 
$/v01/$ & /ax/ \\ 
$/v02/$ & /v/ \\ 
$/v03/$ & /\textopeno\textsci/ \\ 
$/v04/$ & /f/ /\textipa{Z}/ /w/ \\ 
$/v05/$ & /k/ /b/ /d/ /\textipa{T}/ /p/ \\ 
$/v06/$ & /l/ /d\textipa{Z}/ \\ 
$/v07/$ & /g/ /m/ /z/ /y/ /t\textipa{S}/ /\textipa{D}/ /s/ /r/ /t/ /\textipa{S}/ \\ 
$/v08/$ & /n/ /hh/ /\textipa{N}/ \\ 
$/v09/$ & /\textipa{E}/ /ae/ /\textopeno/ /uw/ /\textturnscripta/ /\textsci\textschwa/ /ey/ /ua/ /\textrevepsilon/ \\ 
$/v10/$ & /ay/ /\textscripta/ /\textturnv/ /\textscripta\textupsilon/ /\textupsilon/ /\textschwa\textupsilon/ /\textsci/ /iy/ /\textschwa/ /eh/ \\ 
\hline 
\end{tabular} 
\label{tab:example} 
\end{table} 
 
\section{Step Three: viseme classification} 
\label{sec:visRecog} 
 
Similar to step 1, step 3 involves implementation of 10-fold cross-validation with replacement \cite{efron1983leisurely}, of 200 sentences per speaker, 20 are randomly selected as test samples and these are not included in the training folds. Using the HTK toolkit \cite{htk34} to use Hidden Markov Model (HMM) classes, viseme labelled HMMs are flat-started, re-estimated 11 times over with forced alignment between seventh and eighth estimates. The same HMM prototype is used and a bigram word network supports classification along with the application of a grammar scale factor of $1.0$ (shown to be optimum in~\cite{howellPhD}) and a transition penalty of $0.5$. 
 
The important difference this time around are the viseme classes being used as classification labels. By using these sets of classes which have been shown in step 1 to be confusing on the lips, we now perform classification for each class set. In total this is 45 sets, where the smallest set is of two classes (one with all the vowel phonemes and the other all the consonant phonemes), and the largest set is of 45 classes with one phoneme in each - thus the largest set for each speaker is a repeat of the phoneme classification task but using only phonemes which were originally recognised (either correctly or incorrectly) in step 1. 
 
\section{Searching for an optimum} 
\label{sec:search}
 
\begin{figure} [!ht] 
\centering 
\includegraphics[width=0.95\textwidth]{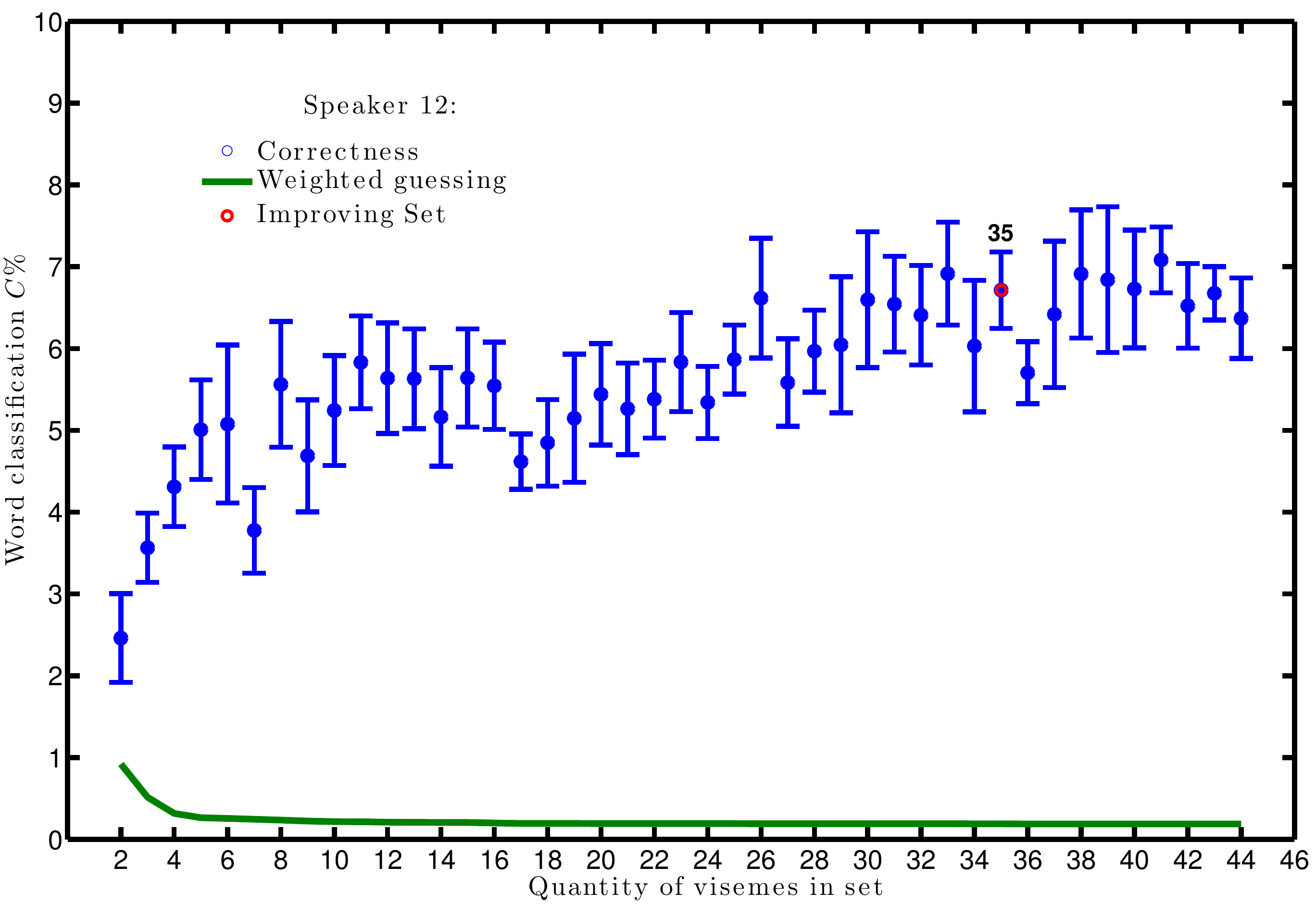} 
\caption{Speaker 12: word classification correctness, $C\pm1\frac{\sigma}{\sqrt{10}}$ for phoneme-to-viseme map sizes 2-44.} 
\label{fig:sp12} 
\end{figure}

In Figures~\ref{fig:sp01} - \ref{fig:sp12}, we show the word correctness, plotted on the $y$-axis for all 12 speakers. Each of the viseme sets, identified by the number of visemes within the set, are plotted in increasing order along the $x$-axis. We have also plotted, in green, guessing weighted by the visual homophones in the transcripts. This has been calculated by:
\begin{equation}
\sum_{i=1}^{i=N} (\frac{TC_i}{W}) * (\frac{1}{N})
\end{equation}
where $TC$ is the total individual token count for that speaker (for each token), $W$ is the total words for that speaker, and $N$ is the number of tokens. $i$ is for all each token where a token is a unique word.

Viseme sets containing fewer visemes produce viseme strings which represent more than one word: homophones. The effect of homophones can be seen on the left side of the graphs in Figures~\ref{fig:sp01} -~\ref{fig:sp12} with viseme sets with fewer than 11 visemes. 
  
  %An example of a homophone in the RMAV data are the words `port' and `bass'. Using Speaker 1's 10-viseme P2V map these both become `$/v5/$ $/v9/$ $/v7/$' i.e. a single identifier for identifying two distinct words. Thus distinguishing between `port' and `bass' is impossible.
  
\begin{figure} [h] 
\centering 
%\setlength{\tabcolsep}{1pt} 
%\begin{tabular} {c c} 
\includegraphics[width=0.95\textwidth]{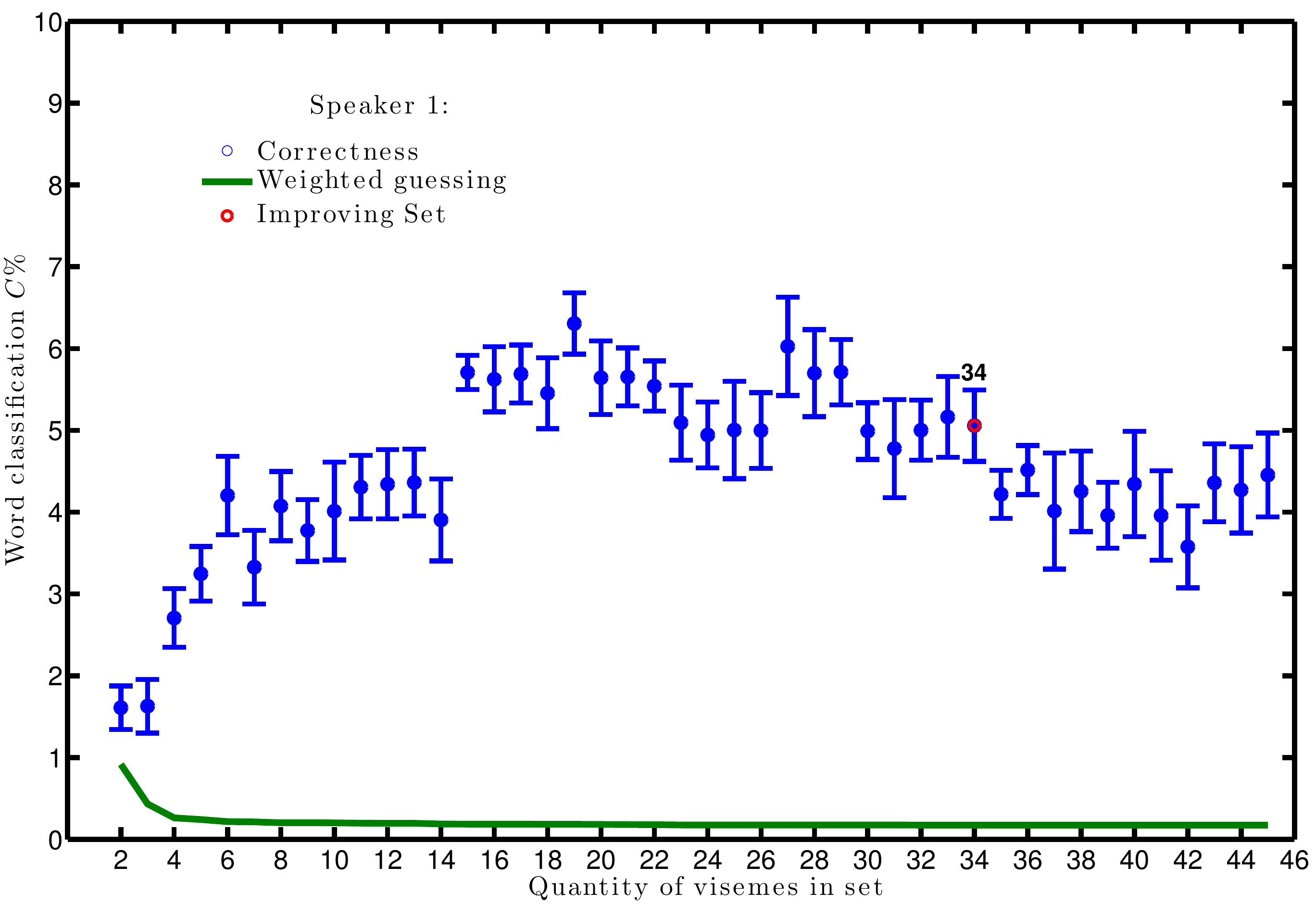} 
\caption{Speaker 1: word classification correctness, $C\pm1\frac{\sigma}{\sqrt{10}}$ for phoneme-to-viseme map sizes 2-45.} 
\label{fig:sp01} 
\end{figure} 
\begin{figure} [!ht] 
\centering 
\includegraphics[width=0.95\textwidth]{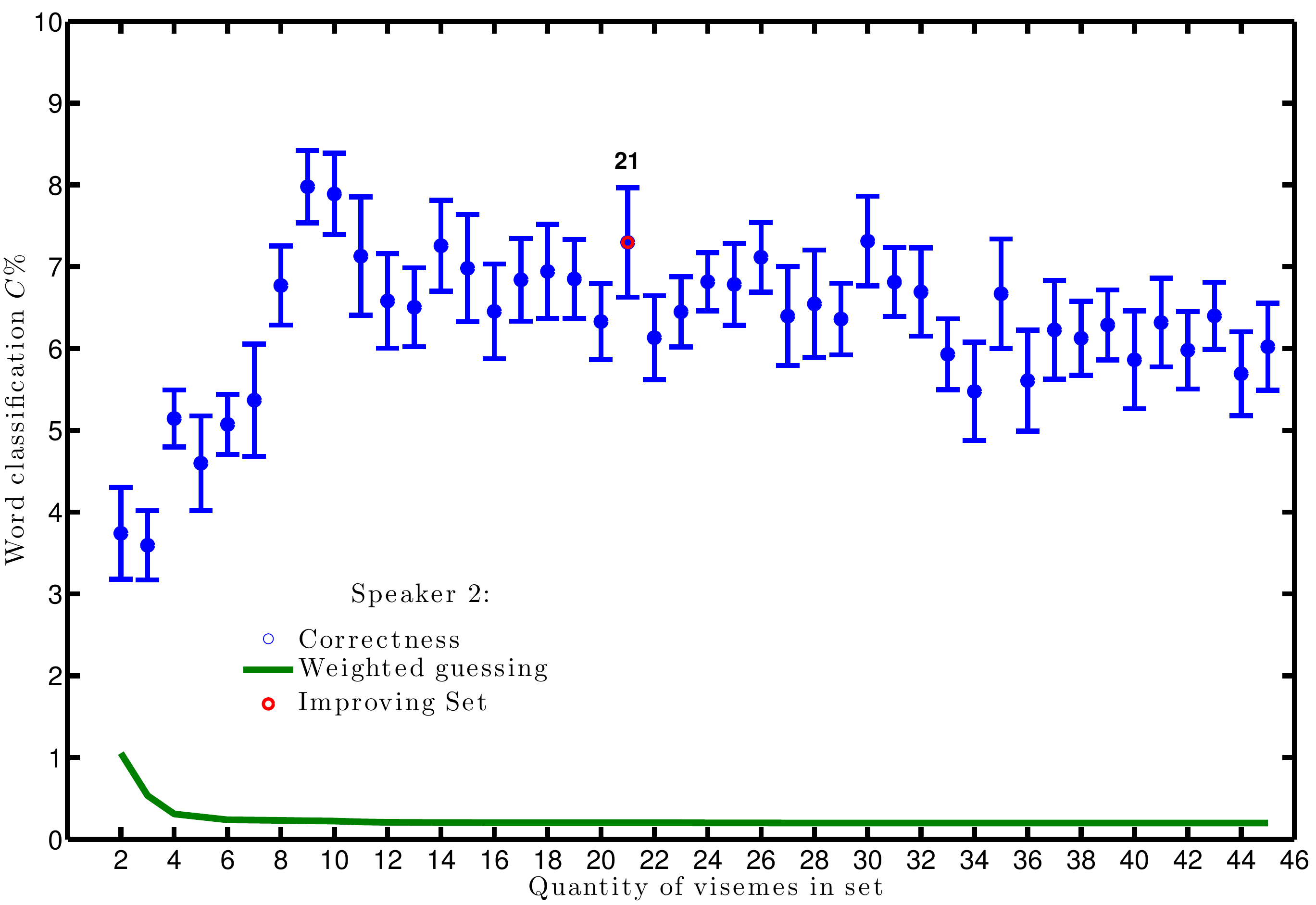} 
\caption{Speaker 2: word classification correctness, $C\pm1\frac{\sigma}{\sqrt{10}}$ for phoneme-to-viseme map sizes 2-45.} 
\label{fig:sp02} 
\end{figure} 
\begin{figure} [!ht] 
\centering 
\includegraphics[width=0.95\textwidth]{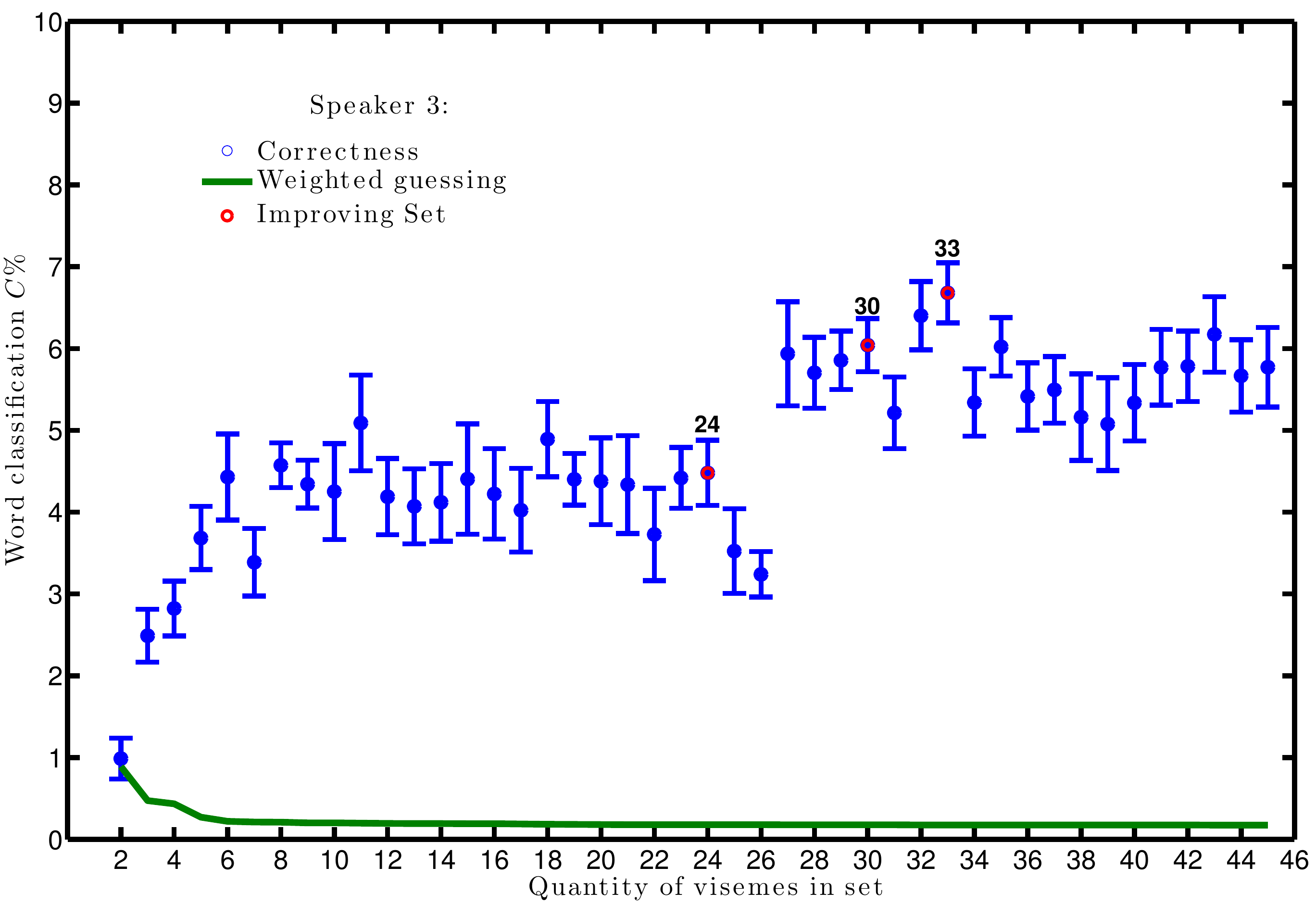} 
\caption{Speaker 3:  word classification correctness, $C\pm1\frac{\sigma}{\sqrt{10}}$ for phoneme-to-viseme map sizes 2-45.} 
\label{fig:sp03} 
\end{figure} 
\begin{figure} [!ht] 
\centering 
\includegraphics[width=0.95\textwidth]{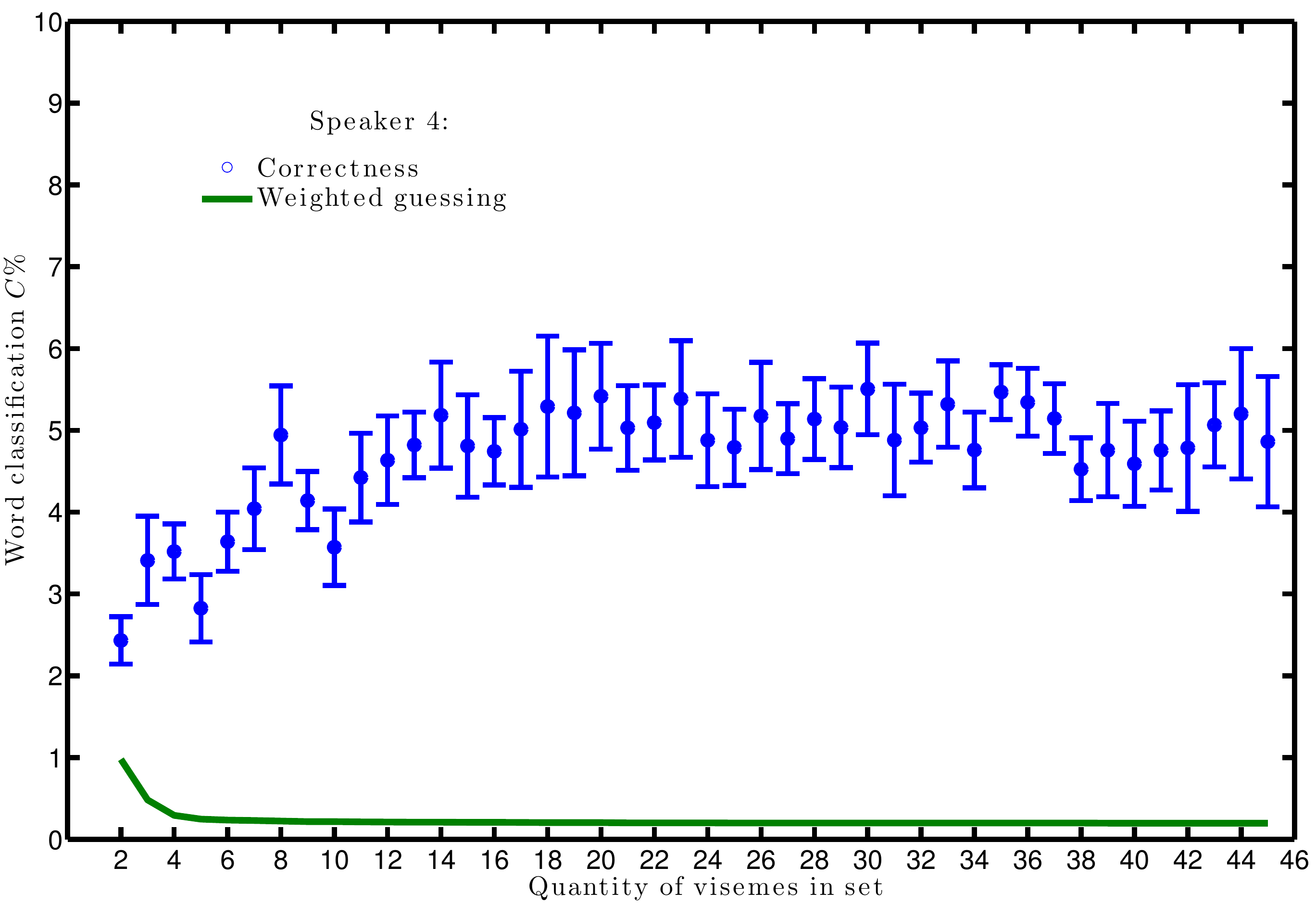} 
\caption{Speaker 4:  word classification correctness, $C\pm1\frac{\sigma}{\sqrt{10}}$ for phoneme-to-viseme map sizes 2-45.} 
\label{fig:sp04} 
\end{figure} 
\begin{figure} [!ht] 
\centering 
\includegraphics[width=0.95\textwidth]{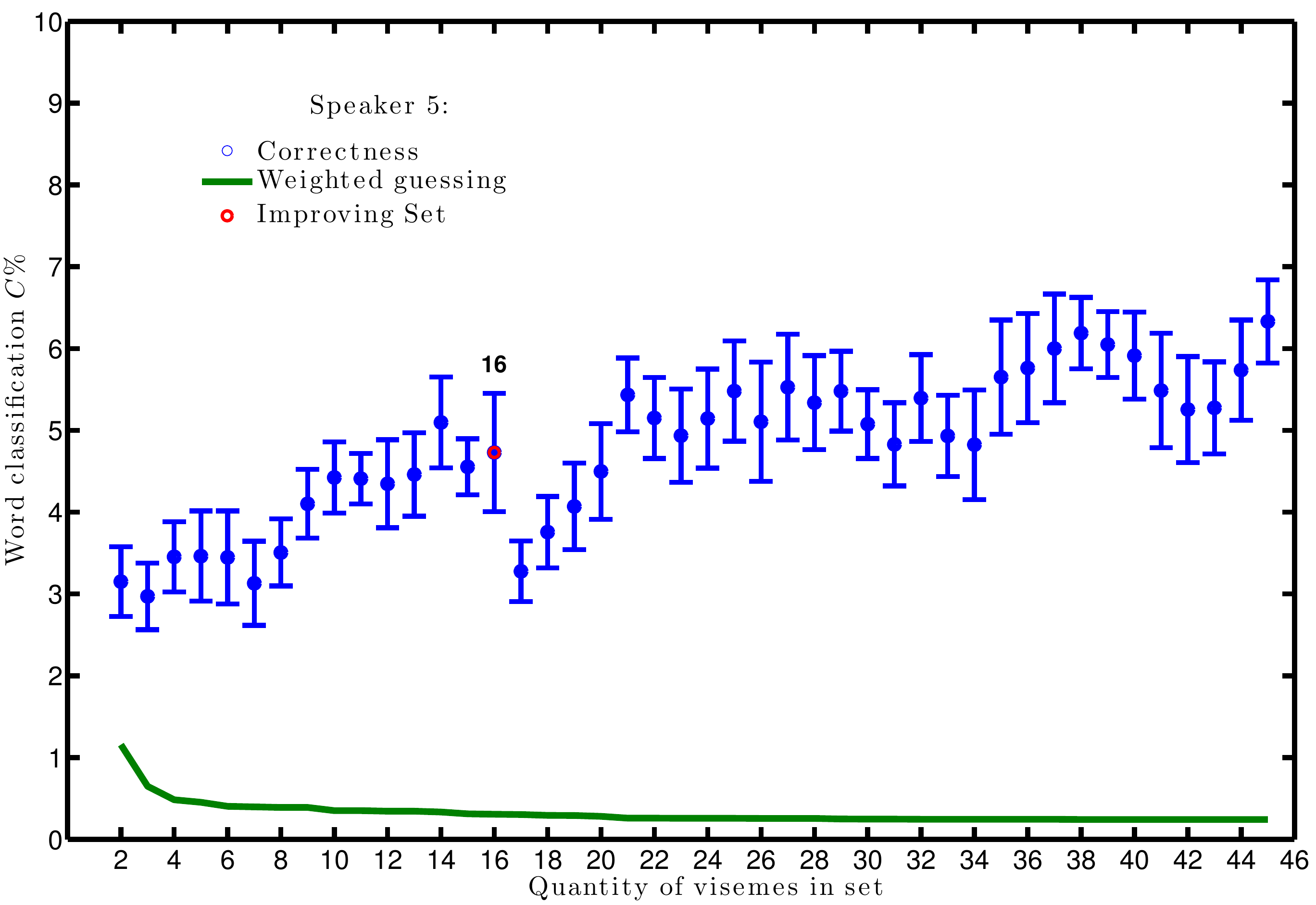} 
\caption{Speaker 5: word classification correctness, $C\pm1\frac{\sigma}{\sqrt{10}}$ for phoneme-to-viseme map sizes 2-45.} 
\label{fig:sp05} 
\end{figure} 
\begin{figure} [!ht] 
\centering 
\includegraphics[width=0.95\textwidth]{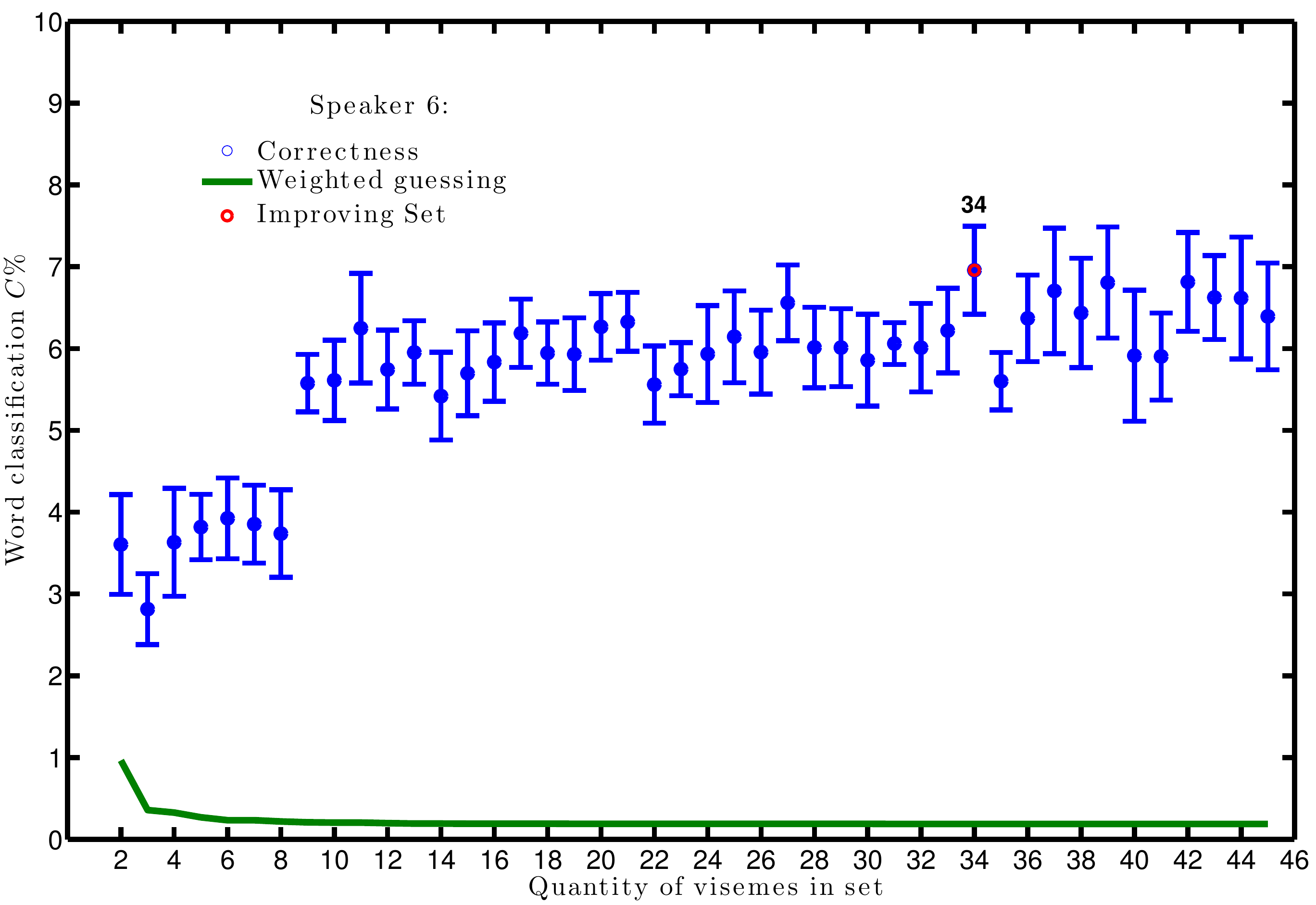} 
\caption{Speaker 6:  word classification correctness, $C\pm1\frac{\sigma}{\sqrt{10}}$ for phoneme-to-viseme map sizes 2-45.} 
\label{fig:sp06} 
\end{figure} 
\begin{figure} [!ht] 
\centering 
\includegraphics[width=0.95\textwidth]{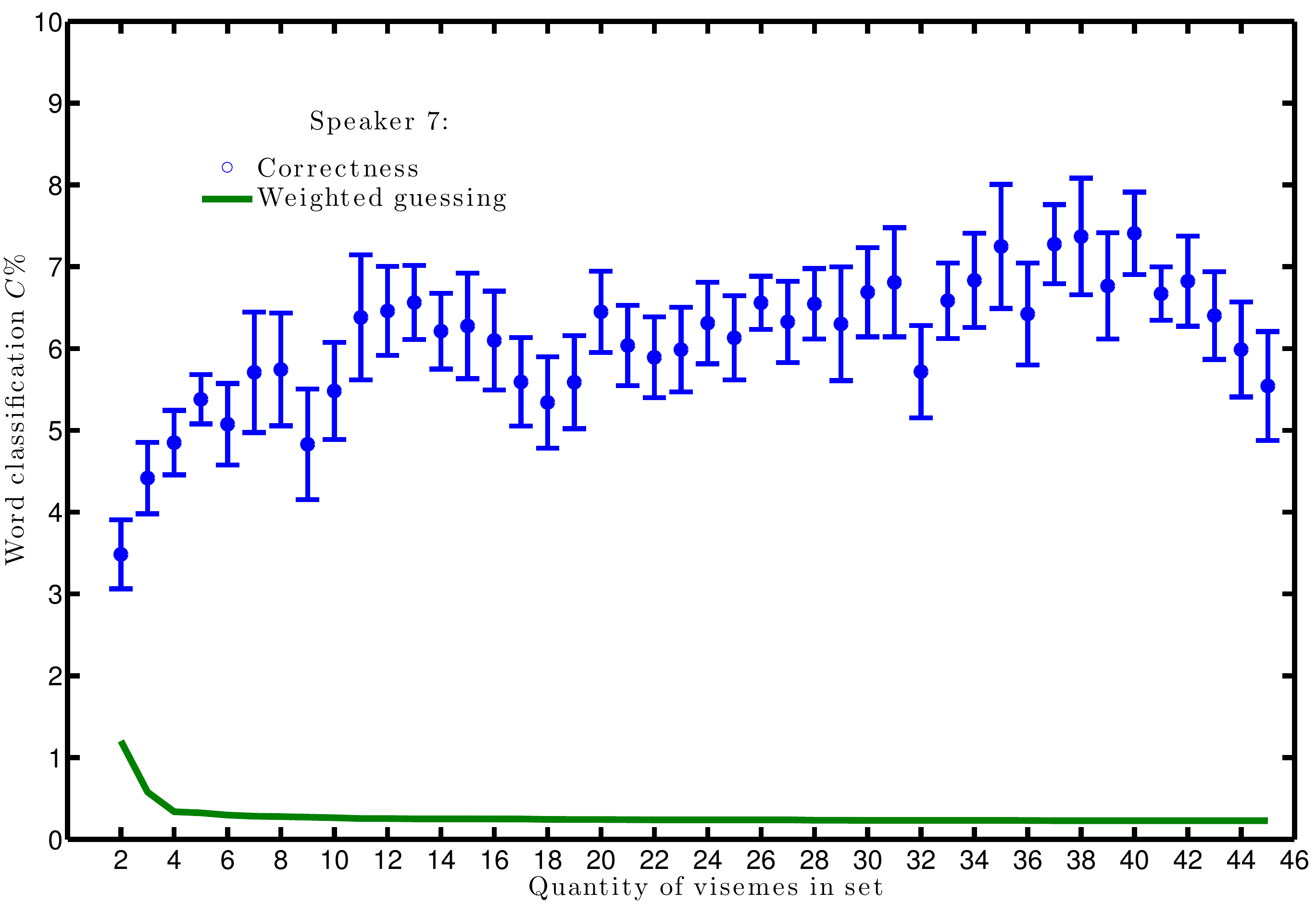} 
\caption{Speaker 7:  word classification correctness, $C\pm1\frac{\sigma}{\sqrt{10}}$ for phoneme-to-viseme map sizes 2-45.} 
\label{fig:sp07} 
\end{figure} 
\begin{figure} [!ht] 
\centering 
\includegraphics[width=0.95\textwidth]{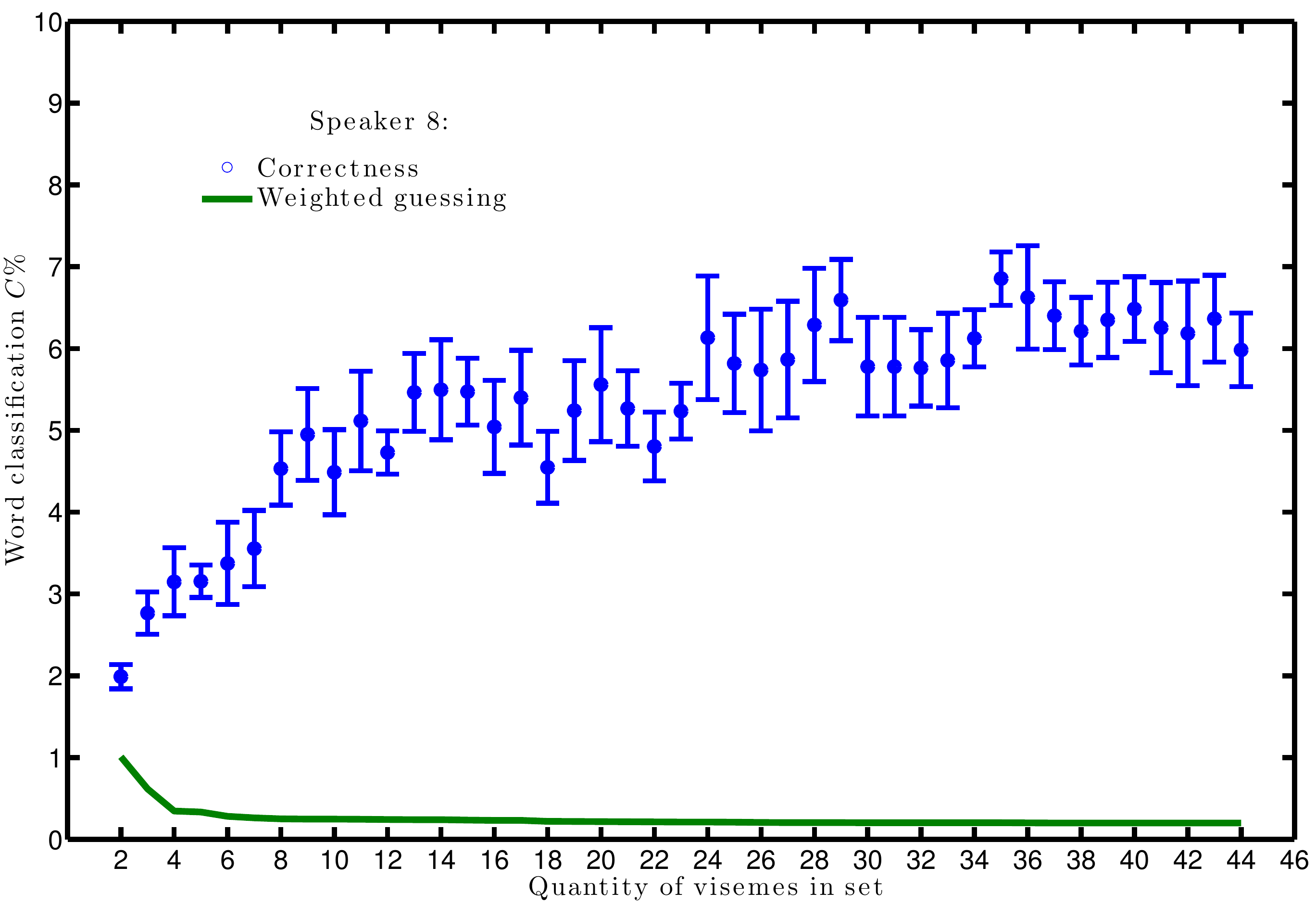} 
\caption{Speaker 8: word classification correctness, $C\pm1\frac{\sigma}{\sqrt{10}}$ for phoneme-to-viseme map sizes 2-44.} 
\label{fig:sp08} 
\end{figure} 
\begin{figure} [!ht] 
\centering 
\includegraphics[width=0.95\textwidth]{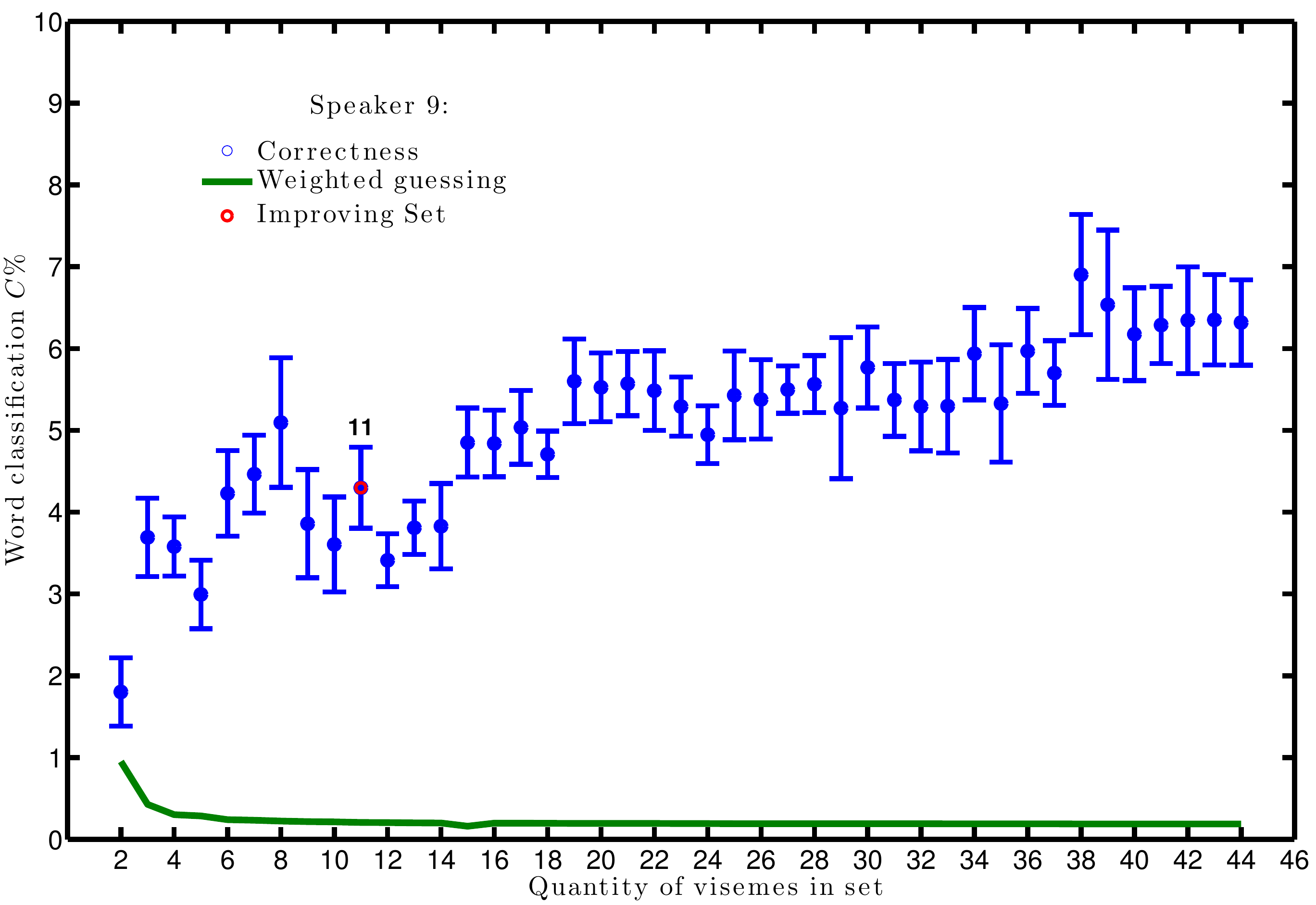} 
\caption{Speaker 9: word classification correctness, $C\pm1\frac{\sigma}{\sqrt{10}}$ for phoneme-to-viseme map sizes 2-44.} 
\label{fig:sp09} 
\end{figure} 
\begin{figure} [!ht] 
\centering 
\includegraphics[width=0.95\textwidth]{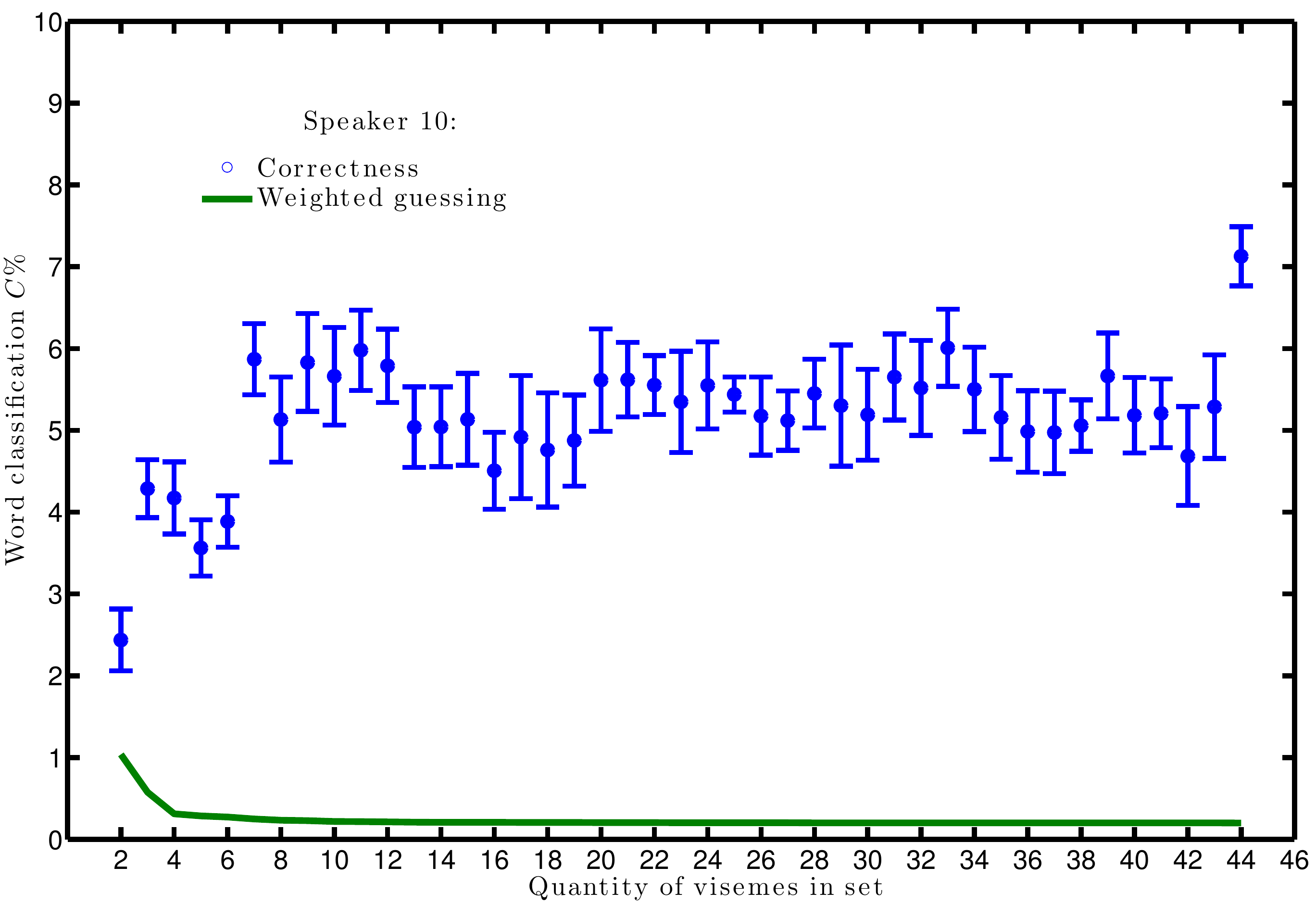} 
\caption{Speaker 10: word classification correctness, $C\pm1\frac{\sigma}{\sqrt{10}}$ for phoneme-to-viseme map sizes 2-44.} 
\label{fig:sp10} 
\end{figure} 
\begin{figure} [!ht] 
\centering 
\includegraphics[width=0.95\textwidth]{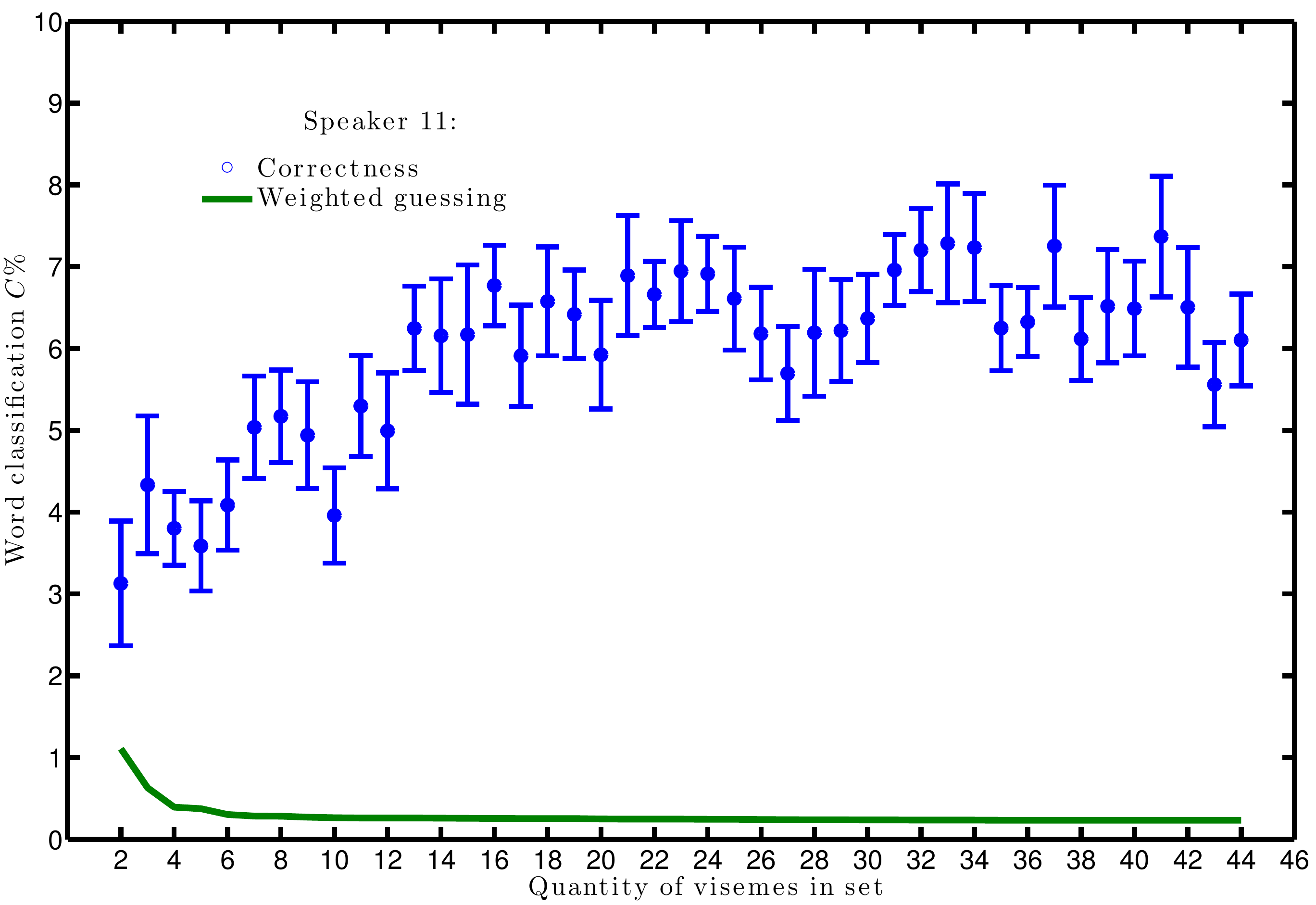} 
\caption{Speaker 11: word classification correctness, $C\pm1\frac{\sigma}{\sqrt{10}}$ for phoneme-to-viseme map sizes 2-44.} 
\label{fig:sp11} 
\end{figure} 
\clearpage
All correctness scores are significantly above chance albeit still low. There is variation between speakers, which is expected but there is a very clear overall trend. Superior performances are to be found with larger numbers of visemes. An important point is some authors report word accuracy as viseme performance when using a word unit language network. This is unhelpful as it masks the effect of homophones by using the network level unit rather than the accuracy of the viseme models themselves. Had we reported this then the effect of needing larger numbers of visemes would not be visible.

Also in Figures~\ref{fig:sp01} - \ref{fig:sp12}, we have highlighted the class sets in red where these show a significant improvement in classification over the adjacent set of units on its right side along the $x$-axis. This is where we can identify the pairs of classes which, when merged into one class, significantly improve classification. Table~\ref{tab:merges} lists these special viseme combinations. Referencing back to speaker demographics (such as gender or age), there is no apparent pattern through these viseme combinations. So we have further evidence to reinforce the knowledge that all speakers are visually unique and we are reminded of how difficult finding a set of cross-speaker visemes is when phonemes require alternative groupings for each individual. 
 
\begin{table}[h] 
\centering 
\caption{Viseme class merges which improve word classification in correctness; $V_n=V_i+V_j$.} 
\begin{tabular}{| l || l | l | l || l | l |} 
\hline 
Speaker & Set No & $V_i$ & $V_j$ & Set No & $V_n$ \\ 
\hline \hline 
Sp01 & 35 & /s/ /r/ & /\textipa{D}/ & 34 & /s/ /r/ /\textipa{D}/ \\ 
Sp02 & 22 & /d/ & /z/ /y/ & 21 & /d/ /z/ /y/ \\ 
Sp03 & 34 & /b/ /t\textipa{S}/ & /\textipa{Z}/ & 33 & /b/ /t\textipa{S}/ /\textipa{Z}/ \\ 
Sp03 & 31 & /\textipa{Z}/ /b/ /t\textipa{S}/ & /z/ & 30 & /\textipa{Z}/ /b/ /t\textipa{S}/ /z/ \\ 
Sp03 & 25 & /p/ /r/ & /\textipa{N}/ & 24 & /p/ /r/ /\textipa{N}/ \\ 
Sp05 & 17 & /ae/ & /eh/ & 16 & /ae/ /eh/ \\ 
Sp06 & 35 & /ae/ /\textturnv/ & /iy/ & 34 & /ae/ /\textturnv/ /iy/ \\ 
Sp09 & 12 & /b/ /w/ /v/ & /d\textipa{Z}/ /hh/ & 11 & /b/ /w/ /v/ /d\textipa{Z}/ /hh/ \\ 
Sp12 & 36 & /\textturnv/ & /\textopeno/ & 34 & /\textturnv/ /\textopeno/ \\ 
\hline 
\end{tabular} 
\label{tab:merges} 
\end{table} 
 
\begin{figure}[!ht] 
\centering 
\includegraphics[width=0.95\linewidth]{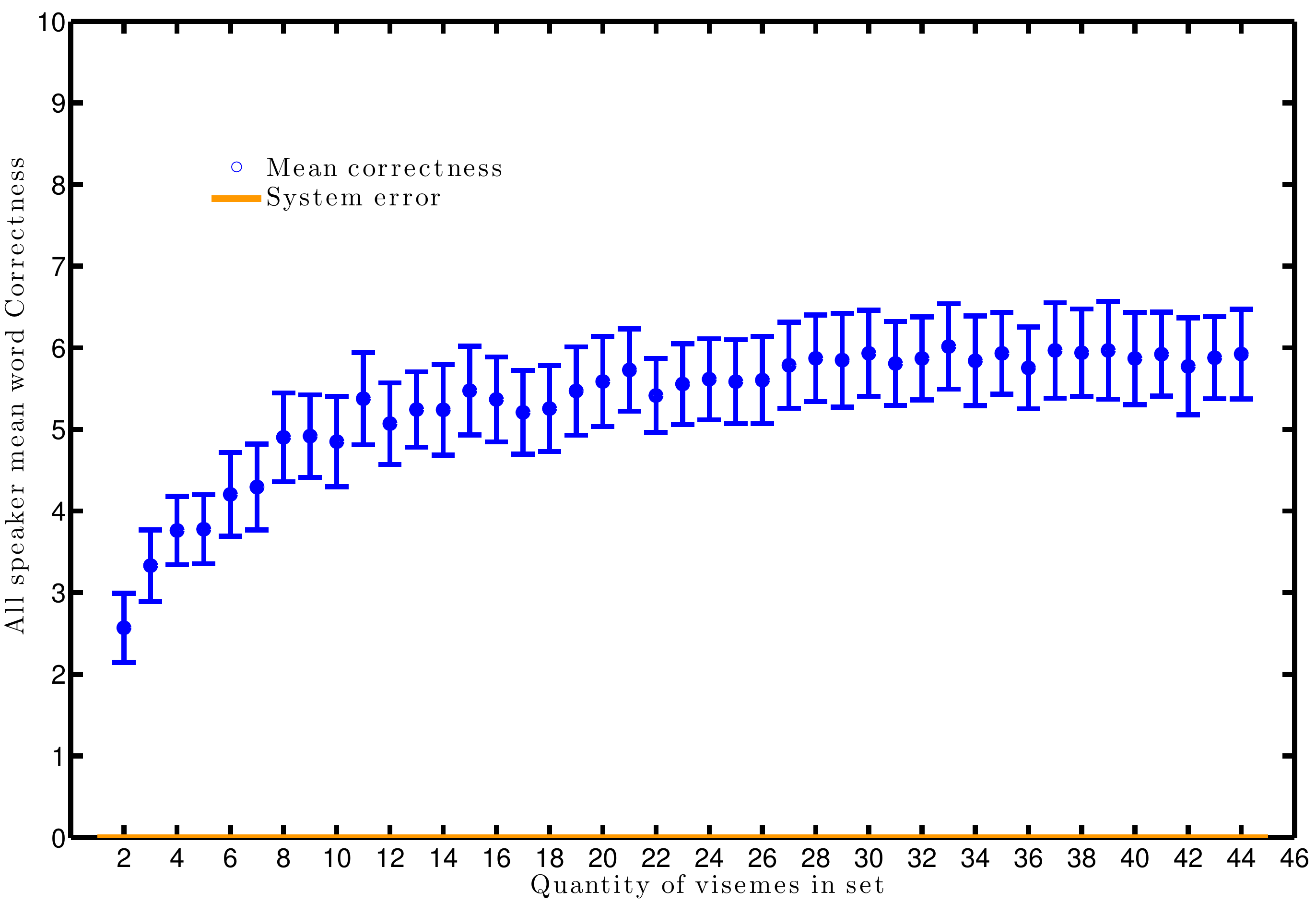} 
\caption{All-speaker mean word classification correctness $C\pm1\frac{\sigma}{\sqrt{10}}$.} 
\label{fig:correctness} 
\end{figure} 

The conventional wisdom, that visemes are needed for lip-reading, (in \cite{hazen2006visual} for example), is countered in our experiments as our phoneme classification is not significantly different from viseme classification. It is however an over simplification to assert better lip-reading can be achieved with phonemes than visemes as this has not been shown here with any significance. Generally speaking, larger numbers of visemes out-perform smaller numbers. However, when classification is aggregated in Figure~\ref{fig:correctness}, which is the mean word correctness, $C$, classification over all speakers, there is, within an error bar, a monotonic trend. In Figure~\ref{fig:correctness} we have also plotted the system error instead of guessing. System error is calculated by using the ground truth transcript of the test data in place of the classifiers output in \texttt{HResults}, in doing so we obtain any errors caused by the system rather than the classifiers. Fortunately, this is zero, demonstrating the robustness of an HMM lip-reading system.
 
In the literature we have already reviewed a number of proposed phoneme-to-viseme maps, typically these generate between 10 and 20 visemes (see subsection~\ref{sec:comparison} for a summary) - the Lee set has six consonant visemes and five vowel visemes \cite{lee2002audio}; Jeffers eight \& three \cite{jeffers1971speechreading} respectively and so on. Figures~\ref{fig:sp01}-\ref{fig:sp12} \&~\ref{fig:correctness} show a definite rapid drop-off in performance for sets which contain fewer than ten visemes but the region between 11 and 20 contains the optimum viseme set for three out of the 12 speakers which is more than chance. This mean, for each speaker we have shown an optimal number of visual units (shown by the best performing result in Figures~\ref{fig:sp01}-\ref{fig:sp12}) but the optimal number is not related to any of the conventional viseme definitions, neither is the number of phonemes. Table~\ref{tab:pr_vals} shows the correctness of each speakers phoneme classification. 
 
\begin{table}[h] 
\centering 
\caption{Phoneme correctness $C$ for each speaker, these are plotted on the right hand side in Figures~\ref{fig:sp01} to~\ref{fig:sp12} as the largest set of visemes (either 44 or 45, subject to the speaker).} 
\begin{tabular}{|l|r|r|r|r|r|r|r|r|r|r|r|r|} 
\hline 
Speaker & 1 & 2 & 3 & 4 & 5 & 6  \\ 
Phoneme $C$ & 0.045 & 0.060 & 0.058 & 0.049 & 0.063 & 0.063 \\ 
\hline\hline 
Speaker & 7 & 8 & 9 &10 & 11 & 12 \\ 
Phoneme $C$ & 0.055 & 0.090 & 0.063 & 0.071 & 0.061 & 0.064 \\ 
\hline 
\end{tabular} 
\label{tab:pr_vals} 
\end{table}

The implication is that, for a few speakers, it is possible to conclude a small number of visemes are optimal. However, when considering all speakers, it is much more likely phonemes provide a better set of classifier labels for classification. 
 
The two factors at play in these graphs are, the underlying accuracy with which the visual units represent the mouth shape and appearances versus the introduction of homophones. For large numbers of visemes these are close to phonetic classification, (with fewer homophones) but they run the risk of visual units which are not visually distinctive - several of the HMM models will ``match" on a particular sub-sequence. This latter problem creates a decoding lattice in which there are several near equal probability paths which, in turn, implies state-of-the-art language models would improve results still further. 

\section{Hierarchical training for weak-learned visemes} 
\label{chap:support network} 
% 
% In machine lip-reading, is the classification of an utterance from a visual-only signal, there are many obstacles to overcome. Some, such as pose \cite{4218129, kaucic1998accurate}, motion \cite{927467,ong2011robust} and resolution \cite{bear2014resolution} are controllable, including the selection of a phoneme-to-viseme mapping \cite{cappelletta2012phoneme, bear2014phoneme} for example. However some, such as what a viseme really is, is still open for debate. Many working definitions have been offered such as; ``A set of phonemes have identical appearance on the lips'' \cite{bear2014phoneme} or ``A visual equivalent of a phoneme'' \cite{bear2014some} but none are yet accepted as the right set of classifier descriptors which, for all speakers, enables robust machine lip-reading. 
 
\begin{figure}[h]
  \centering
\includegraphics[width=0.95\textwidth]{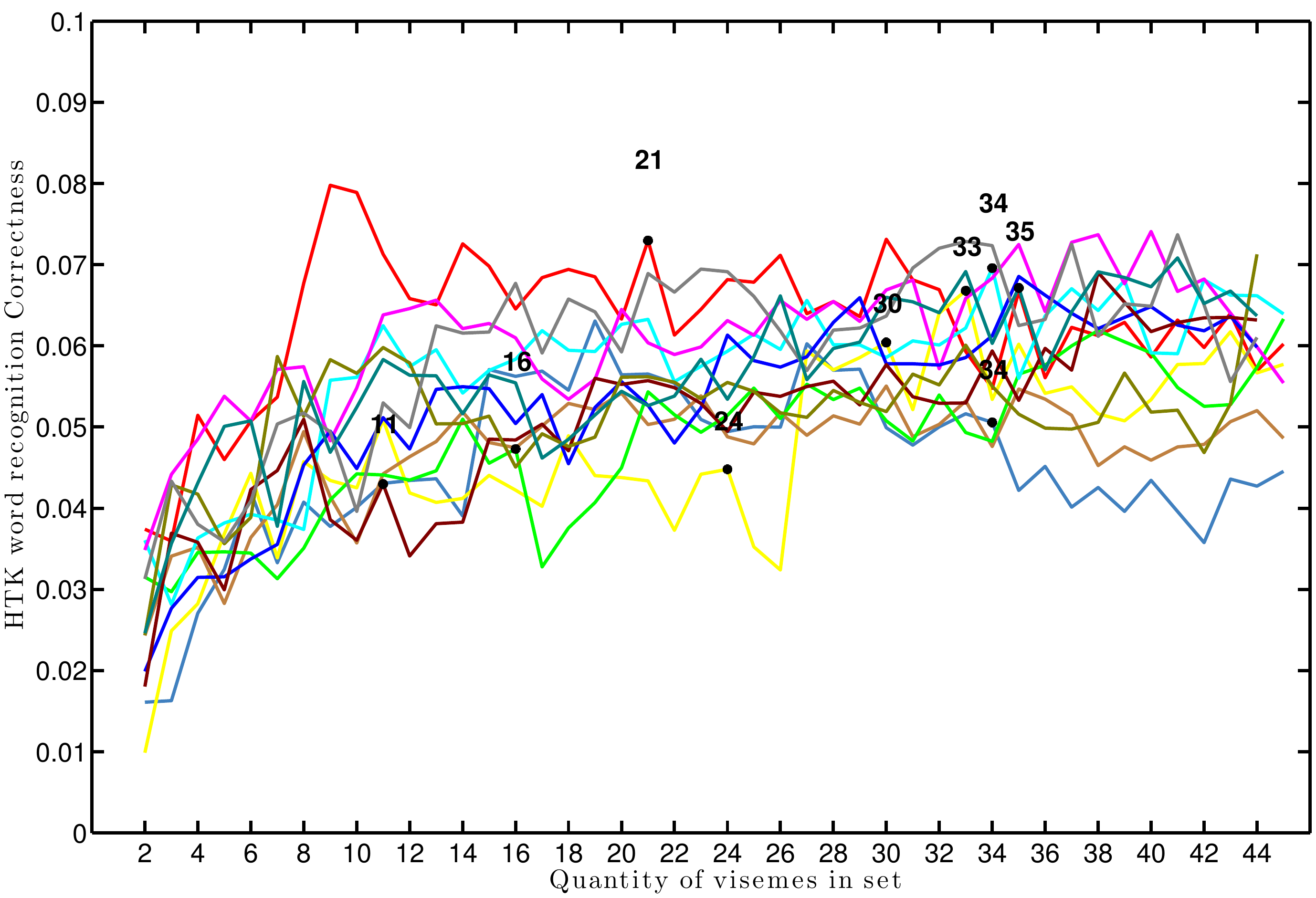}
\caption{Viseme correctness as the quantity of visemes decreases in a set of classifiers for 12 RMAV speakers. Results from \cite{bear2015findingphonemes}.}
\label{fig:biggraph}
\end{figure}

Some recent work presents evidence viseme labels may not be needed because with enough data, classifiers based upon phoneme labels can outperform viseme classification \cite{howellPhD, hazen2006visual}. Additionally, we have now seen there are challenges with using viseme/phoneme labelled classifiers including; the homophone effect, not enough training data per class, and the consequential lack of differentiation between classes when we get too many classes to distinguish between them. These can be seen in Figure~\ref{fig:biggraph} where we have replotted word correctness for 12 speakers from Section~\ref{sec:search} onto one graph. 

Figure~\ref{fig:biggraph} shows our previous results \cite{bear2015findingphonemes}, derived using the algorithm described in \cite{bear2014phoneme}. We were able to generate viseme sets of varying size. Here the $x$-axis runs from 2 to 45. The $y$-axis shows the word correctness of HMM classifiers trained on each viseme in the viseme set. There are 12 lines for the 12 RMAV speakers. 

Figure~\ref{fig:biggraph} also shows for each of our 12 speakers the significantly improving viseme sets listed in Table~\ref{tab:merges}. So we know there are sometimes units between traditional visemes and phonemes which are better for classification of the visual speech signal. Our evidence is pointing towards a larger number of visual units than was previously thought sensible. In the extreme example, if we assume one visual unit per phoneme then there is the problem that identical lip gestures may appear in two separate visemes.  

To examine this, we propose the concept of adopting weak learning for hierarchical classifier training. Our intention is to test if this method can improve phoneme classification without the need for more training data as this approach shares training data across models. This premise avoids the negative effects of introducing more homophones but will assist the identification of the more subtle but important differences in visual gestures representing alternative phonemes. Crucially, this method means we are increasing our valid training data without needing to create or record it. We remember from Chapter~\ref{chap:maps} using the wrong clusters of phonemes is worse than using none. 

Weak learning \cite{schapire1990strength} is an alternative approach to training classification models in lip-reading. Weak learning is traditionally applied in ensembles of classifiers where the sum of the classifiers produces a stronger classifier than that of the independently-weak-trained classifiers \cite{drucker1994boosting}. By acknowledging the poor performance of our viseme labelled classifiers we can assume that they are weakly trained. That is, that whilst they outperform guessing, they are not strongly trained classifiers (confirmed by our dependence on the language model to improve results). Thus, we if we can adopt a method which boosts these weakly trained viseme classifiers into strongly trained phoneme classifiers we hope to achieve significantly higher classification rates. This also encourages use of more training data for the weak-learning phase \cite{EHRENFEUCHT1989247}, and specialised training of specific phoneme samples for the phoneme classifier training phase. 

Therefore our last investigation in this thesis is an attempt to modify the lip-reading process in which we apply weak learning during classifier training, to test if the visual signal can be better translated from visemes to phonemes to better train classifiers with the same volume of visual data, whilst improving the classification. In doing so, our method addresses the challenges identified in this chapter thus far.

An additional benefit of the the revised classification process is because weak learning in the model training phase is before phoneme classification, we no longer need to consider post-classification-processing such as weighted finite state transducers \cite{howell2013confusion} to reverse the phoneme-to-viseme mapping in order to get the real phoneme recognised.

In Figure~\ref{fig:biggraph} the performance of classifiers with small numbers of visemes ($<10$) is poor due to the large number of homophones. Large numbers of visemes ($>35$) do not appear to noticeably improve the correctness: many phonetic variations look similar on the lips. The set numbers printed in black are the significantly improving viseme sets identified by the number of visemes in the set. Therefore we focus on viseme sets in the range 11 to 35 with the same speakers for our experiments using weak learning. 

 \section{Classifier training adaptation}
\begin{figure}[h]
  \centering
\includegraphics[width=0.95\textwidth]{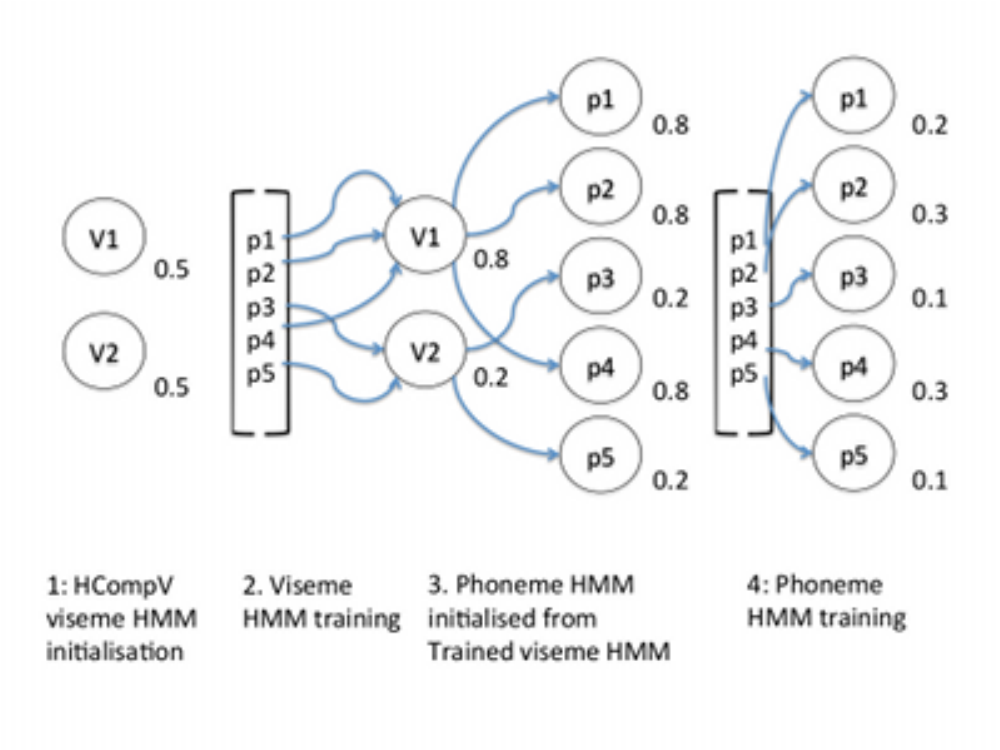}
\caption{Hierarchical training strategy for weak learning of visemes HHMs into phoneme labelled HMM classifiers.}
\label{fig:wlt_process}
\end{figure}

 The basis of the new training approach is to hierarchically train HMM classifiers. Figure~\ref{fig:wlt_process} shows a stylised illustration in which we have five phonemes (in reality there are 45) and two visemes (in reality there will be between 11 and 35). Each phoneme has been assigned to a viseme as in \cite{bear2015findingphonemes} but here we are going to learn intermediate HMMs which are identical to those in \cite{bear2015findingphonemes}. These are the viseme HMMs. We now create models for the phonemes. In this example $/p1/$, $/p2/$ and $/p4/$ are associated with $/v1/$, so are initialised as replicas of HMM $/v1/$. Likewise $/p3/$ and $/p5/$ are initialised as replicas of $/v2/$. We now retrain the phoneme models using the same training data.

In full; we initialise \textit{viseme} HMMs with \texttt{HCompV}, the HTK tool \texttt{HCompV} used for initialising HMMs defines all models equal \cite{young2006htk}. Our prototype HMM is based upon a Gaussian mixture of five components and three states. These are trained 11 times over, including both short pause model state tying (between re-estimates 3 \& 4), and forced alignment between re-estimates 7 \& 8 (this is steps 1 \& 2 in Figure~\ref{fig:wlt_process}). But before classification, these viseme HMM definitions are used as initialised definitions for phoneme labelled HMMs (Figure~\ref{fig:wlt_process} step 3). The respective viseme HMM definition is used for all the phonemes in its relative phoneme-to-viseme map. These phoneme HMMs are retrained and used for classification. As part of the classification, we use a bigram network, apply a grammar scale factor of $1.0$ and apply a transition penalty of $0.5$ (based on \cite{howellPhD}). This is implemented using 10-fold cross-validation with replacement \cite{efron1983leisurely}. 

The advantage of this approach is the phoneme classifiers have seen mostly positive cases therefore have good mode matching, the disadvantage is they are limited in their exposure to negative cases, less than the visemes. 

\subsection {Language network units}
\label{sec:ln}
As we are investigating the correct unit selection for our classifiers, we must not forget about the unit selection for the language network which is used to decode classification transcripts. This means we need to review any effect of the language network unit choice before our final experiment. Using the common process previously described for lip-reading, we perform classification using speaker-dependent visemes \cite{bear2014phoneme}, phonemes and word HMMs with the optional unit networks as listed in Table~\ref{tab:sn_tests}. This means we can answer the question `is there any dependency between the unit choice for the classifier labels and the unit of supporting language network?'. 

\begin{table}[h]
\centering
\caption{Unit selection pairs for HMMs and language network combinations.} 
\begin{tabular}{|l|l|l|}
\hline
Classifier units & Network units & $C$ \\
\hline \hline
Viseme & Viseme & 0.0231 \\
Viseme & Phoneme & 0.1914\\
Viseme & Word & 0.0851 \\
\hdashline
Phoneme & Phoneme & 0.1980\\
Phoneme & Word & 0.1980 \\
\hdashline
Word & Word & 0.1874 \\
\hline
\end{tabular}
\label{tab:sn_tests}
\end{table}
\subsection{Linguistic content} 
 
The linguistic content of any dataset has an impact on a computer lip-reading classification performance. Stylised texts have more structure and restrictions on how a speech or utterance can be organised therefore classification becomes a simpler task. In our case, with the RMAV dataset we have the challenge of lip-reading continuous speech, this is much more difficult as the complexity of the task grows with the size of the variability in what is being said, in what order and how. 
 
As part of the classification task, we ask where does the error rate come from? Which phonemes/visemes are currently recognisable? By this we mean, are there some phonemes which help the classification task more than others, can a classifier place more weight on these phonemes to improve their classification performance? Within HTK classification, grammar networks built on probability statistics of the training data have a priori knowledge of linguistic content at a word or phoneme level to improve lip-reading classification. But when considering natural continuous speech, this makes a word or phoneme/viseme network exceptionally large in order to permit any order combination of utterances. Likewise, a higher-order N-gram language model may improve classification rates but the cost of this model is disproportionate to our intention to develop better classifiers. 
 
Dictionaries help to define the vocabulary to be recognised but in natural speech what happens when a word is uttered which is not previously known? A new slang term for example. A new entry is required to be made up of phonemes which already exist. 

\begin{figure}[!h]
  \centering
\includegraphics[width=0.95\textwidth]{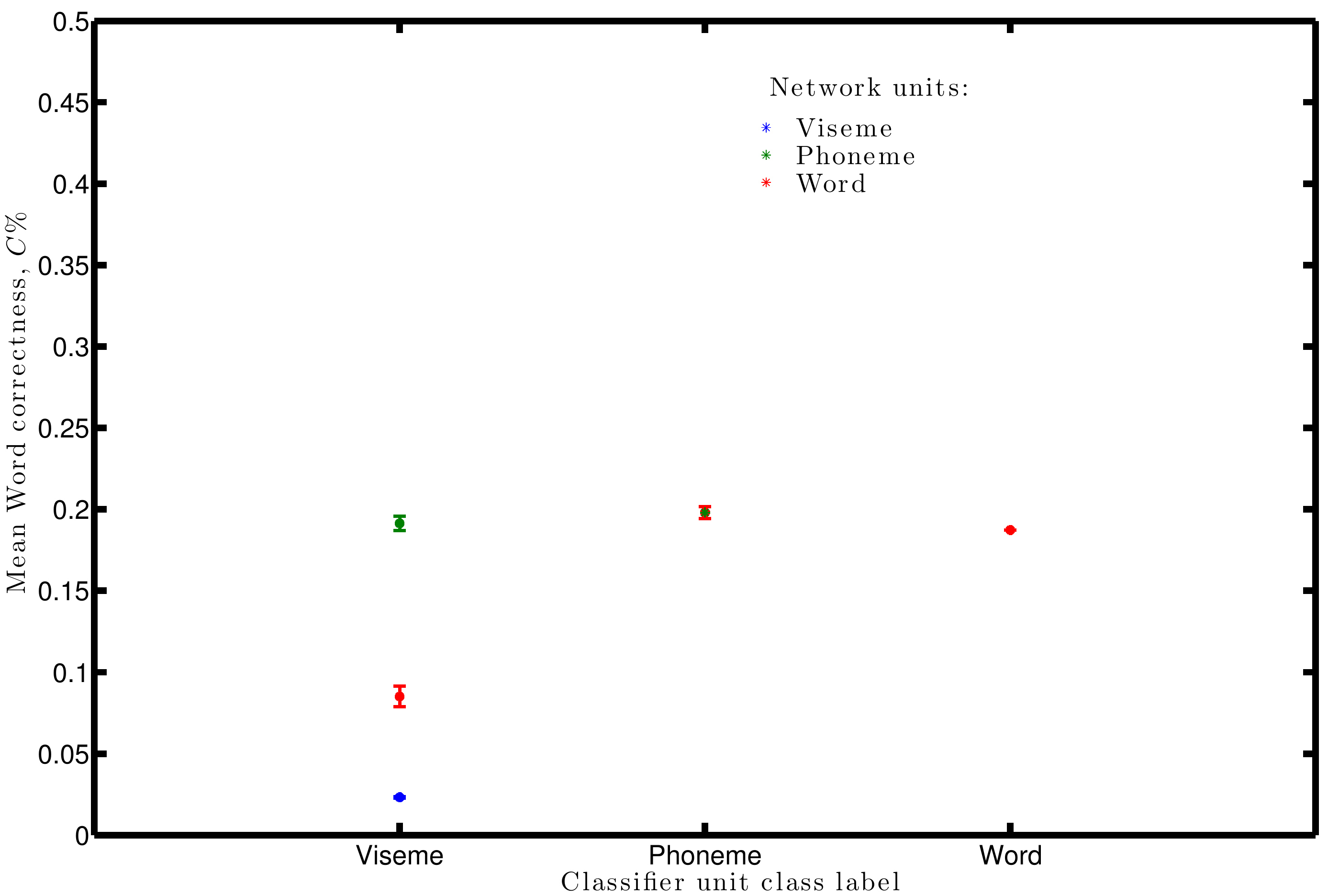}
\caption{Effects of support network unit choice with varying HMM classifier units (along the $x$-axis) measured in all speaker mean correctness, $C$. Units supported by a viseme network are shown in blue, phoneme networks are in green and word networks in red. All \{HMM,network\} pairings are shown in Table~\ref{tab:sn_tests}.}
\label{fig:sn_effects}
\end{figure}
 
\begin{table} 
\centering 
\caption{All-speaker error counts for different combinations of units for HMM classifiers with bigram support networks. HMM units run vertically and network units run horizontally through the table.} 
\begin{tabular}{|l|r|r|r|} 
\hline 
& Viseme & Phoneme & Word \\ 
\hline \hline 
Viseme 	& $0.0005$	& $0.0043$	& $0.0063$		\\ 
Phoneme & \--			& $0.0036$	& $0.0036$		\\ 
Word 	& \--			& \--			& $0.0$		 \\ 
\hline 
\end{tabular} 
\label{tab:sn_errors} 
\end{table} 
 
The effects of the network units are shown in Figure~\ref{fig:sn_effects} which plots the HMM units on the $x$-axis against the classification in Correctness $C$ (defined in \cite{young2006htk}). Error bars show one standard error. Using a viseme network shows the worst classification. This can be attributed to the volume of homophones introduced by translating from words to phonemes to visemes. We no longer consider this option. More interesting are the word and phoneme networks. The phoneme network greatly improves classification for viseme HMMs, more so than a word network. When we use phoneme HMMs, there is no difference at all between an phoneme or word network and the standard error is identical. Thus we use both phoneme and word networks in our final method.

%We elect to test the viseme sets which have previously shown significant improvements between sizes. In \cite{bear2015findingphonemes} and highlighted in Figure~\ref{fig:biggraph} we see that the smallest viseme set with a significant improvement on the set to its right, is $11$ and the largest $35$, therefore in our tests we use viseme sets from $11$ to $36$ for the new method. 
%
\section{Effects of weak learning in viseme classifier training}

\begin{figure}[h]
\centering
\includegraphics[width=0.95\textwidth]{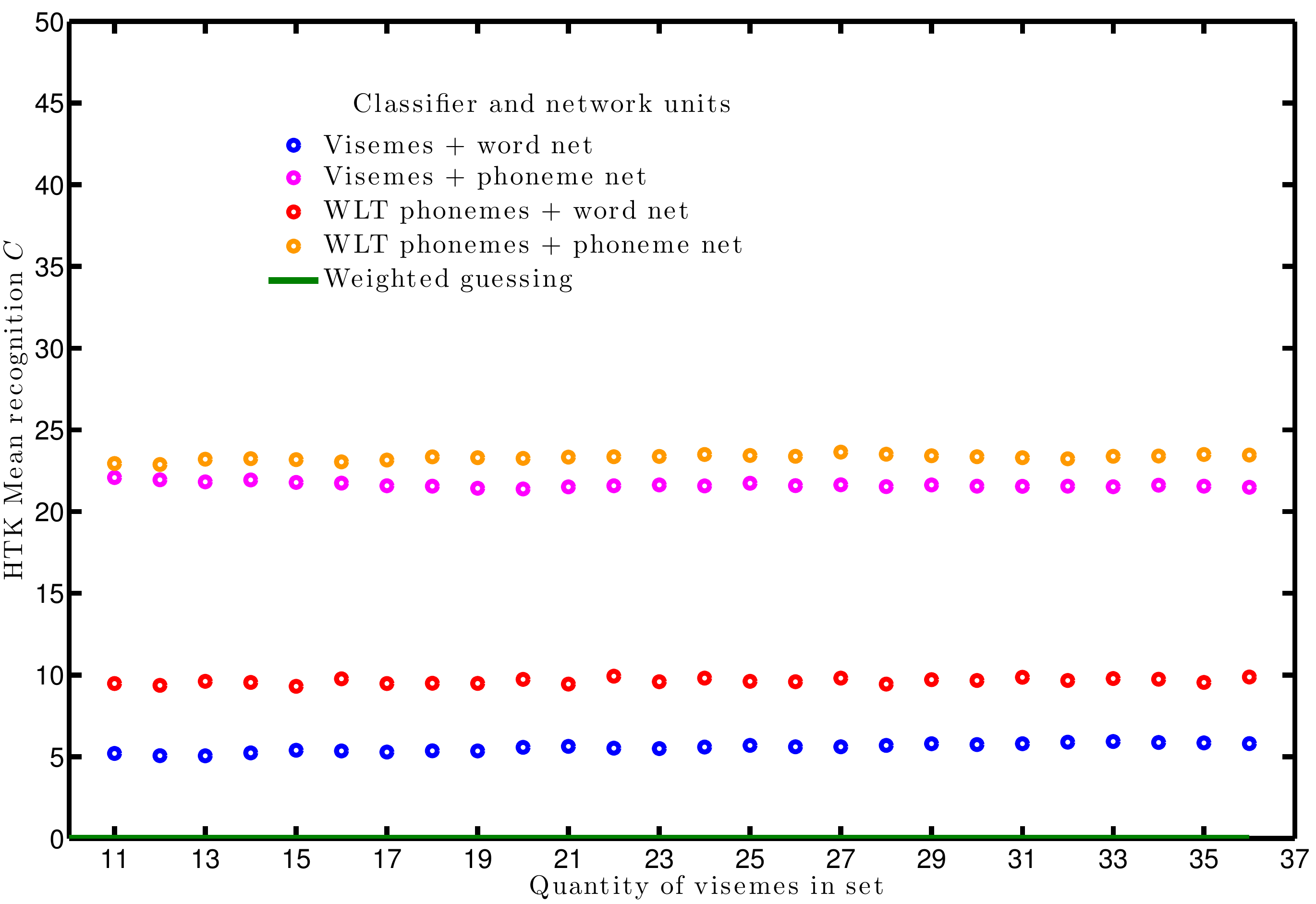}
\caption{HTK Correctness $C$ for viseme classifiers with either phoneme or word language models and weak learned phoneme classifiers with either phoneme or word language models averaged over all 12 speakers.}
\label{fig:res_all}
\end{figure}

In analysing our results, it must be remembered whilst our HMM training is hierarchical, our testing is not. Figure~\ref{fig:res_all} shows the mean Correctness, $C$, for all speakers $\pm1\frac{\sigma}{\sqrt{10}}$ over 10 folds. There are four lines plotted subject to the pairings of our HMM unit labels and the language network unit. 

The $x$-axis of Figure~\ref{fig:res_all} is the size of the viseme sets from Figure~\ref{fig:biggraph} from 11 to 36. We remind the reader this is the range of optimal number of visemes where phoneme label classifiers do not improve classification. The baseline of viseme classification with a word network from \cite{bear2015findingphonemes} is shown in blue and is not significantly different from conventionally learned phoneme classifiers. Based on our unit selection for language network study in section~\ref{sec:ln}, it is not a surprise to see just by using a phoneme network instead of a word network to support viseme classification we significantly improve our mean correctness score for all viseme set sizes for all speakers (shown in pink). Guessing is repeated as per our first previous experiments in this chapter. 

\begin{table}[h]
\centering
\caption{Minimum and maximum all speaker mean correctness, $C$, showing the effect of weak learning on phoneme labelled HMM classification.}
\begin{tabular}{|l|r|r|r|}
\hline
 	& Min & Max & Range \\
\hline \hline
Visemes + word net & 0.0274 & 0.0601 & 0.0327 \\
Phonemes + word net & 0.0905 & 0.0995 & 0.0090 \\
Effect of WLT & 0.0631 & 0.0394 & -- \\
Visemes + phoneme net & 0.2036 & 0.2214 & 0.0179 \\
Phonemes + phoneme net & 0.2253 & 0.2367 & 0.0114 \\ 
Effect of WLT & 0.0217 & 0.0153 & -- \\
\hline
\end{tabular}
\label{tab:meanstats}
\end{table}

More interesting to see is our new weakly-trained phoneme HMMs are significantly better than the viseme HMMs. In the original work of \cite{bear2015findingphonemes} phoneme HMMs gave an all-speaker mean $C = 0.059$ . Here, regardless of the size of the original viseme set, $C$ is almost double. Weakly learnt phoneme classifiers with a word network gain $0.0313$ to $0.0403$ in mean $C$, and when these phoneme classifiers are supported with a phoneme network we see a correctness gain range from $0.1661$ to $0.1775$. These gains are supported by the all speaker mean minimum and maximums listed in Table~\ref{tab:meanstats}. These gain scores are from over all the potential viseme-to-phoneme mappings and show there is little difference in which phoneme-to-viseme map is best for knowing which set of visemes to initialise our phoneme classifiers. All results, including the baseline, are significantly better than guessing (shown in green). 

\begin{figure}[!ht]
  \centering
  \includegraphics[width=0.95\textwidth]{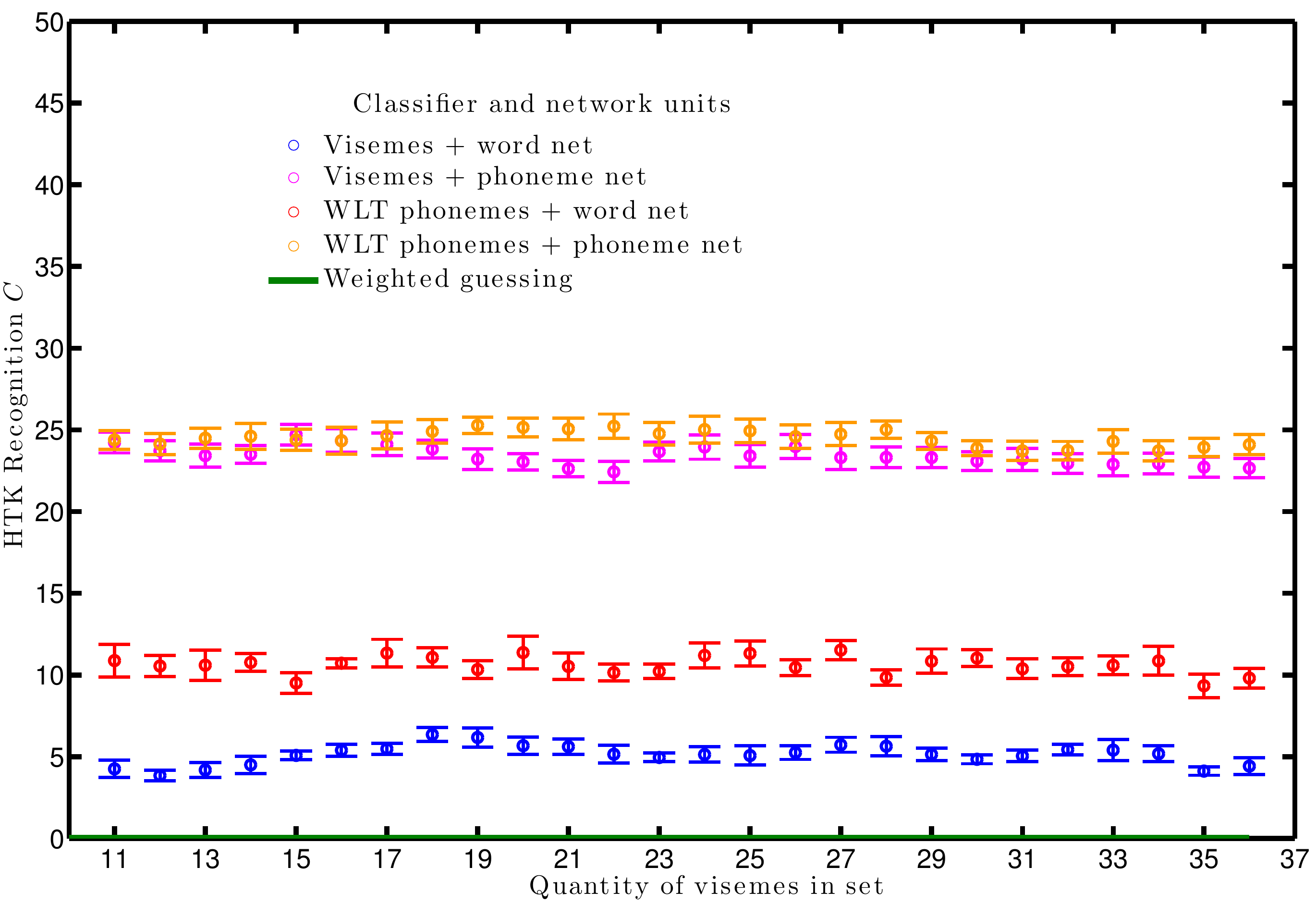}
  \caption{Speaker 1 correctness of viseme sets with a word language model (blue) and the weak learned phoneme classifiers with a phoneme or word network.}
  \label{fig:rmav1wlt}
\end{figure}

\begin{figure}[h]
  \centering
  \includegraphics[width=0.95\textwidth]{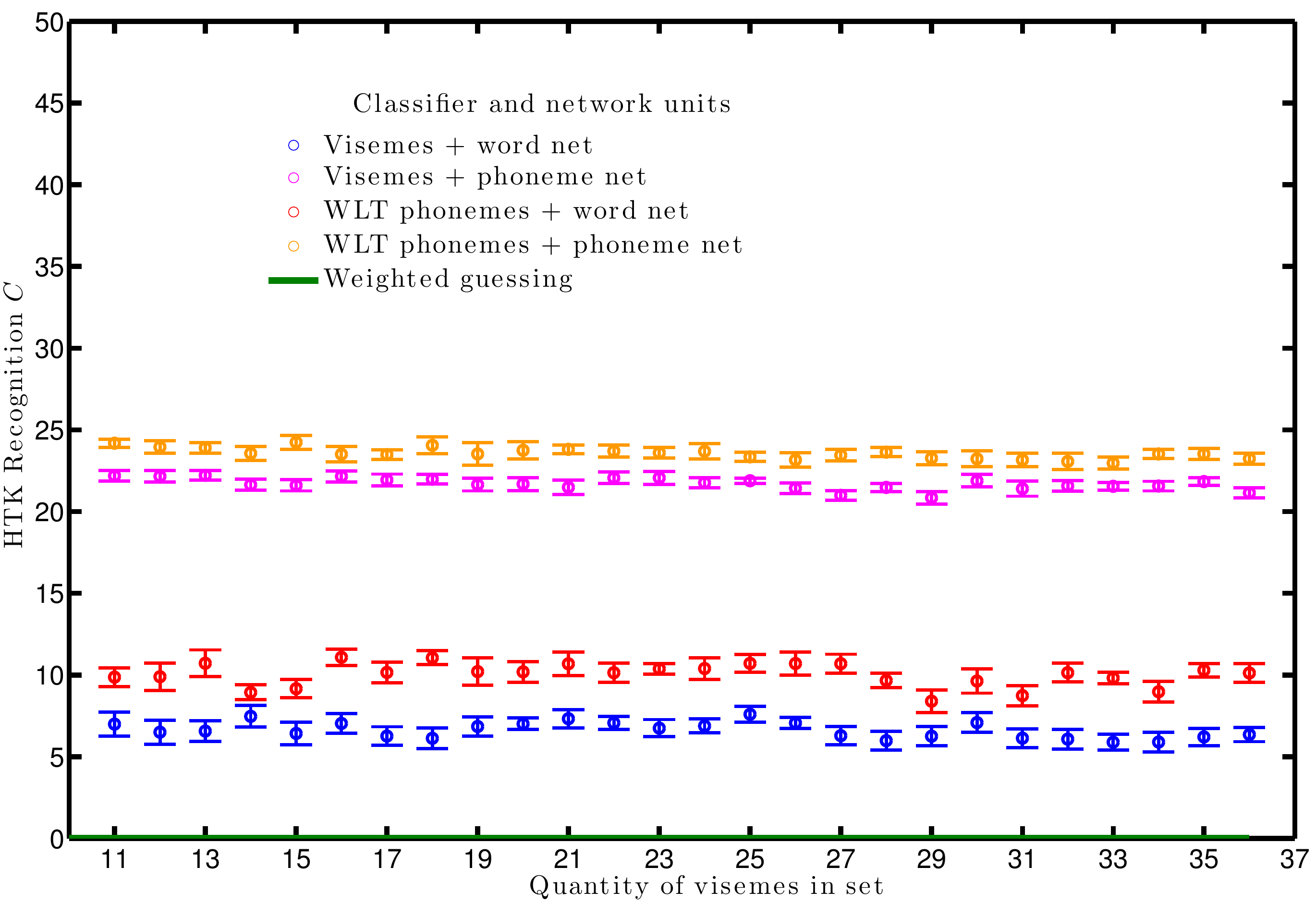}
  \caption{Speaker 2 correctness of viseme sets with a word language model (blue) and the weak learned phoneme classifiers with a phoneme or word network.}
  \label{fig:rmav2wlt}
\end{figure}
\begin{figure}[h]
  \centering
  \includegraphics[width=0.95\textwidth]{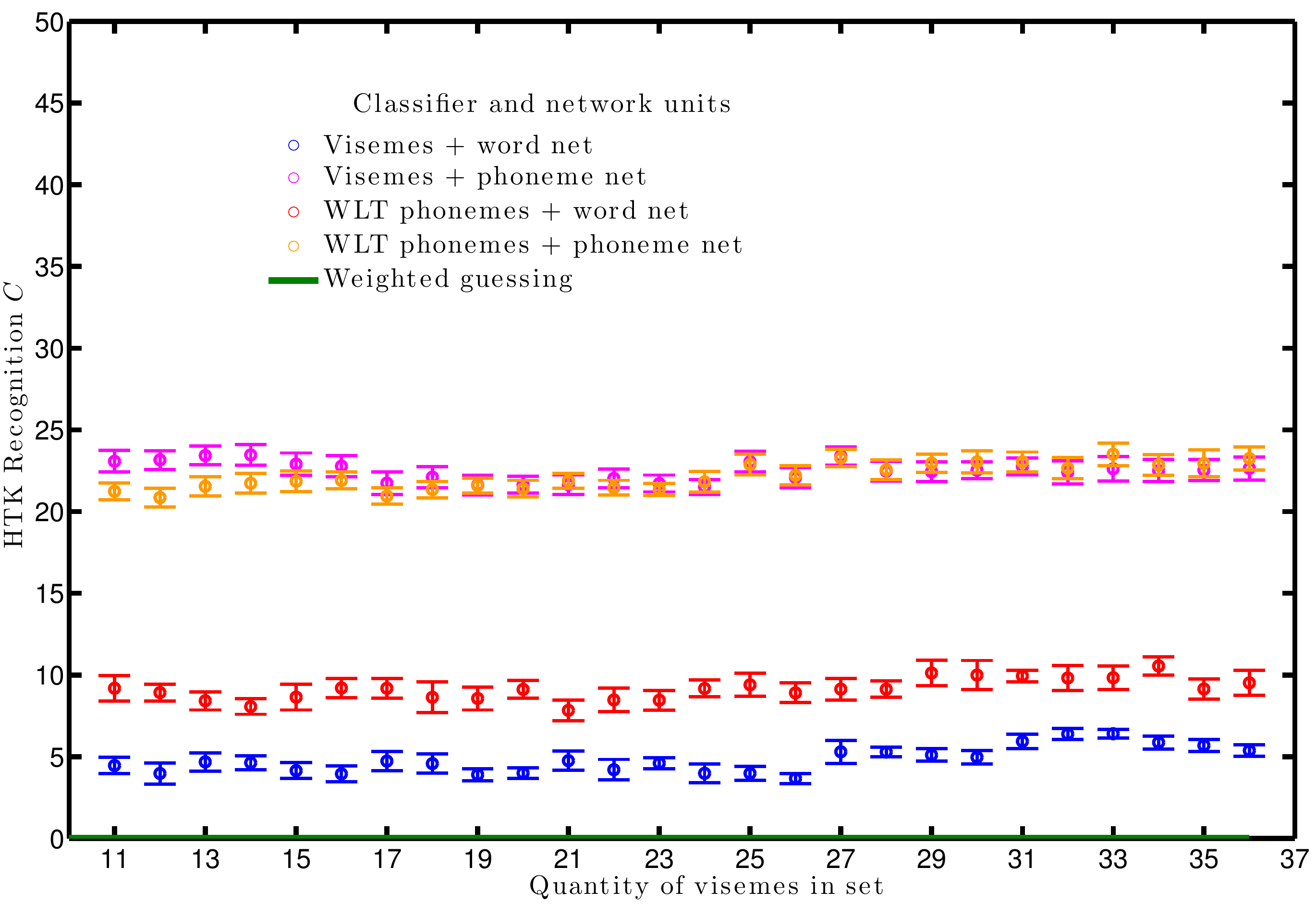}
  \caption{Speaker 3 correctness of viseme sets with a word language model (blue) and the weak learned phoneme classifiers with a phoneme or word network.}
  \label{fig:rmav3wlt}
\end{figure}
\begin{figure}[h]
  \centering
  \includegraphics[width=0.95\textwidth]{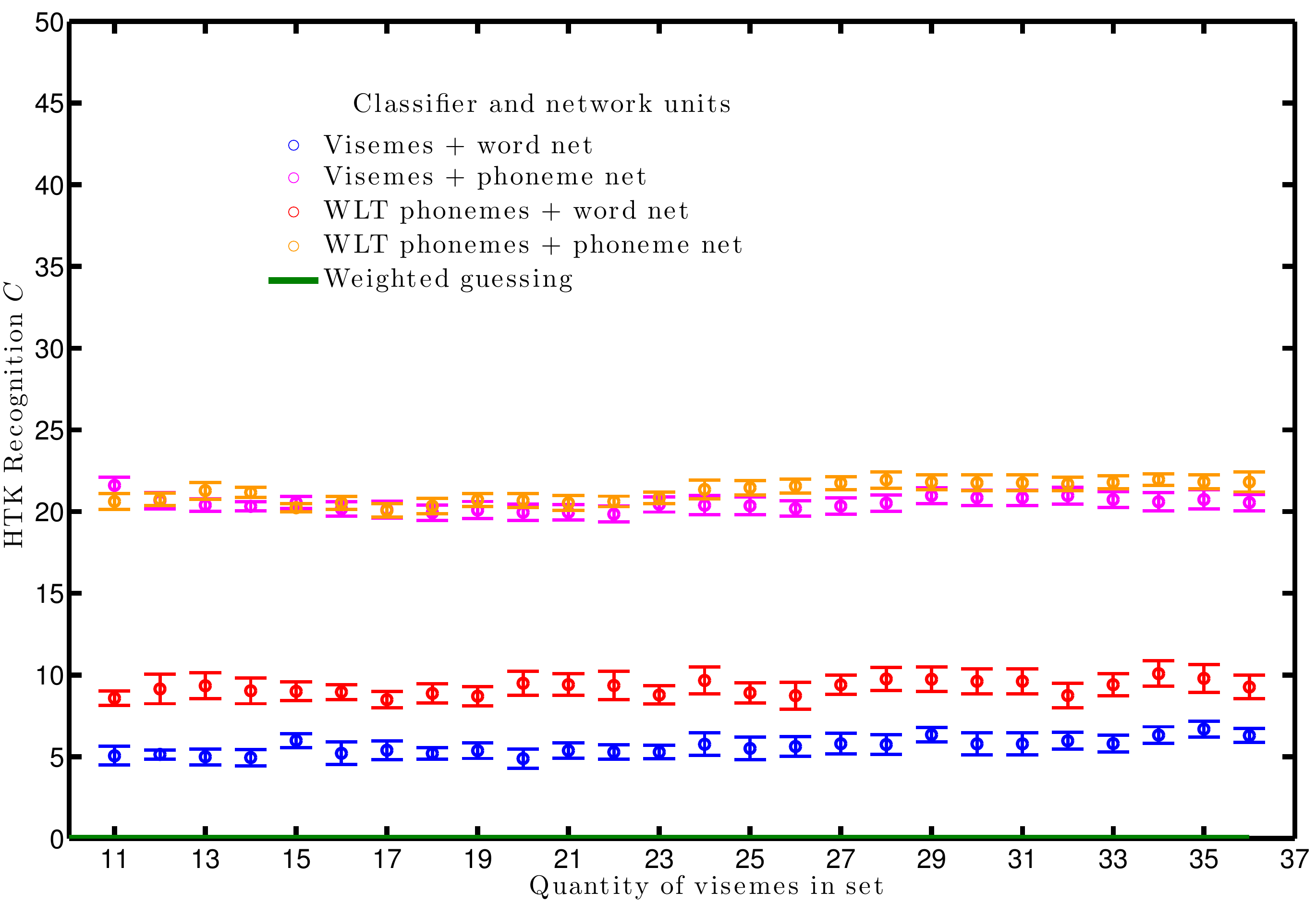}
   \caption{Speaker 4 correctness of viseme sets with a word language model (blue) and the weak learned phoneme classifiers with a phoneme or word network.}
 \label{fig:rmav4wlt}
\end{figure}
\begin{figure}[h]
  \centering
  \includegraphics[width=0.95\textwidth]{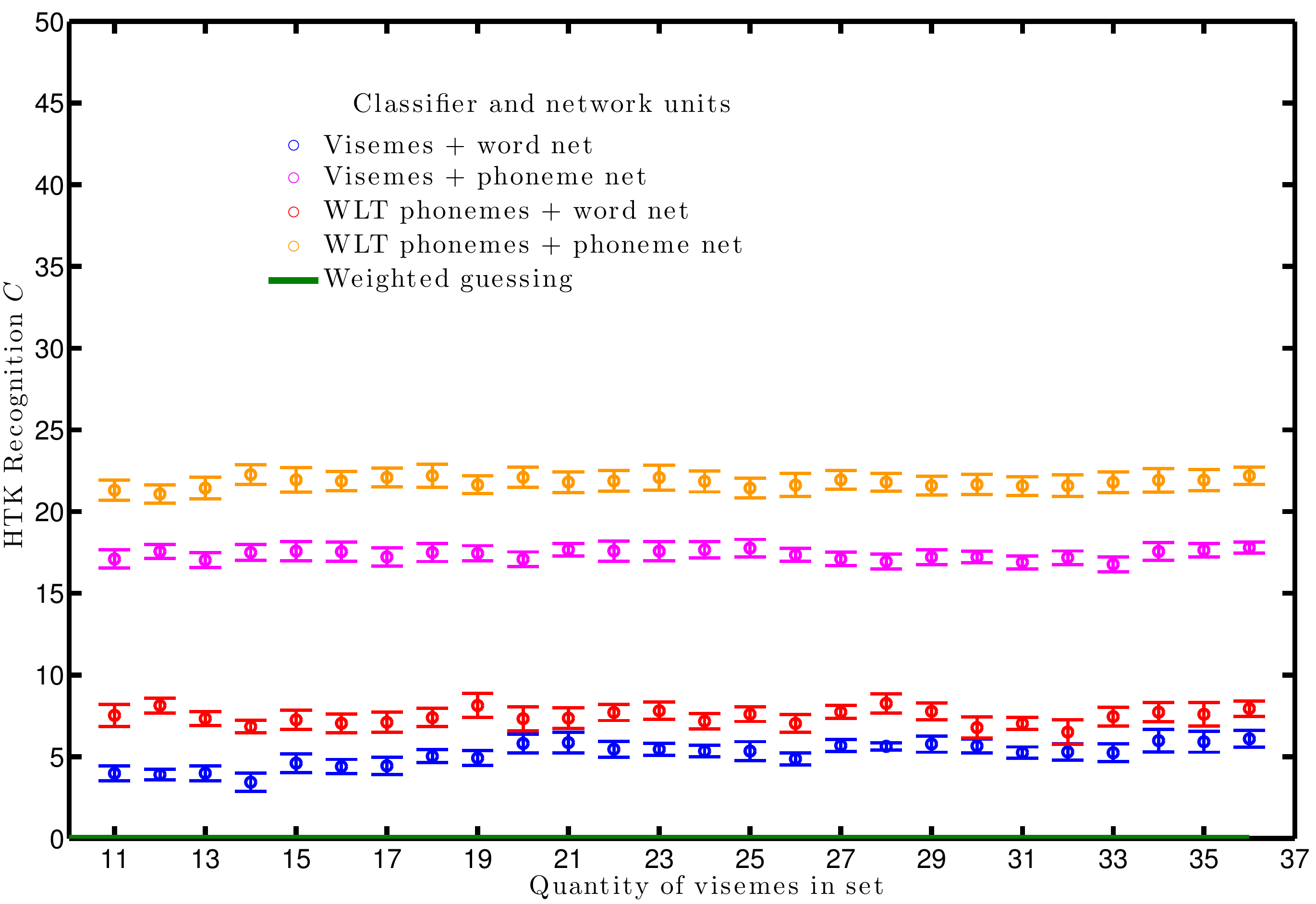}
   \caption{Speaker 5 correctness of viseme sets with a word language model (blue) and the weak learned phoneme classifiers with a phoneme or word network.}
 \label{fig:rmav5wlt}
\end{figure}
\begin{figure}[h]
  \centering
  \includegraphics[width=0.95\textwidth]{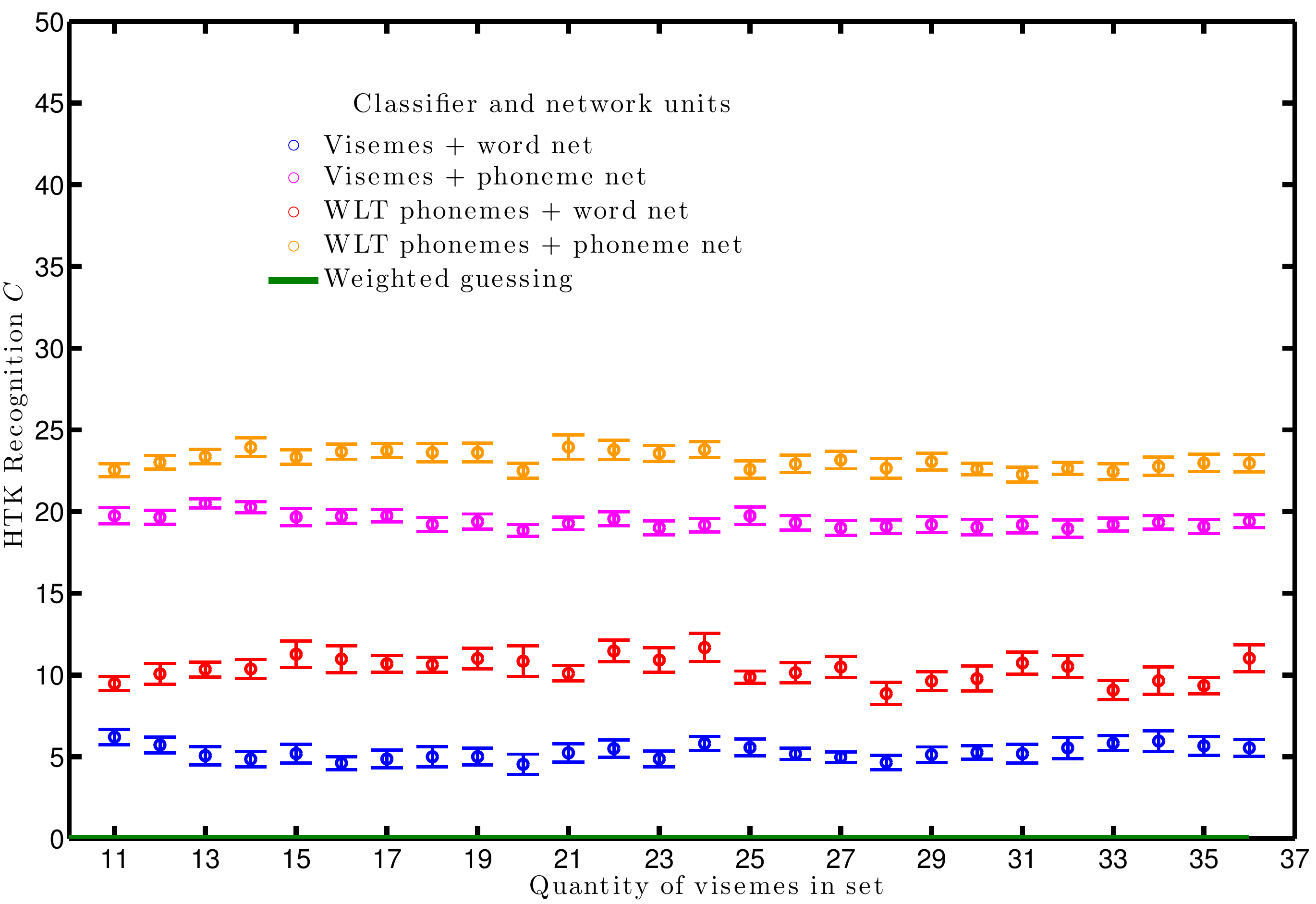}
  \caption{Speaker 6 correctness of viseme sets with a word language model (blue) and the weak learned phoneme classifiers with a phoneme or word network.}
  \label{fig:rmav6wlt}
\end{figure}
\begin{figure}[h]
  \centering
  \includegraphics[width=0.95\textwidth]{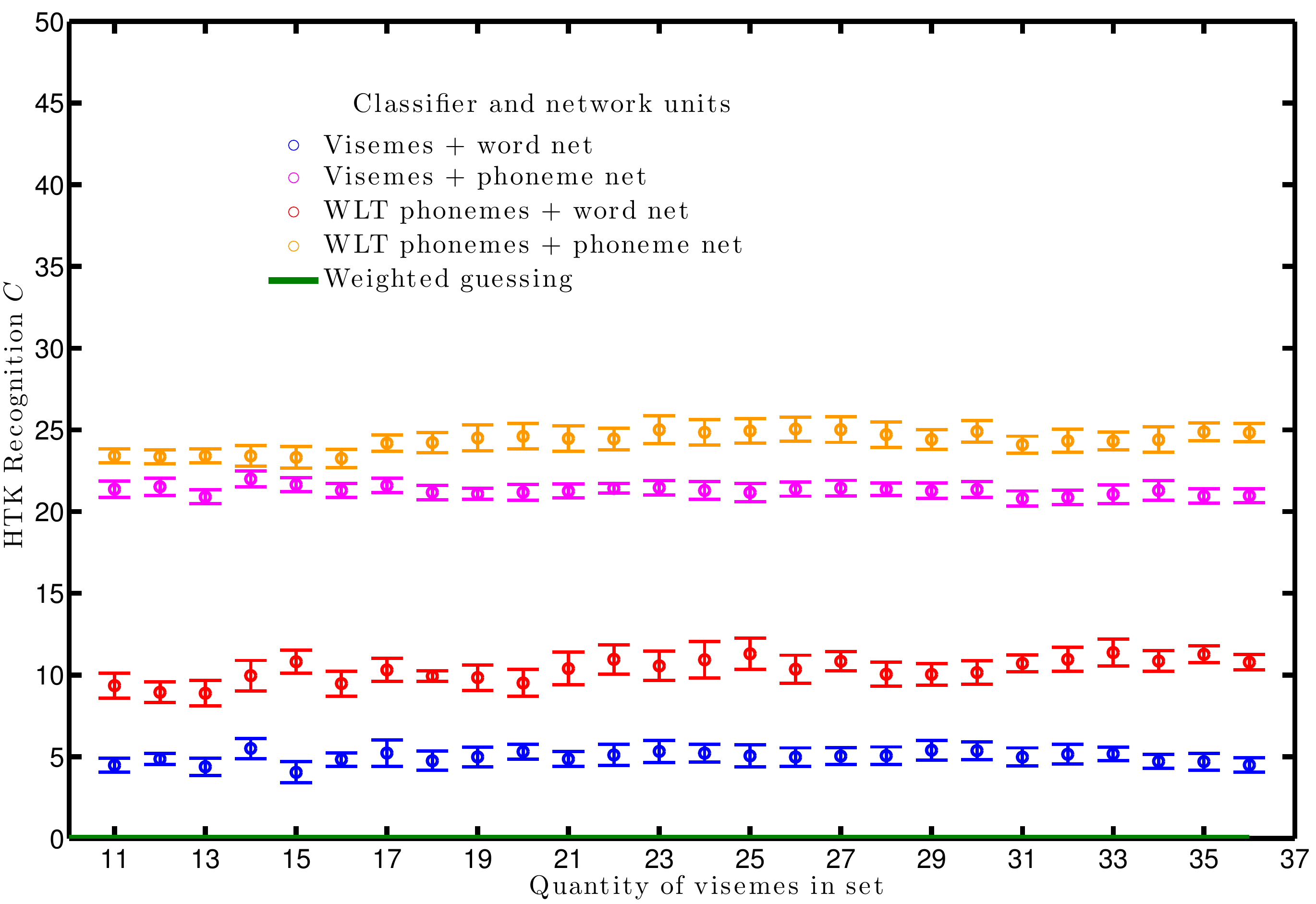}
  \caption{Speaker 7 correctness of viseme sets with a word language model (blue) and the weak learned phoneme classifiers with a phoneme or word network.}
  \label{fig:rmav7wlt}
\end{figure}
\begin{figure}[h]
  \centering
  \includegraphics[width=0.95\textwidth]{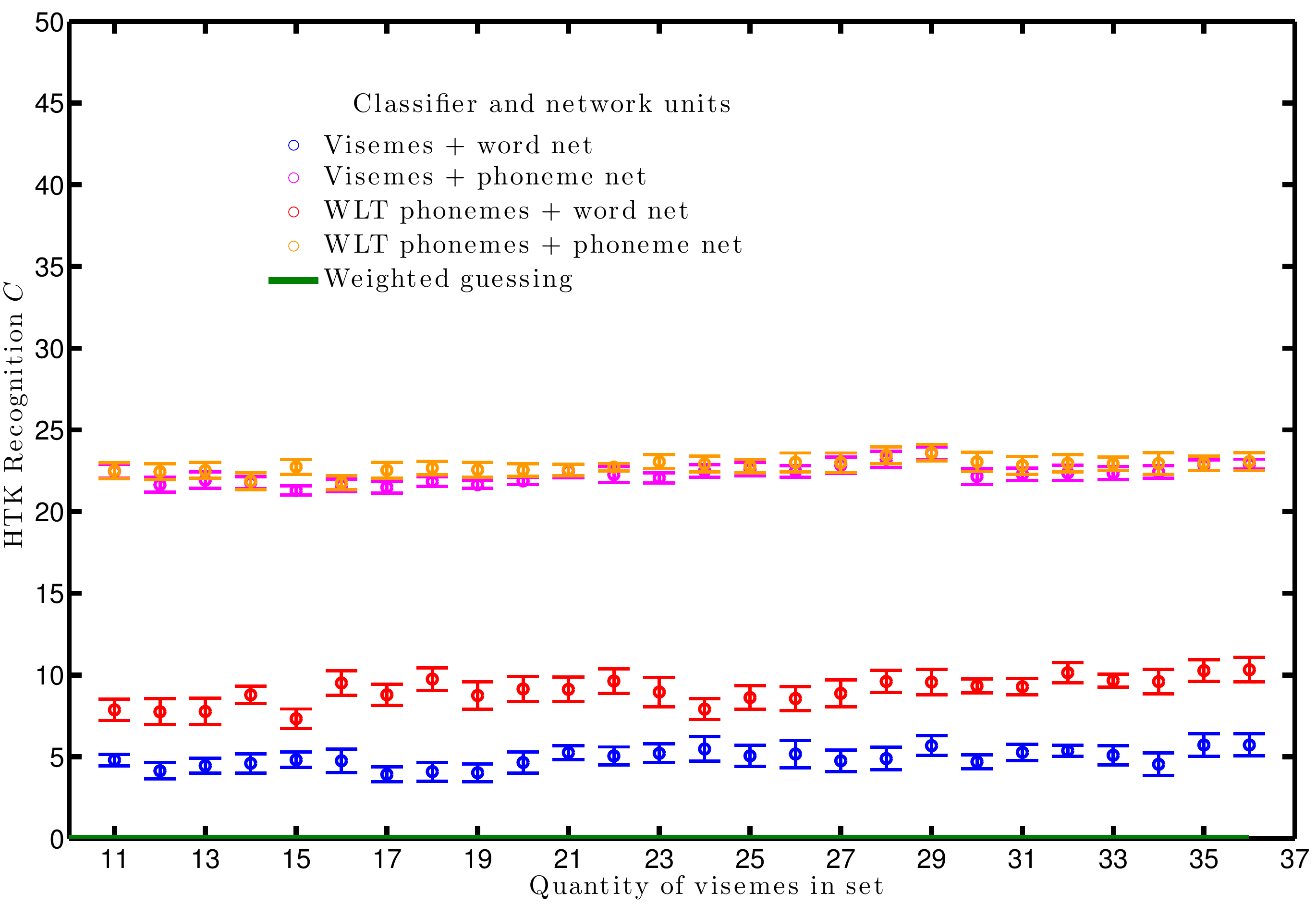}
  \caption{Speaker 8 correctness of viseme sets with a word language model (blue) and the weak learned phoneme classifiers with a phoneme or word network.}
  \label{fig:rmav8wlt}
\end{figure}
\begin{figure}[h]
  \centering
  \includegraphics[width=0.95\textwidth]{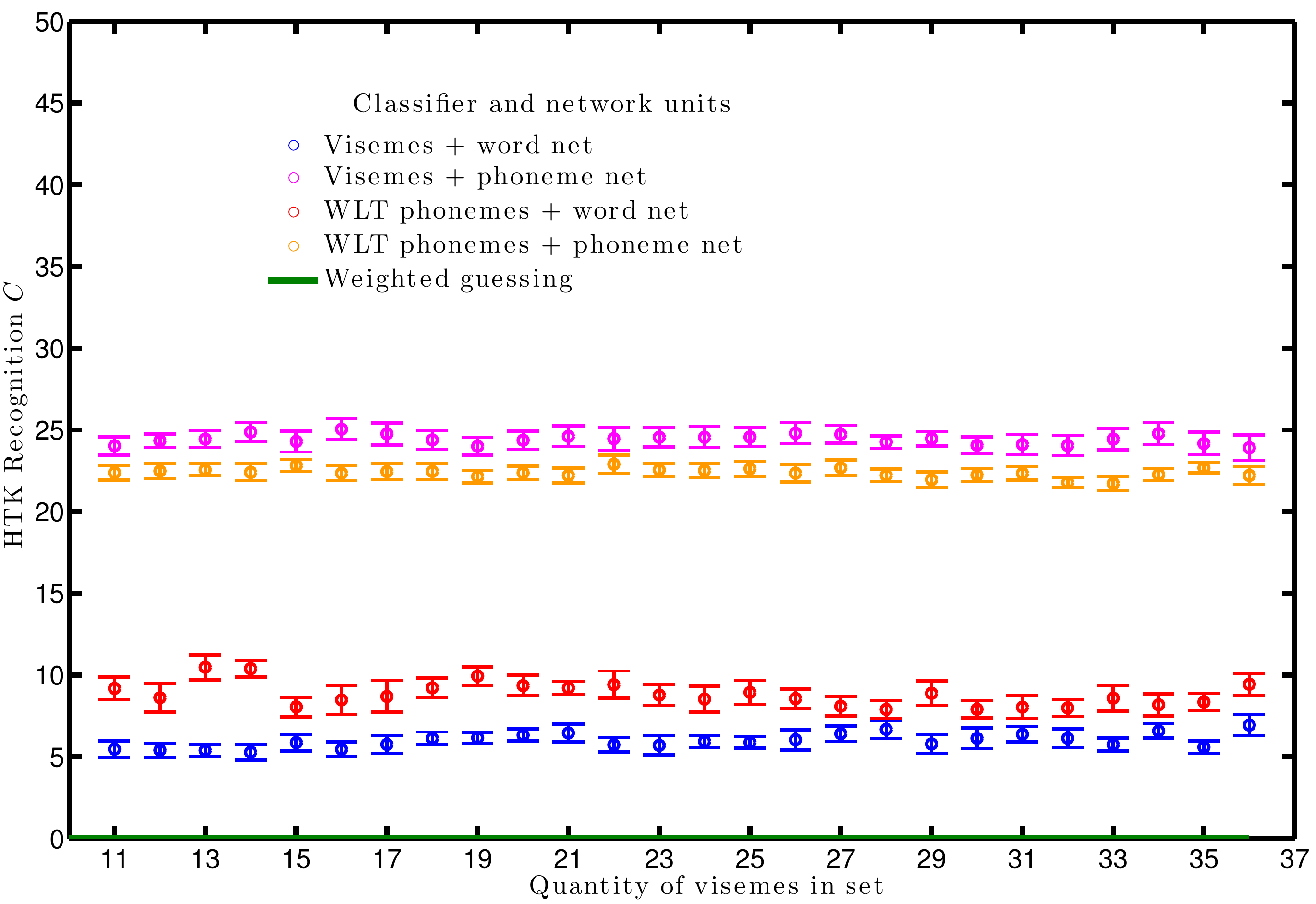}
  \caption{Speaker 9 correctness of viseme sets with a word language model (blue) and the weak learned phoneme classifiers with a phoneme or word network.}
  \label{fig:rmav9wlt}
\end{figure}
\begin{figure}[h]
  \centering
  \includegraphics[width=0.95\textwidth]{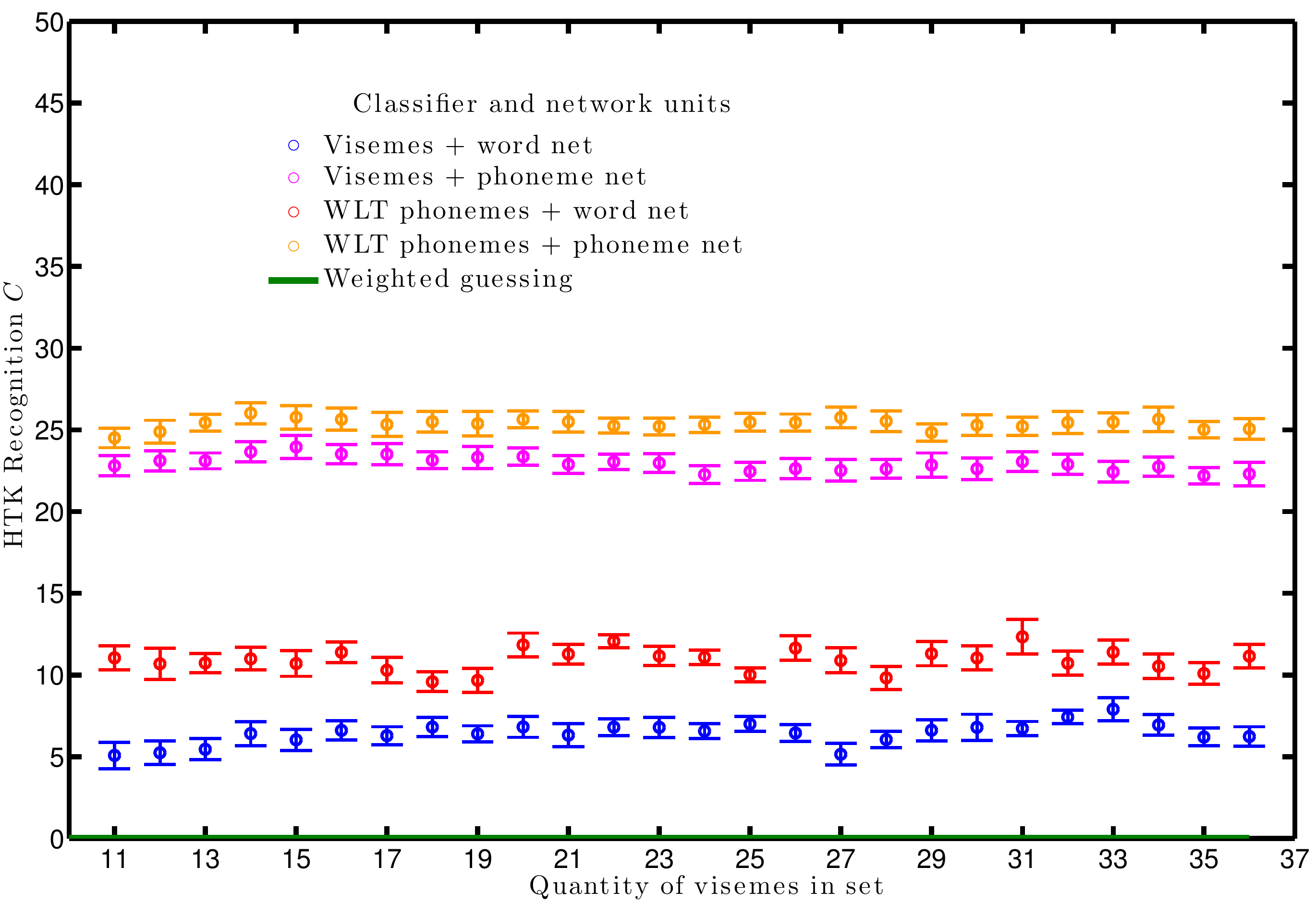}
  \caption{Speaker 10 correctness of viseme sets with a word language model (blue) and the weak learned phoneme classifiers with a phoneme or word network.}
  \label{fig:rmav10wlt}
\end{figure}
\begin{figure}[h]
  \centering
  \includegraphics[width=0.95\textwidth]{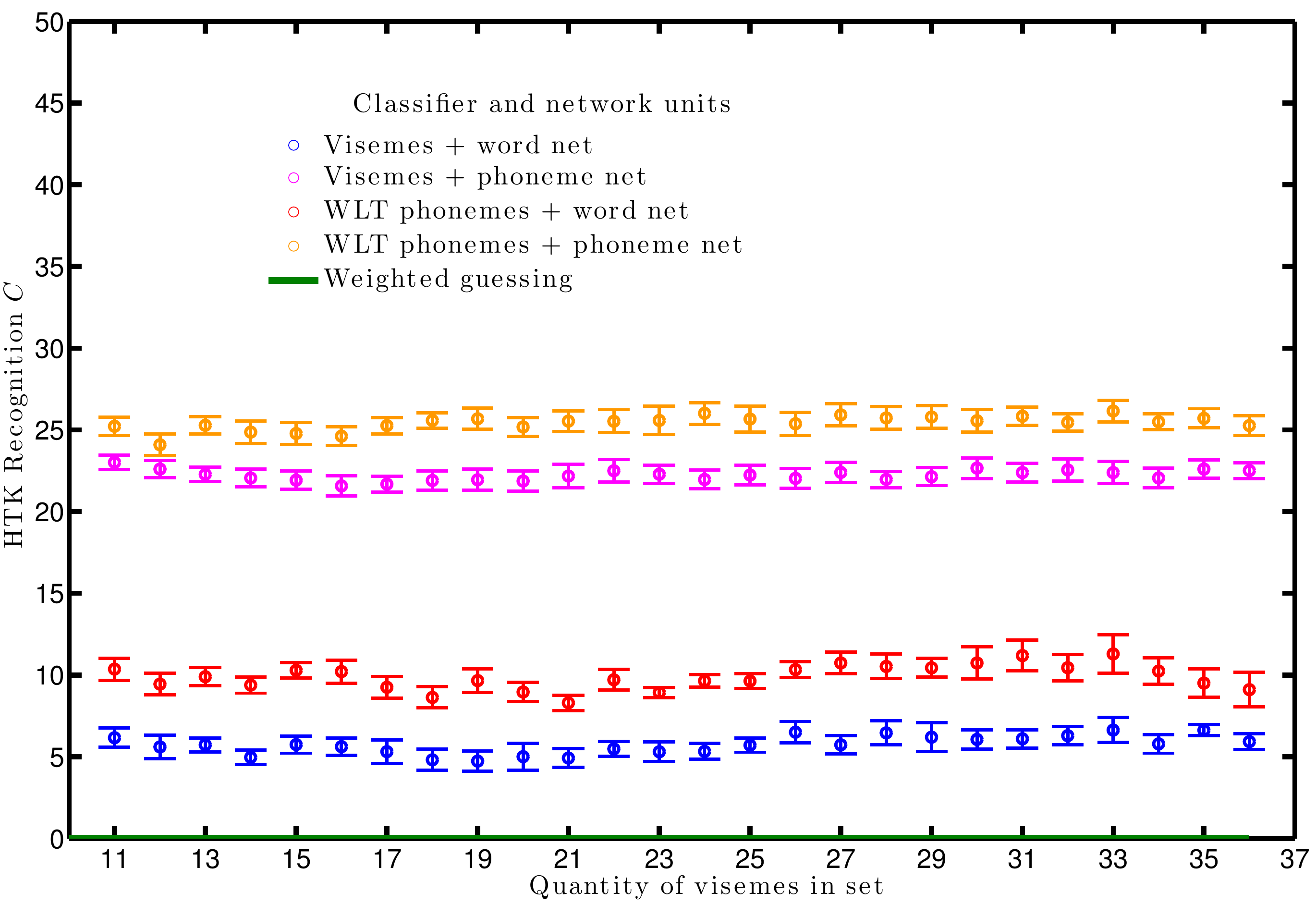}
  \caption{Speaker 11 correctness of viseme sets with a word language model (blue) and the weak learned phoneme classifiers with a phoneme or word network.}
  \label{fig:rmav11wlt}
\end{figure} 
\begin{figure}[h]
  \centering
  \includegraphics[width=0.95\textwidth]{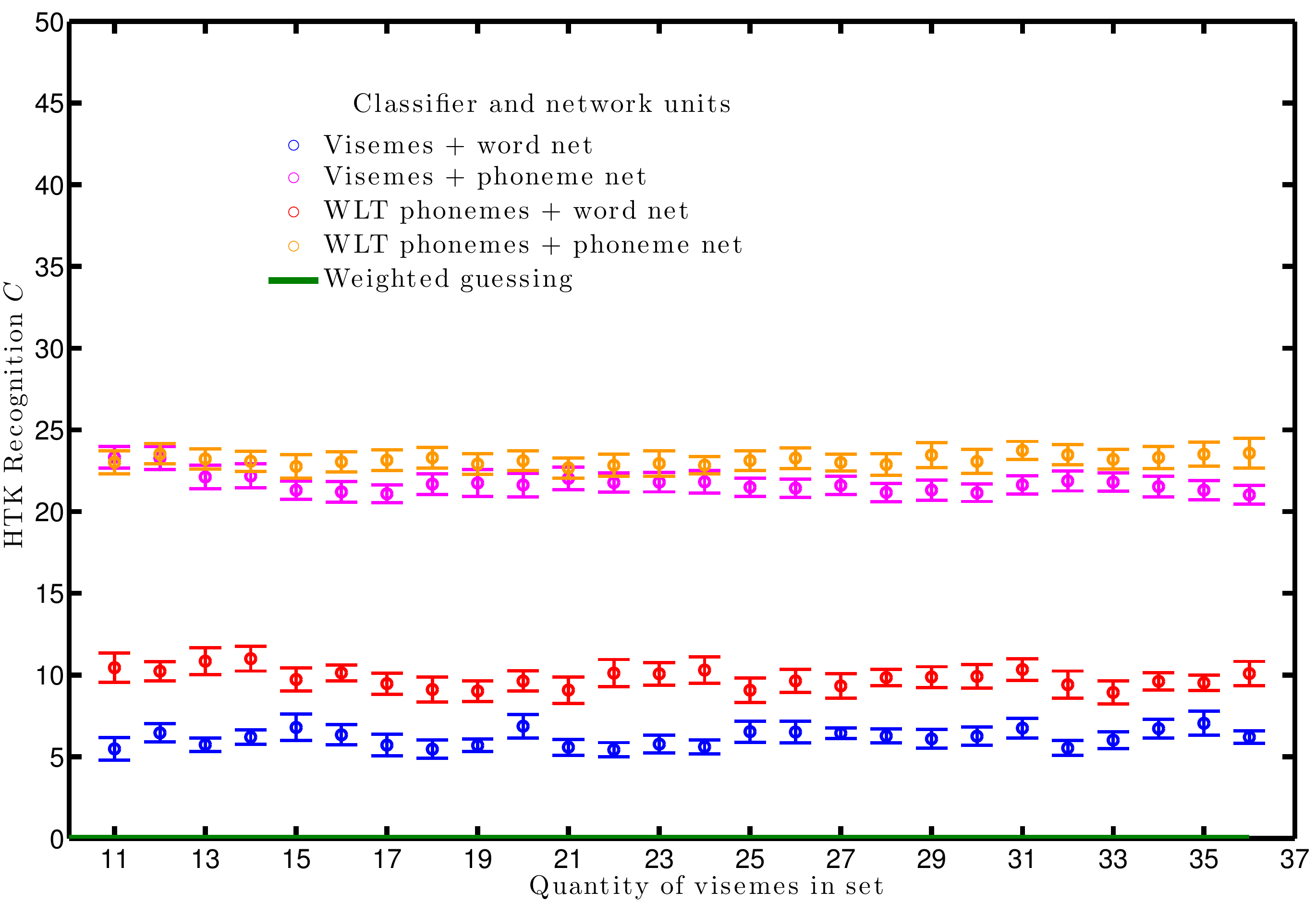}
  \caption{Speaker 12 correctness of viseme sets with a word language model (blue) and the weak learned phoneme classifiers with a phoneme or word network.}
  \label{fig:rmav12wlt}
\end{figure} \clearpage

In Figures~\ref{fig:rmav1wlt} -~\ref{fig:rmav12wlt}, we have plotted for our 12 speakers non-aggregated results showing $C \pm 1 s.e$. Whilst not monotonic, these graphs are much smoother than the speaker-dependent graphs shown in \cite{bear2015findingphonemes}. The significant differences between viseme set sizes shown in Figure~\ref{fig:biggraph} have now disappeared because the learning of differences between visemes, has been incorporated into the training of phoneme classifiers, which in turn are now better trained (plotted in red and orange which improve on blue and pink respectively). 

Our speaker-dependent results with hierarchical learning are intriguing as in \cite{lan2012insights}, with RMAV and published at the start of this thesis, showed an average viseme accuracy of {\raise.17ex\hbox{$\scriptstyle\mathtt{\sim}$}}46\%. Here, we have presented a word accuracy (which we have previously shown to be weaker than viseme accuracy but more useful) of {\raise.17ex\hbox{$\scriptstyle\mathtt{\sim}$}}10\%. We can not present viseme accuracy as our hierarchical training method has transformed the viseme classifiers into phoneme labelled classifiers, but reporting phoneme accuracy provides us with {\raise.17ex\hbox{$\scriptstyle\mathtt{\sim}$}}25\% classification. This is beneficial as phoneme transcripts are both more comprehensible due to less homophones, and reduces our dependency on the language model for comprehension.

An intriguing observation is comparing the use of a phoneme network for visemes and for weakly taught phonemes. For some speakers, the weakly learned phonemes are not always as important as having the right network unit. This is seen in Figures~\ref{fig:rmav1wlt},~\ref{fig:rmav3wlt},~\ref{fig:rmav4wlt},~\ref{fig:rmav8wlt}, and~\ref{fig:rmav12wlt} for Speaker's 1, 3, 4, 8 and 12. By rewatching the original videos to estimate the age of our speakers, we categorise them as either an `older' or 'younger' speaker. The speakers with less significant difference in the effect of weak learning are younger. This implies to lip-read a younger person we need more support from the language model, than for an older speaker. Our own informal observation is young people have more co-articulation than older people, but this is something for further investigation. 

\section{Decoding visemes}

This chapter has described a viseme derivation method which allows us to construct any number of visual units. The reader is reminded this is not a proposal of a new method for the best visemes, the priority objective in this case was a method for enabling comparison of viseme sets in a controlled manner. 
 
The presence of an optimum number of visemes within a set of classes is the result of two competing effects. In the first, as the number of visemes shrinks the number of homophones rises and it becomes more difficult to recognise words (correctness drops). In the second, as the number of visemes rises sufficient training data is no longer available in order to learn the subtle differences in lip-shapes (if they exist), so again, correctness drops. Thus, in theory the optimum number of visual units lies beween 1 and 45. In practice we see this optimum is between the number of phonemes and twelve (the size of one of the smaller viseme sets). 
 
The choice of visual units in lip-reading has caused some debate. Some researchers use visemes as adduced by, for example Fisher \cite{fisher1968confusions} (in which visemes are a theoretical construct representing phonemes should look identical on the lips \cite{hazen2006visual}). Others have noted lip-reading using phonemes can give superior performance to visemes \cite{howellPhD}. 

Here, we supply further evidence to the more nuanced hypothesis there are intermediary units, which for convenience we call visemes, that can provide superior performances provided they are derived by an analysis of the data.  

\begin{figure}[h]
\centering
\includegraphics[width=0.95\textwidth]{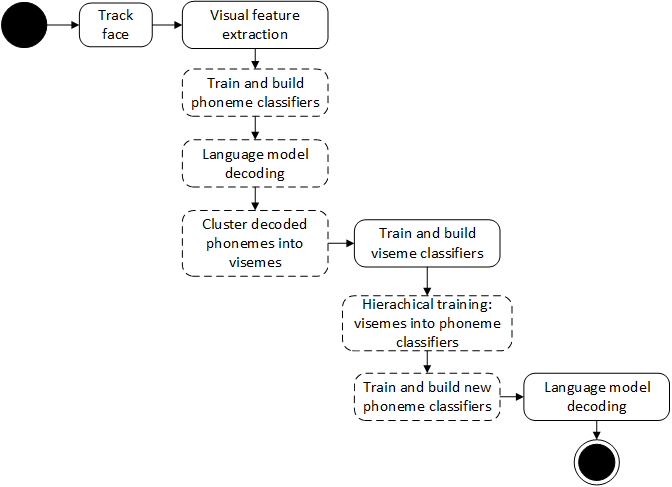}
\caption{Second augmentation to the conventional lip-reading system to include hierarchical training of phoneme-labelled classifiers with visemes.}
\label{fig:augment2}
\end{figure}

Furthermore, we have presented a novel learning algorithm which shows improved performance for these new data-driven visemes when used in hierarchical classifier training. The essence of our method is to re-train the viseme models in a fashion similar to weak learning in order they become better phoneme-labelled classifiers. This produces significantly better classification and is our second augmentation to the lip-reading system. This is shown in Figure~\ref{fig:augment2}, the extra steps are the dash-edged boxes on the right hand side.

 % AVSP - Lilir, finding phonemes and ICASSP??  support network & wlt
%!TEX root = main.tex 
\chapter[Summary of research outputs]{Summary of research outputs} 
\label{chap:nine} 
 
In this final chapter we summarise all we have learned throughout this thesis about decoding visemes. 

\section{Conclusions of research}
Our original research question was how can we further understand visemes in order to augment or replace the current HMM classifiers in conventional automatic lip-reading systems? We have learnt through our experiments that:

There is a lower limit to the resolution at which a machine can lip-read, which is at least two pixels per lip. As long as videos of speakers have at least this then we can achieve some lip-reading. This is important because a high resolution video where a person's face is so far away the pixels per lip are less than two would be worse than a close up low resolution video \cite{bear2014resolution}. 

There is also a limitation on how useful all speaker-independent (or multi-speaker) visemes within a set are towards the overall recognition. A badly trained viseme is worse than no viseme to represent certain phonemes \cite{bear2014some}. When training visemes it is not enough to say we need more data, having bad training data is more detrimental to classification than having less. 

In our comparison of many of the phoneme to viseme maps in literature we have seen there is little difference between each of them but Lee's marginally outperforms all others \cite{bear2014phoneme}. The majority of previous presented P2V maps have been designed from the observations of human lip-readers which are biased towards the individual perception of the human participating. The higher performing maps are more recent presentations and are data driven and/or machine trained. 

When clustering phonemes into visemes we can say with confidence that vowel and consonant phonemes should be isolated. This was shown by our two methods for devising speaker-dependent visemes whereby one permitted mixing of all phonemes, and the second method restricted this clustering. The second method significantly outperformed the former. Speaker individuality is important in visual speech and should be recognised when devising viseme sets due to the variability with which different people use visual gestures whilst talking \cite{bear2014phoneme}.

Viseme sets which are too small, (less than 11), are negatively affected by homophone confusions. The sets which are too large are not able to be trained sufficiently to achieve good classification. This means with the wrong speaker and training volume combination the size of the viseme set is fragile. We have shown a range of optimum sizes from 11 to 35 \cite{bear2015speakerindep}, and demonstrated how this varies by speaker and is higher than the phoneme-to-viseme maps previously presented in literature. We show for speaker dependent recognition there is a range of choices when selecting a set of visual units containing fewer members than the phoneme set, yet these sets outperform phoneme labelled classifiers. It is considered however, for speaker independent recognition, it is still most likely that phonemes are the desirable choice for classifier units as these are consistent across speakers. 

Thus, in speaker-dependent recognition, the right visemes can not just out-perform phoneme-labelled classifiers, but also when used to help train phoneme classifiers, they classify visual speech significantly better \cite{bear2015decoding}. 

To support good classifiers we have seen the effect of different unit labels in the supporting language network. Best results are achieved when the unit labels are the same for both classifiers and the network, but classification is not significantly affected if not. Therefore, for the purposes of decoding phonemes back to the words spoken, the preferred network unit is words \cite{bear2015decoding}.

In conclusion, and to answer our research question, we have improved machine lip-reading by adopting the current HMM classification system to use speaker-specific phoneme confusions within our new clustering algorithm to produce speaker-dependent viseme sets which, in turn, make good prototype HMM classifiers to train phoneme labelled classifiers. These, together with a word labelled language network, mean we can decode visemes to improve machine lip-reading classification. 

\section{Future work}

Machine lip-reading is a large and complicated problem. There are many sub-problems which need to be solved within this challenge to achieve high, consistent classification.  Remaining problems can be grouped into three classes using the Van Trees categories of detection, classification and estimation \cite{van2004detection}: 
\begin{itemize}
\item Some detection problems remaining are: automatically finding a face in an image and re-identifying same speakers between cameras, when is a person speaking/not speaking? Is the face occluded? 
\item Classification problems which remain include: Classification rates still need further improvement to be considered robust and speaker independence between the classifier training and test data is yet to produce good results. 
\item Finally estimation problems such as: how fast is a person speaking? What shapes do the lips form? And what is the the velocity of the lip movement? still need to be addressed. 
\end{itemize}

Speech recognition is a maturing field of research, but if we refer to our introduction (Section~\ref{chap:intro}) we remember our motivation to improve machine lip-reading classification is for two major reasons. 
Firstly the use of such a system would be applicable in a range of areas from entertainment (e.g. sports events) to criminal detection (e.g. CCTV recordings).
Secondly, and this is the main expectation of a lip-reading system, is the integration of such a system into AVSR. A robust lip-reading system could both improve the robustness and accuracy of an AVSR system by better use of the visual channel alone (channel independence) and as a fallback during times when the audio signal drops or is deteriorated by noise. This goal raises a number of significant questions which extend beyond the demands of achieving visual-only speech recognition. Audio-visual signal fusion, environmental noise, camera/microphone movement are three examples of further challenges in AVSR \cite{zhou2014review}.

It is a difficult problem to classify acoustic utterances from a signal of sparse visual cues, so whilst acoustic recognition is achieving ubiquity with commercial applications (in 2015 Google's Translate app added speech recognition as a novel feature to assist travellers to communicate abroad), machine lip-reading is yet to achieve the same level of robustness. Independence to speaker identity, camera view, occlusions and language all still need to be robustly accomplished before we see such technology as a reality. 
 % Conclusions:summary of findings (aka 'we now know' - relate back to research question) future work, recommendations, Summary of research outputs. 

%---------------------------

\begin{spacing}{1}
\addcontentsline{toc}{chapter}{Bibliography}
\bibliographystyle{plain}
\bibliography{thesisBib}  
\clearpage
%!TEX root = main.tex
\chapter{Phonetic notation}
\begin{table} [!ht]
\centering
\caption{For translating vowel phonemes from phonetic symbols to their respective alphabet character representations}
	\begin{tabular}{| c | c | c | c |}
	\hline
	Phonetic Symbol & Character Symbol & Latexipa & Example\\
	\hline
	/ai/ 			& /ay/ 	& /ai/ 	& bo\textbf{uy}\\
	/\textturnv/ 	& /ah/ 	& textturnv 	& \textbf{hu}t \\
	/\ae/			& /ae/	& ae			& p\textbf{a}n \\
	/\textschwa/	& /ax/	& textschwa	& \textbf{a}lbeit \\
	/\textscripta\textupsilon/ 	& /aw/	& textscripta textupsilon 	& cl\textbf{ou}d\\
	/\textopeno/ 	& /ao/ 	& textopeno 	& s\textbf{ou}r\\
	/\textscripta/	& /aa/	& textscripta & c\textbf{ar}d \\	
	/ei/ 			& /ey/ 	& /ei/  		& st\textbf{ay} \\
	/e/ 			& /eh/	& /e/ 			& dw\textbf{e}ll\\
	/\textrevepsilon/ & /er/	& textrevepsilon	& c\textbf{ur}t \\
	/\textipa{E}/	& /ea/	& \{E\} 		& cha\textbf{ir} \\
	/i/ 			& /iy/ 	& /i/ 		& cr\textbf{ee}d\\
	/\textsci/		& /ih/		& textsci 	& k\textbf{i}d\\
	/\textsci\textschwa/	& /ia/ 	& textsci textschwa 	& l\textbf{ea}r\\
	/\textsci/		& /ix/		& textsci & \textbf{i}ll \\
	/\textopeno\textsci/		& /oy/ 	& textopeno textsci 		& c\textbf{oy} \\
	/\textschwa\textupsilon/	& /ow/	& textschwa textupsilon 	& c\textbf{o}de\\
	/\textupsilon\textschwa/	& /oo/	& textupsilon textschwa & p\textbf{ru}de \\
	/\textopeno/	& /oa/ 	& textopeno & g\textbf{oa}t \\
	/\textopeno\textschwa/	& /ou/	& textopeno textschwa & p\textbf{ou}r \\
	 /\textturnscripta/			&	/oh/	&  textturnscripta	& t\textbf{o}t \\
	/u/ 			& /uw/ 	& /u/ 		& c\textbf{ue} \\
	/\textupsilon/	& /uh/ 	& textupsilon	& f\textbf{oo}d\\
	/\textopeno\textschwa/	& /ua/	& textopeno textschwa& c\textbf{or}e \\
	\hline
	\end{tabular}
\end{table}

%\section{Consonants}
%\label{sec:intro}

\begin{table*}[!ht]
\centering
\caption{For translating consonant phonemes from phonetic symbols to their respective alphabet character representations}
	\begin{tabular}{| c | c | c | c |}
	\hline
	Phonetic Symbol & Character Symbol & Latex textipa & Example\\
	\hline
	/\textipa{T}/ 	& /th/ 	& \{T\} 	& \textbf{th}in \\
	/\textipa{D}/ 	& /dh/ 	& \{D\} 	& \textbf{th}ere \\
	/\textipa{S}/ 	& /sh/ 	& \{S\} 	& \textbf{sh}eer \\
	/\textipa{Z}/ 	& /zh/ 	& \{Z\} 	& vi\textbf{s}ual \\
	/d\textipa{Z}/ 	& /jh/ 	& d\{Z\} 	& \textbf{j}udge \\
	/t\textipa{S}/ 	& /ch/ 	& t\{S\} 	& \textbf{ch}runch \\
	/\textipa{H}/ 	& /H/ (or /hh/)	& \{H\} & \textbf{h}unt\\  
	/\textipa{N}/ 	& /ng/ 	& \{N\} 	& ki\textbf{ng}\\
	/\textipa{W}/ 	& /W/ 	& \{W\} 	& \textbf{wh}isky \\
	/b/			& /b/		& /b/ 		& \textbf{b}ar \\
	/d/			& /d/		& /d/ 		& \textbf{d}art \\
	/f/			& /f/		& /f/ 		& \textbf{f}ete \\
	/g/			& /g/		& /g/ 		& \textbf{g}reat \\
	/h/			& /hh/	& /h/ 		& \textbf{h}unt\\
	/k/			& /k/		& /k/ 		& \textbf{c}ane \\
	/l/			& /l/		& /l/ 		& \textbf{l}ake \\
	/m/			& /m/		& /m/ 	& \textbf{m}other \\
	/n/			& /n/		& /n/ 		& \textbf{n}one \\
	/p/			& /p/		& /p/ 		& \textbf{p}ot\\
	/r/			& /r/		& /r/ 		& g\textbf{r}ate \\
	/s/			& /s/		& /s/ 		& \textbf{s}ilk \\
	/t/			& /t/		& /t/ 		& \textbf{t}ack \\
	/v/			& /v/		& /v/ 		& \textbf{v}erge \\
	/w/			& /w/		& /w/		& \textbf{w}eed \\
	/y/			& /y/		& /y/ 		& \textbf{y}aught \\
	/z/			& /z/		& /z/ 		& \textbf{z}ulu \\
	\hline
	\end{tabular}
\end{table*}

%\section{Vowels}
%\label{sec:vows}

 % appendix - phoneme to viseme character representation translations
\end{spacing}

%\begin{appendices}
%\input{appendix1} % appendix - words to E.A.Poes poem "The Raven".

%\input{appendixAa} % appendix - example confusion matrices for phonemes and visemes 
%\input{appendixB} % appendix - extra tables for p2v maps in chapter 7
%\input{appendixC} % appendix - extra tables for  experimental setup in chapter 7
%\input{appendixD} % appendais - publications

%\input{appendixBb} % appendix - all the P2V mappings used in controlled testing of phoneme groupings for measuring effects. 
%\end{appendices}
\end{spacing}

\backmatter
\end{document}